\documentclass[mnsc,nonblindrev]{informs3_modified} % current default for manuscript submission

\OneAndAHalfSpacedXI
\usepackage{statistics}
\usepackage[shortlabels]{enumitem}
\def\wty#1{\textcolor{black}{#1}}
\def\tyw#1{\textcolor{black}{#1}}

\def\IFx{{\rm IF}}

\usepackage{xcolor}

\usepackage{soul}
\soulregister\cite7
\soulregister\ref7
\soulregister\pageref7
\soulregister\Cref7
\def\Tr#1{\text{\rm Tr}[#1]}

\usepackage{natbib}
 \bibpunct[, ]{(}{)}{,}{a}{}{,}%
 \def\bibfont{\small}%
\TheoremsNumberedThrough     % Preferred (Theorem 1, Lemma 1, Theorem 2)
\ECRepeatTheorems

\EquationsNumberedThrough    % Default: (1), (2), ...
\MANUSCRIPTNO{}

\begin{document}

% \RUNAUTHOR{Iyengar, Lam, and Wang}

\RUNTITLE{OIC: Dissecting and Correcting Bias in Data-Driven Optimization}
% \TITLE{Bundling Information Goods of Decreasing Value}
\TITLE{Optimizer’s Information Criterion: Dissecting and\\
Correcting Bias in Data-Driven Optimization}

\ARTICLEAUTHORS{%
\AUTHOR{Garud Iyengar, Henry Lam, Tianyu Wang\footnote{Authors are listed in alphabetical order.}}
\AFF{Department of Industrial Engineering and Operations Research, Columbia University, New York, NY 10027
\EMAIL{garud@ieor.columbia.edu, henry.lam@columbia.edu, tianyu.wang@columbia.edu}}} %, \URL{}}
% \ARTICLEAUTHORS{}

\ABSTRACT{In data-driven optimization, the sample performance of the obtained
decision typically incurs an optimistic bias against the true performance,
a phenomenon commonly known as the Optimizer's Curse and intimately
related to overfitting in machine learning. We
develop a general approach that we call
\emph{Optimizer's Information Criterion (OIC)} to correct this bias. OIC generalizes the celebrated Akaike
Information Criterion from the evaluation of model adequacy, used primarily for model selection, to objective performance in data-driven
optimization which is used for decision selection. Our approach analytically approximates and cancels out the bias that comprises the interplay between model fitting and downstream optimization. As such, it saves the computation need to repeatedly solve optimization problems in cross-validation, while operates more generally than other bias approximating scheme. We apply OIC
to a range of data-driven optimization formulations comprising empirical
and parametric models, their regularized counterparts, and furthermore
contextual optimization. Finally, we provide numerical validation on the
superior performance of our approach under synthetic and real-world
datasets. 

% As such OIC can be used for
% decision selection instead of just model selection. 
% Unlike cross-validation, OIC analytically approximates and cancels the first-order bias, and as such it does not require repeatedly solving additional optimization problems.

% and are therefore computationally expensive. 

% that directly approximates
% the first-order bias and does \emph{not} require solving any additional
% optimization problems. % Our

}

\maketitle

\section{Introduction}\label{sec:intro}
% \begin{outline}
%     \begin{itemize}
%         \item What are some methods in the data-driven area; Data-Driven Optimization Literature; 
%         We would provide more detailed model formulations when we touch on each model later.
%         Note that we are not providing new methodologies but want to provide a criterion that suits different optimization streams for fair comparison; 
%         \item How to compare these methods; Statistical guarantees are not quite useful in comparing these approaches. still resort to data-driven evaluation.
%         \item Empirical evaluation would incur optimistic bias, which is not correct since it prefers these naive approaches instead of regularized approaches. 
%         \item Is there a fair criterion to include evaluate all kinds of approaches mentioned. How we should perform evaluations to see which method outperforms in each scenario. We give a brief introduction here. 
%     \end{itemize}
%     Our contributions are as follows:
%     \begin{itemize}
%         \item A methodology to evaluate decision performances that directly estimate and correct for the optimistic biases that incorporate all kinds of data-driven models;
%         \item This approach is computationally efficient without solving additional optimization problems; We also have closed-form bias, all we need is the function gradient and IF;
%         \item Have some user cases to explain bias more clearly(?) and work well in practical examples.
%     \end{itemize}
% THIS IS THE END OF OUTLINE.
% \end{outline}

We consider % a generic
data-driven stochastic optimization problem of the form
$\min_{x\in\mathcal X}\E_{\P^*}[h(x;\xi)]$, where $h(\cdot;\cdot)$ is a
known cost function, $\mathcal{X} \subseteq \mathbb{R}^{D_x}$ % is the decision variable in
is the set of feasible decisions,
%belongs to the
% space
% $\mathcal
% X$,
{and $\xi$ is a random perturbation.} % The expectation $\E_{\P^*}[\cdot]$ under
The (true) distribution $\P^*$
% that governs
of 
$\xi$ is  unknown; however, the decision maker has access to
% but observable via data
$n$ iid samples
$\Dscr_n:=\{\xi_i\}_{i = 1}^n$. The goal is to % identify a data-driven
use the data $\Dscr_n$ to identify a 
decision
% that gives
with 
the lowest % objective
expected cost with respect to the \emph{true} distribution $\P^*$. % or synonymously,  a decision that has
% the lowest  
% \emph{generalization} performance.
This problem % arises ubiquitously
is ubiquitous --
from empirical
risk minimization in machine learning~\citep{hastie2009elements}, to 
% nature arise % from the machine learning 
% communities to
% various
decision making % applications including
in
supply chain management~\citep{ban2019big,snyder2019fundamentals}, revenue
management~\citep{talluri2006theory,chen2022statistical} and portfolio
optimization \citep{ban2018machine}. 
% % where 
% % % decision maker aims
% % the objective function % is defined as an
% % involves
% % % to optimize the 
% % % expectation
% % an expectation over  % value 
% % % of an objective function
% % % involving
% % random variables.
% %where the goal is to optimize that objective based on a set of samples. 
% % There are several different approaches to data-driven optimization
% % problems to integrate data in the literature.  to integrating data into
% % optimization in the setting described above

Given the importance of the problem, 
a range of different approaches have been proposed to % find
identify 
good data-driven decisions. % have been studied in the literature. 
A natural approach is empirical optimization, or sample average
approximation, where % one replaces the
% unknown
the
expectation with respect to the unknown distribution % wi
% th its
is replaced with its 
empirical % counterpart
estimate~\citep{shapiro2021lectures}. This approach % ,
% however, could
can
lead to % tremendous
to % a
significant 
overfitting in % complex problems with a
% % large
% high dimen
% decision space,
when   cost function $h$ has high complexity and the decision dimension
$D_x \gg 1$,  
or in the contextual
optimization setting where the decision depends on exogenous features. In such
situations, one often resorts % to model-based schemes where the decision
% is
to 
a parameterized decision rule $\{x^*(\theta) \in \Xscr|\theta \in \Theta\}$, where $\Theta
\subset \R^q$, % is of finite dimension,
and the goal reduces to computing  the estimate 
% parameter
$\hat{\theta} \in \Theta$ % such that corresponding decision  and we output a
% decision
using the data $\Dscr_n$. \wty{This parametrized decision rule can be
  indirectly  induced % from a
  by assuming a 
  parametric % distribution model for the data.
  model for the uncertainty $\xi$, i.e., 
% The selection of the decision class $\paranx$ and the estimation
% procedure $\hat\theta$ based on the dataset $\Dscr_n$ is chosen by the
% decision-makers and may recover certain classical data-driven
% optimization methods 
% } 
% can be
% viewed as a choice of the decision class $\{\paranx \in \Xscr| \theta \in \Theta\}$ and an estimation method
% for choosing $\hat{\theta} \in \Theta$ using the dataset $\Dscr_n$. 
% $x^*(\hat\theta)$ with $\hat{\theta}$ learned from data. Finding a
% proper parametric decision rule, i.e., a decision class $\{x^*(\theta),
% \theta \in \Theta\}$ and a good estimate of $\hat\theta$, is at the core
% of data-driven optimization and has spurred many works in recent years.
% In this case,
$x^*(\theta)  = \argmin_{x\in\mathcal
  X}\E_{\P_{\theta}}[h(x;\xi)]$ for $\P_{\theta}$ %  is % a distribution in
% % a
% from a 
% parametric family % , say 
inside a family of distributions  $\Pscr_{\Theta} = \{\P_{\theta}|\theta \in
\Theta\}$. 
% Note that we do not require that  true distribution $\P^*$ % s
% % not necessarily in
% belongs to 
% $\Pscr_{\Theta}$. % some   
% can be derived from a 
% parametric distribution
% model $\Pscr_{\Theta} = \{\P_{\theta}|\theta \in \Theta\}$ , with 
The estimated parameter $\hat\theta$ % obtained via
can be % obtained
computed via}
two-stage estimate-then-optimize~\citep{bertsimas2020predictive,bertsimas2021bootstrap,hu2022fast}
or integrated estimation-and-optimization~\citep{qi2021integrated,elmachtoub2022smart} procedures. 
% Furthermore, optimization problems often amplify errors in the parameter
% estimate, and different approaches attempt to mitigate this
% phenomenon. 
Alternatively, $x^*(\theta)$ % can come from an 
can be represented by an end-to-end % representation
% without necessarily modeling the distribution
architecture~\citep{donti2017task} that maps the data into a decision, and 
$\Theta$ is the set of parameters that defines the function
approximation architecture. 
\wty{% Note that most of the % references cited
  % % here focus
  % previous work focuses
  % on the contextual optimization setting.
  % To align with the contextual
  % optimization framework, our notation
  % We
  Note that in 
  contextual optimization with feature $z$, all these would apply to solution
  % can be extended to
  $x^*(\theta, z)$ that depends on $z$ and generalizes % to incorporate an additional
  $x^*(\theta)$, though we would suppress notation to make our exposition simpler.}
  % , to where $z$ denotes the contextual 
  % covariate. % However, we
  % % omit this to
  % In order to keep the exposition simpler
  % we first develop our results for the non-contextual and extend them
  % later.
  % 
Finally, one can further exploit
% These include   
% On the other hand, to reduce the problem with limited data, 
distributionally robust 
optimization (DRO)~\citep{delage2010distributionally,wiesemann2014distributionally,blanchet2019quantifying,lam2019recovering,duchi2019variance,gao2022distributionally}
and other regularization 
approaches~\citep{maurer2009empirical,srivastava2021data} to control the error propagation due to parameter
estimation and model bias. 

% Despite the abundance of proposed
In spite of a diversity of solution
approaches, the effectiveness of the methodologies is, typically, established using
bespoke analyses relying on 
% most arguments in the
% literature to support their usages are based on individual analyses,
% typically on the 
finite-sample generalization bounds~\citep{el2019generalization,qi2021integrated,iyengar2023hedging}. % While
% insightful, these analyses are based on worst-case calculations and could
% be overly conservative for any particular problem instance.
Consequently, the performance bounds tend to be overly conservative for
any particular problem instance. Furthermore, the conservative % behavior
nature of the bounds
does not allow one to use them to determine the best solution
approach 
% More
% importantly,
% Because of this, % they do not adequately conclude
% these approaches are not adequate for concluding 
% whether a particular
% method is better than another 
for a given problem instance. % To this end, some recent
% works attempt to compare different schemes to obtain data-driven
% decisions,
% To this end, some 
Recent work~\citep{hu2022fast,lam2021impossibility,elmachtoub2023estimatethenoptimize}
% on 
% based on
uses convergence rates or stochastic dominance % ideas
results
to compare different data-driven decision schemes; however, these
approaches impose assumptions that can be difficult to verify and arguably idealized. 
% However,
% taking aside the theoretical assumptions needed in these comparisons that
% can be difficult to verify,
% n summary, 
% % still
% % appears to be 
% these analyses
In a nutshell, these results
do not translate directly into % selection of
practical methods for selecting 
% from the practical selection of 
data-driven decisions % s, because the latter
% requires a more instance-specific evaluation from the data and problem
% structure in hand that these theories are not yet able to capture.
for a particular problem instance and data $\Dscr_n$.

% restricted decision methods via margin conditions or model specification conditions, which
% and finite-sample rates \citep{

% have focused on 

% , most of 

% \wty{Among so many proposed methods, we want to choose the best one from them through practical evaluation using data.} 

 % or asymptotic stochastic dominance \citep{lam2021impossibility,elmachtoub2023estimatethenoptimize}. These works are useful in revealing some insights. However, the theoretical 
 % ; while stochastic dominance arguments, being asymptotic, might lack precision for finite samples. \wty{Besides, they only compare the decision performance of some restricted decision methods via margin conditions or model specification conditions, which are difficult to verify in practice.} 

% Motivated by the above, 
Our goal is to develop a practical methodology to accurately evaluate the % objective
performance of different data-driven decisions % by only 
using only the available
data, i.e.,  % themselves.
% By applying our methodology across different decision rules, a decision
% maker can confidently select the best decision across alternatives.
we provide a \emph{decision selection} criterion. 
This is analogous to the classical statistical problem of \emph{model
  selection}, but instead of focusing % the fitting quality of each
only on how well a probabilistic model fits the data, 
% model, 
we % conduct \emph{decision selection}
select decision rules 
based on the decision's objective % performances
value. Note that this  % latter 
goal involves not only the adequacy of the fit of the probabilistic model,
but also its
interaction with the downstream optimization task.  
More concretely, % our primary criterion to evaluate is
we want to evaluate
\begin{equation}\label{primary}
A_u:=\E_{\Dscr_n}\left[u\left(\E_{\P^*}\left[h(\datax;\xi)\right]\right)\right],
\end{equation}
where $\E_{\P^*}[h(\datax;\xi)]$ is the \emph{true} objective value of the
data-driven solution $\datax$. Note that $\E_{\P^*}[h(\datax;\xi)]$ is a
random variable % where the randomness comes from the
since $\hat{\theta}$ is a function of the random
data $\Dscr_n$. %  in
% driving the solution. 
% To this end,
The user-prescribed deterministic risk function~$u(\cdot)$
% is a deterministic risk function, prescribed by the user, that together with the outer expectation $\E_{\Dscr_n}[\cdot]$ 
captures the distributional aspect of this random variable. For example, if $u(x) = x$, then \eqref{primary} reduces to the expected objective performance, in the form
% . More specifically, this reduces to the estimation of:
% In our setup, 
% %it is natural to consider the expected true performance of $\datax$ given by:
% this risk reduces to: \E_{\Dscr_n}[Z(\datax)] = 
\begin{equation}\label{eq:eval-key}
    A:= \E_{\Dscr_n}\E_{\P^{*}}[h(\datax;\xi)].
\end{equation}
Thus, the smaller $A$ is, the better is the decision $x^*(\hat\theta)$ % is
% in terms of its objective performance
on average.

% with $g$ deteting the risk measure and $X$ denoting the random term 
% $Z(\datax):= $ 

% since $\hat\theta$ depends on the realization of $\Dscr_n$. This criterion can be used in characterizing the moment behavior.
%or to specify different types of stochastic ordering \citep{shaked2007stochastic}. 
% We estimate $A$ within the main body of our discussion, as this estimation serves as a foundation for general function forms $g$. Later in \Cref{sec:general-risk}, we will explore the case of general $g$, referring to the quantity associated with $u(\cdot)$ as general risk measures.
 
 % decision-making objective instead of the model fitting quality. 

 % To turn these decision rules into practical use, an important step is to provide accurate instance-specific evaluations of the true performance so that the 
 
% \wty{Our goal in this paper is to provide a computationally efficient criterion to evaluate and compare various decision rules.} We would like to evaluate the decision performance via 

%Specifically, one natural criterion to evaluate decision performance is as follows. 
%Given an estimator which is the output of some estimation procedure, %\wty{$\hat\theta=T(\hat\P_n)$ for some functional as a particular estimation procedure $T$}

%Note that this criterion depends on both how we parameterize $x^*(\theta)$ and how we estimate $\hat\theta$ from data, the latter possibly intertwined with the former. empirical objective value from the obtained decision  for~\eqref{eq:eval-key} for a fixed parametric decision class $x^*(\theta)$, 

% To get a sense of the
In order to focus on the key 
underlying challenges % that motivates 
our % key
approach needs to address, we focus on~\eqref{eq:eval-key}, % the evaluation of
and later extend the results to~\eqref{primary}. A natural % approach to
% estimating
estimate $\hat{A}_o$ for~\eqref{eq:eval-key} is % to take
the empirical average $\hat{A}_o:= \frac{1}{n}\sum_{i = 1}^n
h(\datax;\xi_i)$. However, % this
$\hat{A}_o$ typically suffers from an \emph{optimistic bias} in that the estimate 
% giving a performance evaluation of the obtained decision $\datax$ that
$\E_{\Dscr_n}[\hat{A}_o] < A$, i.e.,  
% better than
the ground-truth performance. % As a simple example,
To understand the source of this bias, consider the setting where 
% suppose
% our
the 
decision rule % is to find $\hat\theta$
chooses a decision $x$
that minimizes the empirical
objective $\hat{A}_o$, i.e., solves the so-called \emph{empirical optimization}  problem.  
% $\hat{x}$ denote the ERM solution,
Then a simple application of Jensen's inequality establishes that the
bias  $A - \E_{\Dscr_n}[\hat{A}_o]\geq 0$, suggesting that 
% Then we have $\E_{\Dscr_n}[\hat{A}_o] \leq A$ by a simple use of
% Jensen's inequality \citep{mak1999monte}. In other words, because of the
% bias $A - \E_{\Dscr_n}[\hat{A}_o]$, a 
a decision with a low empirical objective value may not necessarily have a
low out-of-sample performance. This is also known as Optimizer's
Curse~\citep{smith2006optimizer}, and % shares the same spirit as
is related to 
overfitting in machine learning \citep{hastie2009elements}.  

% Therefore we are interested in methods that fully utilize all data in constructing solutions. \wty{  is a standard approach to evaluate different methods in practice. However, Our proposed OIC  under the general stochastic optimization setup which is formally defined in \Cref{sec:main}.  by its formula in \Cref{def:oic}
% \begin{new}
\subsection{Our Contributions}
% Our main contribution is a
We propose \emph{Optimizer's Information Criterion (OIC)}  as a 
new methodology
  to evaluate the
true decision performance 
\eqref{eq:eval-key}, and more generally 
\eqref{primary}, for a very general class of decision rules by correcting
the optimistic bias associated with the empirical estimate.
% We call our approach which can be viewed as
OIC % can be viewed as
% s 
% a 
% generalization
is a generalization of the celebrated Akaike Information Criterion
(AIC)~\citep{akaike1974new}, % from the evaluation of model fit, used
% primarily for model selection,
used for model selection, to selecting decision rules 
% to objective performance 
in 
data-driven optimization. % for decision selection.
OIC directly approximates and eliminates the first-order bias % that
% comprises
arising from the interplay between model fitting and downstream optimization
(\Cref{def:oic}), and is both computationally efficient, in the sense that, unlike other methods such as cross-validation, one does
not need to solve additional optimization problems, and is statistically efficient, in the sense that it 
removes 
bias without 
increasing % the
variability~(\Cref{coro:main-moment}). % compared
% with the 
Furthermore, OIC serves as an effective decision selection criterion due
to its similarity to LOOCV when evaluating~\eqref{eq:eval-key}
(\Cref{prop:loocv}). % These
All these
% main
results % arise
follow
from characterizing the
evaluation bias of the empirical average % through
via
a delicate second-order Taylor
analysis, \wty{incorporating a unified influence function expansion} on the data-driven solution performance (Theorem~\ref{thm:main}). We showcase the generality of OIC by applying it to
empirical and parametric decision rules, including regularized,
distributionally robust, and constrained variants. For each decision rule,
we provide explicit and computable expressions for the bias. % formulas
% that can be directly computed. 
To our
best knowledge, OIC offers the most general explicit bias correction
formula to-date, encompassing AIC and other variants in the literature. % These applications All these % developments
These results
are established
% are presented
in \Cref{sec:main}. 

% In \Cref{sec:apply},  we showcase the generality of OIC by applying it to
% empirical and parametric decision rules, including regularized,
% distributionally robust, and constrained variants. For each decision rule,
% we provide \gi{explicit and computable expressions for the bias.} % formulas
% % that can be directly computed. 
% To our
% best knowledge, OIC offers the most general explicit bias correction
% formula to-date, encompassing AIC and other variants in the literature. % These applications of OIC are presented in.  
% Additionally, in \Cref{sec:usercase}, we illustrate our results for specific cost functions.

  % OIC can be used to evaluate general common decision rules in \Cref{sec:apply}. Specifically, 
%Generability: We then apply OIC to common decision rules \Cref{sec:extension} %provides several generalizations of OIC including general risk functions for~\eqref{primary},  Furthermore, OIC is also general in evaluating various performance objectives with a small bias. In Sections~\ref{sec:main} to~\ref{sec:usercase}, we evaluate the objective under~\eqref{eq:eval-key}. However, 

We extend 
% With our developments for evaluating~\eqref{eq:eval-key}, we expand our 
OIC in several significant directions in \Cref{sec:extension}. First, in \Cref{subsec:application-if}, we study OIC for general procedures that include both common data-driven optimization approaches and other non-canonical counterparts. Along the way, we explain the principles in achieving these general results. In \Cref{sec:general-risk},  we generalize OIC to % what we call 
Risk-OIC  % used to evaluate performances
% used
to evaluate % performance for more general risk function $u(\cdot)$ in
decision rules according to~\eqref{primary}. 
Unlike % the expected performance case in 
\eqref{eq:eval-key}, Risk-OIC involves the derivative of the risk function
and variance of the cost. In \Cref{sec:POIC},  % we investigate a
% modification of OIC, which we call
we introduce 
Parametric-OIC to evaluate distributional misspecification errors for some
decision rules. In \Cref{sec:context}, we introduce Context-OIC
that % applies to
addresses
contextual optimization. Our bias correction property
  and other % advantages
  results
  continue to hold for these OIC variants with % the
  respect to their target problems. 
% Risk-OIC and Context-OIC under contextual optimization cases, including for various (constrained) decision rules described in \Cref{sec:apply}. For clarity, we discuss the main results in the non-contextual case.}

After we understand key properties of OIC and its range of variants, we then present provable relative benefits of OIC, in terms of both computation and generality, compared to possible alternatives. These  results, as well as further discussions of our approach, are in Section \ref{sec:challenge}.
% In Section
% \ref{sec:challenge}, we % will
% % We will demonstrate
% discuss
% % in detail 
% the benefits of OIC % offered
% in terms of computation and its generality
% as compared to possible
% alternatives. 
Then, in \Cref{sec:numerical}, we discuss the results of our 
% Finally,  we conduct 
numerical % studies
experiments using 
% on
both synthetic and real-world data with % examples
applications
ranging from portfolio optimization, newsvendor problems to regression
problems. % Our experiments
Both our theoretical and empirical results
show that OIC exhibits a small bias against oracle evaluations and can be computed efficiently across various decision rules. 
% Lastly, in \Cref{sec:discuss}  we % provide
% % some discussions of further utility of OIC.
% discuss some limitations and future % potential utilities and investigation 
% directions of OIC.  
Proofs of all theoretical results, additional discussions, and experimental results are deferred to the Appendix.

\subsection{% Motivating Illustration of 
  Comparison of OIC with Benchmark Approaches}
% Here we give a brief overview of the comparisons between OIC and other alternatives and present an illustrative example. 
Before going into details, we briefly highlight the pros and
cons of OIC relative to the existing methods in evaluating data-driven decision performance, summarized in
Table~\ref{tab:eval}.
% highlights these comparisons. % our
% the
%  discussed above. 
 A well-known % idea
approach 
to address optimistic bias is to % separate
split 
the data into two sets: using one set for % training
computing
the decision % decision
% training and for evaluation. For example, one can simply split the data into two portions, one for constructing the decision
and % one
the other 
for evaluating~\eqref{eq:eval-key}. However,  % but
this approach loses statistical efficiency % not
since not % not
all of the data % are
is 
used to % obtain
compute 
the decision.  
A standard remedy % , one can
is to 
use cross-validation (CV) that repeats the above approach, each time using
a different % division
split
of the data into decision % training
training 
and evaluation. The two main variants are $K$-Fold CV that divides the data
into $K$ batches (where $K$ is small) and each time uses all but one batch
to train the decision, and Leave-One-Out CV (LOOCV) % where the former
                                % (where $K$ is understood to be a small
                                % number), and the latter
that 
uses all but one observation to train the decision. % Despite the
% generality of CV,
Both of these two variants encounter either statistical or computational
challenges: % In particular,
$K$-Fold CV with a small and fixed $K$ incurs a pessimistic bias % due to
because of fewer samples used in each
round~\citep{fushiki2011estimation,gupta2022debiasing}, whereas  % In
% comparison, 
LOOCV involves solving $n$ additional optimization problems, % which makes
rendering it computationally demanding.  
%A number of 
% efficient
%approaches, i.e. 
More recently, the so-called Approximate Leave-One-Out (ALO) methods have been proposed to approximate LOOCV % efficiently by approximating
% each leave-one-out sample 
% by injecting some analytical information to
and 
avoid repeatedly solving optimization problems by injecting some analytical information~\citep{beirami2017optimal,giordano2019swiss,wilson2020approximate,rad2020scalable}. However,
these ALO methods % approximating
are built for specific estimation problems, e.g., for $u(x) = x$, and
unconstrained optimization. % Compared to these methods,
{OIC is more % more
computationally efficient % than
compared to 
LOOCV, since it % does not involve repeatedly
only solves a single % solving
optimization problem, 
more statistically efficient than $K$-Fold
CV, since it does not incur a pessimistic bias, and more general as compared to  % than
ALO
methods. On the other hand, in terms of needed structural information, 
OIC is similar to 
% like
ALO methods in that it does require a level of
analytical knowledge on the decision rule and target optimization
problem.}

% The existing evaluation procedure can be divided into two lines: the first line comprises CV and ALO approaches, while the second line focuses on specific bias correction approaches such as AIC. Within the two aforementioned streams, our work is dedicated to achieving broader applicability of bias corrections for a wider range of data-driven models while maintaining computational efficiency and transparency in our bias correction procedures. Table~\ref{tab:eval} highlights the difference between our procedure and all these existing methods.
%and list the comparison in \Cref{tab:eval} for a problem with $n$ sample points:
\begin{table}[!h]
    \caption{Comparisons among different evaluation methods. ``$\# \hat\theta$ to compute'' and ``$\# x^*(\cdot)$ to compute" refer to the numbers of times we need to compute $\hat\theta$ using (some subset of) data and to evaluate $x^*(\theta)$ at different $\theta$ respectively, ``Model-free'' means that the evaluation method does not require knowledge of the cost function and the decision rule, ``Unknown'' means that the considered case has not been studied in the existing literature. }
    %\wty{(ADD A ROW ON GENERAL RISK FUNCTION?)}} 
    %The optimistic bias may help better in nonsmooth objectives.}
    \label{tab:eval}
    \centering
        % \toprule
        %  & LOOCV & $K$-Fold CV & BC $K$-Fold CV & ALO (JK / NJ) & AIC-type & OIC\\
        %  \midrule
        % \# $\min_{x}\E[h(x;\xi)]$ to solve &$n$&$K$&$K + 1$ & 1  &1& 1 \\
        % \# $x^*(\theta)$ to compute & $n$ & $K$ & $K + 1$ & $n$  &1& 1\\
        % Assumption Conditions & Minimal & Minimal & Minimal & IF & Obj& IF\\
        % Bias Order & $O\Para{\frac{1}{n}}$& $O\Para{\frac{1}{n}}$ & $o\Para{\frac{1}{n}}$ & $o\Para{\frac{1}{n}}$ &  $o\Para{\frac{1}{n}}$ & $o\Para{\frac{1}{n}}$\\
        % Cover all problems & Yes & Yes & Yes & Yes but unknown  &No &  Yes\\
        % Explicit Optimistic Bias &  No & No & No & No & Yes & Yes\\
        % \bottomrule
    % \begin{tabular}{c|cccc|c}
    %     \toprule
    %      & LOOCV & $K$-Fold CV & ALO & AIC-type & OIC\\
    %      \midrule
    %     \# $\min_{x}\E[h(x;\xi)]$ to solve &$n$&$K$ &  $\geq 1$  &1& 1 \\
    %     \# $x^*(\theta)$ to compute & $n$ & $K$  &  $n$  &1& 1\\
    %     Model-free&  \cmark & \cmark & \xmark & \xmark & \xmark \\
    %     Remove $\Theta(1/n)$ bias in~\eqref{eq:eval-key} & \cmark & \xmark & \cmark  & \cmark & \cmark \\
    %     Encompass constrained problems& \cmark & \cmark  & Unknown  & \xmark & \cmark \\
    %     Evaluate for general risk function $u(\cdot)$ & Unknown & Unknown & Unknown & \xmark & \cmark\\
    %     Explicit bias formula&  \xmark & \xmark & \xmark & \cmark & \cmark \\
    %     \bottomrule
    % \end{tabular}
    \begin{tabular}{c|cccc|c}
        \toprule
         & LOOCV & $K$-Fold CV & ALO & AIC-type & OIC\\
         \midrule
         $\# \hat\theta$ to compute & $n$&$K$ &  1  &1& 1 \\
         $\# x^*(\cdot)$ to compute &$n$&$K$ &  $n$  &1& 1 \\
        % \# $x^*(\theta)$ to compute & $n$ & $K$  &  $n$  &1& 1\\
        Model-free&  \cmark & \cmark & \xmark & \xmark & \xmark \\
        Remove $\Theta(1/n)$ bias in~\eqref{eq:eval-key} & \cmark & \xmark & \cmark  & \cmark & \cmark \\
        Encompass constrained problems& \cmark & \cmark  & Unknown  & \xmark & \cmark \\
        Evaluate for general risk function $u(\cdot)$ & Unknown & Unknown & Unknown & \xmark & \cmark\\
        Explicit bias formula&  \xmark & \xmark & \xmark & \cmark & \cmark \\
        \bottomrule
    \end{tabular}
\end{table}
 
 % for lack not computationally efficient in some decision procedure.  When considering specific cost functions, certain problem-dependent bias correction approaches, such as AIC and other variants with a dedicated bias formula, have been proposed. However, they are not applicable to general cost functions. We will provide more discussions on the strengths and drawbacks of these evaluation approaches in Sections~\ref{sec:challenge} and~\ref{sec:discuss}. We call our approach the \textit{Optimizer's Information Criterion (OIC)}, a term in analog to, but also generalizing the celebrated Akaike Information Criterion (AIC) \citep{akaike1974new} from model selection to decision selection that takes into account not only model fitting but also the downstream optimization task. 

% As such, it saves the computation need to repeatedly solve optimization problems in cross-validation, while operates more generally than other bias approximating scheme. 

% which comprises the interplay between model fitting and downstream optimization and as such it aims to address decision selection. 

%provides a reliable estimate of the true performance % while requiringbut 

% \paragraph{Motivating Example.}

\begin{figure}[!htb]
    \centering
    \includegraphics[width = 0.8\textwidth]{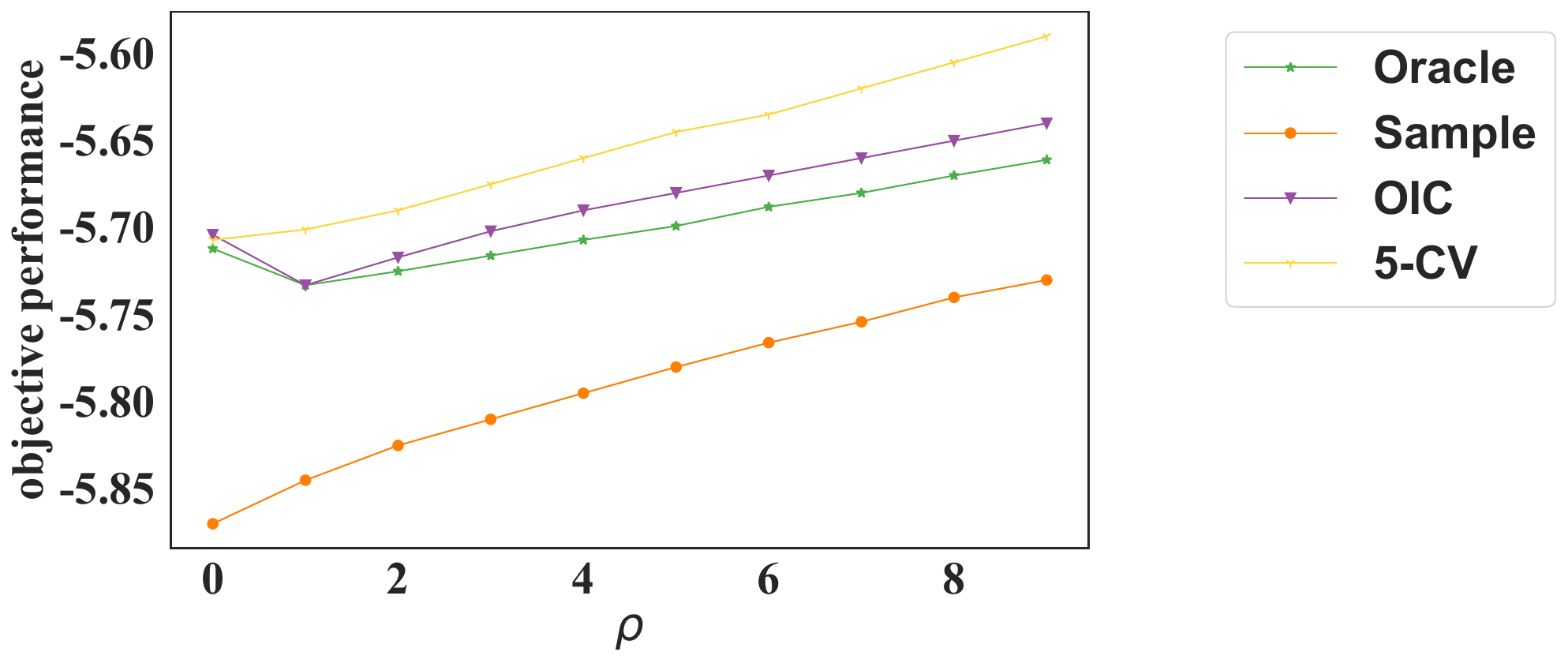}
    \caption{Evaluation of the expected objective performance \eqref{eq:eval-key} for a 40-dimensional portfolio optimization problem (which minimizes a portfolio risk criterion in \Cref{subsec:portfolio}) with $n = 100$, formulated using $\chi^2$-divergence DRO with ball size $\varepsilon = \rho/n$ ($\rho = 0$ corresponds to the empirical optimization). The evaluations are based on the naive empirical average (``Sample"), 5-Fold CV (``5-CV") and OIC, and benchmarked with the true objective value (``Oracle").}
    \label{fig:chi2dro-illustrate}
\end{figure}

%\wty{Recall Minimizer.}
Next, to illustrate the power of OIC in practice, we briefly discuss the results for a
$40$-dimensional portfolio optimization problem % which minimizes a
% portfolio risk criterion and will be elaborated further in
(see~\Cref{subsec:portfolio} for details). % In this example, our goal
% Our goal 
% is to select
The portfolios were  selected 
% best decision 
% obtained from
using DRO with the ambiguity set % constructed
given by a
$\chi^2$-divergence-based % neighborhood
ball % surrounding
of radius~$\rho$ around
the empirical
distribution, and the goal was to select % the ball size
$\rho$
that minimizes the
objective. 
% Here, the ball size, which determines the decision, needs to
% be selected.
In 
\Cref{fig:chi2dro-illustrate} we compare the estimated 
% shows the evaluation result of t
expected objective performance \eqref{eq:eval-key} for the DRO solutions
as a function of the ball size computed using naive empirical
average~(Sample), 5-Fold CV~(5-CV) and
OIC~(OIC) against the true performance ({Oracle}). From the plot
corresponding to the Oracle, it is clear that the true optimal $\rho^* = 1$,
and 
we see
that OIC approximates the true expected objective curve well enough
% therefore
% , is able 
to find the true optimal $\rho^*$. % best
% ball size. 
% This is in contrast to
In contrast, the 
naive empirical average underestimates the objective, and 5-Fold CV
overestimates the objective. % Moreover, though not shown in the
% figure,
Although not shown in the figure, the performance of 
LOOCV % attains
is
similar to % statistical performances as 
OIC, but has a
% requires
substantially % more
higher
computation cost compared % than
to
all the methods shown in
Figure~\ref{fig:chi2dro-illustrate}.  Finally, ALO methods for DRO have not been studied in the literature.
% for % this
% DRO
% have not been studied % at all 
% in the literature. 

 % attaining the largest expected performance with computational efficiency
%as in \cite{gotoh2021calibration} 
% and thus 

 % of $\E_{\Dscr_n}[Z(\datax)]$ (with respect to a hyperparameter $\rho$)

% , and benchmarked with the true objective value (``oracle"). } Through OIC, we can correctly choose the best solution with computational efficiency, while the naive evaluation without bias correction fails. In the case of \Cref{fig:chi2dro-illustrate}, compared with the naive evaluation and cross-validation criterion, 

\paragraph{Notation.} $\nabla_{\theta}$ denotes the gradient
(or subgradient) and $\nabla_{\theta\theta}^2$ denotes the Hessian
matrix with respect to $\theta$. $\Tr{\cdot}$ denotes the trace of a
matrix, $D_{\cdot}$ denotes the dimension of a variable $\cdot$, and $d(\cdot, \cdot)$ denotes a distance between two distributions. $\hat{\P}_n:=\frac{1}{n}\sum_{i = 1}^n\delta_{\xi_i}$ denotes the empirical distribution of $\{\xi_i\}_{i = 1}^n$, where $\delta_{\xi}$ denotes the Dirac measure centered at $\xi$. $1_{D}$ denotes the $D$-dimensional vector with all entries being $1$ and $\mathbf{1}_{\cdot}$ denotes the indicator function. \wty{We use notations $o(\cdot), O(\cdot), \Omega(\cdot), \Theta(\cdot)$ to describe the asymptotic performance of a function.
Specifically, given two functions $f(n)$ and $g(n)$, $g(n) = \Omega(f(n))$ (or $f(n) = O(g(n))$) if there exist positive constants $c_1$, and $n_0$ such that $|g(n)| \geq c_1 |f(n)|$ for all $n \geq n_0$; $g(n) = \Theta(f(n))$ if $g(n) = \Omega(f(n))$ and $f(n) = \Omega(g(n))$; $f(n) = o(g(n))$ if for any positive constant $\varepsilon$, there exists $n_0$ such that $|f(n)| \leq \varepsilon |g(n)|$ for all $n \geq n_0$. Similarly, $f(n) = \tilde o(g(n))$ if $f(n) = o(g(n)\log^k n)$ for some constant $k$. We use $o_p(\cdot), O_p(\cdot)$ to describe the asymptotic convergence of random variables. For a set of random variables $X_n$, $X_n = o_p(f(n))$ if $X_n / f(n) \convp 0$ as $n \to \infty$; $X_n = O_p(f(n))$ if for any $\epsilon > 0$, there exists a finite $M > 0$ and $n_0 > 0$ such that $P(|X_n / f(n)| \geq M) \leq \epsilon, \forall n \geq n_0$.}
% Similarly, $g(n)$ is said to be $\widetilde{\Theta}(f(n))$ if $g(n)$ is $\Theta(f(n) \cdot \log^k n)$ for some constant $k$.
\section{Main Results}\label{sec:main}

Consider the optimization problem $\min_{x\in\mathcal
  X}\E_{\P^*}[h(x;\xi)]$  introduced in
Section~\ref{sec:intro}. Recall that the distribution $\P^*$ of the
uncertain perturbation $\xi \in \Xi$ is unknown,
and we only have $n$ i.i.d. samples $\Dscr_n= \{\xi_i\}_{i \in [n]}$ from
$\P^*$. We show in \Cref{ex:ddo} below 
that the solution computed by the various data-driven optimization
approaches can be expressed in a unified manner in terms of a decision rule
\begin{equation}\label{eq:constrain-prob}
  x^*(\theta) \in \Xscr := \{x \in \R^{D_x} : F_j(x) \leq 0,\ \forall j
  \in J_1, \ F_j(x) = 0, \forall j \in J_2\}.
\end{equation}
that is a known deterministic 
function of a parameter
$\theta\in\Theta$. % With data
% Let $\hat{\P}_n$ denote the empirical distribution of data set
% $\Dscr_n=\{\xi_i\}_{i = 1}^n$, where each
% $\xi_i \in \Xi$ is an observation drawn from the true distribution
% $\P^*$.
% Using the data
% \gi{Let $\P_n = \frac{1}{n} \sum_{i=1}^n \delta_{\xi_i}$ denote the empirical
% distribution of the data 
% $\Dscr_n=\{\xi_i\}_{i = 1}^n$, where each $\xi_i \in \Xi$ is an
% observation drawn % from
% according to the unknown
% $\P^*$.} 
% we obtain a parameter estimate $\hat\theta$
% from $\Dscr_n$. Specifically, we let
% we obtain a parameter
% estimate $\hat\theta$ from $\Dscr_n$. To this end, we
% where r a general constrained
% optimization setting, where:
The decision rule $x^*(\theta)$ is unconstrained
when $J_1 = J_2 = \emptyset$. \wty{Recall $\hat\P_n:= \frac{1}{n}\sum_{i = 1}^n \delta_{\xi_i}$ is the empirical distribution of $\{\xi_i\}_{i = 1}^n$.}
% we have: $\Xscr = \R^{D_x}$, corresponding to an
% unconstrained optimization problem.  
% The specific
% % We compute the
% estimate $\hat\theta = T(\hat\P_n)$, where $T: M(\Xi) \rightarrow \Theta$
% denotes a calibration 
% % procedure
% function
% that maps a probability distribution $\P \in M(\Xi)$  %  the space of
% % distributions with support on $\Xi$,
% to  
% a parameter value in $\Theta$ with $M(\Xi)$ being a set of probability
% distributions supported on $\Xi$.} 
% $\hat\theta =
% T(\hat\P_n)$, where
% Moreover, let 
% Suppose that, , where 
% and recalling $\mathcal{X}$ is the decision space, we represent the decision class 
%which we will specify in \Cref{ex:ddo} later. 
% Each $\theta \in \Theta$ is the output of a functional $T: M(\Xi)
% \rightarrow \Theta$, which takes a probability distribution from
% $M(\Xi)$, a space of distributions with support on $\Xi$. We output the
% decision $x^*(\hat\theta)$ where the parameter $\hat\theta$ is estimated
% through $\Dscr_n$, i.e., $\hat\theta = T(\hat\P_n)$. 
%with the empirical distribution $\hat\P_n = \frac{1}{n}\sum_{i = 1}^n \delta_{\xi_i}$.
%i.e. $T(\hat\P_n) = \hat\theta$.
%We denote its estimate using $\Dscr_n$ by $\hat{\theta}$. 
% The above
% This
% setup covers 
% %The parameterized decision  and estimation of $\hat{\theta}$ above represent 
% a wide range of data-driven optimization procedures.
% including the decision rules $\datax$ and calibration procedures $T(\cdot)$:
% and our proposed OIC applies to all cases in Definition~\ref{ex:ddo}:
\begin{definition}[General Data-Driven Optimization Procedures]
%Specifically, $T(\hat\P_n) = \hat\theta$.
%We consider the following estimation procedure $T$:
$\mbox{}$ \label{ex:ddo} 
\begin{enumerate}[(a),leftmargin=*]
    \item % We parameterize 
      % We parameterize 
      The distribution 
      %The distribution of 
      $\xi$ % via a probability distribution
      % s  parametric family
      %belongs to the 
      % via a
      is assumed to belong to a 
      parametric family
      $\Pscr_{\Theta}:=\{\P_{\theta}|\theta \in \Theta\}$, and the decision
      rule $\paranx\in\argmin_{x \in \Xscr}\E_{\P_{\theta}}[h(x;\xi)]$.
      % and the specific decision $x(\hat{\theta}_n)$ corresponds to the
      % estimate $\hat{\theta}_n = T(\P_n) \in \Theta$.
        % The uncertainty $\xi$ is distributed according to probability
      % distribution $\P \in \Pscr_{\Theta}:=\{\P_{\theta}: \theta \in
      % \Theta\}$, $\Xscr = \R^d$,  and  $\paranx\in \argmin_{x \in
      % \Xscr}\E_{\P_{\theta}}[h(x;\xi)]$.  
    \begin{enumerate}[(1),leftmargin=*]
        \item\emph{Estimate-Then-Optimize (ETO):} % 
          \wty{The estimate $\hat{\theta}\in\argmin_{\theta}\E_{\hat\P_n}[\phi(\theta;\xi)]$, where $\phi(\theta;\xi) \in \R^{D_{\theta}}$ is a function that does not
            depend on the 
          objective $h$. A 
          canonical % example
          choice here is 
          % but not $h$, such as
          maximum likelihood estimation, i.e.,   $\phi(\theta;\xi) = - \ln(p_{\theta}(\xi))$, where $p_{\theta}(\cdot)$ is the probability density function of the distribution $\P_{\theta}.$
          }
        \item\emph{Integrated Estimation-and-Optimization (IEO):}  
          \wty{The estimate $\hat{\theta}\in\argmin_{\theta \in
            \Theta}\E_{\hat{\P}_n}[h(\paranx;\xi)]$, i.e., it
          optimizes the empirical objective performance among the class of solution $x^*(\theta)$ instead of just model fitting. 
    }
    \end{enumerate}
    %\item $\theta$ does not necessarily represent distribution
    %parameter. We set any $\paranx$ that satisfies
    %\Cref{asp:represent}. Then,  
    \item {The decision rule $\{\paranx \in \Xscr | \theta \in \Theta\}$ is any parameterized family of 
      % family of  
      decision functions. }
      %Note that the decision rule where $\Theta = \Xscr$, and $\paranx = \theta \in \Xscr$ is a special case.} 
      % for 
      % $\theta \in \Theta$.
    \begin{enumerate}[(1),leftmargin=*]
    \item\emph{End-to-End (E2E):} $\hat{\theta}\in\argmin_{\theta \in
        \Theta}\E_{\hat{\P}_n}[h(\paranx;\xi)]$. \wty{Note that sample average
      approximation~(SAA) is a special case, corresponding to $\Theta =
      \Xscr$, $x^*(\theta) = \theta$, and $\hat\theta\in  \argmin_{\theta \in 
        \Theta}\E_{\hat{\P}_n}[h(\theta;\xi)]$. }
      % for any $\P$ supported on $\Xi$.
    % The formulation of the standard empirical approach is the same as end-to-end above.
    \item\emph{Regularized E2E (R-E2E):} $\hat{\theta}\in\argmin_{\theta}\{\E_{\hat{\P}_n}[h(\paranx;\xi)] + \lambda
      R(\paranx)\}$ for some regularizer~$R(\cdot)$. % for any $\P$ supported on $\Xi$.
    \item\emph{Distributionally Robust E2E (DR-E2E):} $\hat{\theta}\in\argmin_{\theta}\max_{d(\P, \hat{\P}_n) \leq
        \epsilon}\E_{\P}[h(\paranx;\xi)]$ for some distance metric $d$ and
      $\epsilon = O(1/n)$. 
    \end{enumerate}
    %     We
    % let $\theta^* := T(\P^*)$ denote the \emph{true} parameter value.
\end{enumerate}
\end{definition}
% % Different model parametrizations of $\{\paranx: \theta \in \Theta\}$ result in different decision rules. 
% % % the standard protocol of data-driven optimization is divided into two
% % % stages, i.e. first estimate the uncertainty parameter in the objective,
% % % optimize the cost in terms of the given model. The classical
% % And even with the same parametrization $\paranx$, the approach to
% % estimating $\hat{\theta}$ given $\Dscr_n$ and the cost function 
% % %function mapping from parameter $\theta$ to decision $x$ 
% % can be different depending on the definition of $\theta$.
% % %, especially when $\theta$ is the parameter of a family of
% % distributions.
% \wty{In the optimization problem of each procedure, we do not require
\wty{% Note that the optimization problem defining
  Although we define
  the estimate $\hat{\theta}$ in each case above as an optimal solution of an optimization
  problem, our results hold as long as the estimate is 
  % to be solved exactly,
  % need not be solved to optimality. We allow for an 
  % % while we allow the optimization error of the
  % order 
  $o_p(n^{-1})$ % error
  optimal. 
  % in % terms of 
  % the corresponding first-order
  % optimality conditions. 
  See \Cref{asp:theta-hat} for details.}

% % X % that our subsequent theoretical results are valid
% % the dichotomy into 

Parts (a) and (b) in \Cref{ex:ddo} 
% corresponds to the construction of the
differ in how the 
decision rule $x^*(\theta)$ is constructed. In (a), 
the % basic
% construction
primitive 
is a parametric family of distributions $\Pscr_{\Theta}$, and 
$x^*(\theta) % is the solution to an \emph{oracle} problem 
= \argmin_{x \in \Xscr}\E_{\P_{\theta}}[h(x;\xi)]$.
% $x^*(\theta)$ is the solution to an \emph{oracle} problem $\argmin_{x}\E_{\P_{\theta}}[h(x;\xi)]$ that assumes the distribution is $\P_{\theta}$, i.e., $\theta$ is a parameter in the distribution model. 
%  % that assumes the distribution is $\P_{\theta}$, i.e., $\theta$ is a parameter in the distribution model. To this end,
The two sub-categories (1) and (2) % correspond to two methods to
% calibrate
correspond to whether~(IEO) or not~(ETO) the downstream objective $h(\cdot, \cdot)$ is
taken into consideration in estimating~$\hat\theta$. % namely ETO and 
% IEO, which differ in whether the
% downstream cost function is
% incorporated. 
% In particular, ETO estimates $\theta$ using only the data, without incorporating the optimization objective function, while IEO estimates $\theta$ by minimizing the empirical objective function over the class of decision $x^*(\theta)$. On the other hand, 
In (b), the % basic
% construction
primitive is the decision 
function $\paranx$. % is not necessarily a parameter in a family of distributions and $x^*(\theta)$ is not necessarily the solution to an oracle problem. Instead, we permit $x^*(\theta)$ to be any function of $\theta$ in an Euclidean space. 
In this setting, $\hat{\theta}$ can be estimated through empirical minimization (E2E), % in (b)(1), 
or % with
regularized empirical minimization~(R-E2E), % in (b)(2),
or via a distributionally robust formulation (DR-E2E). \wty{We will only
  consider setting $d$ to an $f$-divergence in the DR-E2E setting since % when
  DR-E2E with  $d$ given by the Wasserstein distance % such DRO problems are known to
  % be
  is approximately equivalent to norm-regularization~\citep{blanchet2019robust,gao2022wasserstein}. % above
% then the 
% result for R-E2E % where we set
% with
% $R(x) = \|x\|_2$ would apply. 
% in (b)(3). 
% In general,
We use 
``end-to-end (E2E)" learning to refer to % training a model that produces
% the
a training process that produces both the 
estimate $\hat{\theta}$ and the decision $x^*(\hat{\theta})$ using  a single, unified
task~\citep{donti2017task,agrawal2019differentiable}, and hence we refer all the
procedures in 
% Here, we refer to
\Cref{ex:ddo}$(b)$ as end-to-end. % as an ``end-to-end'' procedure since it optimizes only
% for one single instance, directly producing $\hat\theta$ and the
% corresponding decision $x^*(\hat\theta)$ without solving additional
% optimization subproblems, unlike the procedures in
% \Cref{ex:ddo}$(a)$.
The {end-to-end training} can be applied to both the non-contextual setting
that we focus on here, and the contextual setting that we discuss
in~\Cref{sec:context}. }
%\hl{Why isn't IEO also an end-to-end process?} \wty{TW: I later think it should be changed to Decision-rule Optimization (DO or DrO) for $(b)$. Let me know how you think the short and terminology should be presented better. Indeed, IEO is also included in E2E. Thanks for the remind! Both DO and IEO should be referred to as E2E for the exact definition.}
% This concept applies to both the non-contextual case
% that we focus on here, and also the contextual optimization case which we
% will study in As a simple example, we can obtain
% sample average approximation (SAA) through the following specification: 
% We defer the standard formulation and discussion of end-to-end learning in the contextual optimization problem to
% corresponds to setting $\paranx = \theta$ and estimating $\hat\theta$
% through E2E in \Cref{ex:ddo}$(b)(1)$.
%if we set $\paranx = \theta$ and estimate $\hat\theta$ through the empirical optimization, 
% this procedure recovers :
% \begin{example}[Sample Average Approximation (SAA)]\label{ex:saa}
%     If we set $\paranx = \theta$ and estimate $\hat\theta$ through E2E in \Cref{ex:ddo}$(b)(1)$, then this setting reduces to SAA.
%   \end{example}
{Next, we use portfolio % optimization problem.
selection to show how to % model
apply the various models described in \Cref{ex:ddo}. 
% We also illustrate the above definitions more generally with a 
} % using \Cref{ex:ddo}.
\begin{example}[Portfolio Optimization]\label{ex:portfolio}
% We present different data-driven optimization procedures based on 
% Consider
  Suppose the cost function $h(x;\xi) = \exp(-\xi^{\top}x) + \gamma
  \|x\|_2^2$ where the random asset return $\xi \in \R^{D_{\xi}}$ and portfolio
  weights $x \in \R^{D_{\xi}}$. In each case below, the dimension of
  $\theta$ may be different % dimension 
  depending on the 
  decision rule $\paranx$. 
%(CAN YOU MAKE BELOW MORE ORGANIZED IN CORRESPONDING TO DEFINITION 1
%ABOVE? FOR EXAMPLE, IT WOULD BE GOOD TO EXPLICITLY INDICATE POINTS 2 AND
%3 ARE EXAMPLES OF PART (B) IN DEFINITION 1. ALSO, WOULD IT MAKE SENSE TO
%LABEL THE BULLET IN THE SAME WAY AS DEFINITION 1 SO THAT READERS KNOW
%WHAT CORRESPONDS TO WHAT EASILY) 
\begin{enumerate}[(a),leftmargin=*]
\item Assume $\xi$ is a Gaussian distribution with independent components,
  i.e., 
% The asset returns % We parameterize 
%   $\xi$ % via a
%   are
%   Gaussian distribution with independent components,
  $\Pscr_{\Theta} = \{\prod_{j \in [D_{\xi}]} N(\mu_j, \sigma_j^2): \theta =
  (\mu, \sigma^2)\}$ and  $\paranx \in \argmin_{x}\{\E_{\P_{\theta}}[\exp(-\xi^{\top}x)] + \gamma\|x\|_2^2\}$.
\begin{enumerate}[(1), leftmargin=*]
    \item % An example of 
      ETO approach: $\hat{\theta} := (\hat{\mu},
      \hat{\sigma}^2)$, % by
      the sample mean and variance of $\{\xi_i\}_{i = 1}^n$, and output $x^*(\hat{\theta})$.
    \item IEO approach: $\hat{\theta} =
      \argmin_{\theta}\{\E_{\hat{\P}_n}[\exp(-\xi^{\top}\paranx)] +
      \gamma \|\paranx\|_2^2\}$, and output $x^*(\hat{\theta})$.
\end{enumerate}
% In $(a)(1)$, In $(a)(2)$, 
%We may consider other parameterizations $\{\P_{\theta}: \theta \in \Theta\}$ with other parametric distributions given some prior knowledge. 
%\vspace{-2mm}
%\noindent Following \Cref{ex:ddo}$(b)$, we may also consider other parametrizations $\paranx$ through E2E. And the associated R-E2E and DR-E2E can be derived similarly.
\item Equal-weighted portfolio, i.e., $\paranx = \theta 1_{D_{\xi}}$, and
  unconstrained-weighted portfolio, i.e., $\paranx = \theta$, are two
  % examples
  possibilities of $x^*(\theta)$ that follows the approach in \Cref{ex:ddo}$(b)$.
  \begin{enumerate}[(1), leftmargin=*]
  % f we apply 
  \item E2E approach: % then we would use
    $\hat{\theta} = \argmin_{\theta}\{\E_{\hat\P_n}[\exp(-\theta \sum_{j
      \in [D_{\xi}]}\xi_{\cdot,j})] + \gamma D_{\xi}^2 \theta^2\}$ for
    equal weights, and $\hat{\theta} =
    \argmin_{\theta}\{\E_{\hat{\P}_n}[\exp(-\xi^{\top}\theta)] + \gamma
    \|\theta\|_2^2\}$ for unconstrained weights (i.e., SAA).
    %where the latter one is
    % SAA in \Cref{ex:saa}.
    \item[(2,3)]We can also apply R-E2E or DR-E2E % analogously as (b)(1)
      % right above
      in an analogous manner
      % similarly
  to calibrate $\hat\theta$ and the decision $x^*(\hat{\theta})$. 
  \end{enumerate}
%(YOUR PREVIOUS WRITING IS UNCLEAR BECAUSE YOU MIX THE CONCEPT OF ARBITRARILY PARAMERIZED FAMILY $x^*(\theta)$ WITH E2E AND, ADDING TO THE CONFUSION, YOU NEVER MENTION E2E. I CHANGED, HOPE IT'S BETTER)
\end{enumerate}
\end{example}
%This can also be generalized to the regression case, where each decision $x$ in $h(x;\xi)$ represents a measurable function. The raw problem of finding the best measurable function to fit $\P^*$ is intractable since $\mathcal{X}$ is infinite dimensional so it is necessary to parameterize $x$ to be $x^*(\theta)$ as a function parameterized by $\theta$. For example, $x^*(\theta) = \theta(\xi)$ where $\theta$ denotes the parameters of a linear model or a neural network. }
% \wty{In terms of $\paranx$, 
% %even fixing the parameter within a space of distributions, the mapping from parameter to decisions can be different. 
% The classical approach is the two-stage approach where one uses data to estimate 
% % is first estimate
% the % model
% parameter $\hat{\theta}$, % i.e. from the unknown distribution,
% and then obtains $\datax$ by optimizing under the estimated parametric distribution, i.e. $\E_{\P_{\hat{\theta}}}[h(x;\xi)]$~\citep{bertsimas2020predictive,hu2022fast}.  % from the observed data
% % On one hand, to encode the parametric uncertainty from limited data
% % directly into the
% A more recent approach combines parameter estimation and decision
% making in an end-to-end
% manner~\citep{donti2017task,agrawal2019differentiable}. 
% % indicating that downstream aware approaches
% Such approaches also go by 
% operational statistics~\citep{liyanage2005practical} and
% smart-predict-then-optimize~\citep{elmachtoub2022smart}.  }
% impose the following smoothness assumptions on

\noindent Next, we describe smoothness and regularity  assumptions needed to establish
general results.
\wty{The first assumption is on the smoothness of the composite
function $h(\paranx;\xi)$.  Note that the conditions in this assumption are satisfied by
all % four
the % decision procedures when applied to the 
portfolio selection procedures 
% setting
discussed in \Cref{ex:portfolio}. } 
% We make the following assumptions on the composite function $h$ and decision
% rule $x^*(\theta)$:
% \begin{assumption}[Smoothness of Decision]\label{asp:represent}
% The decision rule $x^*(\theta)$ is twice differentiable, i.e, $\nabla_{\theta} \paranx$ and $\hessianp \paranx$ exist for any $\theta \in \Theta$.
% \end{assumption}
% \begin{assumption}[Smoothness of Cost Function]\label{asp:func}
%   % For any $\xi \in \Xi$ almost surely, 
%  The composite function $h(\paranx;\xi)$ is measurable
%   % function that is
%   and 
%   twice continuously differentiable with respect to
%   $\theta$ for any $\theta \in \Theta$, $\P^*$-a.s.
% %(IN WHAT SPACE, LOCALLY AROUND $x^*$ OR IN THE ENTIRE $\Theta$?) %Moreover, the following exchangeability between expectation and derivative holds: $\nabla_{\theta}\E_{\P^*}[h(\paranx;\xi)] = \E_{\P^*}\Paran{\nabla_{\theta} h(\paranx;\xi)}$.
% \end{assumption}
% \hl{I have removed Assumption 1. I think we can subsume it in
%   Assumption~2. See below.}
% a wide class of smooth functions in operational
% practice. In~\Cref{subsec:nonsmooth}, we relax~\Cref{asp:func} to
% incorporate some nonsmooth cases. \hat{x} is uniquely represented by
% $\theta$. We abbreviate it as  
% We generalize
% \wty{\Cref{asp:func} is not necessary for our results. It can be relaxed 
% % \Cref{asp:func} 
% to % incorporate
% allow for 
% some non-smooth objective functions. 
\wty{\begin{assumption}[Smoothness of Cost Function]\label{asp:nonsmooth}
    The composite function $h(\paranx;\xi)$ is of the form $h(\paranx;\xi) =
    f(g(\paranx;\xi))$, where
    \begin{enumerate}[(i)]
    \item $g(x^*(\theta);\xi)$ is measurable
  % function that is
      and 
      twice continuously differentiable with respect to
      $\theta$ for any $\theta \in \Theta$, $\P^*$-a.s.
      % satisfies \Cref{asp:func} %  with $h$ replaced by
      % $g$;
    \item 
    $f(z)$ is locally Lipschitz % (i.e., $\forall M > 0, |f(x) - f(y)|
    % \leq L_M|x - y|, \forall x, y \in [-M, M]$)
    %(DON'T USE $A$ BECAUSE YOU'VE DEFINED IT ALREADY?) 
    and twice differentiable % almost everywhere
    a.e. 
    with finitely many
    non-differentiable points.
  \end{enumerate}
\end{assumption}}
\noindent
% The condition in \Cref{thm:nonsmooth} extends the ones in
% Theorems \ref{thm:main} and \ref{coro:main-moment} to some % important
% objectives that are important for operations research applications,
% including the newsvendor cost where % we 
% % set
% $f(z) = c z^{+} + (p - c)z^{-}$ and $g(x;\xi) = x - \xi$. 
\wty{% Assumptions~\ref{asp:func} and~
  \Cref{asp:nonsmooth} requires that  % smoothness of 
  the composite function
$h(\paranx;\xi)$ is sufficiently smooth with respect to $\theta$. % , which generally depends on
% both the smoothness of $h(\cdot;\xi)$ and $x^*(\cdot)$.
% Although we typically consider the setting where
Typically, 
the function $f$ in 
\Cref{asp:nonsmooth} is the identity function, % we allow
but we allow for a general $f$ to accommodate 
% generalizes smooth objectives such as exponential
% utility optimization in \Cref{ex:portfolio} to
% extended the set of allowed functions to 
% include
% for
some non-smooth objectives % (e.g., piecewise linear objectives)
that are important in operations research and machine learning
applications. % These include 
These include newsvendor loss, conditional
value-at-risk loss, 
% portfolio optimization, 
quantile regression using SAA under
\Cref{ex:ddo}$(b)$, and parametrized decisions with $\P_{\theta}$ belonging
to exponential family distribution classes under \Cref{ex:ddo}$(a)$. In Appendix~\ref{app:discuss}.
% will
% provide more transparent
we discuss sufficient conditions for \Cref{asp:nonsmooth} that are easier to verify. 
}

\wty{The parameter estimation procedure in each decision rule
  of~\Cref{ex:ddo} implicitly maps the empirical distribution $\hat{\P}_n$ to
$\Theta$. We formally define the map by extending it to all $\P \in
M(\Xi)$, the set of probability distributions on $\Xi$.
\begin{definition}[Calibration function $T$]
  The calibration function $T$ corresponding to a procedure in \Cref{ex:ddo} is
  the map $T: M(\Xi) \mapsto \Theta$ such that the estimate $\theta = T(\P)$ % is the
  % estimate 
  when % the distribution over $\Xi$ is given by
  the uncertain parameter 
  $\xi \sim \P$. 
  % , we define $T(\P)$ as the set of
  %   the exact optimal solution of each optimization procedure under $\P$,
  %   i.e., a map $T$ from $M(\Xi)$, a set of probability distributions on
  %   $\Xi$, to $\Theta$, the set of 
  %     admissible parameters, evaluated at the distribution $\P$ (If
  %     $\hat\theta$ is solved to optimality, then $\hat\theta \in
  %     T(\hat\P_n))$. 
      Concretely, for ETO with MLE: $T(\P) := 
      \argmax_{\theta \in \Theta}\E_{\P}[\ln 
      p_{\theta}(\xi)]$; IEO / E2E: $T(\P) := \argmin_{\theta \in
        \Theta}\E_{\P}[h(\paranx;\xi)]$; R-E2E: $T(\P) := \argmin_{\theta \in
        \Theta}\{\E_{\P}[h(\paranx;\xi)] + \lambda R(\paranx)\}$; DR-E2E: $T(\P) := \argmin_{\theta \in
        \Theta} \max_{d(\Q,\P) \leq \epsilon} \E_{\Q}[h(\paranx;\xi)]$. Here, we assume the argmin operator of each procedure is unique. 
\end{definition}}
% \gi{The calibration function $T$ is introduced in  \Cref{ex:ddo}(c) order to define the
% influence function in~\Cref{def:if}.}
\wty{Next, we define the regularity conditions required for the calibration
  function~$T$ at the parameter $\theta^* = T(\P^*)$ corresponding to the
  true (unknown) distribution $\P^*$.} % is
  % the true (unknown) distribution of the random variable $\xi$. 
\begin{assumption}[Regularity Condition of $T(\P^*)$]\label{asp:theta-star}
  % In each procedure of \Cref{ex:ddo}, we have:
  % Let $\theta^* =
  % T(\P^*)$ denote the map of the unknown true distribution $\P^*$. 
  % The true parameter 
  % $\theta^*$ % is the solution of appropriate first order conditions.
  % satisfies the following conditions as a function of the decision
  % rule.
  \wty{The parameter value  $\theta^* = T(\P^*)$ % satisfies the following regularity
    corresponding to the true (unknown)
    distribution $\P^*$ satisfies the following
  conditions.}
  \wty{\begin{enumerate}[(i),leftmargin=*]
  \item ETO: $\theta^*$ is a root of the  equation
    $\nabla_{\theta}\E_{\P^*}[\phi(\theta;\xi)] = 0$, and $x^*(\theta^*)$ % satisfies the first order
    % conditions
    % {$\nabla_x
    % \Para{\E_{\P_{\theta^*}}[h(x;\xi)] + \sum_{j \in
    %     B_{\alpha(\theta^*)}}\alpha_j(\theta^*) F_j(x)} = 0$, where
    % $\alpha(\theta^*) \leq 0$ is the Lagrangian multiplier, and
    % $B_{\alpha(\theta^*)} = \{j \in J: F_j(x^*(\theta^*)) = 0\}$ denotes
    % the set of active constraints}.
    is a Karush-Kuhn-Tucker~(KKT) point for the optimization problem:
      \[\min_x \paran{\E_{\P_{\theta^*}}[h(x;\xi)], ~\text{s.t.}~F_j(x) \leq 0, j \in J_1; F_j(x) = 0, j \in J_2}.\]
      In particular, when $\Xscr = \R^{D_x}$, $x^*(\theta^*)$ is a root of the equation $\nabla_x \E_{\P_{\theta^*}}[h(x;\xi)] = 0$.
    \item IEO / E2E / R-E2E / DR-E2E: $\theta^*$
    % uniquely solving the equation 
    is a % root of the equation $\gradp
    % \Para{\E_{\P^*}[h(x^*(\theta);\xi)] +  \sum_{j \in B^*}\alpha^*
    %   F_j(x)} = 0$, where $\alpha^* \leq 0$ is the Lagrangian multiplier, and $B^* 
    % = \{j \in J: F_j(x^*(\theta^*)) = 0\}$ denotes the set of active
    % constraints.
    KKT point for the optimization problem in $T(\P^*)$. For example, for IEO / E2E, $\theta^*$ is a KKT point for the optimization problem:
    \[\min_{\theta \in \Theta}\paran{\E_{\P^*}[h(x^*(\theta);\xi)],~\text{s.t.}~F_j(x^*(\theta)) \leq 0, j \in J_1; F_j(x^*(\theta)) = 0, j \in J_2}.\]
    In particular, when $\Xscr = \R^{D_x}$, $\theta^*$ is a root of the equation $\nabla_{\theta} \E_{\P^*}[h(x^*(\theta);\xi)] = 0$.
\end{enumerate}}
\end{assumption}
 % \[\min_\theta \paran{\E_{\P^*}[h(x^*(\theta);\xi) + \lambda R(x^*(\theta))], ~\text{s.t.}~F_j(x^*(\theta)) \leq 0, j \in J_1; F_j(x^*(\theta)) = 0, j \in J_2.}
 %    \]
  % $\theta^*$ is a root of the equation {$\gradp
  % \E_{\P^*}[h(x^*(\theta);\xi)] + \lambda \gradp R(x^*(\theta)) = 0$}. 
\wty{\Cref{asp:theta-star} characterizes  $\theta^*$ in terms of the first-order optimality condition from the calibration function~$T$. When the optimization problem in \Cref{asp:theta-star} admits a unique solution (i.e., a unique KKT point), a convenient sufficient condition is that the objective for each procedure (i.e., $\E_{\P_{\theta^*}}[h(x;\xi)]$ with respect to $x$ in ETO and $\E_{\P^*}[h(x^*(\theta);\xi)]$ with respect to $\theta$ in the other cases) is strictly convex on a convex decision set $\Xscr$~\citep{boyd2004convex}. However, global convexity of the objective, or even the uniqueness of the optimal solution, is not required for Assumption~\ref{asp:theta-star} to hold. In settings with non-convex objectives or non-uniqueness of optimal solution, \Cref{asp:theta-star} can be satisfied with a specific local optimum $\theta^*$ to which our estimate $\hat\theta$ naturally converges as sample size $n$ increases. We will discuss this in detail later after \Cref{asp:theta-asymptotics} in this section.}
\begin{assumption}[Moment Condition]\label{asp:moment}
Each entry of the gradient
$\nabla_{\theta}\E_{\P^*}[h(\bestx;\xi)]$ and Hessian
$I_h(\theta^*) :=\nabla_{\theta\theta}^2\E_{\P^*}[h(\bestx;\xi)]$ is finite, and % $I_h(\theta) :=
% \nabla_{\theta\theta}^2\E_{\P^*}[h(\bestx;\xi)]$
$I_h(\theta^*)$ 
  is \wty{invertible}.  
\end{assumption}
% \begin{assumption}[Moment Condition]\label{asp:moment}
% %$\E_{\P^*}[\nabla_{\theta} h(\bestx;\xi) \nabla_{\theta}
% %h(\bestx;\xi)^{\top}] < \infty$
% % The calibration function $T$ maps the unknown distribution~$\P^*$ for
% %  $\xi$ to a \wty{unique} value   
% % $\theta^* := T(\P^*) \in \Theta$. 
% % is a \wty{unique} value, where $T$
% % denotes the calibration function.
% Each entry of the gradient
% $\nabla_{\theta}\E_{\P^*}[h(\bestx;\xi)]$ and Hessian
% $I_h(\theta^*) :=\nabla_{\theta\theta}^2\E_{\P^*}[h(\bestx;\xi)]$ evaluated at $\theta^*$
% is finite and % $I_h(\theta) :=
% % \nabla_{\theta\theta}^2\E_{\P^*}[h(\bestx;\xi)]$
% $I_h(\theta^*)$
%   is \wty{invertible}. 
% %\E_{\P^*}[\hessianp h(\bestx;\xi) \hessianp h(\bestx;\xi)] < \infty.
% \end{assumption}
%(PUT EXAMPLES HERE, WRITTEN IN A GENERAL WAY)
%(ARE THERE MORE SUFFICIENT CONDITIONS TO CHECK ASSUMPTIONS 1 AND 2? THAT
%IS, IF I'M A USER WITH AN $h$ AND $\mathbb P_\theta$ OR $x^*(\theta)$,
%HOW DO I KNOW IF ASSUMPTIONS 1 AND 2 ARE SATISFIED. WE MAY PUT THESE
%SUFFICIENT CONDITIONS IN APPENDIX AND INDICATE HERE)  
%Assumption \ref{asp:represent} is on the smoothness of $x^*(\cdot)$ only. 
\wty{Assumption \ref{asp:moment} imposes moment conditions on the
  composite function $\E_{\P^*}[h(\paranx;\xi)]$.} % under a unique limiting
  % point of the optimization output
  % at the unique $\theta^* = T(\P^*)$.
% $\theta^*$.
\wty{Note that % our results do not apply to linear optimization problems,
  % as they lack the invertibility property of
  the Hessian matrix is not invertible for linear optimization problems,
  and therefore, our general results do not directly apply.  That being
  said, we show in \Cref{prop:oic-entropic-lp}  in \Cref{subsec:application-if} % we can apply
  that 
  our results apply provided  the linear objective is appropriately
  smoothed using an  entropic regularization. % such that the
% Hessian is invertible and discuss the corresponding results in
% in.
% Moreover, while Assumption \ref{asp:func}
% assumes smoothness on $h(x^*(\cdot);\xi)$, later in this section we % will
% % show a
% introduce a
% generalization to allow for non-smooth costs.
% Note that for each data-driven model, the map $\theta \mapsto \paranx$
% is different but deterministic. It can even be the identity mapping when
% $\Xscr$ is of finite dimension.
%Before introducing assumptions on the calibration of $\hat\theta$, 
% OIC removes bias for cost functions that satisfy
% Assumptions~\ref{asp:func} (or~\ref{asp:nonsmooth})
% and~\ref{asp:moment}, % which excludes the linear objective case.
% and therefore, it does not apply to linear objective functions.
Existing approaches that address 
unbiased
evaluation for linear
objectives~\citep{ito2018unbiased,gupta2021small,gupta2022debiasing,gupta2024decision}
% has been investigated in existing 
% literature~,
% with approaches that differ significantly from ours due to properties of
% linear optimization.
use significantly different techniques % due to
that leverage
specific properties of
linear or weakly coupled objectives. We compare these methods with OIC in
Appendix~\ref{subsec:compare-linear}.  % We review their methods and 
% discuss
% the differences with ours 
} 
%IN CONSTRAINED CASE, WE DESIGN ENTROPIC REGULARIZATION TO DEMONSTRATE IT MAY WORK UNDER SOME TILED CASE; ALSO ADD SOME DISCUSSION FOR THE PORTFOLIO CASE MENTIONED.}

Next, we define % the notion of 
influence functions that quantify the impact of a single data point on the
calibration function $T$, \wty{which allow one to express
  the evaluation bias of the different procedures in \Cref{ex:ddo} in a
  unified manner.}
\begin{definition}[Influence Function]\label{def:if}
%and $T(\P^*) = \theta^*$. 
%be the asymptotic value of some estimator sequence $\{T_n\}_{n \in \mathbb{N}}$. 
% We define
%Let $\theta^*= T(\P^*)$. Then 
The~\emph{influence function}
$\IFx(\cdot)$ is defined as follows: 
\[
  \IFx(\xi; T, \P^*) = \lim_{\epsilon \to
    0^{+}}\frac{T(\epsilon\delta_{\xi} + (1-\epsilon)\P^*) -
    T(\P^*)}{\epsilon}.
\]
\end{definition}
\wty{Influence functions were introduced originally in the robust
  statistics literature~\citep{hampel1974influence,cook1982residuals}. % and can be interpreted as 
% % the functional derivative of an estimator with respect to the underlying
% % distribution $\P^*$. Intuitively, it
% the impact of a point $\xi$
% on the calibration map $T(\cdot)$.
Recall that the calibration function $T$ is different
for each cost function and the optimization
% model selection
procedure. % which implies that
Therefore, 
the corresponding $\IFx(\cdot)$ is also different in each case. We abbreviate
$\IFx(\xi;T, \P^*)$ as $\IFx(\xi)$ when there is no confusion.  
We
provide % formulas
expressions 
for the influence functions % across
for 
various optimization
procedures:}

\begin{definition}[Influence Functions of Procedures in \Cref{ex:ddo}]\label{coro:oic-procedure}
~~~~~~~~
\begin{enumerate}[(i),leftmargin = *]
    \item ETO:  $\IFx(\xi) := -(\E_{\P^*}[\nabla_{\theta\theta}^2\phi(\theta^*;\xi)])^{-1}\nabla_{\theta}\phi(\theta^*;\xi)$.
    \item IEO / E2E / DR-E2E: ${\IFx}(\xi) := - P
      (I_{h,\alpha}(\theta^*))^{\dagger} P \nabla_{\theta} h(\bestx;\xi)$,
      where ${I}_{h,\alpha}({\theta}^*) = 
 \nabla_{\theta\theta}^2\E_{\P^*}[h(\bestx;\xi)] +
\sum_{j \in {B^*}} \alpha_j^* \nabla_{\theta\theta}^2 F_j(\bestx)$,
$\alpha$ is the Lagrangian multiplier corresponding to the binding
constraint $B^* = \{j \in J: F_j(\bestx) = 0\}$, and $P = I - 
{C}^{\top}({C}{C}^{\top})^{\dagger}{C}$, with $C = [\nabla_{\theta}
F_j(x^*(\theta^*))]_{j \in B^*}$ and $^{\dagger}$ denotes the
pseudo-inverse. Specifically, when $\Xscr = \R^{D_x}$, $\IFx(\xi) :=
-I_h(\theta^*)^{-1}\nabla_{\theta}h(\bestx;\xi)$. 
    \item R-E2E: ${\IFx}(\xi)$ is the same as in $(ii)$, with  
      $h(x^*(\cdot);\xi)$ replaced by $h(x^*(\cdot);\xi) + \lambda R(x^*(\cdot))$.
\end{enumerate}
\end{definition}
\noindent
% \wty{In \Cref{sec:usercase}, we apply them % into
% to
% newsvendor and
% portfolio optimization problems.}
%which gives rise to the following assumption:
% \begin{new}
% Borrowing the decomposition $\hat\theta - \theta^*$ using influence
% functions, we express the general performance characterization of any
% data-driven optimization in \Cref{ex:ddo} as follows: 
\wty{Recall that we had noted in Definition \ref{ex:ddo} that we do not 
  need the estimate $\hat{\theta}$ to be an optimal solution of the optimization problem. The following assumption specifies the degree of suboptimality that can be allowed without impacting our general results.
\begin{assumption}[Computing $x^*(\hat\theta)$]\label{asp:theta-hat}
   The estimate $\hat{\theta}$ and
    the output solution $\datax$ for each procedure in \Cref{ex:ddo} satisfies the following conditions:
    \begin{enumerate}[(i),leftmargin=*]
        \item ETO: $\E_{\hat{\P}_n}[\nabla_{\theta}\phi(\hat\theta;\xi)] =
          o_p(n^{-1})$ and $\E_{\P_{\hat\theta}}[h(\datax;\xi)] = \min_{x \in \Xscr}\E_{\P_{\hat\theta}}[h(x;\xi)] + o_p(n^{-1})$;
          % $\left.\nabla_{x}\big(\E_{\P_{\hat\theta}}[h(x;\xi)] + \sum_{j \in B_{\alpha(\hat\theta)}}\alpha_j(\hat\theta) F_j(x)\big)\right|_{{x =
          %     x^*(\hat\theta)}} = o_p(n^{-1})$.
        \item IEO / E2E: $\E_{\hat\P_n}[h(\datax;\xi)] = \min_{\theta \in \Theta} \E_{\hat\P_n}[h(\paranx;\xi)] + o_p(n^{-1})$;
    % $\wty{\nabla_{\theta} \E_{\hat\P_n}[h(\datax;\xi)]} + \sum_{j \in \hat{B}}
    % \hat{\alpha}_j \nabla_{\theta} F_j(\datax) = o_p(n^{-1})$; 
        \item R-E2E: $\E_{\hat\P_n}[h(\datax;\xi) + \lambda R(\datax)] = \min_{\theta \in \Theta} \E_{\hat\P_n}[h(\paranx;\xi) + \lambda R(\paranx)] + o_p(n^{-1})$;
    %     $\gradp \E_{\hat\P_n}[h(\datax;\xi)] + \lambda \gradp R(x^*(\hat\theta)) + \sum_{j \in \hat{B}}
    % \hat{\alpha}_j \nabla_{\theta} F_j(\datax)= o_p(n^{-1})$;
        \item DR-E2E: $\sup_{d(\P, \hat\P_n)\leq \epsilon}\E_{\P}[h(\datax;\xi)] = \min_{\theta \in \Theta}\sup_{d(\P, \hat\P_n)\leq \epsilon} \E_{\P}[h(\paranx;\xi)] + o_p(n^{-1})$.
        %$\gradp \sup_{d(\P, \hat\P_n)\leq \epsilon} \E_{\hat\P_n}[h(\datax;\xi)] + \sum_{j \in \hat{B}} \hat{\alpha}_j \nabla_{\theta} F_j(\datax) = o_p(n^{-1})$.
    \end{enumerate}
% In each case above, $\alpha_j(\hat\theta)$ (or $\hat\alpha_j$) denotes the KKT multipliers associated with the $j$-th constraint of the optimization problem whose solution defines the exact minimizer $T(\hat\P_n)$.
%(i.e., RHS = 0 for the equations above).
% \hl{I still have an issue with the last comment. It appears that you need
%   the KKT multipliers at the optimal solution -- but the whole point here
%   is to assert that one does NOT have to compute the optimal
%   solution. If one is only computing the $o(n^{-1})$-optimal solution, how
%   does one get the KKT multipliers at the optimal solution? One should be
%   able to make do with the KKT multipliers at the suboptimal solution.}
\end{assumption}
\Cref{asp:theta-hat} requires \tyw{each procedure to output a solution
  $x^*(\hat\theta)$ whose objective value is within $o_p(n^{-1})$ of the optimum; in the ETO case, it further requires
  $\hat\theta$ to satisfy the first-order stationarity condition of the
  estimation criterion within an $o_p(n^{-1})$ error.} In
Appendix~\ref{subsec:comp}, we discuss the applicability of OIC that
accounts for the computational limitations of stochastic approximation
algorithms when solving the target decision problems, which typically
require a polynomial number of iterations with respect to $n$. In
Appendix~\ref{subsec:comp}, we also present an alternative assumption
(i.e., \Cref{asp:theta-hat2}) that relates to \Cref{asp:theta-hat}, which
characterizes the degree of suboptimality via the first-order optimality
error when the decision rule is
unconstrained. Assumption~\ref{asp:theta-hat2} is easier to verify for the
unconstrained nonconvex objectives and can serve as a replacement for
Assumption~\ref{asp:theta-hat} such that the subsequent statistical
results hold. } 
% In order to 
% % highlight
% focus on the main ideas,
 % (ARE THESE SUFFICIENT CONDITIONS FOR ASSUMPTION EC.6 OR ASSUMPTION 4 OR PLAY OTHER ROLES? CAN'T TELL WHAT THESE CONDITIONS ARE FOR) }

To construct OIC, we first present the following result.
\begin{theorem}[Bias Characterization]\label{thm:main}
For each procedure in \Cref{ex:ddo}, suppose
Assumptions~\ref{asp:nonsmooth},~\ref{asp:theta-star},~\ref{asp:moment},~\ref{asp:theta-hat}
and the technical conditions in~\Cref{asp:additional-regular} hold. Then the true expected performance $A = \E_{\Dscr_n}\E_{\P^*}[h(\datax;\xi)]$ of the decision $\datax$ % given by, 
can be expressed as:
\begin{equation}\label{eq:bias}
  % \E_{\Dscr_n}\E_{\P^*}[h(\datax;\xi)]
  A
= \E_{\Dscr_n}\Big[\underbrace{\frac{1}{n}\sum_{i = 1}^n
  h(\datax;\xi_i)}_{=: \hat{A}_o}\Big]\underbrace{-\frac{\E_{\P^*}[\gradp h(\bestx;\xi)^{\top}
    \IFx(\xi)]}{n}}_{=: A_c} + o\Para{\frac{1}{n}}. 
\end{equation}
\end{theorem}

In \eqref{eq:bias}, the first term % in the RHS 
is the expected value of the empirical estimator $\hat{A}_o =
\frac{1}{n}\sum_{i = 1}^n h(\datax;\xi_i)$. Thus, the bias of $\hat{A}_o$ against $A$ up to
$o(n^{-1})$ is given by $A_c := - n^{-1} \E_{\P^*}[\gradp h(\bestx;\xi)^{\top}
\IFx(\xi)]$. {Note that in the context of ETO / IEO, the bias formula holds
  % regardless of whether the
  whether or not the 
  distribution model is well-specified, i.e., regardless of whether $\P^*
  \in \Pscr_{\Theta}$.} 
\wty{For unconstrained E2E, % case% and
$A_c = n^{-1} \Tr{I_h(\theta^*)^{-1} J_h(\theta^*)}$ with $J_h(\theta^*) =
\E_{\P^*}[\nabla_{\theta} h(\bestx;\xi) 
  \nabla_{\theta} h(\bestx;\xi)^{\top}]$. When $h(x^*(\theta);\xi)$ is
  convex in $\theta$,  $I_h(\theta^*)$ is positive semidefinite, and
  therefore $A_c > 0$, i.e., the bias in $\hat A_o$ is optimistic (see discussion in Section~\ref{sec:intro}). In
general, however, the bias $A_c$ can take any sign and our approach applies regardless of its sign.} The bias expression in the presence of constraints
(see~\eqref{eq:constrain-prob}) is given by $n^{-1}\text{Tr}\Paran{({P} 
  {I}_{h,{\alpha}}({\theta}^*){P}))^{\dagger} 
  {J_h}(\theta^*)}$. Since ${I}_h({\theta^*})  \preceq
{I}_{h,{\alpha}}({\theta}^*)$, it follows that 
% we see that including constraints can % eliminate more
% lead to lower
% bias as compared to the unconstrained case, i.e., 
$\text{Tr}\Paran{({P}
  {I}_{h,{\alpha}}({\theta}^*){P}))^{\dagger} 
  {J_h}(\theta^*)} \leq \text{Tr}\Paran{{I}_h({\theta^*})^{-1}
  {J}_h({\theta^*})}$, % since ${I}_h({\theta^*})  \preceq
% {I}_{h,{\alpha}}({\theta}^*)$.
i.e., presence of constraints lowers bias~\citep{jagannathan2003risk}, where $I_{h,\alpha}$ is defined in \Cref{coro:oic-procedure}(ii). \wty{Besides, Theorem 1 requires some
% we have summarized some other 
standard technical conditions, i.e., \Cref{asp:additional-regular} in Appendix~\ref{app:apply}. We do not mention it explicitly in the main body since \Cref{asp:additional-regular} is natural and commonly satisfied by general data-driven optimization procedures~\citep{gotoh2021calibration,lam2021impossibility,elmachtoub2023estimatethenoptimize} and constrained stochastic
optimization~\citep{duchi2021asymptotic,kallus2023stochastic}.} 
%implies that $\Tr{\hat{P} (\hat{I}_{h,\hat{\alpha}}(\hat{\theta}))^{\dagger} \hat{P}} \leq \Tr{\hat{I}_h(\hat\theta)}$. 
% In the
% which is positive from 
% \wty{and $A_c > 0$ if $I_h(\theta)$ in \Cref{asp:moment} is further
% positive definite}, i.e., the bias % and corresponds to the  
% in $\hat{A}_o$ is \emph{optimistic} (see discussion in
% \Cref{sec:intro}).  
% %\hl{Shouldn't $A_c
% %   < 0$ for an optimistic bias? Am I missing something here?}
% In general, however, the bias $A_c$ % s not necessarily positive;
% can take any sign and 
% % nonetheless, 
% our approach applies regardless of its % direction
% sign.  
%(IS THIS ALWAYS AN OPTIMISTIC BIAS? THIS WOULD BE IN READER'S MIND WHEN
%THEY FIRST SEE OUR FORMULA).  
%And the empirical estimator $\hat A_o = \frac{1}{n}\sum_{i = 1}^n
%h(\datax;\xi_i)$ incurs a bias of the order $O(1/n)$. 

Motivated by \Cref{thm:main}, we define our bias-corrected
estimator  % we introduce
OIC $\hat{A}$ % defined in \eqref{eq:oic} as
for \eqref{eq:eval-key} \wty{by replacing the unknown $\theta^*$, $\IFx$ and $\P^*$ with empirical
or estimated
counterparts. }
% an estimator to estimate that bias term:
\begin{definition}[Optimizer's Information Criterion (OIC)]\label{def:oic} 
  % We define OIC as:
  The OIC
  \begin{equation}\label{eq:oic}
    \hat{A} := % \underbrace{\frac{1}{n}\sum_{i = 1}^n
    % h(x^*(\hat{\theta});\xi_i)}_{\hat{A}_o}\ \underbrace{-
    % \frac{1}{n^2}\sum_{i = 1}^n
    % \nabla_{\theta}h(x^*(\hat{\theta});\xi_i)^{\top}
    % \widehat{IF}(\xi_i).}_{\hat{A}_c}
    \underbrace{\E_{\hat\P_n}[h(x^*(\hat{\theta});\xi)]}_{\hat{A}_o}\
    \underbrace{- \frac{1}{n}
      \E_{\hat\P_n}[\nabla_{\theta}h(x^*(\hat{\theta});\xi)^{\top}
      \widehat{\IFx}(\xi)].}_{\hat{A}_c},
  \end{equation}
  \wty{where % \wty{In \Cref{def:oic}, a natural candidate to approximate the influence function is: 
% % \wty{
% %   \begin{definition}[Estimated Influence Function]\label{defn:estimate-if}
% %     The estimated influence function is given by:
% % Define
\begin{equation}
  \label{defn:estimate-if}
  \widehat{\IFx}(\xi) := \IFx(\xi;\hat{T}, \hat\P_n),
\end{equation}
% i.e., the unknown $\P^*$ in $IF(\xi)=IF(\xi;T, \P^*)$ is replaced with
and % and the exact minimizer $T(\cdot)$ is replaced by $\hat
% T(\cdot)$, i.e., an approximated mapping of $T(\cdot)$, which allows
% $\hat\theta$ to replace $\theta^*$.
$\hat{T}$ is the map that takes a distribution $\P$ to the
$o_p(n^{-1})$-optimal parameter $\hat{\theta}$.}
\end{definition}
\wty{% Following this, we obtain the
  % estimated influence function procedures for each procedure in
  In Definition \ref{def:oic}, the approximate influence function $\widehat{\IFx}$ is defined generally in
  terms of  the
  approximate map $\hat{T}$. This can be expressed more explicitly for different data-driven optimization procedures described in
\Cref{ex:ddo}.}
% can be expressed explicitly.} the approximate influence functions for the 
\wty{
  \begin{definition}[Estimated Influence Functions of Procedures in
    \Cref{ex:ddo}]\label{coro:oic-procedure-2} 
    ~~~~~
    % Suppose
    % Assumptions~\ref{asp:nonsmooth},~\ref{asp:moment},~\ref{asp:theta-star}
    % and~\ref{asp:theta-hat} hold. Then we have the following. 
    % The influence and estimated influence functions can be computed as follows:
    \begin{enumerate}[(i),leftmargin=*]
    \item ETO:  $\widehat{\IFx}(\xi) =
      -(\E_{\hat{\P}_n}[\nabla_{\theta\theta}^2\phi(\hat{\theta};\xi)])^{-1}
      \nabla_{\theta}\phi(\hat{\theta};\xi)$; 
    \item IEO / E2E / DR-E2E: $\widehat{\IFx}(\xi) = -\hat P I_{h,\hat\alpha}(\hat\theta)^{\dagger}
      \hat P \nabla_{\theta} h(\datax;\xi)$, where $\hat{P} = I -
      \hat{C}^{\top} (\hat{C} \hat{C}^{\top})^{\dagger}\hat{C}$, $\hat C =
      [\nabla_{\theta} F_j(\datax)]_{j \in \hat B}$ and $\hat{B} = \{j \in J:
      F_j(\datax) = 0\}$ denotes the set of active constraints at
      $\hat{\theta}$. Any
      approximation $I_{h,\alpha}(\hat\theta)$ is valid as long as
      $\E_{\Dscr_n}[\|\hat I_{h,\alpha}(\hat\theta) -
      I_{h,\alpha^*}(\theta^*)\|_2] = o(1)$ (e.g., kernel density estimate in
      \Cref{subsec:numerical-newsvendor}). When $h$ is smooth
      (i.e., $f(z) = z$ in \Cref{asp:nonsmooth}),
      $\hat{I}_{h,\alpha}(\hat{\theta}) = 
      \frac{1}{n}\sum_{i = 1}^n \nabla_{\theta\theta}^2 h(\datax;\xi_i) +
      \sum_{j \in \hat{B}} \hat{\alpha}_j \nabla^2 F_j(\datax)$, where
      $\hat{\alpha}$ denote the Lagrange multipliers. The expression reduces
      to $\frac{1}{n}\sum_{i = 1}^n\nabla_{\theta\theta}^2 
      h(\datax;\xi_i)$ when $\Xscr = \R^{D_x}$.
      % $\frac{1}{n}\sum_{i = 1}^n\nabla_{\theta\theta}^2h(\datax;\xi_i)$. 
    \item R-E2E: $\widehat{\IFx}(\xi)$ is the same as in $(ii)$, with  
      $h(x^*(\cdot);\xi)$ replaced by $h(x^*(\cdot);\xi) + \lambda R(x^*(\cdot))$.
    \end{enumerate}
  \end{definition}
  Then, the bias formula of each procedure in \Cref{ex:ddo} can be directly calculated by plugging each estimated influence function into~\eqref{eq:oic}. We now present the main properties of OIC as follows.
  }

\begin{theorem}[Statistical Properties of OIC]\label{coro:main-moment}
Suppose
Assumptions~\ref{asp:nonsmooth},~\ref{asp:theta-star},~\ref{asp:moment},~\ref{asp:theta-hat}
and the technical conditions in \Cref{asp:additional-regular} hold. 
%Suppose Assumptions~\ref{asp:nonsmooth},~\ref{asp:theta-star},
%~\ref{asp:theta-asymptotics}
%and~\ref{asp:empirical-if} hold. 
Then the following results hold.
\begin{enumerate}[(1),leftmargin=*]
\item The OIC estimator $\hat{A}$ satisfies $\E_{\Dscr_n}[\hat{A}] = A + o\Para{1/n}$.
\item The bias correction component $\hat A_c$ satisfies $\E[\hat{A}_c] =
  A_c + o\Para{1/n} = O\Para{1/n}$, for $A_c$ % is defined
  in~\eqref{eq:bias}. % $A_c =
  % -\E_{\P^*}[\nabla_{\theta} h(\bestx;\xi)^{\top} IF(\xi)]/n$; 
  % $\hat A_c \overset{p}{\to} A_c(:=-\frac{\E_{\P^*}[\nabla_{\theta} h(\bestx;\xi)^{\top} IF(\xi)]}{n})$;
\item Suppose $\E_{\P^*}[(\nabla_{\theta}
  h(\bestx;\xi)^{\top}\IFx(\xi))^2] < \infty$, for all
  $\theta \in \Theta$.  
  % (DO YOU NEED $\theta \in \Theta$?). 
  Then $\var[\hat{A}_c] = \Theta\Para{1/n^2}$.
\end{enumerate}
    %, \quad n \hat{A}_c \convp -\E_{\P^*}[\nabla_{\theta}
    %h(\bestx;\xi)^{\top}IF(\xi)].\]  
\end{theorem}
% \wty{
\Cref{coro:main-moment} states that $\hat A$ % unbiasedly
is an unbiased estimate for 
% estimates
$A$ up to % error
$o\Para{1/n}$, and $\hat{A}_c$ % being
is an unbiased estimator of the bias up to the same order of error. We will discuss the interpretation of the bias correction term in \eqref{eq:oic}, which comes from \eqref{eq:bias}, momentarily. In the remainder of this subsection, we briefly discuss the proofs of Theorems~\ref{thm:main} and~\ref{coro:main-moment}. The proof of \Cref{thm:main} relies on the second-order Taylor expansion
centered at $\theta^*$ of both the \emph{sample optimality gap}
$\frac{1}{n}\sum_{i = 1}^n \Para{h(\datax;\xi_i) - h(\bestx;\xi_i)}$ and
\emph{true optimality gap} $\E_{\P^*}[h(\bestx;\xi)] -
\E_{\P^*}[h(\datax;\xi)]$.  For nonsmooth objectives satisfying \Cref{asp:nonsmooth}$(ii)$, the key 
%to establishing Theorem \ref{thm:nonsmooth} 
is to % establish the
% fact
show
that as long as the \emph{expected} value of the cost function is twice
differentiable in $\theta$, the correction term in OIC can still be
computed. \wty{In both cases, the result relies on the following expansion of the difference between $\hat\theta - \theta^*$}, which stipulates that $\sqrt{n}(\hat{\theta} -
\theta^*)$ is asymptotically linear and normal.
\begin{lemma}[Statistical Properties of Parameter
  Estimator]\label{asp:theta-asymptotics}
  Suppose
Assumptions~\ref{asp:nonsmooth},~\ref{asp:theta-star},~\ref{asp:moment},~\ref{asp:theta-hat}
and the technical conditions in~\Cref{asp:additional-regular} hold. Then there exists $\theta^*$ such that: 
  % the error in 
  % parameter estimate $\hat\theta = T(\hat\P_n)$ is of the form
  \begin{equation}\label{eq:theta-if}
    \hat{\theta} - \theta^* = \frac{1}{n}\sum_{i = 1}^n \IFx(\xi_i)  + o_p\Para{\frac{1}{\sqrt{n}}}.
  \end{equation}
  % We % further assume. Furthermore,
  % assume that 
Furthermore, $\E_{\P^*}[\IFx(\xi)] = 0$ and the sequence $\sqrt{n}(\hat\theta - \theta^*)$ is asymptotically normal with mean zero and covariance matrix $\Psi(\theta^*) := \E_{\P^*}[\IFx(\xi)\IFx(\xi)^{\top}]$.  %Intuitively, the influence function captures the impact of one single point $\xi$ on the estimated value $\hat\theta$.
%Denote $\hat{\theta}$ to be the parameter estimated under $\Dscr_n$, 
%We have $\sqrt{n}(\hat{\theta} - \theta^*) \convd N(0, \Psi(\theta^*))$, where $\|\E_{\Dscr_n}\Paran{(\hat{\theta} - \theta^*) (\hat{\theta} - \theta^*)^{\top}} - \Para{\frac{\Psi(\theta^*)}{n}}\|_1 = o\Para{\frac{1}{n}}$ and $\|\cdot\|_1$ is the 1-norm of matrix induced by the vector norm. \wty{$\Psi(\theta^*) := \E_{\P^*}[IF(\xi; T, \P^*)IF(\xi; T, \P^*)^{\top}]$. TODO: Synscrnize the notations of $\Psi(\theta), \hat\theta - \theta^*, T(\cdot)$.}
\end{lemma}
% \Cref{eq:theta-if} and 
% This is satisfied by many
% standard estimators in classical statistics
% \citep{bickel1993efficient,van2000asymptotic}, and can be shown to hold
% for procedures in \Cref{ex:ddo}. 
\wty{We show \Cref{asp:theta-asymptotics} holds for each procedure by leveraging the $M$-estimator theory to justify the asymptotic properties of $\hat\theta$, which is valid when it is only an approximate solution as described in Assumption~\ref{asp:theta-hat}. Note that \eqref{eq:theta-if}
  % to hold.
  continues to hold even when the set of optimal solutions is not a singleton, 
  as long as % certain
  % regularity properties hold
 % ,
  % hold, we can identify some $\theta^* \in T(\P^*)$ such that
  $\hat\theta\convp \theta^*$ for some $\theta^*$ satisfying \Cref{asp:theta-star}. Theorem 5.14 in \cite{van2000asymptotic} implies that the presence of multiple optima does not pose a fundamental issue as long as $\hat\theta$ under a proper choice of selection rule (e.g., optimization algorithms with the same initialization across $n$), converges to one of the points. Under this setup, consistency still holds and one can apply Theorem 5.21 in \cite{van2000asymptotic} (e.g., in ETO, unconstrained IEO / E2E / R-E2E) to show the corresponding asymptotic normality. % n this way,
  % Thus,
  % \Cref{asp:theta-asymptotics} may % still
  % hold even when $T(\P^*)$ is not a
  % singleton. 
  We also note that  % which is also assumed in 
  the convergence $\hat\theta\convp \theta^*$ appears explicitly in Definition 1 and Assumption 3 in \cite{bertsimas2020predictive}, in settings where the set of optimal solutions is not unique. } 

\wty{We now provide a proof sketch on decomposing the performance difference. Consider the Taylor expansion of the sample optimality gap $\E_{\hat\P_n}[h(x^*(\theta^*);\xi)] - \E_{\hat\P_n}[h(\datax;\xi)]$ at $\theta^*$:
\begin{align*}
    \E_{\hat\P_n}[h(x^*(\hat\theta);\xi)] - \E_{\hat\P_n}[h(\bestx;\xi)] = \nabla_{\theta}\E_{\hat\P_n}[h(\bestx;\xi)]^{\top}(\hat\theta - \theta^*) + \Theta(\|\hat\theta - \theta^*\|_2^2),
\end{align*}
where we can show the remainder term $\Theta(\|\hat\theta - \theta^*\|_2^2)$ cancels out with the true optimality gap. For $\nabla_{\theta}\E_{\hat\P_n}[h(\bestx;\xi)]^{\top}(\hat\theta - \theta^*)$, if we take the expectation over $\Dscr_n$ further, we have, by Lemma \ref{asp:theta-asymptotics}:
\[\E_{\Dscr_n}[\nabla_{\theta}\E_{\hat\P_n}[h(\bestx;\xi)]^{\top}(\hat\theta - \theta^*)] \approx \E_{\Dscr_n}\Paran{\frac{1}{n^2}\sum_{i,j} \nabla_{\theta} h(\bestx;\xi_i)^{\top} \IFx(\xi_j)} = \frac{1}{n}\E_{\P^*}[\nabla_{\theta} h(\bestx;\xi)^{\top} \IFx(\xi)],\]
which gives rise to the bias correction terms in \eqref{eq:bias}. This gives Theorem \ref{thm:main}. In the literature, the Taylor expansion of the sample optimality gap is often at $\hat\theta$ and requires $\nabla_{\theta}\E_{\hat\P_n}[h(\datax;\xi)] = 0$, which only holds in the specialized case where we have an exact optimizer, for IEO / E2E, in an unconstrained problem. In contrast, our Taylor expansion centers at $\theta^*$ and uses an expansion of the unified influence function from \Cref{asp:theta-asymptotics} to obtain the formula of the evaluation bias.}

  \wty{To show \Cref{coro:main-moment}, 
% order to % derive guarantees on the use of
% establish guarantee on the 
% OIC, 
we % make a the mild 
verify
%Under suitable regularity conditions in $M$-estimation,
  the estimated influence function defined in \eqref{defn:estimate-if} % would result in consistency,
  is consistent:}
  % and % furthermore
  % in addition,
  % asymptotically % normality
  % normal
  % (Section 3.4 in \cite{cook1982residuals}; Section 5 in
  % \cite{van2000asymptotic}).
  % that satisfy \Cref{asp:empirical-if}.
\begin{lemma}[Influence Function Estimation]\label{asp:empirical-if}
%There exists an estimate $\widehat{IF}$ for the influence function such that  
Suppose
Assumptions~\ref{asp:nonsmooth},~\ref{asp:theta-star},~\ref{asp:moment},~\ref{asp:theta-hat}
and the technical conditions in~\Cref{asp:additional-regular} hold. Then for all $\xi \in \Xi$, we have:
%with empirical estimator $\hat{\theta}$ with: 
\begin{equation}\label{eq:empirical-if}
    \E_{\Dscr_n}[\|\widehat{\IFx}(\xi)\|_2^2] - \|\IFx(\xi)\|_2^2  = o(1).
    %(\widehat{IF}(\xi) - IF(\xi))^{\top}] = o(1), %\hat{\Psi}(\hat{\theta}) = \Psi(\theta^*) + o_p(1), 
\end{equation}
\end{lemma}

For each procedure, to show Theorems~\ref{thm:main} and~\ref{coro:main-moment}, the key 
% In each case of \Cref{ex:ddo}, the key 
is to verify that Lemmas~\ref{asp:theta-asymptotics}
and~\ref{asp:empirical-if} hold based on varieties of $M$-estimator theories implied from Assumptions~\ref{asp:theta-star},~\ref{asp:moment},~\ref{asp:theta-hat}. We provide proof details in \Cref{app:apply}.

\subsection{Unpacking the OIC}
In this subsection, we explain some insights on the bias correction term in OIC, in particular highlighting the interaction between the parameter space
$\Theta$ and the decision space $\Xscr$. First, we can rewrite: 
\[
  \hat{A}_c = \E_{\hat\P_n}\Paran{\|\nabla_{\theta}
  h(x^*(\hat{\theta});\xi)\|_2 \|\widehat{\IFx}(\xi)\|_2 \text{cos}(-\nabla_{\theta} 
  h(x^*(\hat{\theta});\xi), \widehat{\IFx}(\xi))},
\]
where $\text{cos}(-\nabla_{\theta} 
  h(x^*(\hat{\theta});\xi), \widehat{\IFx}(\xi))$ is the cosine of the angle between $-\nabla_{\theta} 
  h(x^*(\hat{\theta});\xi)$ and $\widehat{\IFx}(\xi)$. \tyw{This expansion provides a useful interpretation of OIC as correcting the self-influence of the training sample on the evaluated decision. The vector \(\widehat{\IFx}(\xi_i)\) 
captures how much the $i$-th sample $\xi_i$ changes the fitted parameter \(\hat\theta\), while 
\(\nabla_\theta h(x^*(\hat\theta);\xi_i)\) captures how sensitive the
evaluated objective at \(\xi_i\) is to this change. The inner product,
therefore, quantifies the extent to which a particular
observation influences both the estimated parameter and the evaluated
objective value. } 

This decomposition also clarifies when the bias is large. First, the bias is possibly high when the decision rule is ``complex'' in that % $\nabla_{\theta}x^*(\cdot)$
% triggers a function class
$\{\|\nabla_{\theta}
h(x^*(\theta);\cdot)\|_2:\theta\in\Theta\}$ % that
has a high complexity
measure, % such as the
i.e., a large
Vapnik–Chervonenkis dimension, metric entropy, or
localized Rademacher complexity such that $\E_{\hat\P_n}[\|\nabla_{\theta} h(x^*(\hat\theta);\xi)\|_2]$ is large. Second, a high-dimensional parametric space $\Theta$ also causes a large bias, since it typically increases $\E_{\hat\P_n}[\|\widehat{\IFx}(\xi)\|_2]$, as a single sample can significantly affect the parameter estimate. \tyw{Finally, the magnitude of the bias depends on the cosine similarity between $\widehat{\IFx}(\xi)$ and $\nabla_{\theta} h(\datax;\xi)$. When these two vectors are closely aligned, a sample tends to influence the fitted parameter in a direction that also impacts the decision objective, and the bias is large. Conversely, the bias can be small if the vector fields $\nabla_{\theta}h(\datax;\xi)$
and $\widehat{\IFx}(\xi)$ are approximately orthogonal, which corresponds to the case where observations tend to perturb the fitted parameter in directions that have little effect on the evaluated objective. }

Similarly, the expected bias term can be similarly decomposed
into a product of three terms:
\begin{equation}\label{eq:bias-illustration}
  A_c = \E_{\P^*}\Big[\|\nabla_{\theta}
  h(x^*(\theta^*);\xi)\|_2\|\IFx(\xi)\|_2\text{cos}\big(-\nabla_{\theta}
  h(x^*(\theta^*);\xi), \IFx(\xi)\big)\Big].
\end{equation}
%\wty{Will change aligned with Ex 1 and previous descriptions.}
% Consider the unconstrained
Here, we visualize the interactions of these terms in the context of the portfolio optimization problem in \Cref{ex:portfolio}. Let 
% objective
$h(x;\xi) = \exp(-\xi^{\top}x) +
0.2\|x\|_2^2$, i.e., set $\gamma = 0.2$ in \Cref{ex:portfolio}.
% \hl{Why did you pick $\gamma = 0.2$? Do you additional numerical
%   experiments in the appendix for other values?}
Suppose the
true distribution $\P^* = N\left(\mu^*, \Sigma\right)$ with $\mu^* =
(1,2)^{\top}$ and $\Sigma = \bigl(\begin{smallmatrix}4 & 2\\ 2 &
  4\end{smallmatrix} \bigr)$. We specifically compare the bias terms $A_c$
for the following methods by 
demonstrating the interactions of the two components $-\nabla_{\theta} 
h(\bestx;\xi)$ and $\IFx(\xi)$ in $A_c(=\E_{\P^*}[(-\nabla_{\theta}h(\bestx;\xi))^{\top}\IFx(\xi)])$. 
%\hl{How did the negative sign suddenly appear here?}
% there specifically:
% then based on \Cref{ex:portfolio}, we have:
%so the comparison of the true performance of these two is more complicated depending on the size of the in-sample performance and the bias.}
% \begin{itemize}
  
\textbf{SAA}: % Consider $x^*(\theta)$ in \Cref{ex:saa}.
% $x^* = \argmin_{x \in \R^2}\E_{\hat{\P}_n}[h(\paranx;\xi)]$.  % in \Cref{ex:saa}. 
Here $\IFx(\xi) = -(\E_{\P^*}[\nabla_{\theta\theta}^2
h(\theta^*;\xi)])^{-1} \nabla_{\theta} h(\theta^*;\xi)$, where $\theta^* =
\argmin_{\theta \in \R^2}\E_{\P^*}[h(\theta;\xi)]$. The gradient
$\nabla_{\theta}h(\theta;\xi) = -\xi \exp(-\xi^{\top}\theta) + 0.4
\theta$.  
% where
% $x^* = \argmin_{\theta \in \R^2}\E_{\P^*}[h(\paranx;\xi)]$.
% The
% gradient $\nabla_{x}h(x^*;\xi) = -\xi \exp(-\xi^{\top} x)
% + 0.4 x$. 
In this problem instance, $x^*=
(0.04, 0.41)^{\top}$ and is optimal in the sense that it minimizes the expected cost $\E_{\P^*}[h(x;\xi)]$.

\textbf{ETO}: % Consider
% Suppose
Consider $\paranx$ as the decision rule in \Cref{ex:ddo}$(a)(1)$ with the 
misspecified distribution class $\Pscr_{\Theta}=\{ N(\theta, 4 I_{2
  \times 2}): \theta \in \mathbb{R}^2\}$ and $T$ given by the MLE. 
Thus, the estimated $\hat{\theta} = \frac{1}{n} \sum_{i = 1}^n \xi_i$, and $\IFx(\xi) = \xi - \theta^*$, where 
$\theta^* = \mu^*$. The gradient $\nabla_{\theta} h(x^*(\theta);\xi) =
\nabla_{x} h(x;\xi) \nabla_{\theta} x^*(\theta)$, where the $i$-th
component of $\nabla_{\theta}x^*(\theta)$ is obtained from the first-order
optimality condition in ETO, i.e., $\nabla_{x}(\E_{\P_{\theta}}[h(x;\xi)])
= (-\theta  + 4x)\exp(-\theta^{\top}x + 2\|x\|_2^2) + 0.4 x  
= \mathbf{0}$ through the chain rule applied to the implicit function. 
% and  
% Then the first-order condition in ETO is  $f(\theta;x) (:=
% \nabla_{x}(\E_{\P_{\theta}}[h(x;\xi)]) = \mathbf{0}$. Then following the
% chain rules of the implicit function result in
% Appendix~\ref{app:discuss} to $f_i(\theta;x)$, i.e., $i$-th component of
% $f(\theta;x)$ yields the expression of $\nabla_{\theta}(x^*(\theta))_i$,
% i.e. the $i$-th gradient component of $x^*(\theta)$.  
In this problem instance, $x^*(\theta^*) = (0.21, 0.42)^{\top}$.
% \end{itemize}

\textbf{IEO}: Consider
$\paranx$ as the decision rule in \Cref{ex:ddo}$(a)(2)$ with the 
misspecified distribution class $\Pscr_{\Theta}=\{ N(\theta, 4 I_{2 
  \times 2}): \theta \in \mathbb{R}^2\}$. The solution $x^*(\theta)$ satisfies the first order conditions $f(\theta;x):=(-\theta + 4x) \exp(-\theta^{\top}x + 2\|x\|_2^2) + 0.4 x = \bm{0}$.  Since both $\nabla_{\theta}f(\theta, x)$, and $\nabla_x f(\theta,x)$ are positive definite, $\nabla x^*(\theta)$ and $\nabla_{\theta\theta}^2
x^*(\theta)$ exist from inverse function theorem
(\Cref{lemma:implicitfunction} in Appendix~\ref{sec:further discussions}). We can compute $\nabla x^*(\theta)$ and $\nabla_{\theta\theta}^2 x^*(\theta)$ using~\eqref{eq:datax} and~\eqref{eq:inverse-general} in Appendix~\ref{sec:further discussions} and obtain the gradient $\nabla_{\theta} h(x^*(\theta);\xi) = \nabla_{x} h(x;\xi) \nabla_{\theta} x^*(\theta)$  and influence function $\IFx(\xi) = -(\E_{\P^*}[\nabla_{\theta\theta}^2 
h(x^*(\theta^*);\xi)])^{-1} \nabla_{\theta} h(x^*(\theta^*);\xi)$. \tyw{In this problem instance, although $\P^* \notin \Pscr_{\Theta}$, the parametric family $\Pscr_{\Theta}$ remains sufficiently flexible to reproduce the optimal decision, and therefore IEO and SAA yield the same decision. We have $x^*(\theta^*) = (0.04, 0.41)^{\top}$ with $\theta^* = (0.17,0.91)^{\top}$. Note that multiple values of $\theta^*$ satisfy $f(\theta^*;x^*(\theta^*)) = 0$; among them, we report the one minimizing $\|\theta\|$.}

\begin{figure}[!htb]
    \centering
    \subfloat[SAA Bias $(A_c = 2.72)$]
    {
        \begin{minipage}[t]{0.32\textwidth}
            \centering
            \includegraphics[width = 0.99\textwidth]{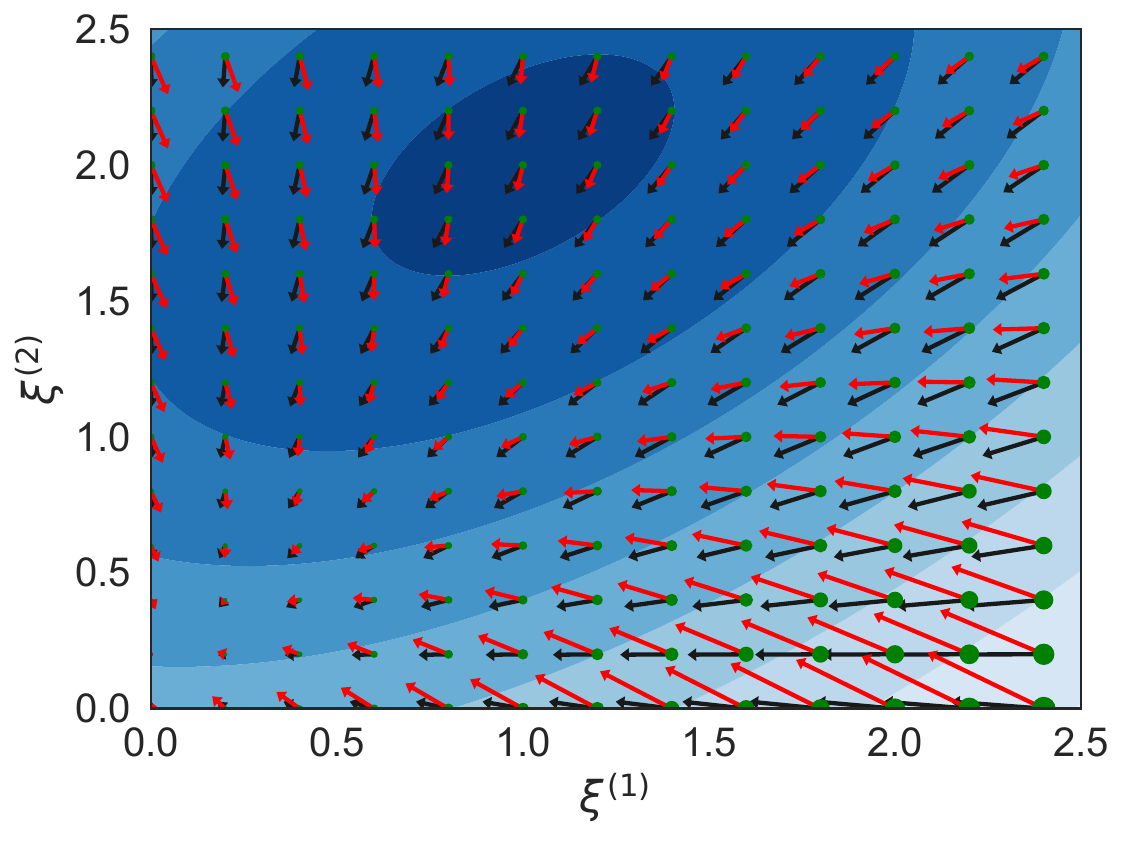}
        \end{minipage}
    }
    \subfloat[ETO Bias $(A_c = 1.11)$] 
    {
        \begin{minipage}[t]{0.32\textwidth}
            \centering
            \includegraphics[width = 0.99\textwidth]{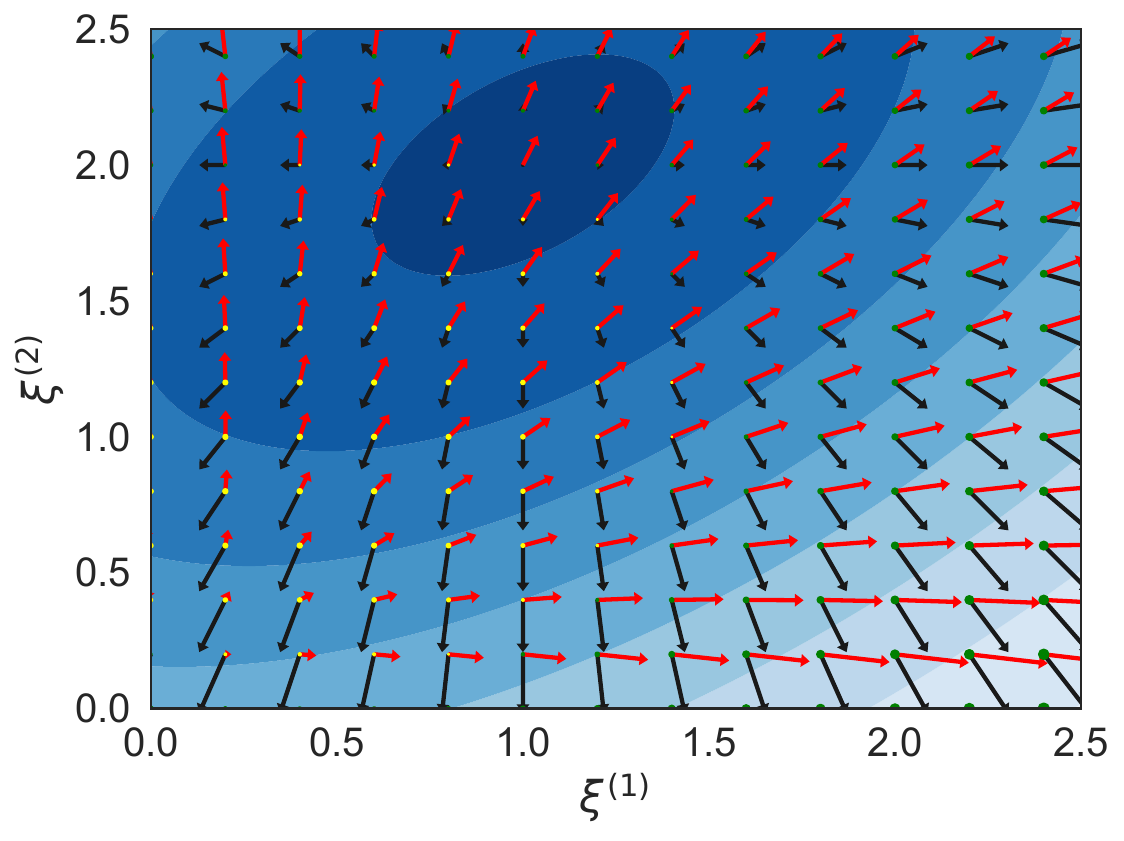}
        \end{minipage}
    }
    \subfloat[IEO Bias $(A_c = 2.72)$]
    {
        \begin{minipage}[t]{0.32\textwidth}
            \centering
            \includegraphics[width = 0.99\textwidth]{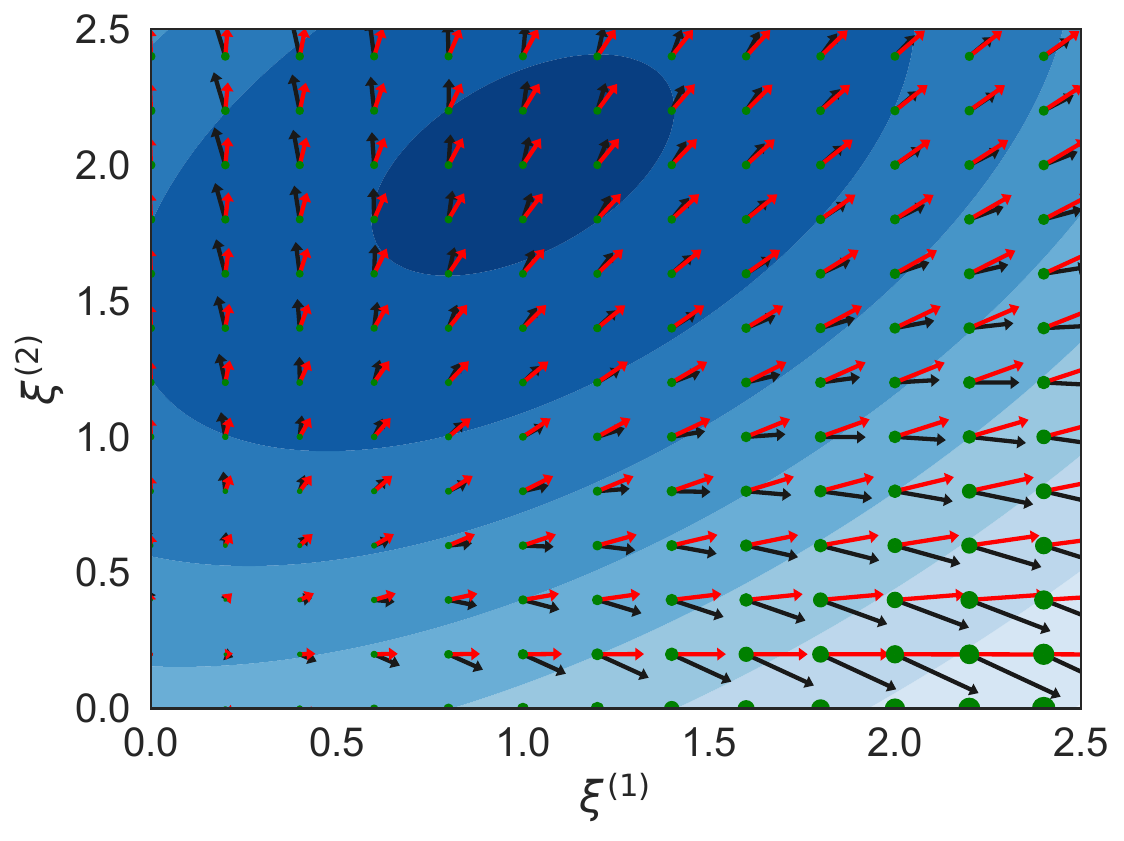}
        \end{minipage}
    }
    \caption{Vector field plot interaction of the negative function gradient $-\nabla_{\theta} h(x^*(\theta^*);\xi)$ and influence function $\IFx(\xi)$ under $\xi = (\xi_1, \xi_2)^{\top}$, where the blue background is the density of $\xi \sim \P^*$. The green (resp. yellow) points represent
the points where the inner product of $-\nabla_{\theta}
h(x^*(\theta^*);\xi)$ (red) and $\IFx(\xi)$ (black) is positive
(resp. negative), and the size of these points represents the absolute
value of that inner product.  } 
    \label{fig:gradient}
\end{figure}

The values of $-\nabla_{\theta} h(x^*(\theta^*);\xi)$ and
$\IFx(\xi)$ for SAA, ETO and IEO as a function of $\xi \in \Xi$ are
plotted in \Cref{fig:gradient}. \tyw{Although SAA and IEO produce the same decision
  and share the same $A_c$, the gradient and influence function vectors are different because the solutions are computed using 
  different optimization procedures and correspond to
  different parameterizations.  % representations.
} It is clear that the norm of the
gradient term $\|\nabla_{\theta} h(x^*(\theta^*);\xi)\|_2$ and that of the
influence function term $\|\IFx(\xi)\|_2$ for SAA, ETO and IEO are both
large.  However, the angle between the two vectors is larger in ETO as
compared to \tyw{SAA and IEO}. From~\eqref{eq:bias-illustration}, these
observations explain why the bias in ETO is lower than that in \tyw{SAA
  and IEO}. 

% \wty{Furthermore, we notice that both norms of the gradient term
% $\|\nabla_{\theta} h(x^*(\theta);\xi)\|$ and the influence function term
% $\|IF(\xi)\|$ in ETO can be large from the comparable black and red
% vector length between ETO in $(a)$ and SAA in $(b)$. However, since the
% bias $A_c$ is computed by: 
% \[\E_{\P^*}[\|\nabla_{\theta}
% h(x^*(\theta);\xi)\|_2\|IF(\xi)\|_2\text{cos}(\nabla_{\theta}
% h(x^*(\theta);\xi), IF(\xi))],\] 
% }

%   Furthermore we present the
% The
% estimated influence functions
% $\widehat{IF}(\xi)$ for the procedures as defined 
% given $\Dscr_n$ and estimated $\hat\theta$ for each
% procedure above. 
\wty{Finally, we include the mathematical forms of estimated influence functions for completion. Since the cost function and optimization procedures satisfy the conditions in \Cref{coro:oic-procedure}, it follows that $\widehat{\IFx}(\xi) = -(\E_{\hat\P_n}[\nabla_{\theta\theta}^2
h(\hat\theta;\xi)])^{-1} \nabla_{\theta} h(\hat\theta;\xi)$  for 
SAA, 
$\widehat{\IFx}(\xi) = \xi - \hat\theta$ for ETO, and  $\widehat{\IFx}(\xi) =
-(\E_{\hat\P_n}[\nabla_{\theta\theta}^2 h(x^*(\hat\theta);\xi)])^{-1}
\nabla_{\theta} h(x^*(\hat\theta);\xi)$ % where
% $\nabla_{\theta\theta}h(\datax;\xi)$ and
% $\nabla_{\theta}h(\datax;\xi)$.
for IEO. Since all these procedures satisfy
\Cref{asp:empirical-if}, the specific expressions in these formulas can be further computed using~\eqref{eq:datax}
and~\eqref{eq:inverse-general} in 
Appendix~\ref{sec:further discussions}.} 

\section{Generalizations of OIC}\label{sec:extension}
In this section, we present several generalizations of the OIC framework.  % including the
% incorporation of the
In particular, we discuss applying OIC beyond canonical procedures in Definition~\ref{ex:ddo} (Section~\ref{subsec:application-if}), the
extension of the framework to general 
%to include 
risk functions $u$ % that we have promised in criterion
% \eqref{primary} 
(Section \ref{sec:general-risk}),  an alternative
evaluation criterion and modified OIC methodology that can help evaluate
the impact of model misspecification~(\Cref{sec:POIC}), extension to
contextual optimization~(\Cref{sec:context}) and extension to
nonparametric decision rules~(\Cref{subsec:np-decision}). 
% suits parametric models the model evaluation using OIC over the general risk case beyond the expected decision performance in \Cref{sec:general-risk}. Then we describe an alternative evaluation criterion against OIC under some decision rules in \Cref{sec:POIC}, which also helps calculate the effects of parametric model misspecification error on the downstream decision rule and extend the OIC result to the evaluation under the contextual optimization problem in \Cref{sec:context}.

\subsection{Applying OIC beyond Canonical Procedures in~\Cref{ex:ddo}}\label{subsec:application-if} 
Although we have focused on the  optimization
procedures in \Cref{ex:ddo}, from the proof of Theorems~\ref{thm:main}
and~\ref{coro:main-moment} it follows that the  % n fact, our general  
bias correction % formulas
expressions hold
% hold as long as the conditions of Theorems~\ref{thm:main}
% and~\ref{coro:main-moment} -- most notably
as long as~\eqref{eq:theta-if} and~\eqref{eq:empirical-if} are
satisfied. This provides the most general % vehicle to assert
setting for the statistical guarantees of OIC. % On the other hand,
\begin{proposition}[Validity of OIC beyond \Cref{ex:ddo}]\label{prop:oic-general}
    For any $x^*(\theta)$, suppose Assumptions~\ref{asp:nonsmooth} and~\ref{asp:moment}, and~\eqref{eq:theta-if} and~\eqref{eq:empirical-if}  hold for $\hat\theta$ and $\theta^*$. Then the conclusions of Theorems~\ref{thm:main} and~\ref{coro:main-moment} hold.
\end{proposition}
This indicates our results hold for procedures that are not covered by~\Cref{ex:ddo}.

Next, we discuss general principles for verifying~\eqref{eq:theta-if} and~\eqref{eq:empirical-if}. Equation \eqref{eq:theta-if} regarding the estimator $\hat\theta$ is typically satisfied by a range of statistical estimators, including $M$-estimators, robust statistics, $U$-statistics (e.g., Chapter 12 in \cite{van2000asymptotic}, and random forest~\citep{wager2018estimation}), and semiparametric estimators (e.g., \cite{kennedy2024semiparametric}) since they all have the asymptotic normality properties. 

\tyw{The verification of \eqref{eq:empirical-if} depends on how the influence function is estimated:
\begin{enumerate}[(i)]
    \item When the estimated influence function is computed using the
      plug-in form \eqref{defn:estimate-if},~\eqref{eq:empirical-if} can
      be verified via a standard decomposition: 
    \begin{equation}\label{eq:eif-consisteny-bound} 
    \|\widehat{\IFx}(\xi) - \IFx(\xi)\| \leq \|\IFx(\xi;\hat T, \hat\P_n)
    - \IFx(\xi;\hat T, \P^*)\| + \|\IFx(\xi;\hat T, \P^*) - \IFx(\xi;T,
    \P^*)\|. 
  \end{equation}
  The first term on the right of the inequality in
  \eqref{eq:eif-consisteny-bound} reflects the error from replacing population quantities with empirical counterparts. It is
  controlled by the consistency of the empirical gradients, Hessian,
  multipliers and the estimated parameter $\hat\theta$. The second term
  captures the error from replacing the exact calibration function
  $T(\cdot)$ by % its approximate implementation
  approximation
  $\hat T(\cdot)$. This
  term is controlled since the
  optimization accuracy is of the order of $o_p(n^{-1})$. Then the plug-in form~\eqref{defn:estimate-if}
  satisfies~\eqref{eq:empirical-if} under standard consistency
  conditions. 
\item Besides the plug-in form in~\eqref{defn:estimate-if}, one may also
  use some problem-specific approaches to estimate the influence function
  % while
  that 
  avoid 
  computing 
  the inverse of the Hessian. % inverse.  
  For example, one may use iterative linear-algebra methods that approximate the
  inverse-Hessian-vector product without forming the inverse
  explicitly~\citep{koh2017understanding,stephenson2020approximate}. To
  verify~\eqref{eq:empirical-if} for such approximations, it suffices to
  show that the approximation error relative to the plug-in estimator is
  asymptotically negligible, and then combine this with the plug-in
  bound in~\eqref{eq:eif-consisteny-bound}. We provide a concrete
  example based on conjugate gradient methods~\citep{saad2003iterative}
  \tyw{with error guarantees} in Appendix~\ref{app:application-if}. 
\end{enumerate}}

Finally, when Assumption~\ref{asp:nonsmooth} or~\ref{asp:moment} is not satisfied, we can still apply our results provided we can appropriately smooth the objective so that \eqref{eq:theta-if} and~\eqref{eq:empirical-if}  hold for some $\hat\theta$ and $\theta^*$. Here, we demonstrate how OIC can be applied to approximations of linear objectives by verifying~\eqref{eq:theta-if} and~\eqref{eq:empirical-if} hold for linear objectives suitably regularized by a proximal function. For example, SAA in the linear objective setting results in the linear program % of the form: 
\begin{equation}\label{eq:lp}
    \hat\theta \in \argmin_{\theta}
    \paran{\E_{\hat\P_n}[\xi]^{\top}\theta,~\text{s.t.},~ F\theta = g, \theta
      \geq 0}.
  \end{equation}
  It is clear that the objective of the linear program violates \Cref{asp:moment}. Instead, consider the following entropy-regularized convex optimization problem 
  % The entropic regularization is proposed to solve:
\begin{equation}\label{eq:lp-entropic}
  \hat\theta(\eta_n) \in
  \argmin_{\theta}\paran{\E_{\hat\P_n}[\xi]^{\top}\theta + \eta_n^{-1}
    \big(\sum_{i}\theta_i \log \theta_i\big), ~\text{s.t.}~F\theta = g},  
\end{equation} 
where $\eta_n > 0$ is a penalization parameter. \tyw{The entropy-regularized objective $\tilde h(x;\xi) = \xi^{\top}x + \eta_n^{-1}\sum_i x_i \log x_i$ satisfies \Cref{asp:moment}. Moreover, for an appropriate choice of $\eta_n$, the entropy-regularized solution is close to the original LP solution. Motivated by this observation, we propose a computable OIC for the entropy-regularized objective to approximate
$A = \E_{\Dscr_n}\E_{\P^*}[h(x^*(\hat\theta);\xi)]$.} %  (i.e., the exact 
\begin{corollary}[OIC for Entropy-Regularized Linear Program]\label{prop:oic-entropic-lp}
  {% Consider
    Suppose $h(x;\xi) = \xi^{\top}x$, and let  $\hat{\theta}(\eta_n)$
    denote the solution of the entropy-regularized SAA
    procedure~\eqref{eq:lp-entropic}. Define
    \[
      \hat A(\eta_n) =  \E_{\hat\P_n}[\xi]^{\top}\hat\theta({\eta_n}) - \frac{1}{n^2}\sum_{i
      = 1}^n \xi_i^{\top}\widehat{\IFx}(\xi_i) - \eta_n^{-1}\sum_{j \in [D_{\xi}]}
    \hat\theta_j(\eta_n) \log\hat\theta_j(\eta_n),
  \]
  where $\widehat{\IFx}(\xi) = -
  \eta_n \cdot \text{diag}(\hat\theta({\eta_n}))\cdot(I - F^{\top}(F
  \text{diag}(\hat\theta({\eta_n}))F^{\top})^{-1} F
  \text{diag}(\hat\theta({\eta_n})))(\xi - \E_{\hat\P_n}[\xi])$ and 
  $\eta_n = 2\log n$.
 Suppose the decision space $\Xscr = \{x: Fx = g, x
\geq 0\}$ is bounded and % some regularity conditions hold (i.e.,
 \Cref{asp:additional-entropic-lp} in
 Appendix~\ref{app:application-if} holds.  Then 
  $\E[\hat A({\eta_n})] = A + \tilde o(1/n)$.} 
\end{corollary}
This result is established by (i) applying OIC to the smoothed objective
$\tilde h(x;\xi) = \xi^{\top}x + \eta_n^{-1}\sum_i x_i \log x_i$ under
$\hat\theta(\eta_n)$ \tyw{through verifying the conditions of
  \Cref{prop:oic-general}; and (ii) subtracting the entropy term
  $\eta_n^{-1}\sum_i x_i \log x_i$ to recover the performance evaluation
  of the original linear objective under $\hat\theta$. {Note that for this problem, the expression of the influence function can be simplified because the Lagrangian multiplier for this constrained problem has a closed-form expression as a function of problem parameters.} %  which allows us to simplify
  % the influence function formula. 
  We choose $\eta_n = \Theta(\log n)$ to
  balance the stochastic and smoothing errors. % simultaneously
  This choice
  ensures (i) the corresponding influence function $\IFx(\xi)$ grows only
  logarithmically in $n$ such that the stochastic
  expansion~\eqref{eq:theta-if} remains valid up to logarithmic factors;
  (ii) $\|\hat\theta(\eta_n) - \hat\theta\|_1 = O(\exp(-\eta_n)) = o(1/n)$
  (from Corollary 9 in~\cite{weed2018explicit}) such that replacing
  $\hat\theta$ with its entropy-regularized counterpart
  $\hat\theta(\eta_n)$ incurs only a $\tilde o(1/n)$ difference in the
  evaluated objective value.} 
% the calibration procedure for some non-smooth 
% objectives do not satisfy 

% n general optimization procedures,

\tyw{% Overall, 
  This subsection % clarifies
  extends the applicability of OIC beyond
  the canonical procedures in \Cref{ex:ddo}. The key requirements are the asymptotic linear expansion~\eqref{eq:theta-if} and the consistency condition~\eqref{eq:empirical-if}. Thus,
  for % a new
  any
  data-driven optimization procedure, OIC can be applied by
  verifying these two conditions. When the original problem violates the
  smoothness or curvature assumptions, as in linear optimization, one may
  instead apply OIC to a smooth approximation, and then % eliminate
  subtract  the
  approximation error to recover the evaluation of the original problem.}

\subsection{General Risk Functions}\label{sec:general-risk}
Section \ref{sec:main} focuses on evaluating the
bias for % evaluation
% of 
the expected objective performance~\eqref{eq:eval-key}.
% As
% described in the introduction, 
Here we discuss bias evaluation for the more general
objective~\eqref{primary}.
Since the realized objective
% performance
value 
of a
data-driven decision is a random variable, it is helpful to % understand
evaluate 
it
% from a distributional view, including
using a ``risk''  metric. % that emphasizes 
% % for instance 
% the ``risk'' of the random variable.
% We have moved some regularity conditions to ensure $\E[((1/n) \sum_{i = 1}^n h(\paranx;\xi_i) -
% \E_{\P^*}[h(\paranx;\xi)])^2]$ is well-defined to
% Appendix~\ref{app:general-risk}.  
% \Cref{asp:additional-risk} includes % some
% regularity conditions to ensure $\E[((1/n) \sum_{i = 1}^n h(\paranx;\xi) -
% \E_{\P^*}[h(\paranx;\xi))]^2]$ is well-defined. 
% tail distribution, e.g., 
% % of the objective performance.
% To this end, we are interested in evaluating
% \eqref{primary} for any given risk function $u(\cdot)$, where $u(x)=x$
% reduces to the case studied in Section \ref{sec:main}. 
%     A := \E_{\Dscr_n}\E_{\P^{*}}[h(\datax;\xi)].
% \end{equation}
% Recall $Z(\datax) = \E_{\P^*}[h(\datax;\xi)]$, $A = Z(\bestx) = \E_{\P^*}[h(\bestx;\xi)]$ and $\hat A_o = \E_{\hat\P_n}[h(\datax;\xi)]$. In this section, 
% %we estimate more deterministic quantities from the distribution $Z(\datax)$ under different forms of $u(\cdot)$. 
% our target quantity of the decision performance becomes:
% \begin{equation}\label{eq:risk-formulation}
%     A_u := \E_{\Dscr_n}\left[u(Z(\datax))\right] = \E_{\Dscr_n}\left[u\Para{\E_{\P^*}[h(\datax;\xi)]}\right],
% \end{equation}
% where we call $u(\cdot): \R \mapsto \R$ a risk function. The expected method performance $A$ we discuss so far is a special case with $u(x) = x$. 
% We have the following result:
\begin{theorem}[Risk-Incorporated Optimizer's Information Criterion
  (Risk-OIC)]\label{thm:risk-oic}
 Suppose Assumptions~\ref{asp:nonsmooth},~\ref{asp:moment},~\ref{asp:additional-risk} in Appendix~\ref{app:general-risk}, and~\eqref{eq:theta-if},~\eqref{eq:empirical-if} hold. 
 % When 
 Also suppose $u(x)$ has continuous second-order derivatives everywhere. Define the Risk-OIC
 \[
   \hat A_u := u(\hat A_o) + u'(\hat A_o)\hat A_c - (1/(2n)) u''(\hat
   A_o)\hat\sigma^2,
 \]
 where $\hat \sigma^2 =
 (1/n)\sum_{i = 1}^n (h(\datax;\xi_i) - \hat A_o)^2 $, and $\hat A_c$ is
 the bias correction term in % defined in
 \Cref{def:oic}. 
 Then $\E[\hat A_u] = A_u + o\Para{1/n} = \E[u(\hat A_o)] + u'(A) A_c - (1/2n)
u''(A)\text{Var}[h(\bestx;\xi)] + o\Para{1/n}$. 
\end{theorem}
% Under different forms of $u(\cdot)$, we can estimate more quantities from the distribution $Z(\datax)$. 
% Here
% To understand Theorem
% \ref{thm:risk-oic}, first 

Note that the bias in evaluating $A_u$ % via
using the 
% natural
empirical estimator % namely 
$u(\hat A_o)$ is given by $A_u -
\E[u(\hat A_o)] = u'({A}) A_c - (1/2n)
u''(A)\text{Var}[h(\bestx;\xi)] + o(1/n)$. 
%This is because 
%Like the case $u(x)=x$ that we studied in Section \ref{sec:main}, 
Since $A_c = O(1/n)$, the bias of the empirical estimator $u(\hat A_o)$ is
of order  $O(1/n)$ and reduces to $o(1/n)$ if and only if $2n u'(\hat{A})
A_c = u''(\hat{A})\text{Var}[h(\bestx;\xi)]$, which is seldom the case.  
% the case. In this regard, a bias-corrected estimator
% $\hat A_u$ should have $\E[\hat A_u] = A_u + o\Para{1/n}$, and this is
% achieved precisely in 
Theorem \ref{thm:risk-oic} % We can see that
computes a $o(1/n)$ estimator by adding two terms to $u(\hat A_o)$.
% In this
% Note that the term $\hat A_c$ introduced in % OIC of 
% \Cref{def:oic} still plays an important role in 
% % this risk-incorporated
% Risk-OIC. % formula, 
% However,
Compared to \Cref{def:oic}, the % correction to fully remove that
bias correction term in Risk-OIC has 
% requires
two % additional
% manipulations
modifications:
% , one being the amplification of
% first, 
first, $\hat A_c$, introduced in \Cref{def:oic},  is multiplied by the
first-order derivative of the risk function, $u'(\hat 
A_o)$; second, % one being
% second, 
there is 
an additional % bias correction 
term $-(1/2n)u''(\hat
A_o)\hat\sigma^2$ consisting of  both the second-order derivative of
$u(\cdot)$ at $\hat A_o$ and the variance of
$h(x^*(\hat\theta);\xi)$. % The proof of
These modifications arise because 
\Cref{thm:risk-oic} is established by taking a second-order 
Taylor expansion of $u(\cdot)$, in addition to the optimality gap. % to utilize the existing bias formula
% $\hat A_c$ as well as bringing about the variance $\E[(\hat A_o -
% \E_{\P^*}[h(\datax;\xi)])^2]$.
% leading to these two terms.
% We note that, among all possible
% $u(\cdot)$,
% . However, we need to take the second-order Taylor expansion to remove the $\Theta(1/n)$ bias since the variance term $\E[(\hat A_o - Z(\datax))^2]$ in the second-order expansion is $\Theta(1/n)$.
% a crucial difference arises from the additional Compared with our results in Section \ref{sec:main},  the optimistic bias of $\hat A_o$ is thus amplified with the gradient $g'(\cdot)$ for determining the bias order of general risk measures (I COPIED HERE FROM YOUR PREVIOUS WRITING AND IT WOULD NEED REVISION. IT'S IMPORTANT TO TELL READER THE CONCEPTUAL DIFFERENCE BETWEEN RISK OIC AND OIC, WHICH I THINK IS THE ADDITIONAL BIAS THAT INVOLVES THE DERIVATIVE  VARIANCE ETC. UNFORTUNATELY, I DON'T SEE A CLEAR EXPLANATION AND JUSTIFICATION YET IN THE PREVIOUS VERSION). 
% \begin{corollary}\label{coro:data}
% %Recall $A = Z(\bestx)$, then t
% $\E[u(\hat A_o)] = A_u + o\Para{1/n}$ if and only if $2n u'(A) A_c =
% u''(A)\text{Var}[h(\bestx;\xi)]$. 
% %, where $\sigma^2 = \frac{1}{n}$.
% \end{corollary}
% The condition $u'(A) A_c = \frac{1}{2n} u''(A)\text{Var}[h(\bestx;\xi)]$ from \Cref{coro:data} is seldom justified. 
% We can also conduct 
% We can also conduct
We end this section with an example of Risk-OIC applied to  specific risk functions.
%as well as observations that $Z(\datax) = Z(\bestx) + O_p\Para{\frac{1}{n}}$ and 
%$\hat\sigma^2$ consistently estimates the error . 
\begin{example}[Objective Performance Evaluation for different $u(\cdot)$]
  Suppose $h(x;\xi) > 0$, for all $x, \xi$. 
  %\hl{Why do we need this assumption?}
  % By using
  Then setting 
  $u(x) = x^2$ % the
  % decision-maker evaluates the
  results in a risk-averse evaluation of the realized
  objective~\citep{mas1995microeconomic}. % performance in a risk-averse
  % manner 
  In this case, $\hat A_u = \hat A_o^2 + 2
  \hat A_o \hat A_c - 2 \hat\sigma^2$.
  Setting $u(x) = \sqrt x$ results in a  risk-seeking evaluation of the
  realized objective. 
  % decision-maker is risk-seeking, 
  In this case $\hat A_u = \sqrt{\hat A_o} +
  \frac{1}{2}\hat A_c\cdot(\hat A_o)^{-1/2} + \frac{1}{2}\hat
  \sigma^2\cdot(\hat A_o)^{-3/2}$. % With these formulas, if we just
  Comparing the bias terms
  $2 \hat A_o \hat A_c - 2 \hat\sigma^2$ and
  $\frac{1}{2}\hat A_c\cdot(\hat A_o)^{-1/2} + \frac{1}{2}\hat
  \sigma^2\cdot(\hat A_o)^{-3/2}$, we see that % the decision-maker favors
  % a
  a smaller $\hat A_o$ is preferred in the % when
  risk-averse setting and vice versa in the
  risk-seeking setting.
\end{example}

\subsection{Parametric-OIC and Model Misspecification}\label{sec:POIC}
The goal of Parametric-OIC is to \tyw{provide an alternative estimator of
  the decision performance} and evaluate the impact of distributional misspecification on the downstream decision-making when using the parametric approach in~\Cref{ex:ddo}$(a)$. 
% where $\theta$ represents the parameter of the distribution.
%This leverages on the construction of another performance estimator.
% First, 
We start with the case where the true distribution is well-specified,
i.e., $\P^* \in \Pscr_{\Theta}$, and thus 
it is legitimate to use the parametric model itself in the
evaluation. % This gives our Parametric-OIC: 
% , which leads to an alternative construction of OIC. 
%In the well-specified case of ETO (i.e.  when ), 
% we can design an alternative estimator for $A$ with the same principle as before but, instead of using the bias corrected form from the empirical counterpart $\hat A_o$, we use the bias corrected version from the parametric counterpart which is now statistically consistent since $\P^* \in \Pscr_{\Theta}$
% We can obtain another asymptotical unbiased estimator of $A$ when $\P^*
% \in \Pscr_{\Theta}$, which does not use EM as the base estimator
% $\hat{A}_o$.
\begin{assumption}[Optimality Condition for \Cref{ex:ddo}$(a)$]\label{asp:pf-fit-optimal}
%The first-order optimality condition holds with 
  % Suppose $x^*(\hat\theta)$ satisfies 
  % % uniquely solving the equation 
  % the equation $\nabla_x \E_{\P_{\hat\theta}}[h(x;\xi)] = \wty{o_p(n^{-1})}$. 
  Suppose Assumption~\ref{asp:theta-star} holds. In addition, when 
  \begin{enumerate}[(i)]
  \item % if
    $\hat\theta$ % is obtained from
    is estimated using ETO: 
    Assumptions~\ref{asp:theta-hat}$(i)$ and~\ref{ex:theta-asymptotics} in
    Appendix~\ref{app:eto}  hold.
  \item $\hat\theta$ is estimated using IEO:
    Assumptions~\ref{asp:theta-hat}$(ii)$ and~\ref{asp:ierm-additional}
    in Appendix~\ref{app:ieo} hold.
  \end{enumerate}
  % The second-order optimality condition holds with 
%And the Hessian $\hessianp \E_{\P^*}[h(\bestx;\xi)]$ is positive definite.
%And for the empirical distribution $\hat{\P}_n$ associated with $\hat{\theta}$, 
\end{assumption}
\begin{theorem}[Parametric Optimizer's Information Criterion (Parametric-OIC)]\label{prop:pf-fit} 
% (THIS, AS WELL AS THE THEORY ON CONTEXT-OIC, SHOULD BE THEOREMS INSTEAD OF PROPOSITION. I HAVEN'T CHANGED THESE YET BECAUSE IT MAY INDUCE MORE NEEDED CHANGES..)
 Define the Parametric-OIC 
  \begin{equation}\label{eq:pf-fit}
    \hat{A}_p := \E_{\P_{\hat{\theta}}}[h(x^*(\hat{\theta});\xi)] +
    \frac{1}{2n}\Tr{I_h(\hat{\theta}) \Psi(\hat{\theta})},  
  \end{equation}
where $I_h(\theta) = \hessianp \E_{\P^*}[h(\paranx;\xi)]$ and
$\Psi(\hat{\theta})$ is defined in \Cref{asp:theta-asymptotics}.
 Suppose
  Assumptions~\ref{asp:nonsmooth} and~\ref{asp:pf-fit-optimal} hold, $\P^* = \P_{\theta^*}\in \Pscr_{\Theta}$ 
%$\E[\hat{\theta}] = \theta$, and the solution when $\theta$ denotes the parametric distribution is given by $\datax$ (see \Cref{ex:ddo}$(a)$), 
  and $\E[\hat{\theta}] = \theta^* + o(n^{-1})$.
Then $\E[\hat{A}_p] = A + o\Para{1/n}$.
%where the definition of $I_h(\theta)$ is the same as in~\Cref{coro:ierm}.
\end{theorem}
% \begin{definition}[Parametric Optimizer's Information Criterion (Parametric-OIC)]\label{def:OIC-P}
% Parametric-OIC $\hat{A}_p$ is defined as follows:
% % \begin{equation}\label{eq:pf-fit}
% %     \hat{A}_p = \E_{\P_{\hat{\theta}}}[h(x^*(\hat{\theta});\xi)] + \frac{1}{2n}\Tr{I_h(\hat{\theta}) \Psi(\hat{\theta})}, 
% % \end{equation}
% % where $I_h(\theta) = \hessianp \E_{\P^*}[h(\paranx;\xi)]$ and $\Psi(\hat{\theta})$ is defined in \Cref{asp:theta-asymptotics}.
% \end{definition}
% Compared to the OIC in \Cref{def:oic},

Similar to  OIC, Parametric-OIC also consists of two terms: the plug-in estimate (similar
to $\hat A_o$ in \Cref{def:oic}) and the bias correction term (similar to
$\hat A_c$ in \Cref{def:oic}). The distinction % arises is our usage of
is that we use 
$\P_{\hat\theta}$, instead of the empirical distribution $\hat\P_n$, in
defining Parametric-OIC. % to approximate the true distribution $\P^*$. 
The bias correction term
in \eqref{eq:pf-fit} is designed to alleviate the error introduced by
plugging in $\P_{\hat\theta}$. % We show our bias correction result as
% follows:  
% \Cref{prop:pf-fit} shows that Parametric-OIC can reduce the bias up to $o(1/n)$ when the distribution is well-specified. 
As an application, we apply Parametric-OIC to the newsvendor problem.
\begin{example}[Newsvendor Problem]\label{ex:newsvendor2} 
Recall the newsvendor problem in \Cref{subsec:numerical-newsvendor}. % Consider the 
%being the family of exponential distributions parameterized by the mean $\theta$, 
%Then for
The
ETO solution $\paranx = \ln(p/c) \theta$ % with
for
the parametrization $\Pscr_{\Theta} = \{\text{Exp}(\theta): \theta \geq
0\}$.  % (see \Cref{ex:ddo}$(a)(1)$). 
We have $\Psi(\hat{\theta}) = \hat{\theta}$ and $I_h(\hat{\theta}) = \hessianp \E[h(\paranx;\xi)] = c [\ln(p/c)]^2$.
This recovers the result in Corollary 5 of \cite{siegel2021profit}.
%in \Cref{ex:ddo}$(a)(1)$, which is the same as result in Corollary 5 of \cite{siegel2021profit}.
\end{example}
% In general 

When $\P^* \notin \Pscr_{\Theta}$, our Parametric-OIC can be used to measure the impact of distributional misspecification. In this case, $\hat A_p$ in \eqref{eq:pf-fit} is an estimate for $\E_{\Dscr_n}\E_{\P_{\theta^*}}[h(\datax;\xi)]$, but not an estimate for 
the true decision performance 
$\E_{\Dscr_n}\E_{\P^*}[h(x^*(\hat{\theta});\xi)]$.
% when $\P^*
% \notin \Pscr_{\Theta}$,
However, $\hat A$ in
\eqref{eq:oic} % gives
is 
a bias-corrected estimate for the latter. {Let 
\[
  \Escr(\paranx;\Pscr_{\Theta}) := \Big(\E_{\Dscr_n}\E_{\P^*}[h(x^*(\hat{\theta});\xi)] -
  \E_{\Dscr_n}\E_{\P_{\theta^*}}[h(x^*(\hat{\theta});\xi)]\Big)
\]
denote the decrease in decision performance associated with approximating the true
distribution of $\xi$ using the parametric family
$\Pscr_{\Theta}$. \tyw{The quantity $\Escr(\paranx;\Pscr_{\Theta})$
  measures the discrepancy between the expected objective value of the data-driven decision $x^*(\hat{\theta})$ under the true distribution $\P^*$ and the misspecified parametric distribution
  $\P_{\theta^*}$. 
%(MAY NOT BE FULLY CLEAR WHAT DECISION LOSS HERE MEANS.. THIS SEEMS TO BE THE DISCREPANCY BETWEEN THE TRUE OBJECTIVE PERFORMANCE AND THE MISSPECIFIED OBJECTIVE VALUE, EVALUATED AT THE SAME DATA-DRIVEN SOLUTION) incurred by approximating the true distribution $\P^*$ within the misspecified parametric family $\Pscr_{\Theta}$. 
Unlike classical goodness-of-fit measures, $\Escr(\paranx;\Pscr_{\Theta})$ quantifies misspecification through its
impact on the downstream performance instead of 
% through
distributional discrepancy alone.}
%the decision loss attributable to model misspecification and helps diagnose whether the performance degradation of a data-driven optimization procedure is caused by limited data or by an inadequate parametric model.} 
The difference % between $\hat A_p$ and 
$\hat{A} - \hat{A}_p$ % quantifies the % effect
% impact of
% the parametric distribution misspecification
is a good estimate for $\Escr(\paranx;\Pscr_{\Theta})$.}
% error with respect to the
% downstream decision-making problem.
% on the performance in the downstream decision-making problem.
\begin{corollary}[Parametric Misspecification Error]\label{coro:eto-misspecification}
    {Suppose Assumptions~\ref{asp:nonsmooth},~\ref{asp:moment},~\ref{asp:theta-hat} and~\ref{asp:pf-fit-optimal} hold, and $\E[\hat\theta] = \theta^* +
    o(1/n)$. Then $\E[\hat A - \hat A_p] = % \Escr(\paranx;\Pscr_{\Theta}) 
    \Escr(\paranx;\Pscr_{\Theta}) +
    o(n^{-1})$.} % where \Escr(\paranx;\Pscr_{\Theta}) =
    % is the
    % misspecification error.
\end{corollary}
\tyw{The misspecification error can be used to assess whether
  a parametric model is adequate for a % downstream
  decision-making problem. For
  example, several parametric families could be candidates for modeling the demand in an inventory control problem, e.g., exponential, gamma, or log-normal distributions. A large value of $\hat A-\hat A_p$ for a parametric family indicates a substantial discrepancy in the downstream decision-making, and suggests that 
  an alternative model should be considered. Conversely, a small value of
  $\hat A-\hat A_p$ indicates that the parametric family is sufficiently
  accurate from a decision-making perspective, even if it is not close under a goodness-of-fit metric.} 

\tyw{More broadly, when the optimal objective value is normalized to zero, Parametric-OIC decomposes performance degradation into: (i) statistical estimation error arising from finite samples, captured by $\hat A_p$; and (ii) distributional misspecification error arising from an inaccurate parametric model, captured by $\hat A-\hat A_p$. This decomposition helps diagnose whether poor performance is primarily due to finite-sample estimation or to an inappropriate modeling assumption.}

% And the difference $\hat{A} - \hat{A}_p$ 
% %defined in \Cref{thm:main} and \Cref{prop:pf-fit}, respectively, yields an 
% is an estimator of the model misspecification error $\Escr(\paranx;\Pscr_{\Theta})$ with a bias of $o\Para{\frac{1}{n}}$, where $\Escr(\paranx;\Pscr_{\Theta})$ is
% defined as follows: 
%(to the downside optimization task) in terms of $\paranx$ parameterized by $\P_{\theta}$:
% \[.\]
% That is, when we evaluate different fitted parametric models, the term $\hat A - \hat A_p$ for these models serve to compare their model misspecification errors. 
% Note that in this subsection we have focused on the evaluation of
% \eqref{eq:eval-key}, but
{A Parametric-OIC for general risk functions $u$ can also be established by combining techniques developed in this section with those 
% techniques described 
% similarly as 
in \Cref{sec:general-risk}.} 

%in these models. 

%In Appendix, we show how to estimate the misspecification error of ETO in the contextual optimization setting. So that we can evaluate the model misspecification error in the downstream optimization tasks.

\subsection{Parametric Contextual Optimization}\label{sec:context}
%Finally, we extend our model formulation to include exogenous covariates
%that aims to reduce uncertainty .
% In this section, 
% we % generalize
% extend
% our results in Section~\ref{sec:main} to
In 
contextual
stochastic optimization problems, % where
% In the contextual optimization setup, 
the distribution of $\xi$ % depends on some covariates
is a function of a covariate
$z \in
\R^{D_{z}}$~\citep{bertsimas2020predictive,elmachtoub2022smart}.
The ground-truth distribution $\P_{\xi|z}^*$ is unknown; instead, the
decision maker only has
data $\Dscr_n= \{(z_i, \xi_i)\}_{i = 1}^n$ consisting of iid samples from the joint distribution 
$\P_{(x,z)}^*$.  The decision maker observes the covariate $Z = z$  %and the decision maker would oerve the concurrent $z$ before making decisions $x$. 
before making the decision. 
%each decision-making instance, we can 
%Denote the true conditional distribution of $\xi$ be 
%Let $\P^*_{\xi|z}$ denote the true conditional. 
In this setting, we parameterize 
the decision $x^*(\theta, z)$ as  a function of the parameter $\theta \in
\Theta$ and the % associated 
covariate $z$, and estimate $\hat\theta$ using $\Dscr_n$. The goal is to
estimate $A_{con} :=
\E_{\Dscr_n}\E_{\P_z^*}\E_{\P_{\xi|z}^*}[h(x^*(\hat\theta,z);\xi)]$ for
decision $x^*(\hat\theta,z)$.  
 % and the new covariate $z$
%Here classes of policy we can evaluate include the integrated-estimation-optimization and estimate-then-optimize procedure described in \cite{hu2022fast,elmachtoub2023estimatethenoptimize} as well as the regularized version. 
% In the contextual stochastic optimization problem, 
% With all model assumptions of $h, x$ the same as those used for previous non-contextual optimization problems,
%where the derivative conditions in $\paranx$ in \Cref{asp:represent} there are changed with $\nabla_{\theta}x^*(\theta,z), \nabla_{\theta\theta}^2 x^*(\theta,z)$. 
%However, now the underlying distribution $\P_{\xi|z}^*$ of the random variable $\xi$ depends on the covariates $z$. 
%with $\P_{\xi|z}^*$ being the ground-truth data generating marginal distribution of $z$.
%And we want to evaluate the decision performance $A_{con}$ accurately with the considered decision estimated from data being $x^*(\hat\theta, z)$. 
% each
\Cref{ex:ddo-c} is the extension of
% calibration procedure in 
\Cref{ex:ddo} % can be generalized as:
for the contextual optimization. 
\begin{definition}[General Data-Driven Contextual Optimization Procedures]
$\mbox{}$\label{ex:ddo-c}
\begin{enumerate}[(a),leftmargin=*]
\item 
  % We parameterize 
  The
  distribution of $\xi$ conditioned on the covariate $z$ is assumed to belong to the family of  
  % The distribution  % uncertainty
  % of
  % $\xi$ conditioned on the covariate $z$ % via a
  % belongs a family of   
  conditional probability distributions 
  $% \P \in 
  \mathcal{\P}_{\Theta|z}:=\{\P_{\theta|z}| \theta \in \Theta\}$, % (conditioned on the covariate $z$) 
  and 
  $x^*(\theta, z)\in\argmin_{x} \E_{\P_{\theta|z}}[h(x;\xi)]$.  
  \begin{enumerate}[(1),leftmargin=*]
  \item \textit{ETO}: $\hat{\theta}$ is estimated % from some
    using a 
    statistical approach that only uses the  data $\Dscr_n$,
    e.g., maximum likelihood, i.e., $\hat\theta \in \argmax_{\theta \in \Theta} 
    \frac{1}{n}\sum_{i = 1}^n \ln p_{\theta}(\xi_i \mid z_i)$.
  \item \textit{IEO}: $\hat{\theta} \in \argmin_{\theta\in
      \Theta}\frac{1}{n}\sum_{i = 1}^n h(x^*(\theta, 
    z_i);\xi_i)$. 
  \end{enumerate}
\item $x^*(\theta, z)$ is any parameterized family of decision
  functions. Let $\hat{\P}_n := \frac{1}{n}\sum_{i = 1}^n \delta_{(z_i,
    \xi_i)}$ denote the empirical joint distribution. 
  \begin{enumerate}[(1),leftmargin=*]
  \item \textit{E2E}: $\hat{\theta} \in \argmin_{\theta}\E_{(z,\xi)
      \sim \hat{\P}_n}[h(x^*(\theta, z);\xi)]$;
  \item \textit{R-E2E}: $\hat{\theta} \in
    \argmin_{\theta}\big\{\E_{(z,\xi) \sim \hat{\P}_n}[h(x^*(\theta,
    z);\xi)  
    + \lambda R(x^*(\theta, z))]\big\}$ for some regularizer $R(\cdot)$; 
  \item \textit{DR-E2E}: $\hat{\theta} \in \argmin_{\theta}\max_{d(\P, \hat{\P}_n) \leq
      \epsilon}\E_{(z,\xi) \sim \P}[h(x^*(\theta, z);\xi)]$ for some
    metric $d$. 
  \end{enumerate}
\end{enumerate}
\end{definition}
% We illustrate the above definitions with a portfolio optimization example:
% \begin{example}[Contextual Portfolio Optimization]
%     Continuing from \Cref{ex:portfolio}. % Let
%     % Then we can take 
%     Let $\P_{\theta|z} = N(\theta z, I_{d_x \times d_x})$ in
%     \Cref{ex:ddo-c}$(a)$ or $x^*(\theta,z) = \theta z$ in
%     \Cref{ex:ddo-c}$(b)$ when $\theta \in \R^{d_x \times d_z}$, where $z$
%     denotes some observable market factors.
%     % information. 
%     In \Cref{ex:ddo-c}$(b)$, we can also set $x^*(\theta,z)$ as a neural
%     network where $\theta$ is the collection of parameters of the neural
%     network. 
% \end{example}
The following result % shows the use of our generalization of OIC, which
% we call Context-OIC, to evaluate the decision
% performance $A_{con}$ in the contextual setting.
extends OIC to the contextual setting. 
% \begin{definition}[Contextual Optimizer's Information Criterion (Context-OIC)]\label{def:OIC-con}
% Context-OIC $\hat{A}_{con}$ is defined as follows:
% \begin{equation}\label{eq:context-general}
%     \hat{A}_{con}= \frac{1}{n}\sum_{i = 1}^n h(x^*(\hat{\theta},z_i);\xi_i) \underbrace{- \frac{1}{n^2}\sum_{i = 1}^n \nabla_{\theta} h(x^*(\hat{\theta},z_i);\xi_i)^{\top} \widehat{IF}(z_i, \xi_i)}_{\hat A_{con-c}}.
% \end{equation}
% \end{definition}
\begin{theorem}[Contextual Optimizer's Information Criterion (Context-OIC)]\label{coro:context}
% We define the Context-OIC 
% $\hat{A}$ as:
% We
  Define
  Context-OIC
  \begin{equation}\label{eq:context-general}
    \hat{A}_{con}:= \frac{1}{n}\sum_{i = 1}^n
    h(x^*(\hat{\theta},z_i);\xi_i) \underbrace{- \frac{1}{n^2}\sum_{i =
      1}^n \nabla_{\theta} h(x^*(\hat{\theta},z_i);\xi_i)^{\top}
    \widehat{\IFx}(z_i, \xi_i)}_{\hat A_{con-c}}.
\end{equation}
Assume that the conditions in \Cref{thm:main} hold,  and $(\IFx(\xi),
\widehat{\IFx}(\xi))$ and $(\paranx, 
h(\paranx;\xi))$ are replaced with  
$(\IFx(z, \xi), \widehat{\IFx}(z, \xi))$ and % with
$(x^*(\theta, z), h(x^*(\theta, z);\xi))$, respectively. 
Then $\E[\hat{A}_{con}] = A_{con} + o\Para{1/n}$ and $\E[\hat A_{con-c}] =
-\E_{z, \xi}[\nabla_{\theta} h(x^*(\theta^*,z);\xi)^{\top} \IFx(z, \xi)]/n +
o(n^{-1})$. 
\end{theorem}
% The good news of this by-product compared with general contextual
% optimization tasks is when the model is well-specified, we can directly
% estimate the performance of the model under given (new observed) $z$,
% instead of the average expected decision performance. 
% \paragraph{Bias in Contextual Optimization.}  as the interplay between
% function complexity and distribution complexity 
%(FIX WRITING.. NOT FULLY UNDERSTANDABLE. AND IS THIS FOR CONTEXT-OIC OR
%ALSO APPLY TO OIC?)  
% The bias correction term in~\eqref{eq:context-general} converges to 
% Denoting the bias part $\hat A_{con-c}:=- \frac{1}{n^2}\sum_{i = 1}^n
% \nabla_{\theta} h(x^*(\hat{\theta},z_i);\xi_i)^{\top}
% IF_{\hat{\theta}}(z_i, \xi_i)$, we can also show that .  
Like % \Cref{thm:main},
$\hat{A}_c$, 
the form of bias estimate $\hat A_{con-c}$ shows the source of the  bias,
and provides suggestions for  % a better calibration
improved 
design. For example, in ETO, if the parametric distribution
space is enlarged % in
such that the % associated
direction of the associated  influence function
$\IFx(z,\xi)$ is nearly orthogonal to the gradient
$\nabla_{\theta}h(x^*(\theta,z);\xi)$ of the downstream optimization 
gradient, % then
% we would incur a
% smaller 
bias is likely to be smaller. This gives us insight into how to design % good
improved
ETO methods for
contextual optimization. 
%in the contextual stochastic optimization problem, 

For each procedure in \Cref{ex:ddo-c}, % the detailed
an explicit
bias formula can be derived in a manner similar to the results in 
\Cref{sec:main} by replacing $x^*(\theta)$ % there
with $x^*(\theta,z)$. For
example, we can obtain the bias characterization in
\Cref{ex:ddo-c}$(b)(1)$ as follows.
\begin{example}[Context-OIC for IEO / E2E]
  Suppose the % same
  conditions in \Cref{coro:oic-procedure}$(ii)$ hold for
  $x^*(\theta,z)$, i.e., {Assumptions~\ref{asp:nonsmooth},~\ref{asp:theta-star}$(ii)$,~\ref{asp:moment},~\ref{asp:theta-hat}$(ii)$ 
  and \Cref{asp:ierm-additional} hold when $\big(h(\paranx;\xi),
  x^*(\theta)\big)$ % there are
  are replaced with $\big(h(x^*(\theta,z);\xi),
  x^*(\theta,z)\big)$.} Then when $\Xscr = \R^{D_x}$,  
  % the optimistic bias is:
  % imposing the same conditions except . Similar to \Cref{coro:ierm}, the
  % corresponding performance estimator $\hat{A}_{con}$ would be: 
  $\hat{A}_{con-c} = 
  % \frac{1}{n}\sum_{i = 1}^n h(x^*(\hat{\theta}, z_i);\xi_i) +
  \Tr{\hat{I}_{h,z}(\hat{\theta})^{-1} \hat{J}_{h,z}(\hat{\theta})}/n$, where 
  % the term $\text{Tr}\Paran{\hat{I}_{h,z}(\hat{\theta})^{-1} \hat{J}_{h,z}(\hat{\theta})}$ characterizes the optimistic bias with:
  % \begin{equation}\label{eq:ieo-c}
  $\hat{I}_{h,z}(\hat{\theta}) = \frac{1}{n}\sum_{i = 1}^n \hessianp h(x^*(\hat{\theta},z_i);\xi_i), \hat{J}_{h,z}(\hat{\theta}) = \frac{1}{n}\sum_{i = 1}^n \nabla_{\theta}h(x^*(\hat{\theta},z_i);\xi_i) \nabla_{\theta}h(x^*(\hat{\theta},z_i);\xi_i)^{\top}.$
\end{example}
%\end{equation}
% In other cases of \Cref{ex:ddo-c}, we can obtain similar bias characterization forms under the corresponding conditions as \Cref{sec:usercase} by replacing all $x^*(\theta)$ there with $x^*(\theta, z)$. 
%and also applies to general OIC without contexts.
%\paragraph{Evaluate Model Misspecification Error.}
% Furthermore, we can extend
The analysis for 
model misspecification error % analysis
in 
% from
\Cref{sec:POIC} % for
can be extended to 
the decision rules in \Cref{ex:ddo-c}$(a)$. % Similar to~
Let
\[
  \Escr(x^*(\theta, z);\Pscr_{\Theta|z}) := \E_{\Dscr_n}\E_{{z}}
  \E_{\P_{\xi|{z}}^*}[h(x^*(\hat\theta,z);\xi)] -
  \E_{\Dscr_n}\E_{z}\E_{\xi \sim
    \P_{\theta^*|{z}}}[h(x^*(\hat\theta,z);\xi)]
\]
denote the misspecification error for the decision rules in
\Cref{ex:ddo-c}(a).
% \gi{On can establish a result analogous to \Cref{prop:pf-fit} % , if we 
% % %parameterize the conditional distribution by $\P_{\theta| z}$ and 
% % When the calibration procedure in \Cref{ex:ddo-c}$(a)$ is used, we can
% % obtain a different performance estimator
% where the cost function is
% evaluated under the parameterized distribution $\P_{\hat\theta|z}$.}  
%However, we would incur the ``price" of model misspecification as in the non-contextualized case.
% Together with Context-OIC, we can evaluate the distribution
% misspecification error for downstream contextual optimization as
% follows:
The following result is analogous to \Cref{coro:eto-misspecification}. 
% given by the term:
% % 
%through $\hat{B}_{con} = \hat{A}_{con} - \frac{1}{n}\sum_{i = 1}^n A_{z_i}$
%More formally, the result is given by:
\begin{proposition}[Contextual Misspecification Error]\label{prop:contextual-misspecified}
Suppose
Assumptions~\ref{asp:nonsmooth},~\ref{asp:theta-star},~\ref{asp:moment},~\ref{asp:theta-hat},~\ref{asp:pf-fit-optimal}
hold % , but all
% with
when 
$\big(h(\paranx;\xi), x^*(\theta)\big)$ % here are 
is replaced with $\big(h(x^*(\theta,z);\xi), x^*(\theta,z)\big)$, and $\E[\hat\theta] =
\theta^* + o(1/n)$. Define 
\[
  \hat B_{con} := \hat A_{con} - \frac{1}{n}\sum_{i = 1}^n \hat A_{z_i},
\]
where $\hat{A}_z := \E_{\P_{\hat{\theta}|z}}[h(x^*(\hat{\theta},z);\xi)] +
\frac{1}{2n}\Tr{\hat{I}_{h,z}(\hat{\theta})\hat{\Psi}(\hat{\theta})}$ with $\hat{I}_{h,z}(\theta) = \frac{1}{n}\sum_{i = 1}^n
\nabla_{\theta\theta}^2 h(x^*(\theta,z);\xi_i)$ and $\hat{\Psi}(\theta) =
\frac{1}{n}\sum_{i = 1}^n \widehat{\IFx}(z_i, \xi_i)
\widehat{\IFx}(z_i, \xi_i)^{\top}$.
Then
\[
  \E[\hat B_{con}] = \Escr(x^*(\theta, z);\Pscr_{\Theta|z}) +
  o(n^{-1}).
\] 
% For the decision rules in \Cref{ex:ddo-c}$(a)$, denote the
% misspecification error as 
% If $\E[\hat\theta] = \theta^* + o(1/n)$, then for, we have: 
%     where $\hat{A}_{con}$ is defined in~\eqref{eq:context-general} and 
%     % \begin{equation}\label{eq:context-well}
%    $\hat{A}_z := \E_{\P_{\hat{\theta}|z}}[h(x^*(\hat{\theta},z);\xi)] + \frac{1}{2n}\text{Tr}[\hat{I}_{h,z}(\hat{\theta})\hat{\Psi}(\hat{\theta})]$ with 
   %- \frac{\nabla_{\theta}\Para{\E_{\P_{\hat{\theta}}}[h(x^*(\hat{\theta};z);\xi)]}^{\top} C(\P^*, \theta^*)}{n}
    % \end{equation}
%where 
\end{proposition}
% This provides an approach to evaluate distribution misspecification errors of parametric distributions on the downstream contextual stochastic optimization. 

To establish this result, we first notice that for each $z$, $\hat A_{z}$
satisfies $\E[\hat{A}_z] = \E_{\Dscr_n}\E_{\xi \sim
  \P_{\theta^*|z}}[h(x^*(\theta,z);\xi)] + o\Para{1/n}$ using the same
arguments as those used to establish \Cref{prop:pf-fit}. Then we apply the same proof as
\Cref{coro:eto-misspecification}.

\subsection{Nonparametric Contextual Optimization}\label{subsec:np-decision}
\wty{In this section, we discuss nonparametric contextual optimization
procedures. %  While we have focused on formulations from
% Definitions~\ref{ex:ddo} and~\ref{ex:ddo-c} that rely on parametric
% decision rules, our debiasing procedures and % formulas
% expressions
% can be % considerably
% used in nonparametric % optimization problems.
% procedures as well.
} 
% Here, we discuss
% nonparametric optimization procedures more broadly in contextual
% optimization due to their usefulness especially in this latter scenario. 
% Following the setup in \Cref{sec:context}, we denote
\wty{% Let
% $x(z)$ % as the
% denote the 
% decision % depending on
% as a function of the covariate~$z$.
  In particular, 
we consider the class of nonparametric optimization procedures % with $\hat x(z)$ obtained from the elegant formulation
% developed
in \cite{bertsimas2020predictive}.}
\wty{\begin{definition}[Data-Driven Nonparametric Optimization Procedures]\label{defn:np-opt}
  % Consider $\hat x(z)$ obtained through:
  The empirical decision rule
  \begin{equation}\label{eq:np-optimizer}
    \hat{x}(z) \in \argmin_{x \in \Xscr} \sum_{i \in [n]} w_{n,i}(z)h(x;\xi_i),    
  \end{equation}
  where $\{w_{n,i}(z)\}_{i \in [n]}$ are weights determined by $\Dscr_n$ and $z$. 
\end{definition}}
\wty{Formulation \eqref{eq:np-optimizer} includes as special cases
%   several of the most  nonparametric
% optimization procedures such as
k-Nearest Neighbors (kNN)
\citep{bertsimas2020predictive}, Nadaraya-Watson kernel estimators
\citep{ban2019big} and random forests \citep{kallus2023stochastic}. % as
% special cases.
Like in previous settings, 
% Even for the formulation in~\eqref{eq:np-optimizer}, 
% such a formulation, 
the empirical performance estimate $\hat A_o = \frac{1}{n}\sum_{i
  = 1}^n h(\hat x(z_i);\xi_i)$ % also
suffers from % a similar problem of
an optimistic bias when $h(x;\xi)$ is convex in $x$.  For these nonparametric optimization procedures, the influence function expansion and bias characterization follow % derived
from Lemma 1 in \cite{iyengar2024cross}. Specifically, for each covariate
$z$ (see~\Cref{asp:theta-asymptotics}) there exists a sequence of influence
functions % (at the covariate $z$)
$\{\IFx_{n,z}((\tilde z,\tilde\xi))\}_{n =
  1}^{\infty}$ and fixed functions $\{x_n(z)\}_{n = 1}^{\infty}$ (the % their
dependence on $n$ arises % from their dependence on
because of the $n$-dependence of the weights
$w_{n,i}(z)$)) such that: 
\[
  \hat x(z) - x_n(z) = \frac{1}{n}\sum_{i = 1}^n \IFx_{n,z}((z_i, \xi_i)) +
  o_p(n^{-\gamma_v}),
\]
where $\gamma_v \in (0, \frac{1}{2}]$. 
% This way, we can obtain the bias characterization same as \Cref{thm:main}. 
%Building on these results, 
%for a number of nonparametric optimization procedures. 
% For some nonparametric optimization procedures there, 
 % can explicitly derive their
The expressions for the influence function formulas, and subsequently,
expressions for the bias (see, e.g., Theorems~\ref{thm:main} and
\ref{coro:context}) can be explicitly characterized for some
nonparametric procedures.} 
\wty{\begin{example}[Nonparametric-OIC for kNN] % Optimization procedure]
    % Under mild conditions, when $\hat z(\cdot)$ refers to 
  The kNN optimization procedure belongs to the class defined in~\eqref{eq:np-optimizer}
  % corresponds to
  with weights
  $w_{n,i}(z) = \mathbf{1}_{\{i \in
    \Nscr_{\Dscr_n,z}(k_n)\}}$, for $\Nscr_{\Dscr_n,z}(k) = \{i \in
  [n]~|~z_i~\text{is a kNN of}~z\}$. % with a hyperparameter $k_n$
  % in~\eqref{eq:np-optimizer}, define:
  When $\Xscr = \R^{D_x}$, define
  \begin{eqnarray}
    \hat A_{con} & := & \frac{1}{n}\sum_{i = 1}^n h(\hat x(z);\xi_i)
                        \notag \\
    && \mbox{} -
    \frac{1}{n^2}\text{Tr}\Big[\nabla_{xx}^2\E_{\P^*}[w_{n,i}(z) h(\hat
    x(z);\xi)]\Big(\sum_{i = 1}^n w_{n,i}(z_i) \nabla_x h(\hat
    x(z_i);\xi_i)\nabla_x h(\hat
       x(z_i);\xi_i)^{\top}\Big)\Big]. \label{eq:np-oic-knn}
  \end{eqnarray}
  Then $\E[\hat A_{con}] = \E_{\Dscr_n}\E_z\E_{\P_{\xi|z}^*}[h(\hat x(z);\xi)]
  + o(n^{-2\gamma_v})$. When $h(x;\xi)$ is twice differentiable with
  respect to $x$, $\nabla_{xx}^2\E_{\P^*}[w_{n,i}(x) h(\hat x(z);\xi)]$
  in~\eqref{eq:np-oic-knn} can be replaced by $\frac{1}{n}\sum_{i = 1}^n
  w_{n,i}(z_i) \nabla_{xx}^2 h(\hat x(z_i);\xi_i)$.  
\end{example}}
% However, 
\wty{We do not include these optimization procedures in our general
framework defined in Definitions~\ref{ex:ddo} and~\ref{ex:ddo-c},
because the order of the bias, $O(n^{-2\gamma_v})$ % in general nonparametric
% models, i.e.,
is specific to each procedure, and 
% , differs and 
can be much larger than that for parametric
optimization procedures. % when $\gamma_v$ is close to zero. 
% Besides
Additionally, the
comparison between direct evaluation approaches for nonparametric optimization, not even bias-corrected, and cross-validation methods
is complicated, as hinted in \cite{iyengar2024cross}. We thus leave the full
investigation to future work.}  
%In terms of the difference with Theorem~\ref{thm:main} and~\ref{coro:main-moment}, There, the main difference is that the main bias order is different. $\Theta(n^{-2\gamma})$ with $\gamma \in (0, \frac{1}{2})$; Furthermore, it is challenging to derive formulations of some nonparametric models. }

\section{Comparison with Existing Methods}\label{sec:challenge}
In this section, we 
% elaborate the challenges in the existing methods to
% evaluate \eqref{primary} (or~\eqref{eq:eval-key} specifically) % rom
% % there we
% and then 
describe the rationale and benefits of OIC in more detail by elaborating the challenges in the existing methods to evaluate \eqref{primary} and \eqref{eq:eval-key}.

\subsection{Dedicated Bias Correction Expressions} 
% To explicitly characterize and remove the bias, the other stream of evaluation methods is problem-dependent, yielding specific bias formulas derived for some special problem instances. In machine learning problems, these include 
% Famous
% metrics including evaluating \eqref{eq:eval-key} to identify the optimal configurations In operational problems, It is worth noting that these criteria only correct the bias for 
Some methods aim to devise explicit bias correction % formulas
expressions
for certain
estimation tasks to assist in model selection. Prominent examples include Akaike Information Criterion (AIC)~\citep{akaike1974new} in maximum likelihood ratio, Takeuchi Information Criterion (TIC)~\citep{takeuchi1976distribution,murata1994network,thomas2020interplay} \tyw{and Generalized Information Criterion (GIC)~\citep{konishi1996generalised}.} OIC % , in a sense,x
can be % considered
viewed 
as a
substantial
generalization of these approaches. 

First, the classical AIC designed primarily for model selection is a special case of OIC. % % by using the
% obtained by setting the objective to the 
% log-likelihood % as the objective 
% function, and use E2E as
% the calibration procedure. 
% protocol:
\begin{example}[AIC]\label{ex:aic}
Consider the parametric family of densities $\Pscr = \{p_{\theta}(\cdot) \mid
\theta \in \Theta\}$. \wty{Suppose $\P^*$ is
  well-specified, i.e., the corresponding p.d.f. $p^* \in \Pscr$.}
Then AIC is OIC applied to E2E (see \Cref{ex:ddo}(b)(1)) using the
function $h(x^*(\theta);\xi) = -\ln p_{\theta}(\xi)$, and $A_c =
D_{\theta}/n$, where $D_{\theta}$ denotes the dimension of $\theta$.
% where $p_{\theta}(\cdot)$
% is the probability density function (p.d.f.) of the distribution
% parameterized by $\theta \in \Theta$. \wty{We assume that $\P^*$ is
%   well-specified, meaning that its p.d.f. belongs to
%   $\{p_{\theta}(\cdot)|\theta \in \Theta\}$}. Under this specification,
% AIC is the OIC for E2E.  
% with $\hat{\theta}_k \in \argmin_{\theta \in
% \Theta_k}\E_{\hat{\P}_n}[h(\paranx;\xi)]$ for each distribution
% parameter class $\Theta_k$. as the downstream optimization protocol  
\end{example}
% \wty{Since $I_h(\theta^*) = J_h(\theta^*)$ for the AIC setting,  the first-order expected bias in Corollary~\ref{coro:oic-procedure}$(ii)$ 
% %, and the associated first-order expected bias $A_{c} = \frac{1}{n}\text{Tr}[I_h(\theta^*)^{-1} J_h(\theta^*)]$ from \Cref{thm:main} with $I_h(\theta) = \hessianp\E_{\P^*}[ h(\paranx;\xi)]$ and $J_h(\theta) = \E_{\P^*}[\gradp h(\paranx;\xi)\gradp h(\paranx;\xi)^{\top}]$, 
% recovers 
% % existing statistical criteria as special cases. 
% %$p_{\theta}(\cdot), \theta(\cdot)$ represent a subset of functions in the space which has been shown in previous literature \citep{thomas2020interplay}. Plugging in different cost function and refined assumptions, we show this estimator can be helpful.
% %\begin{example}[AIC]
% % In particular, for AIC, we have 
% %Continued from \Cref{ex:aic}, we have: 
% % AIC as a special case with 
% $A_c = D_{\theta}/n$, \wty{where $D_{\theta}$ denotes the dimension of $\theta$}. 
% %\end{example}
% % For \Cref{ex:portfolio}, $\hat{A}_c$ for all settings other than ETO can be computed using \Cref{coro:ierm}. 
% %(AS A GOOD WRITING PRACTICE, IF YOU'RE GOING TO TALK ABOUT EXAMPLE 1
% HERE, YOU SHOULD TALK ABOUT EXAMPLE 1 FOR ALL OTHER COROLLARIES SO THAT
% EXAMPLE 1 IS REGARDED AS A RUNNING EXAMPLE. THIS MAKES YOUR DISCUSSION
% MORE SYSTEMATIC AND ORGANIZED.. ACTUALLY, I WOULD SAY WE DON'T NEED TO
% BRING THIS EXAMPLE UP HERE).

\wty{In contrast to methods like AIC that focuses on model selection,  
% Compared to
% model-focused evaluation methods such as AIC, 
OIC % s goal of comparing
can be applied to a much larger set of 
% decision rules allows it to be applied to a broader range of cases
procedures, e.g.,  \Cref{ex:ddo}$(b)$, instead of those  % -- than decision rules 
based on distribution estimation, e.g., \Cref{ex:ddo}$(a)$.  % when applying
% AIC. 
Moreover, even for procedures in \Cref{ex:ddo}$(a)$, OIC % provides
is a more effective decision selection % compared
criterion as compared to AIC. While AIC may select 
% fitted
a distribution with % better
higher
likelihood, there is no guarantee that the decision rule % based on
% this distribution model often results in inferior decision
% performance.
corresponding to the best-fit distribution leads to the best performance. 
% n this case,
Instead, 
OIC % provides a better decision rule by
directly corrects the bias in the objective value. In~\Cref{fig:main_nv}$(d)$ of \Cref{subsec:numerical-newsvendor}, % we
% advantages of
we show that OIC outperforms AIC in selecting a better ETO solution under the exponential or normal distribution class.}

When $\P^*$ is not well-specified for the general loss,
the bias correction term $A_c$ (or $\hat A_c$) for the unconstrained E2E recovers the 
TIC and GIC.
\begin{example}[TIC]
For the general loss $h(\paranx;\xi)$ and $\Xscr = \R^{D_x}$, TIC is OIC applied to IEO / E2E with $\hat A_c = \text{Tr}[\hat I(\hat\theta)^{-1}
J(\hat\theta)]/n$ where $\hat I(\hat\theta) = \frac{1}{n}\sum_{i = 1}^n
\nabla_{\theta\theta}^2 h(\datax;\xi)$ and $J(\hat\theta) =
\frac{1}{n}\sum_{i = 1}^n \nabla_{\theta} h(\datax;\xi_i)\nabla_{\theta}
h(\datax;\xi_i)^{\top}$.
\end{example}
To the best of our knowledge, except for the unconstrained IEO and E2E settings where we overlap with the existing TIC, all of our OIC results are new. Note that applying TIC to the other cases incorrectly estimates bias. % estimates
This is because  for optimization procedures other than the unconstrained IEO/E2E, establishing bias correction results requires techniques based on a different Taylor expansion. 
Specifically, the key steps in the proof of TIC rely on
\Cref{asp:theta-star}(ii) ($\Xscr = \R^{D_x}$) to % perform
do a Taylor expansion around
$x^*(\hat\theta)$ and $x^*(\theta^*)$, ensuring that the first-order term vanishes, and hence, the bias can be derived 
using the second-order term. However, \Cref{asp:theta-star}(ii) does not hold for many optimization procedures, e.g., ETO when $\P^*$ is not well-specified, and  general regularized or constrained E2E settings. Consequently, computing the bias for % general
these procedures requires novel expansions centered at $x^*(\theta^*)$, as we describe in the proof sketch of Section~\ref{sec:main}.

\tyw{\begin{example}[GIC]
     For $\Xscr = \R^{D_x}$, GIC is OIC applied to the log likelihood cost function $h(\paranx;\xi) = -\ln p_{\theta}(\xi)$ with $A_c = \E_{\P}[\nabla_{\theta} \ln p_{\theta^*}(\xi)^{\top}\IFx(\xi)]/n$.
\end{example}
Compared with OIC, the main difference is that GIC corrects the evaluation bias in a fitted density model instead of the downstream objective
value of a data-driven decision $x^*(\hat\theta)$. Thus, OIC extends GIC
to general losses and decision rules, including constrained decision rules
and spaces, contextual optimization, and general risk
transformations. Technically, GIC assumes second-order differentiability
of $T(\cdot)$. In contrast, OIC does not require this, as it is derived
through the first-order asymptotic linear expansion \eqref{eq:theta-if}. The expansion~\eqref{eq:theta-if} suffices since we identify the bias through the interaction between the first-order influence function and the gradient, while the remaining higher-order terms can be shown to be
negligible. % Equation
The expansion~\eqref{eq:theta-if} is satisfied by a wider range of
calibration maps, e.g., constrained problems and distributionally robust optimization problems, where $T(\cdot)$ may involve active constraints or projections onto tangent spaces. Relaxing the second-order
differentiability assumptions broadens the class of influence functions
and their estimators that can be incorporated into the OIC
formula. Moreover, OIC also accommodates certain nonsmooth cost functions
through smooth-approximation arguments. 
}

Besides AIC, TIC and GIC, other examples of dedicated bias correction formulas include covariance 
penalty~\citep{efron1986biased} in linear regression, and regression via deep neural networks \citep{murata1994network}. 
% More
% recently, \cite{koh2017understanding} examined various 
% empirical optimization approaches from the perspective of $M$-estimation,
% shedding light on how individual points can affect overall model
% prediction 
% performance. 
\cite{novak2018sensitivity} empirically established a link
between the bias and the norm of the input-output Jacobian of a
network. 
\cite{siegel2021profit, siegel2023data} derived bias formulas
% under
for 
well-specified parametric models and empirical optimization for the
newsvendor problem. These methods % share a similar spirit as
are similar in spirit to 
OIC in that they provide an explicit bias correction formula; however, they % only
focus on very
specific  cost functions and decision
classes~\citep{anderson2004model}. 
%, e.g., AIC and TIC as we see in \Cref{subsec:cv}.

% in fact, we % where, as will be seen, can be 
% show that several of these approaches are 
% % encompass some of them as 
% special cases.
% Consequently, they are unable to incorporate general procedures as
% discussed in the context of our problem setups. 
% \end{new} 

\subsection{Cross Validation (CV)}
% The two main variants of CV:
There are two main variants of CV:
$K$-Fold CV and LOOCV. % , operate as follows
In $K$-Fold CV, % where typically 
$K$ is a small number that is fixed % compared with
as a function of 
$n$. The dataset is divided into $K$ subsets. For each of the $K$ subsets,
the procedure % epeatedly
% uses a different
uses the remaining 
$K-1$ subsets to train the decision, and evaluates the performance % using
% the corresponding remaining
on the given 
subset. The final output is the average of % all these
the $K$ evaluations. LOOCV % operates similarly but with
is equivalent to $n$-Fold CV,
% $K$ taken to be $n$, 
i.e., each time the procedure uses all but one observations to train the
decision and % evaluates
the performance is evaluated using the left-out
observation. These methods are % advantageousl y
``model-free'', in the sense of not requiring specific structural knowledge
regarding  the decision % rule
class 
or the target optimization problem, and are, therefore, % thus making them
very
flexible. % to use. 
However, a limitation of $K$-Fold CV is that it introduces a pessimistic
bias of order $\Theta(1/n)$ in estimating
$A$~\citep{fushiki2011estimation}. This bias arises % due to the
because each iteration uses a
smaller amount of training data, which is also implied from \Cref{thm:main}: % implies the following bias property of $K$-Fold CV:
% establish
% imply
% the following result 
% of $K$-Fold CV.
\begin{proposition}[Bias of $K$-Fold CV]\label{coro:K-Fold}
  % Under the same conditions as in \Cref{thm:main}, 
  Suppose Assumptions~\ref{asp:nonsmooth} and~\ref{asp:moment}, and~\eqref{eq:theta-if} hold. Then the expected value of the
  bias of $K$-Fold CV $\hat A_{kcv}$ is given by
  \[
    \E[\hat A_{kcv}] - A = \frac{B}{(K - 1)n} + o\Para{\frac{1}{n}}
  \]
  for any $K$ with $\frac{n}{K}\in \mathbb{Z}$, where $B = \Tr{I_h(\theta^*)\Psi(\theta^*)}/2 +
  (\nabla_{\theta}\E_{\P^*}[h(\bestx;\xi)])^{\top}n(\E[\hat\theta] -
  \theta^*)$. If $\E[\hat\theta] - \theta^* = O(1/n)$, then $\E[\hat
  A_{kcv}]- A = O\Para{\frac{B}{(K-1)n}}$. 
\end{proposition}
% \hl{I don't understand the qualification -- doesn't it follow from the
%   expression of $\hat{A}_{kcv}$.} \wty{TW: This gives the bias of k-cv.}
%$\E[\hat A_{kcv}] - A =
%\E_{\Dscr_{(K-1)n/K}}\E_{\P^*}[h(x^*(T(\Dscr_{(K-1)n/K}));\xi)] -
%\E_{\P^*}[h(\bestx;\xi)]$. That is,  
% This result is established by observing that
Note that 
the bias of $K$-Fold CV is
equal to the \emph{true optimality gap} when % we estimate
the decision is estimated using $(K-1) n / K$ observations. 
% \hl{How is ``optimality gap'' defined? This is the first time it is being
%   used.} \wty{TW: it is already defined after Theorem 1.}
Then \Cref{coro:K-Fold} follows
% the same idea in
the techniques used to approximate the true optimality gap in the proof of
\Cref{thm:main}. % From this corollary, we see that
This result shows that, unlike OIC, the $K$-Fold CV
incurs a pessimistic bias of an order $O(1/n)$ for a fixed $K$. \cite{fushiki2011estimation} proposed % to correct this bias
% \textit{bias-corrected $K$-Fold CV}
% estimator as
using 
a % linear
convex
combination of $K$-Fold CV and the empirical average $\hat A_o$
to correct this bias. % , to mitigate the $\Theta(1/n)$ bias.
However, the optimal weight of this % linear
convex
combination % still needs to be
% determined based on the
depends on the 
estimation procedure for $\hat\theta$, and needs to be empirically
estimated. 

On the
other hand, if we set $K = n$ in \Cref{coro:K-Fold}, LOOCV removes the $\Theta(1/n)$ bias,
even for high-dimensional
problems~\citep{beirami2017optimal,wang2020debiasing}. 
We show that the asymptotic performance of OIC % performs
% similar to
and 
LOOCV % in terms of
% asymptotic behavior.
are similar, indicating \tyw{OIC} is a good decision selection criterion:
% \end{new} (where each $\hat{\theta}_{-i} = T(\Dscr_n \backslash \{\xi_i\})$) 
\begin{corollary}[Similarity between LOOCV and OIC]\label{prop:loocv}
Suppose Assumptions~\ref{asp:nonsmooth},~\ref{asp:theta-star},~\ref{asp:moment},~\ref{asp:theta-hat} and~\ref{asp:additional-regular} hold. Define  $\hat{A}_{loocv}:=
  \frac{1}{n}\sum_{i = 1}^n 
  h(x^*(\hat{\theta}_{-i});\xi_i)$, where $\hat{\theta}_{-i} =
  T(\Dscr_n \backslash \{\xi_i\})$. 
  % Then for  under \Cref{ex:ddo}, the
  % performance of LOOCV $\hat{A}_{loocv} := \frac{1}{n}\sum_{i = 1}^n
  % h(x^*(\hat{\theta}_{-i});\xi_i)$ is similar to OIC $\hat A$ in the
  % sense that
  Then $\hat{A}_{loocv} - \hat{A} = o_p\Para{\frac{1}{n}}$. In
  contrast, for the empirical objective $\hat A_o$, we have
  $\hat A_o  
  - \hat A_{loocv} = O_p\Para{\frac{1}{n}}$.   
\end{corollary}
\Cref{prop:loocv} generalizes the classical result in
\cite{stone1977asymptotic} on the equivalence to LOOCV from AIC to OIC. However, % comes
% with the drawback of being
LOOCV comes with a high 
computation % expensive due to the need to perform
cost of solving 
$n$ % additional
optimization problems. 

Furthermore, existing CV methods only focus on
assessing \eqref{eq:eval-key} and do not address the evaluation problem
for general $u(\cdot)$ in \eqref{primary}. Applying LOOCV to the risk setting requires additional modification as follows:
% The bias correction for
% LOOCV can be extended to  the risk % incorporated
% setting, % which % which
% % also requires the second-order derivative of $u(\cdot)$:
% is defined as follows.
\begin{corollary}[Risk-Incorporated LOOCV]\label{coro:risk-loocv}
  Suppose the % same
  conditions in \Cref{thm:risk-oic} hold. Then $\hat
  A_{u-loocv} := u(\hat A_{loocv}) - (1/(2n)) u''(\hat
  A_{loocv})\hat\sigma_{loocv}^2$, where $\hat\sigma_{loocv}^2 = \frac{1}{n}\sum_{i = 1}^n
  (h(x^*(\hat\theta^{(-i)});\xi_i) - \hat A_{loocv})^2$,  satisfies $\E[\hat A_{u-loocv}] = A_u +
  o(1/n)$.
\end{corollary}
The strength of OIC compared to LOOCV in the risk-incorporated setting is 
% inherited from
the same as 
that in the basic setting, i.e. $u(x)=x$,  % compared
% to LOOCV,
in that 
OIC does not require solving additional optimization
problems.

% used in each iteration.
% Motivated by this, 

%this approach requires additional evaluation steps (WHAT DOES ADDITIONAL
%EVALUATION STEPS MEAN? YOU MEAN THE TUNING OF THE LINEAR COMBINATION
%WEIGHT REQUIRES ADDITIONAL EVALUATION EFFORT?).  
% In comparison, by setting $K = n$, 
% which complicates the computation due to the additional nonlinearity
% arising from the curvature of $u(\cdot)$. 
% since it needs to evaluate additional terms. 
%(I THINK THE WORD "STEP" IS UNCLEAR, JUST LIKE THE WORD "MODEL" THAT YOU
%HAVE KEPT OVER-USING (YOU USED IT TO REFER TO TOO MANY THINGS; I HAVE
%REMOVED MOST OF THE WORD "MODEL" IN THE PARTS THAT I PASSED THROUGH..)
%TRY TO CHANGE IT HERE, E.G., COMPUTATION COMPLICATION, EVALUATION
%EFFORT). 
%\wty{(CAN CV APPLY TO GENERAL RISK FUNCTION? IF NOT, THIS IS ANOTHER
%DRAWBACK OF CV THAT WE NEED TO MENTION)} 
%that LOOCV is computationally expensive. 

\subsection{Approximate Leave-One-Out (ALO)}
% Since LOOCV is known for its debiased generalization performance, particularly in high-dimensional problems \citep{beirami2017optimal,wang2020debiasing}, researchers recently investigate some 
ALO methods are designed to avoid the computational bottleneck of
repeatedly solving optimization problems in LOOCV. They can be categorized
into two categories depending % on how approximate the
on how 
LOO solution is approximated depending on the specific problem structure: % from $\datax$, one based on 
Newton's step (NS) based methods~\citep{beirami2017optimal} and % he other on the 
infinitesimal jackknife (IJ) based methods~\citep{giordano2019swiss}. % Despite their computational gain
% Despite their computational gain \citep{wang2018approximate, rad2020scalable, wilson2020approximate}, these methods require access to the so-called influence function \citep{cook1982residuals} that depends on specific problem structures. 
% The ALO reduces the computational cost by 
% ~\citep{wang2018approximate, rad2020scalable, wilson2020approximate}, % these methods require access to the
% leveraging the knowledge of the 
% so-called influence function~\citep{cook1982residuals} that have to
% recomputed for each  % that depends on 
% specific problem structure. % In particular,
Despite these developments,
% In particular, 
existing ALO literature primarily focuses on standard or penalized
$M$-estimation that are prevalent in machine learning, and it is unclear if these methods can be applied to various formulations that arise in the
data-driven optimization more generally. In fact, we can show that a generalization of ALO with IJ (ALO-IJ)
\citep{giordano2019swiss} yields similar performance guarantees as OIC. 
\begin{proposition}[Bias of ALO-IJ Approaches]\label{coro:alo-bias}
    Suppose Assumptions~\ref{asp:nonsmooth} and~\ref{asp:moment}, and~\eqref{eq:theta-if} and~\eqref{eq:empirical-if} hold. Let $\hat A_{alo} =
    \frac{1}{n}\sum_{i = 1}^n h(x^*(\tilde\theta_{-i});\xi_i)$, where
    $\tilde\theta_{-i} = \hat\theta
    -\frac{1}{n}\widehat{\IFx}(\xi_i)$. Then $\E[\hat A_{alo}]
    = A + o\Para{\frac{1}{n}}$ and $\hat A_{alo} - A =~o_p\Para{\frac{1}{n}}$. 
  \end{proposition}
  
  ALO-IJ % operates % similarly
  % is a manner similar to OIC in that it 
  also subtracts the influence
  function to correct bias. % However, we should point out
  Note
  % that the
  % statistical property for ALO-IJ displayed in 
  \Cref{coro:alo-bias} is a
  % itself
  new result, since~\cite{giordano2019swiss} and the follow-up work \citep{wilson2020approximate,stephenson2020approximate} only % consider
  focus on 
  $\hat\theta$ generated from (penalized) unconstrained $M$-estimators
  while we establish the result for a much broader class of  
  % which restricts the class of influence functions derived from
    calibration functions~$T(\cdot)$. ~\Cref{coro:alo-bias} is established 
  using the first-order Taylor expansion % from our OIC result in
  and
  \Cref{coro:main-moment}.  
  % However, this expansion explicitly presents a bias formula in OIC.
  % (I DON'T UNDERSTAND WHAT ONE-STEP TRANSFORMATION MEANS). 
  Moreover, despite % their
  similarity in the form of the estimator % form
  and statistical properties, unlike OIC (see~\Cref{sec:general-risk} for
  details), ALO still suffers from the following two problems:
%   ALO does not address the evaluation problem for 
%   general~$u(\cdot)$. % , which can be handled by OIC as we will describe in
%   % Section \ref{sec:general-risk}. 
%   Furthermore, since % ALO still requires
%   % evaluating
%   $x^*(\tilde{\theta}_{-i})$ has to be computed for $\tilde\theta_{-i}$, $\forall i \in [n]$, %  as shown
%   % in Corollary \ref{coro:alo-bias},
%   ALO can be computationally more % costly
%   expensive 
%   for the ETO and IEO formulations in Definition~\ref{ex:ddo}(a) since
%   computing each $x^*(\cdot)$ requires solving an optimization problem.  

% Moreover, even if ALO can be
% applied, we will see that for some data-driven approaches, e.g., 
% estimate-then-optimize, the computation benefits are not materialized.
% % which are prevalent in the data-driven optimization area. %Furthermore, ALO approaches only 
% %Regardless of the input requirements made by the aforementioned evaluation methods to remove the bias,
% Furthermore, ALO methods attempt to approximate LOOCV instead of the
% actual bias term in the objective value evaluation. This % convoluted goal
% means that the ALO methods % , coupled with the lack of an explicit analytical formula,
% % means that they 
% may not be able to reduce the order of original bias, which
% is typically $\Theta(1/n)$, % with respect to the true objective
% % performance under
% for 
% % the
% general $u(\cdot)$ in~\eqref{primary}.

\textbf{Generality of problem settings and conditions.} % First,
% none of
The ALO literature does not 
% discuss
address the performance evaluation problem for 
general risk functions~$u$. % for performance evaluation.
% (i.e., our
% general form of $u(\cdot)$ in \Cref{sec:general-risk}) besides the
% expected decision performance.
% Estimating such a
Since $\E_{\P^*}[h(\datax;\xi)]$ is random, evaluating the performance
using a risk measure is natural; however, estimating the out-of-sample
performance 
% general risk-incorporated
% performance is natural 
% variable, but 
is nontrivial since the variance % of the sample objective
over $\Dscr_n$ becomes a significant component. % In particular, plugging
% in
Simply plugging in 
ALO or LOOCV there directly 
% i.e., estimating performance by $u\big(1/n\sum_{i = 1}^n
% h(x^*(\hat\theta_{-i});\xi_i)\big)$, % ignores this important variation that has
% been shown to be
leads to a
$\Theta\Para{1/n}$ bias (see~\Cref{coro:risk-loocv}).
% \hl{Why can't we consider a $h(x^*(\hat\theta_{-i});\xi_i)$ as a random
%   variable, and estimate the performance for the risk function using the
%   $n$ samples?} \wty{TW: Not quite understand. Then it is $\E_{\P^*}[u(\cdot)]$ instead of $u(\E_{\P^*}[\cdot])$ considered.}
% Second,
Furthermore, 
even % considering the
for the % the
expected performance, i.e., $u(x) = x$, % previous
the
ALO literature
% works
focuses on % special setups in
very specific 
unconstrained problems where standard
$M$-estimators (i.e., Chapter 5 in \cite{van2000asymptotic}) apply. One of
our main
contributions %  of % our paper
% this paper
is to provide a % more
general framework that can
be applied to a variety of data-driven stochastic optimization
formulations. % in addition to these standard $M$-estimators.
In the general constrained case (including DR-E2E and constrained decision space), 
% constrained problems in \Cref{subsec:constraint-e2e}), 
the influence
function term $\widehat{\IFx}$ we estimate does not come from the
standard $M$-estimator, and we % do not see
are not aware of 
any existing % developments
% (including assumptions and results) of
work in the 
ALO literature addressing this important case.   

     % and approximate obtain leave-one-out  $x^*(\tilde\theta_{-i})$ in ALO. This saves substantial computation time when $x^*(\cdot)$ is expensive to evaluate, especially in a two-stage problem like ETO. 
\textbf{Computational efficiency in ETO and IEO settings.} Although in the
ALO approach, one 
only  
needs to compute $\hat\theta$ once to approximate all leave-one-out $\tilde\theta_{-i}$, $i \in [n]$, one still needs to compute 
% it still needs to evaluate
$x^*(\tilde\theta_{-i})$ % $n$ times, one 
for 
each of the approximated leave-one-out parameter $\tilde\theta_{-i}$, $i
\in [n]$. \wty{In ETO and IEO settings where typically $x^*(\cdot)$ % does not have an
does not have an analytical form explicitly},  one has to compute one 
% computing
% one
  $x^*(\cdot)$ % typically involves
  by
% latter amounts to 
solving % $n$
an
optimization problem.  % which can be
% expensive,
Consequently, ALO approaches are likely to be computationally inefficient 
% especially in a two-stage problem like ETO.
for ETO and IEO settings. 
In contrast, 
OIC avoids the need to repeatedly solve optimization problems by using the
closed form expressions provided in Corollary~\ref{coro:oic-procedure}. % For
% instance,
\wty{As a concrete example, 
consider the portfolio optimization problem in \Cref{ex:portfolio}.
% ,
In this setting, it is very efficient 
to estimate the empirical mean and
variance, i.e., the parameter estimate $\hat{\theta}$; however, 
%in the first stage.
% However, we
one 
needs to solve a portfolio
optimization problem % when we evaluate
to compute 
$x^*(\tilde{\theta}_{-i})$ for each
$\tilde\theta_{-i}$. % in the second stage. 
% This means that the
Therefore, the 
computation costs of the current ALO
approaches % do not save much computational burden from
are no better than
LOOCV. % since they
% still need to solve $n$ optimization problems in the second stage.
In contrast, OIC only solves only one  % second-stage
optimization
problem.} 

\subsection{Summary in Our Position to Existing Methods and Further Discussions} 
\wty{In general, OIC is a decision-focused evaluation approach that
% incorporates optimistic bias to assess decision performance, leading to
% bias-correction methodology that attempts to compute a
computes an accurate estimate for the out-of-sample % attained
performance of a chosen decision by removing the bias introduced by
in-sample performance estimate. } OIC % can be viewed as
is 
the most general bias correction formula known to-date that applies across
different data-driven optimization problems with general decision rules
and cost functions, and is able to handle general risk functions as
in~\eqref{primary}. It recovers previous similar formulas like AIC and its
variants as special cases, and therefore, inherits the benefits of these
previous formulas, but with the additional % strength
benefit
of being substantially
more general % to allow its use
in that it can be used for 
decision selection instead of % merely
just
model selection. 
% models, where ``model-free'' denotes that the approach is free of any knowledge from the optimization problem besides an oracle to solve them
%This scope sets our work apart from the second stream.  while keeping the computational efficiency and explainability of the bias correction procedure.
\wty{% Given the
  Since OIC is
  similar to LOOCV, it can also % serve as a % classical
% model selection tool, aiding in the
be used to select optimal hyperparameters well. In~\Cref{sec:numerical} we
discuss using OIC for hyperparameter selection in portfolio selection (see
\Cref{subsec:portfolio}) and model class selection for training ETO models
in newsvendor problem (see \Cref{subsec:numerical-newsvendor}).} 
% applies to various settings, including model class selected for training
% ETO models as we will see \Cref{fig:main_nv}$(d)$ in
% \Cref{subsec:numerical-newsvendor}, and parameter selection in the
% objective such as DRO and regularization in statistical learning
% problems. We present corresponding examples in the portfolio allocation
% problem in \Cref{subsec:portfolio}, with results shown in
% \Cref{fig:portfolio-saa-complexity}$(a)(b)$. }Unlike LOOCV that requires the solution of $n$ optimization problems, OIC
% does not need to solve any additional optimization 
% problems, and is, therefore, more computationally % more
% efficient. $K$-Fold CV % can
% % educe the computation costs of CV to
% only solves
% $K$ % repetitions in solving the 
% optimization problems, but % at the
% comes with the 
% cost of a $\Theta(1/n)$ bias. 
% On the other hand, the CV methods are ``model-free'',
% Certainly
OIC requires
access to the gradient to compute the bias, and sometimes the Hessian
of the cost function and the influence function of the decision rule with respect to its parametrization. % instead of being. 
Nonetheless, we % will see how a
show that for a wide range of data-driven optimization formulations, % are indeed
% explicitly analyzable
these quantities are explicitly computable so that
% so that 
OIC % becomes
is
an efficiently implementable approach.

Note that % in spite of the lack of improvement on the
  OIC does not improve the  mean squared error % rate of the mean squared error
  over the naive empirical estimator $\hat{A}_o$ --  % since 
  this is because any estimator 
  $\tilde A$
    based on $\Dscr_n$ incurs $\E_{\Dscr_n}[(\tilde A - A)^2] =
    \Omega\Para{1/n}$~\citep{wainwright2019high} due to the statistical 
    variability in $\Dscr_n$. Nonetheless, from \Cref{coro:main-moment}
    we have that % our
    % OIC $\hat A$ % does not worsen the rate of the mean squared
    % error $\Theta\Para{1/n}$ compared to the naive estimator. %To compare
    % % models, we evaluate the performance
    % % through some deterministic criteria of
    % % the random variable, i.e., $A$ in
    % % \eqref{eq:eval-key} and $u(\cdot)$
    % % further in \eqref{primary}.
    % That is, 
    OIC alleviates the evaluation bias  $A - \E[\hat A_o]$ without
    hurting the mean squared error. The bias correction % is still useful for
    % still
    % clearly 
    % provides an improvement over the
    leads to a marked improvement in decision selection. 
  % empirical estimate $\hat{A}_o$
  % for 
  % % the important task of 
  % decision selection. % To explain, note that
  Without the bias correction provided by OIC, the empirical estimate 
  % selection criterion using the naive approach, which simply selects the
  % decision with the smallest
  $\hat A_o$ % trivially
  always favors %  empirical
  % optimization
  SAA
  % The problematic performance of this latter approach has been
  % well evidenced especially
  which is not the optimal decision rule when the sample size is only
  moderately large~\citep{kuhn2019wasserstein,van2020data}. 
  This fact is illustrated in Figure~\ref{fig:chi2dro-illustrate}, 
  where  the naive empirical estimate (i.e., $\hat{A}_o$) incorrectly picks $\rho  =0$,
  i.e. SAA, as optimal, whereas OIC identifies the true optimal $\rho^* = 1$. 
  % we also show
  % that empirical optimization, which corresponds to $\rho = 0$, is rightly
  % ranked worse using our OIC but is selected as the best if we apply the
  % naive evaluation.
  % As shown in \Cref{prop:loocv}, OIC is close to LOOCV, % which has been 
  % generally % known to be
  % as 
  % a good model 
  % selection criterion, without the additional associated computational
  % cost. 

% \subsection{\wty{Computing Influence Functions} in OIC}
In terms of limitations and future work, the main computation overhead of OIC is computing the
% , we need to compute the 
% We need to estimate the 
%sample or the 
empirical estimate $\widehat{\IFx}$ for the influence function. %  in
% terms of $\theta$ through our analysis to obtain OIC, while 
% For that, we need to
For most problems % from
in \Cref{sec:main},
this requires computing the gradient, and the inverse of the empirical
estimate of the 
Hessian of the cost function. % would need to. 
For ETO, one may also need to compute
$\nabla_{\theta} \datax$ % via inverse functions under
via the inverse function theorem. Thus, OIC is most suitable for problems where computing the gradient and inverse of the Hessian is more efficient than solving additional optimization problems. 
\tyw{% For high-dimensional
  {When the problem dimension is large relative to the sample size, as in high-order polynomial regression and overparameterized
  neural networks, the empirical Hessian estimate that appears in
  \Cref{coro:oic-procedure-2}  % may deviate substantially from the true one
  % in the influence function,
  may have substantial error; therefore, 
  % and obtaining an accurate
  estimating the 
  influence function
  % as in~\eqref{eq:empirical-if} 
  can be challenging. }% At the same time, if
  Similarly, if
  the optimization problem is not solved to near optimality, the
  approximate optimality condition in \Cref{asp:theta-hat} and the linear
  expansion~\eqref{eq:theta-if} may also fail to hold with sufficient accuracy. % in
  % practice.
  To address the influence-function estimation issue, we introduce problem-specific
  corrections to the raw influence-function estimates in
  \Cref{subsec:regression} and observe improvements in both statistical
  and computational performance. We anticipate that advances in the
  influence function estimation and early stopping results will allow more accurate application of OIC to high-dimensional problems in the future.}

% of such linear optimization problems 

% and are not computationally efficient. 
%In specific operations models, some specific model evaluation approaches
%are discussed in portfolio optimization~\citep{siegel2007performance} and
%newsvendor problem~\citep{siegel2021profit}. However, these approaches
%build on the assumption that the true distribution is well-specified
%within some parametric distribution classes in these concrete objectives,
%while our results do not require such a well-specification
%assumption. The procedure of \cite{siegel2023data} is restricted for the
%solution structure of SAA in the newsvendor problem as the order
%statistics, thus cannot be generalized further.  and thus can be
%potentially incorporated into the objective. That is, it can  

% \subsection{Incorporating OIC in the Training Procedure}
% As discussed, 

% n selecting the better ETO solution underfor the newsvendor problem 
% From another viewpoint, due to its

% applies to various settings, including model class selected for training
% ETO models as we will see \Cref{fig:main_nv}$(d)$ in
% \Cref{subsec:numerical-newsvendor}, and parameter selection in the
% objective such as DRO and regularization in statistical learning
% problems. We present corresponding examples in the portfolio allocation
% problem in \Cref{subsec:portfolio}, with results shown in
% \Cref{fig:portfolio-saa-complexity}$(a)(b)$. 
% , and constrained problems when choosing the best constraint parameter

% Furthermore, 
Finally, as another future direction, since OIC % provides an explicit form for the bias,
leads to an explicit formulation for the bias, 
one can % potentially
% be
% incorporated
use OIC as the
training objective, instead of using it for the post-training model
evaluation and selection. Given a decision mapping $x^*(\theta)$, we % consider the
can solve the 
following
optimization problem
\[
  \min_{\theta}\left\{\frac{1}{n}\sum_{i = 1}^n h(x^*(\theta);\xi_i) -
  \frac{1}{n^2}\sum_{i = 1}^n \nabla_{\theta} h(x^*(\theta);\xi_i)^{\top}
  \widehat{\IFx}(\xi_i)\right\},
\] 
%This also distinguishes us from the cross-validation and their approximation literature. 
% In other words,
where
the first-order bias estimator $\hat A_c$ acts as a
regularization term. This approach is similar to % distributionally robust
% optimization which suggests training on objectives with
DRO with gradient regularization~\citep{gao2022distributionally,shafieezadeh2019regularization} and variance regularization \citep{duchi2019variance,lam2016robust,gotoh2018robust} that has been shown to be favorable for certain problem classes (e.g., low-variance problems). 
Similarly, OIC-regularization could be potentially 
favorable for some problems, and would be worth exploring in future work.

{\subsection{Managerial Implications and Scope of Applicability}
OIC helps % support
decision evaluation and selection in data-driven settings
% optimization
by providing a bias-corrected estimate of the out-of-sample performance of
the chosen 
decision. % performance.
{Furthermore, in practice, the decision procedure is not fixed; rather, it is a choice. % Instead, 
OIC can evaluate and choose among multiple decision pipelines % as described in
(see~\Cref{ex:ddo}) in a manner that is both accurate and computationally efficient.} This perspective leads to several managerial implications.

\tyw{First, OIC provides a more credible assessment of the risk associated with
deploying the current decision: it has a smaller bias as compared to other evaluation procedures, and a similar variance. Therefore, OIC can better assess whether a data-driven decision is reliable, since its data-driven estimate closely reflects its true out-of-sample performance. This improved reliability can, in turn, strengthen stakeholders' confidence in the decision.}

Second, OIC can guide the selection of:
\begin{enumerate}[(i),leftmargin=*]
\item hyperparameters, % such as
  e.g.,  
  regularization or robustness
  parameters. For example, in portfolio selection, capacity allocation, or
  resource planning problems, % different
  the robustness parameter corresponds
  to different levels of conservatism. A % overly
  small value for robustness
  parameter % may
  looks attractive in-sample, but leads to poor out-of-sample
  performance; conversely, an overly large % parameter
  value 
  may be too
  conservative. OIC provides a data-driven method to choose % this
  % tradeoff by
  the parameter by 
  directly comparing the impact of the parameter on the overall out-of-sample
  performance.  
\item % model classes, such as 
  distribution model classes. For example, in the context of 
  inventory control problems, {the choice of parametric models for the demand should not be driven solely by goodness of fit, but also by operational costs of the corresponding decision. OIC evaluates models through the out-of-sample performance of the corresponding
  decisions, thereby providing a decision-focused alternative to likelihood-based model selection.}
  % criteria
\end{enumerate}
%(IS THIS BY PARAMETRIC-OIC?)
Third, when the decision rule involves several components, such as
parametric distribution models and estimation procedures, % together with
% Parametric-OIC in \Cref{sec:POIC}, 
OIC (see Parametric-OIC in \Cref{sec:POIC}) % also
provides a diagnostic tool to
understand the source of poor decision performance. OIC separates the
effects of limited data from inappropriate modeling assumptions on
performance degradation. A large estimated misspecification error suggests
that the decision-maker should enrich or change the parametric model,
while a small misspecification error suggests that the chosen parametric
family is adequate for the downstream % decision
performance. } 

{
  OIC requires (i) a known decision procedure and cost function in explicit form; (ii) regularity of the problem, i.e.,
  the cost function coupled with the decision rule is twice differentiable
  a.e. and the Hessian of the cost function under $\P$ is nonsingular; (iii) a nonsingular KKT matrix and strict
  complementarity for constrained problems. We summarize the application scope of OIC in
  \Cref{tab:oic-scope}. % Compared with all other evaluation procedures, 
  OIC
  is particularly useful when (i) repeatedly solving the optimization problem is substantially more expensive than computing
  gradients, Hessians,  or influence-function approximations, and (ii) one
  is interested in % evaluating
  contextual optimization, general risk
  functions, % and in other cases
  or other settings 
  not covered by existing bias-correction  % formulas
  results.

OIC is not intended to be universally applicable. It is not directly applicable to non-smooth problems including linear optimization without smoothing,  
  problems with degenerate optimizers, or % highly
  overparameterized models for which
  influence-function estimates are unstable or computationally
  prohibitive. In these settings, cross-validation, bootstrap, smoothing, or
  nonsmooth sensitivity analysis may be more appropriate. Thus, OIC should
  be viewed as an accurate and computationally efficient evaluation tool for
  smooth or locally regular data-driven optimization problems, rather than
  as a replacement for all validation methods. 
  \begin{table}[t]
    \centering
\caption{Application scope of OIC.}
\label{tab:oic-scope}
\begin{tabular}{p{0.28\linewidth}p{0.34\linewidth}p{0.30\linewidth}}
\toprule
Components & OIC is suitable when & OIC is less suitable when \\
\midrule
Objective &
The cost function is twice differentiable a.e. and has a nonsingular Hessian under $\P$. & The cost function is discontinuous or without an explicit function form. \\
\midrule
Decision Rule in \Cref{ex:ddo}$(a)$ & (i) $\P_{\theta}$ is three-times differentiable; and (ii) The binding constraint set does not change around $x^*(\theta)$ and standard regularity conditions hold, e.g., strict complementarity with nonsingular KKT matrix.& (i) $\P_{\theta}$ is not differentiable; or (ii) Some constraints switch between binding and non-binding regimes, or strict-complementarity fails. \\
\midrule
Decision Rule in \Cref{ex:ddo}$(b)$ & $x^*(\theta)$ is twice differentiable and feasible (by construction) or the composite $h(\paranx;\xi)$ is twice differentiable a.e.& $h(\paranx;\xi)$ is not twice differentiable a.e.\\
% Model-class selection &
% Several parametric or decision-rule classes are plausible, and the best statistical fit may not induce the best decision. &
% The candidate model class is very high-dimensional and Hessian/influence-function estimation is unreliable. \\

\bottomrule
\end{tabular}
\end{table}
}
%unless more structural assumptions are in place.
% not computationally efficient. 

% Furthermore, the expected debias procedure of OIC can be extended to estimate the general risk (characterized by $u(\cdot)$ in~\eqref{eq:eval-key}) of decision rules while the corresponding estimators derived from existing CV and ALO approaches are unknown in literature. And OIC exhibits an explicit bias characterization whereas the literature in the cross-validation and the ALO do not offer.  
%We defer more detailed discussions in \Cref{sec:discuss}. \wty{TODO: expand these a bit.}

% Like these previous formulas, OIC explicitly characterizes the bias term

% . The analysis of existing bias corrections along the previous stream of work only covers a subset of decision rules in some particular cost functions, whereas we extend our methodology to include other decision rules including ETO, constrained cases in more general problem instances such as nonsmooth objectives and contextual optimizations. 

\section{Numerical Study}\label{sec:numerical}
% Throughout numerical studies,
In this section, we report the results of our numerical experiments
exploring
% We numerically illustrate 
the performance of
OIC and the associated bias correction term $\hat{A}_c$.
%From the analysis between $\hat{A}_c$, we see that incorporating the bias correction term can give a true decision performance of the given data-driven solution. Then for each estimator of $\hat{A}$, we mainly focus on its performance against $\hat{A}_o$. 
We % benchmark
 compare
OIC with  the simple empirical method (EM) that plugs in
the data-driven decision into the empirical objective, % and other
% evaluation methods including 
cross-validation, ALO and bootstrap
approaches (see~Appendix~\ref{app:btsp-jcnf}).  % he oracle method (if we know 
%in terms of the difference between the true . 
%in each table highlighting the expected true performance for each decision rule $\E_{\Dscr_n}\E_{\P^*}[h(x^*(\hat{\theta});\xi)]$. for each problem instance 
We
use several different families of decision rules to
% incorporate
test several 
% all
representative scenarios. % discussed so far.in the cost or time entries

In each table in this section, boldfaced values %  indicate that the
% corresponding
% highlight
indicate 
the 
evaluation method % is the best in terms of having
with 
the
% smallest
lowest
bias, i.e., closest to the ``oracle'' $A$ in~\eqref{eq:eval-key}, 
or the lowest computation time, as the case might be. The computation time for each instance is % measured
the sum of the time % taken
to compute 
% by
% one complete cycle of obtaining
the solution $x^*(\hat{\theta})$, %  which
% involves solving one optimization problem and performing the cost
and the time % required
to compute the bias correction
% evaluation to obtain each 
$\hat A$; and therefore, it includes tasks such as
computing bias in OIC and solving $K$ additional optimization problems in
$K$-Fold CV.   
Detailed setups 
and full experimental results can be found in Appendix~\ref{app:numeric}. 
%the boldfaced value in each table 
%- $\E_{\Dscr_n}[(1/n)\sum_{i = 1}^n h(x^*(\hat{\theta});\xi_i)]$ if not specifically mentioned. 
%in this case.
%\Cref{app:numeric}.  

%The oracle represents the true error between true costs and in-sample costs, i.e.. 

%We are interested in the error / complexity computed across evaluation approaches against the oracle then.

\subsection{Portfolio Allocation}\label{subsec:portfolio}
Consider an unconstrained mean-variance portfolio optimization problem (e.g.,
\cite{blanchet2022distributionally}) with the cost function being
$h(x;\xi) := x^{\top} (\xi - \E\xi)(\xi - \E\xi)^{\top} x - 
                                    \xi^{\top} x + 2 x^{\top}x$,
% % with a cost function % without 
% % constraints:
% \[
%   \begin{array}{rl}
%     \mbox{min}_{x \in \mathbb{R}^{D_{\xi}}} &h(x;\xi) := x^{\top} (\xi - \E\xi)(\xi - \E\xi)^{\top} x -
%                                     \xi^{\top} x + 2 x^{\top}x,
%   \end{array}
% \]
where the last term 
% penalizes the allocation weight into the objective.
penalizes large portfolio weights.
%In the base case, we let the space $\Xscr = \R^d$ to be unconstrained. 
%We also include variants of objectives in the Appendix to show the generalization properties of our approach. 
% Here, we simulate 
The asset returns $\xi = (\xi_A, \xi_B)^{\top}$ where the returns  $\xi_A$
and $\xi_B$ for the two asset classes  are independent with $\xi_A \sim N(\mu_A,
\Sigma_A)$, $\xi_B \sim N(\mu_B, \Sigma_B)$, and the number of assets in
each class $D_{\xi_A} = D_{\xi_B} = \frac{1}{2}D_{\xi}$. % denoting two classes of assets.

We consider the following decision rules: % \wty{where some have been
% discussed in \Cref{ex:portfolio} except
% differences in loss functions.}
\begin{enumerate}[(i)]
  % \textbf{Decision Classes.}
\item E2E with different mappings $\paranx$:
  \begin{enumerate}[(a)]
  \item SAA-U: Equal weights on all assets, i.e., $x^*(\theta) = \theta
    \bm{1}_{D_{\xi}}$.
  \item SAA-B: Equal weights on assets in a particular class, i.e.
    $\paranx = (\theta_A \bm{1}_{D_{\xi_A}}, \theta_A \bm{1}_{D_{\xi_B}})$ but
    $\theta_A$ and $\theta_B$ need not be equal.
  \item SAA: unconstrained weights, i.e. $\paranx = \theta \in \mathbb{R}^{D_{\xi}}$.
  \end{enumerate}
\item ETO with $\Pscr_{\Theta} = \{\prod_{i \in [D_{\xi}]}
  N(\mu_i,\sigma_i^2)^{\top}: \theta = (\mu, \sigma^2)\}$, i.e.,  $x^*(\theta)$ follows
  \Cref{ex:portfolio}$(a)$).
\item DR-E2E with the distance measure $d$ given by the $\chi^2$-divergence with different ambiguity 
  levels $\epsilon = \rho/n$ and $\paranx = \theta
  \in \mathbb{R}^{D_{\xi}}$. 
% set to be Gaussian models with independent margins (\wty{$x^*(\theta)$ the same in} \Cref{ex:portfolio}$(a)$); (iii)  
\end{enumerate}
% equally weighted across all assets (SAA-U) (\wty{$x^*(\theta) =
% \theta$}, i.e., equal weights in \Cref{ex:portfolio}$(b)$); equally
% weighted for 
% assets in one class (SAA-B) (\wty{$\paranx = (\theta_1
% {1}_{D_{\xi_A}}^{\top}, \theta_2 {1}_{D_{\xi_B}}^{\top})^{\top}$}); SAA
% (\wty{$x^*(\theta) = \theta$}, i.e., Unconstrained weights in
% \Cref{ex:portfolio}$(b)$); %\hl{Don't understand what  per-asset (SAA)
% means.}  

%Fixing sample size $n = 50$ and varying $D_{\xi}$, we estimate the complexity term in OIC for the classical SAA policy and compare the result with the oracle as well as the 5-fold CV.

\begin{table}[htb]
\small
    \centering
\renewcommand{\arraystretch}{1.0}
\renewcommand{\tabcolsep}{3mm}
    \caption{\wty{Difference between the 
      estimated bias $\hat A - \hat A_o$ and the true expected bias $A_c$ of each method averaged over 200 problem instances when $n =
      50, D_{\xi} = 10, \rho = 3$,  where $\hat A_o$ is the empirical average and $\hat A$ is output by each evaluation method. A ``+''(resp. ``-'') indicates 
      the estimated objective is larger (smaller) than the oracle. The quantity in the parentheses denotes the average running seconds per instance.
      % ``+" (``-'') means the estimated bias of the evaluation method is
      % larger (smaller) than the oracle model. 
      The boldfaced value in each row 
      % entry represent
      is the one with lowest bias (computational time).}
      % the one with the smallest bias difference (in
      % absolute values) and computational time.
    } 
    \label{tab:portfolio-complexity}
    \resizebox{\textwidth}{!}{
    \begin{tabular}{c|c|cccc}
    \toprule
          & Oracle  bias $A_c$       & OIC             & 5-CV          & 10-CV &  ALO-IJ  \\
    \midrule
 %    SAA         & 0.091 & \textbf{0.096} \scriptsize $\pm 0.026$ (\textbf{1.64e-3}) & 0.116 \scriptsize $\pm 0.031$ (2.07e-3) & 0.109 \scriptsize $\pm 0.022$ (3.84e-3) & 0.080 \scriptsize $\pm 0.020$ (4.10e-3)  \\
 %    SAA-U & {0.036}  & \textbf{0.050} \scriptsize $\pm 0.015$ (\textbf{1.01e-3}) & 0.060 \scriptsize $\pm 0.021$ (1.89e-3) & 0.055 \scriptsize 
 % $\pm$0.013 (3.67e-3) & 0.053 \scriptsize $\pm 0.019$ (4.09e-3)  \\
 %    SAA-B & {0.070} & {0.078} \scriptsize $\pm 0.022$ (\textbf{4.42e-3}) & 0.092 \scriptsize $\pm 0.028$ (1.68e-1)  & \textbf{0.067} \scriptsize  $\pm 0.015$ (3.36e-1) & 0.074 \scriptsize $\pm0.017$ (8.09e-3)\\
 %    ETO       & {0.030}  & \textbf{0.035} \scriptsize $\pm 0.022$ (5.50e-1) & 0.058 \scriptsize $\pm 0.025$ (\textbf{2.69e-1}) &0.036 \scriptsize $\pm 0.012$ (4.18e-1) & \textbf{0.035} \scriptsize $\pm 0.018$ (3.23e-0)  \\
 %    $\chi^2$-DRO         & {0.089}  & 0.083 \scriptsize $\pm 0.022$ (\textbf{1.64e-3}) & 0.114 \scriptsize $\pm 0.030$ (2.94e-0) & \textbf{0.085} \scriptsize $\pm 0.022$ (6.59e-0) & 0.066 \scriptsize $\pm 0.017$ (4.29e-3)\\
    SAA         & 0.091 & +\textbf{0.005}  (\textbf{1.64e-3}) & +0.025 (2.07e-3) & +0.018  (3.84e-3) & -0.011 (4.10e-3)  \\
    SAA-U & {0.036}  & +\textbf{0.014}  (\textbf{1.01e-3}) & +0.024 (1.89e-3) & +0.019  (3.67e-3) & +0.017 (4.09e-3)  \\
    SAA-B & {0.070} & +{0.008}  (\textbf{4.42e-3}) & +0.022 (1.68e-1)  & -\textbf{0.003} (3.36e-1) & +0.004 (8.09e-3)\\
    ETO       & {0.030}  & +\textbf{0.005} (5.50e-1) & -0.008 (\textbf{2.69e-1}) &+0.006 (4.18e-1) & +\textbf{0.005} (3.23e-0)  \\
    $\chi^2$-DRO         & {0.089}  & -0.006 (\textbf{1.64e-3}) & +0.025 (2.94e-0) & -\textbf{0.004} (6.59e-0) & -0.023 (4.29e-3)\\
    \bottomrule
    \end{tabular}
    }
\end{table}

\Cref{tab:portfolio-complexity} reports the estimated bias and
computational cost % time of each decision rule across evaluation methods
% including
for OIC, 5-CV, 10-CV and ALO-IJ for each of the decision rules listed above.
%the bias the OIC computed value, and the value computed by 5-CV. 
% We first get a quantitative understanding of the bias correction across
% methods. 
% We see that 
% our
OIC % can attain
% almost similar bias to the oracle method.
% is accurate in estimating the oracle.
accurately estimates the oracle value -- it has the lowest bias for $3$
of the $5$ cases.
OIC is also computationally efficient % and usually runs faster than other
% evaluation methods across each decision
% rule.
-- it has the lowest computational time in $4$ out of the $5$ cases.
Note that although the Gaussian parametric
models are misspecified with $2D_{\xi}$ parameters to estimate,
the total bias is still small, % in this case, 
% indicating
illustrating the
benefits of % applying
incorporating parametric models in selecting decisions 
\citep{iyengar2023hedging,elmachtoub2023estimatethenoptimize}.
% Comparing
% the bias for 
% SAA, SAA-U, SAA-B, % $\hat{A}_c$
% it is clear that OIC, unlike AIC,  
% does not scale linearly with the dimension
% of decision space; rather, it % like AIC 
% % but
% depends on the size of 
% % depends on 
% the decision class and cost
% function. % Besides,
% Furthermore, we see that 
% % imposing
% DRO % approaches
% can reduce the model bias a bit
% (also from \Cref{fig:portfolio-dro-complexity}).  
%Furthermore, 
We defer the estimation of the risk-incorporated criterion that is 
% Risk-OIC applied to this problem is 
discussed in \Cref{sec:general-risk} to Appendix~\ref{app:experiment-risk}, where Risk-OIC also gives an accurate corresponding estimate.

\begin{figure}[!htb]
    \centering
    % \subfloat[Estimated Bias of SAA varying $D_{\xi}$ and $n$] 
    % {
    %     \begin{minipage}[t]{0.5\textwidth}
    %         \centering
    %         \includegraphics[width = 0.99\textwidth]{figs/newfigs/portfolio_test_saa2.pdf}
    %     \end{minipage}
    % }
    \subfloat[Estimated Costs of $\chi^2$-DRO with $(n, D_{\xi}) = (100, 20)$.]
    {
        \begin{minipage}[t]{0.5\textwidth}
            \centering
            \includegraphics[width = 0.99\textwidth]{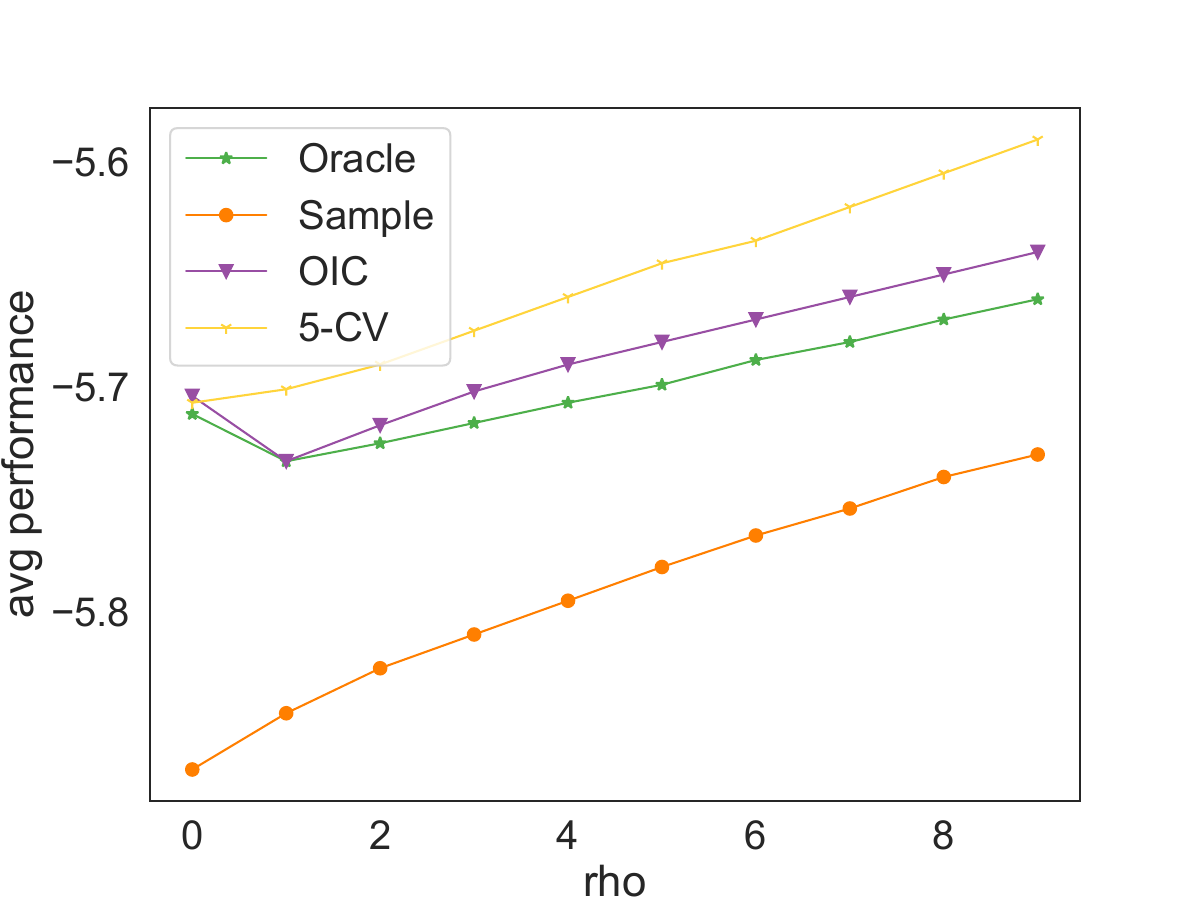}
        \end{minipage}
    }
    \subfloat[Estimated Costs of SAA with $(n, D_{\xi}) = (100,5)$]
    {
        \begin{minipage}[t]{0.5\textwidth}
            \centering
            \includegraphics[width = 0.9\textwidth]{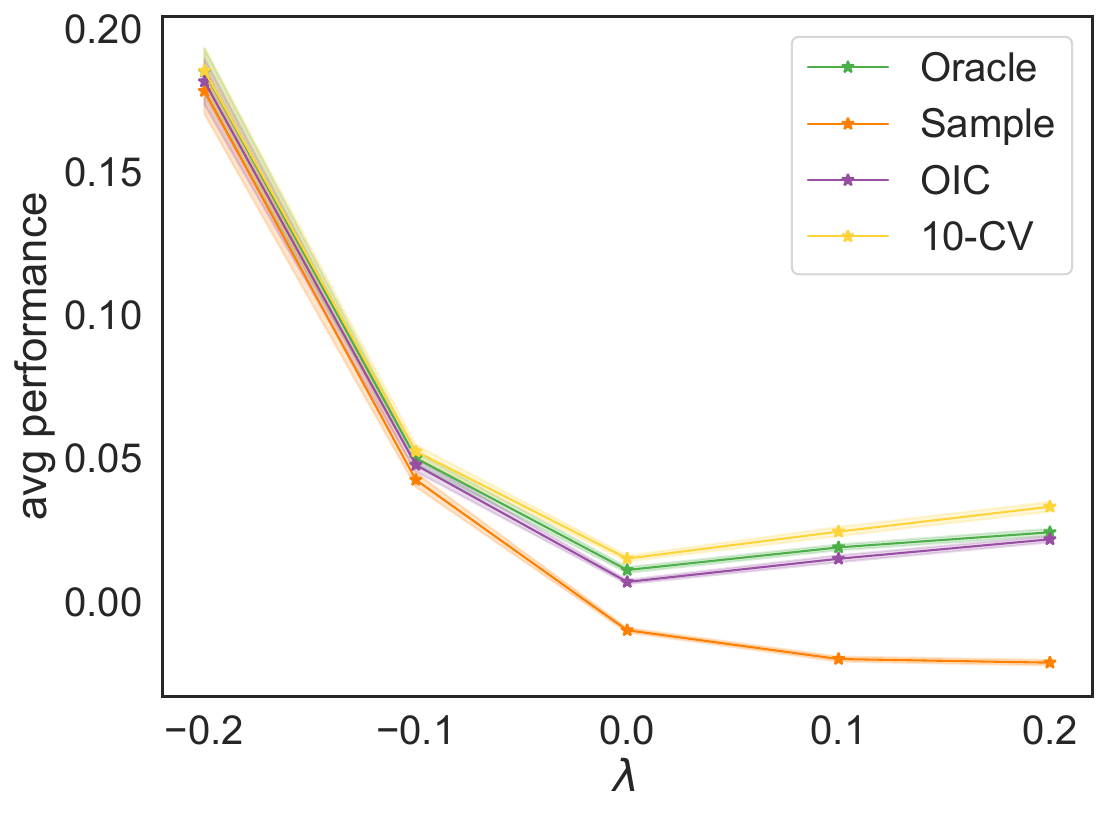}
        \end{minipage}
    }
    \caption{Performance evaluations of specific decision rules given by \wty{variants of} $\chi^2$-DRO and SAA.}
    \label{fig:portfolio-saa-complexity}
\end{figure}

% The computational time for % evaluating the $\chi^2$-DRO
% % approach by
% computing the 
% OIC was much smaller than that for 5-CV  evaluation approach. }

% Then we turn to some
% specific decision rules to further
{In
\Cref{fig:portfolio-saa-complexity}(a) % shows the performances of OIC in
we plot the performance of OIC 
% evaluating
% for
% SAA and 
with respect to DR-E2E as a function of the hyperparameter $\rho$ that sets the radius of the ambiguity set. We find that OIC correctly identifies $\rho^* = 1$.}
%Specifically, we focus on the bias correction performance . 
% n \Cref{fig:portfolio-saa-complexity}$(a)$ we plot the estimated costs as a function of 
% % In terms of decision selection of 
% $\rho$ for
% $\chi^2$-DRO. We see that OIC % can also
% corrects the optimistic bias of EM, attaining almost the same
% % same curve of estimated 
% performance as the oracle. % Therefore
% Furthermore, 
% % , the bias
% % correction term in 
% OIC % helps we identify correct decisions regions
% correctly identifies the % decision
% best hyperparameter $\rho$ such that DRO performs the best 
% % $\rho
% % \in (0, 2]$
% % where DRO methods % achieve
% % perform
% % better than the SAA ($\rho = 0$ in \Cref{fig:portfolio-saa-complexity}$(b)$) 
% while 5-CV cannot. % Besides,
\wty{In \Cref{fig:portfolio-saa-complexity}(b), we plot the estimated
objective value of $\E_{\Dscr_n}\E_{\P^*}[x^*(\hat\theta)^{\top}(\xi - a)(\xi -a)^{\top}x^*(\hat\theta) - \xi^{\top}x]$ for the constrained SAA portfolio optimization procedure $x^*(\theta) = \theta$ with $\hat\theta$ computed by:
\begin{equation*}
  \begin{array}{rl}
  \min & \E_{\hat\P_n} [\theta^{\top}(\xi - \E[\xi])(\xi - \E[\xi])^{\top}\theta -
         \xi^{\top}\theta],\\ [0.5em]
    \text{s.t.} & \|\theta\|_{\infty} \geq -\lambda.
  \end{array}
\end{equation*}
as a function of the hyperparameter $\lambda$. We observe that OIC
correctly identifies the optimal $\lambda$, and its performance is the
same as $10$-CV, but at a fraction of the computational cost. 
% We observe that when adjusting the parameter in the constrained SAA, OIC
% performs well, identifying the optimal $\lambda$ with the smallest
% objective cost $\E_{\P^*}[x^{\top}(\xi - a)(\xi - a)^{\top} -
% \xi^{\top}x]$ and achieving similar evaluation performance to the oracle
% compared to 10-CV.
Furthermore, OIC is able to correctly identify the optimal $\lambda$ is
$65$\% of the $200$ trials, whereas EM is only able to identify the
optimal $\lambda$ in $6$\% of the cases.
% -- % \% hl{What is the performance of $10$-CV?}
% % Furthermore, we compute the correct selection
% probability of identifying the best $\lambda$ for each evaluation
% procedure. By correcting the bias, the selection probability increases
% significantly from 6\% under EM to 65\% with OIC, which is comparable to
$10$-CV correctly identifies $\lambda$ in 
63\% % achieved by 10-CV across 200 problem instances.
of the instances but at approximately 10 times the computational cost.}

%and estimate associated gradient and hessian here. 

% To understand the complexity here with decision space, tnteraction between Decision Space Complexity + Cost Function: It does not only scale with the number of parameters in the decision space. Although it does not appear to be a simple term from the direct calculation of the expression, we conduct a experiment to investigate how the bias-correction term grows with number of the dimension of $\xi$. Varying the dimension of $\xi$ here.

%Besides, we find the criterion can offer us select a good model and attain relative good decision performance, intuitively it keeps the same order of hyperparameter in $\chi^2$-DRO models. average Cost here (x-axis: $\rho$ y-axis: estimated value for EM, OIC, and 5-CV with the oracle).

\subsection{Newsvendor Problem}\label{subsec:numerical-newsvendor}

Consider the newsvendor cost function $h(x;\xi) = cx - p\min\{\xi, x\}$, where $c$
is the (known) cost per unit and $p$ is the (known) price per unit. % We aim
% to obtai
Our goal is to compute the bias term $A_c$ and its estimator $\hat A_c$
for a range of different data-driven optimization procedures. \wty{We consider the following four decision rules and compute the bias formula $\hat A_c$ for each case:
\begin{enumerate}[(i)]
\item SAA: From \Cref{coro:oic-procedure}$(ii)$, we have $I_h(\theta) =
\hessianp\E_{\P^*}[h(\paranx;\xi)] = pf(\theta)$, where $f$ is the
true p.d.f. of $\P^*$. Therefore, it follows that $A_c =
\frac{1}{n}\Tr{I_h(\theta^*)^{-1} J_h(\theta^*)} = \frac{c(p-c)}{n p
  f(\theta^*)}$. Thus, we recover the main result in
\cite{siegel2023data}. We take $\hat A_c = \frac{c(p - c)}{n p\hat
  f(\hat\theta)}$ while using kernel density estimate $\hat 
f(\hat\theta)$ % to approximate
to approximate $f(\theta^*)$. 
\item ETO with $\Pscr_{\Theta} = \{N(\mu,\sigma^2): \theta =
  (\mu,\sigma^2)^{\top}\}$ and the
  calibration function $T$ given by the MLE: The decision rule is $x^*(\theta) = \mu +
  \sqrt{\sigma^2} \Phi^{-1}(1 - \frac{c}{p})$, % from
                                % \Cref{ex:ddo}$(a)(1)$,  
  where $\Phi$ is the c.d.f. of the standard normal distribution,  % function
  and the estimate $\hat{\theta} = (\hat{\mu}, \hat{\sigma}^2)^{\top}$, where
  $\hat\mu$, $\hat\sigma^2$ is the sample mean and variance, 
  respectively, 
  of the data $\Dscr_n = \{\xi_i\}_{i \in [n]}$. Then,
  $\widehat{\IFx}(\xi_i) = (\xi_i - \hat{\mu}, (\xi_i - 
  \hat{\mu})^2 - \hat{\sigma}^2)^{\top}$. 
\item ETO with $\Pscr_{\Theta} = \{\text{Exp}(\theta): \theta \geq 0\}$ and the
  calibration function $T$ given by the MLE: The decision
  rule % is
  $x^*(\theta) = \log (p / c) \theta$ % from \Cref{ex:ddo}$(a)(1)$
  where the estimate $\hat{\theta} = \hat{\mu}$, and $\hat\mu$ is the
  sample mean
  of the data $\Dscr_n = \{\xi_i\}_{i \in [n]}$. Then, 
  $\widehat{\IFx}(\xi_i) = \xi_i - \hat{\mu}$.
\item Operational Statistics with $\Pscr_{\Theta} = \{\text{Exp}(\theta):
  \theta \geq 0\}$ and the calibration function $T$ given by operational
  statistics~\citep{liyanage2005practical}: The decision rule $\paranx = n[(p/c)^{1/(n + 1)} -
  1]\theta$ %  from the operational statistics
  % and
  and $\hat{\theta} = \hat{\mu}$, where $\hat\mu$ is the
  sample mean of the data $\Dscr_n = \{\xi_i\}_{i \in [n]}$, and 
  $\widehat{\IFx}(\xi_i) = \xi_i - \hat{\mu}$. The influence function here
  is the same as ETO but the mapping $\theta 
  \mapsto \paranx$ is different since the operational statistics method
  incorporates the downstream objective differently. 
\end{enumerate}
For the decision rules (ii) - (iv), we compute $\hat A_c$  via~\eqref{eq:oic}. In~\eqref{eq:oic}, $\nabla_{\theta} h(\datax;\xi) = \nabla_x h(\datax;\xi) \nabla_{\theta} \datax$ from the chain rule where $\nabla_x h(x;\xi) = c\mathbf{1}_{\{x >
  \xi\}} + (c - p) \mathbf{1}_{\{x \leq \xi\}}$. 
}

Note that \cite{siegel2021profit} also discusses the estimation of the
evaluation bias in these three decision rules. However, their results only
apply when $\Pscr_{\Theta}$ includes the true distribution $\P^*$, i.e.,
$\P^* \in \Pscr_{\Theta}$.  
% since in their evaluation goal, 
%true evaluation objective is $\E_{\Dscr_n}\E_{\P_{\theta^*}}[h(\datax;\xi)]$ instead of 
In contrast, the OIC bias correction procedure % of is suitable regardless
can be applied even when 
% of whether 
$\P^* \not\in \Pscr_{\Theta}$. This is because in the
evaluation goal of \cite{siegel2021profit}, the random variable $\xi$ is
evaluated under $\P_{\theta^*}$ instead of $\P^*$ in \eqref{eq:eval-key}
and the two are the same only if $\P^* \in \Pscr_{\Theta}$.

\begin{figure}
    \centering
    \subfloat[$\xi_S \sim N(100, 40)$, $n = 50$]
    {
        \begin{minipage}[t]{0.4\textwidth}
            \centering
            \includegraphics[width = 1.1\textwidth]{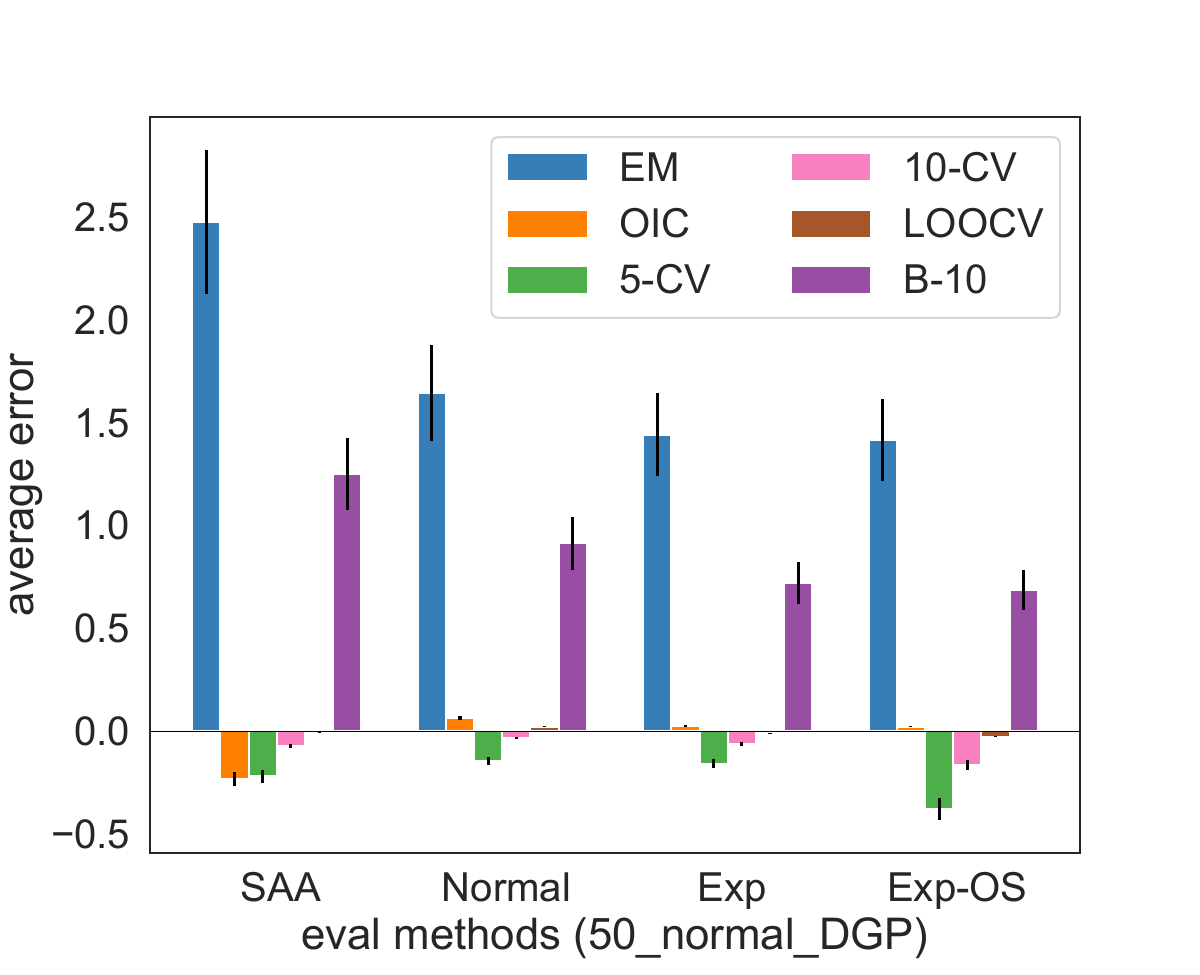}
        \end{minipage}
    }
    \subfloat[$\xi_S \sim \text{Exp}(100)$, $n = 100$]
    {
        \begin{minipage}[t]{0.4\textwidth}
            \centering
            \includegraphics[width = 1.1\textwidth]{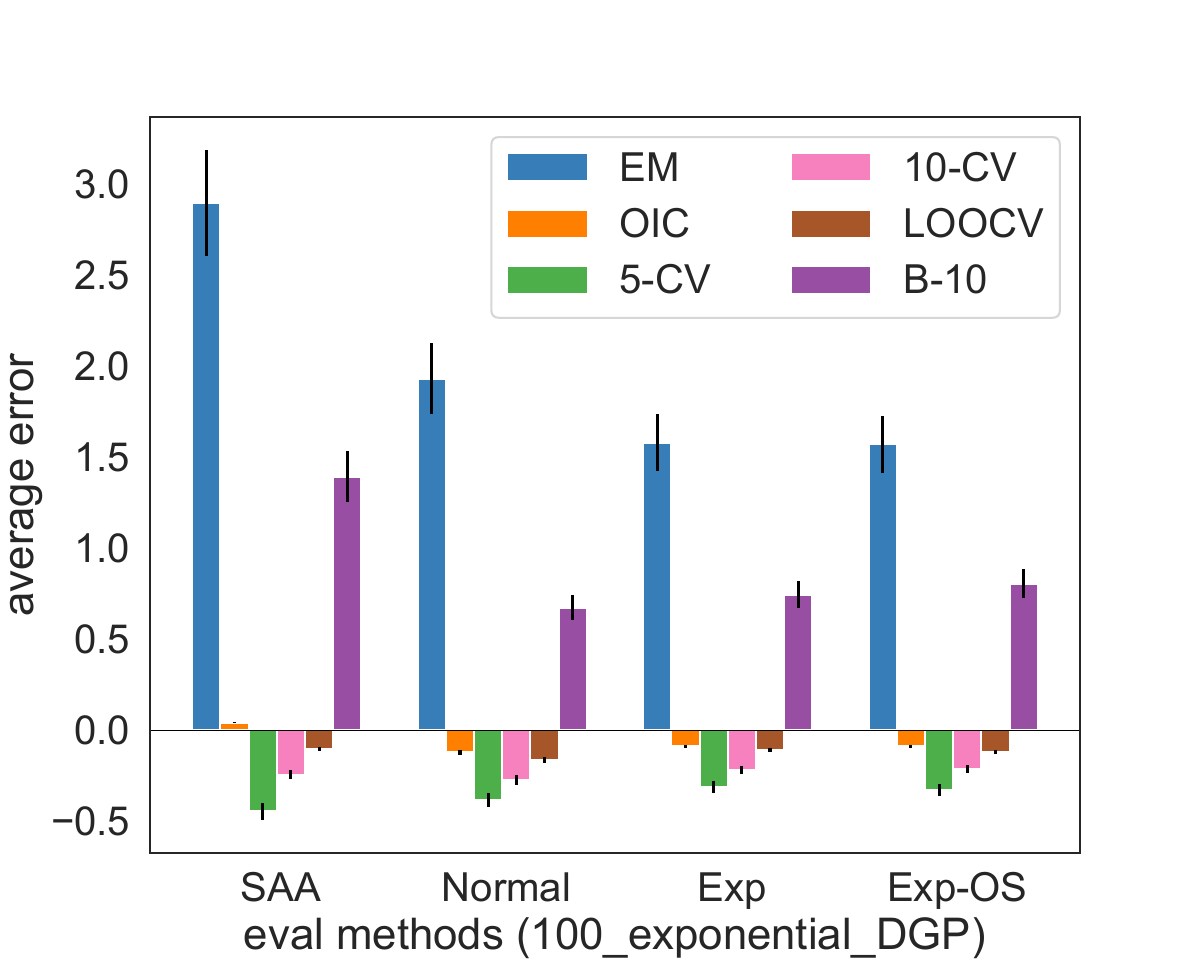}
        \end{minipage}
    }
    
    \subfloat[Evaluation of Misspecification Error]
    {
        \begin{minipage}[t]{0.4\textwidth}
            \centering
            \includegraphics[width = 1.1\textwidth]{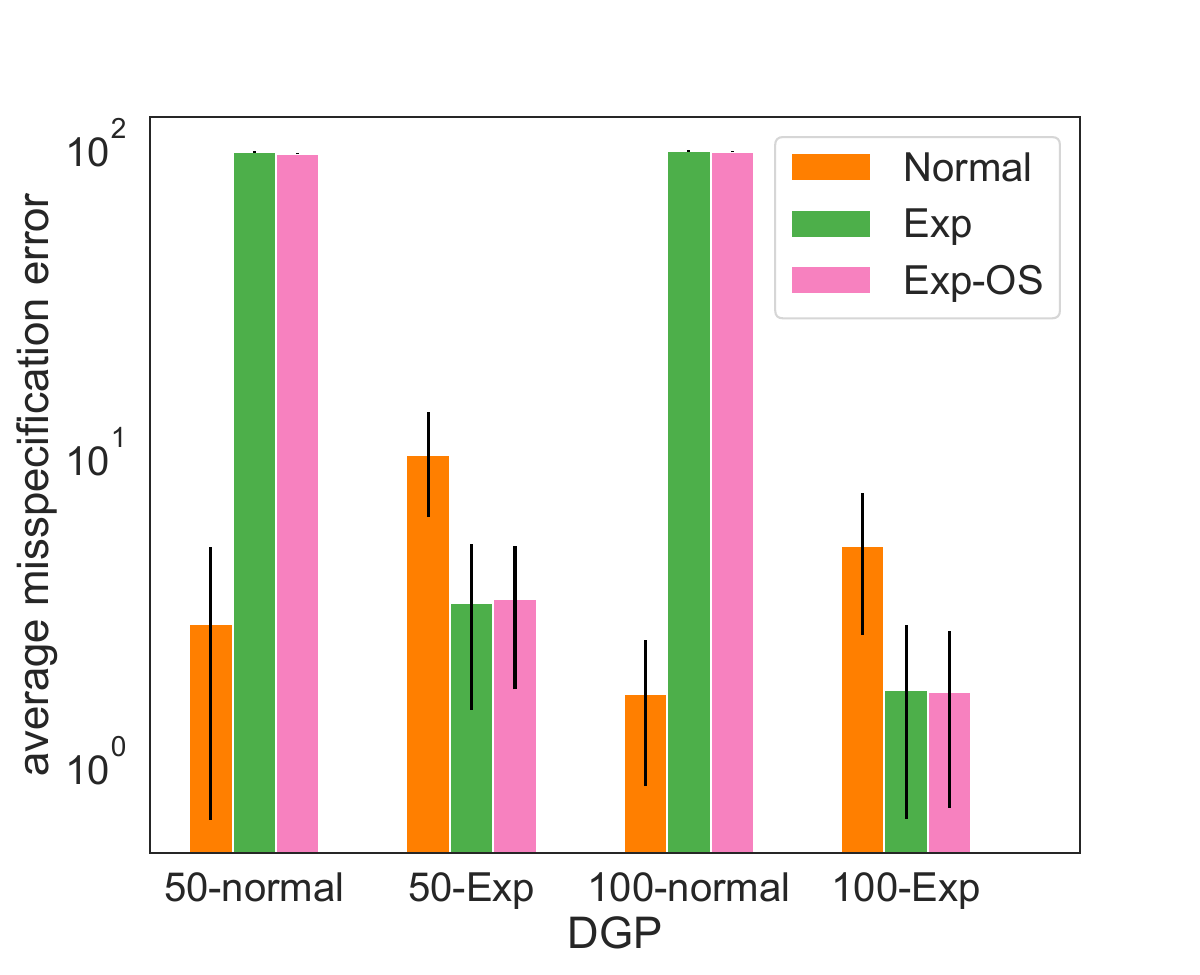}
        \end{minipage}
     }
     %of selecting the best optimization model
    \subfloat[Optimality Gap over Evaluation Procedures] 
    %the Best (the decision rule that minimizes $A$)]
    {
        \begin{minipage}[t]{0.4\textwidth}
            \centering
            \includegraphics[width = 1.1\textwidth]{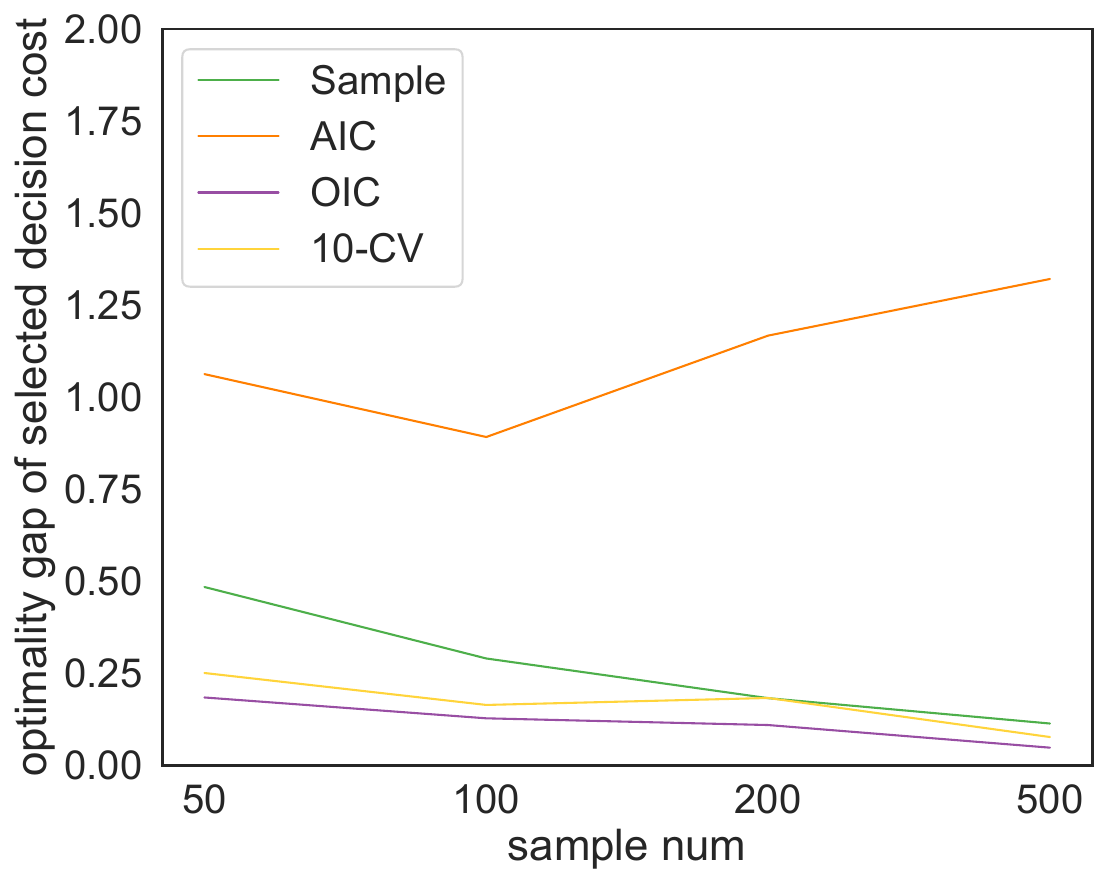}
        \end{minipage}
     }
     \caption{{In plots (a) (b), we report the mean and standard error of the
       % difference
       % between the true and estimated performances
       error over $100$ replications 
       for normal
       (plot (a)) and exponential (plot (b)) data generating process.}
       % under
       % different evaluation methods, where we run over 100 independent
       % runs for 100 rounds, for two different data generating processes
       % (DGPs). 
       In plot (c), we report the estimated model misspecification
       error of approaches in \Cref{def:oic}$(a)$. \tyw{In plot (d), we report
       the optimality gap of the selected decisions using different evaluation procedures. 
       %(SAY OPTIMALITY GAP TO BE MORE PRECISE?)
       }} 
    \label{fig:main_nv}
\end{figure}

% We consider 
In our numerical example, we set $c = 2, p = 5$. We consider the random variable
with the demand
$\xi \overset{d}{=} \xi_S + \xi_N$, where $\xi_S$ and $\xi_N$ are
independent, $\xi_S \sim N(100, 40)$ or $\text{Exp}(100)$ and $\xi_N \sim U(-5, 
5)$. 
% \paragraph{Decision Classes.}
%\wty{with respect to how $x^*(\theta)$ and $\hat\theta$ are obtained in each case.}
% (i) SAA,  ETO with (ii) Normal model, (iii) Exponential,  and (iv)
% Exponential models with Operational Statistics \citep{liyanage2005practical}.
% %Detailed models and expressions of the model evaluation are shown in Appendix. 
% Note that \cite{siegel2021profit} also consider  the bias of 
% decision classes (ii)-(iv) % were also considered 
% % in % but they required
% % under the condition that 
% % the true distribution to be
% % well-specified for each of the model class (i.e. 
% when $\xi$ is well-specified without the noise % part
% $\xi_N$. % therefore does not match this case.
We benchmark the performance of OIC against EM,  $5$-CV, $10$-CV, LOOCV,
and B-$10$ (Bootstrap with $B = 10$). We do not consider ALO approaches
since it is computationally efficient to run LOOCV. 
% estimation error across several evaluation methods with 
% good
The results of the comparison are plotted in
\Cref{fig:main_nv}. In~\Cref{fig:main_nv} (a) we plot the average error
when $\xi_S \sim N(100, 40)$, and in~\Cref{fig:main_nv} (b) we plot the
error when $\xi_S \sim \text{Exp}(100)$.
The results in these two figures clearly indicate that  OIC % can
achieves a smaller evaluation error when compared to % with cross
% validation with $K = 5, 10$ across empirical SAA and different parametric
% models.
to 5-CV and 10-CV for most decision rules. Furthermore, OIC % can enjoy similar 
performance is similar to LOOCV when the sample size is
large, % which indicates our previous
supporting the asymptotic equivalence result of the two methods
in \Cref{prop:loocv}. 
% Then
We also show that OIC can find better decisions with a smaller $A$
compared with EM in Appendix~\ref{app:newsvendor-additional}.  

% existing evaluation methods in
% \Cref{fig:main_nv} (a)-(b), quantify misspecification error (discussed in \Cref{sec:POIC}) in
% \Cref{fig:main_nv}(c) \wty{and evaluate OIC's performance in model selection in \Cref{fig:main_nv}(d)}. 
%Due to page limitation, we only consider the relative good of all
%nonparametric evaluation methods mention as the benchmarks, i.e. 5,
%10-fold CV, LOOCV, bootstrap with $B = 10$ here.  
%\texttt{EM} is denoted as the empirical average across samples. Besides
%the approach OIC in \Cref{thm:main} and OIC-P (specifically for the
%parametric models) in \Cref{prop:pf-fit}.  

% From \Cref{fig:main_nv}(a)-(b), % it is clear
% we see 
% that OIC % can
% achieves a smaller evaluation error compared % with cross
% % validation with $K = 5, 10$ across empirical SAA and different parametric
% % models.
% to 5-CV and 10-CV for most decision rules. Furthermore, OIC % can enjoy similar 
% performance is similar to LOOCV when the sample size is
% large, % which indicates our previous
% supporting the asymptotic equivalence result of the two methods
% in \Cref{prop:loocv}. 
% % Then
% We also show that OIC can find better decisions with a smaller $A$ compared with EM in Appendix~\ref{app:newsvendor-additional}. 
%Besides, cross-validation is known to leads to pessimism results (with negative mean estimation) due to fewer samples trained than the total number while our approach circumvents that.  
% And
%the less one method deviates from zero, the better performance this would gain. 
%Although cross-validation methods with $K = 5, 10$ have been used widely in practice \citep{hastie2009elements}, 

By comparing the results for OIC and Parametric-OIC in~\Cref{fig:main_nv}(c), we can 
quantify the model misspecification errors of different parametric models
introduced in~\Cref{sec:POIC}. Despite the fact that the two data-generating
processes are not fully parametric, we find that the misspecification error of using the normal 
model % incur much smaller misspecification error 
is significantly smaller than using the exponential model % under
when $\xi_S \sim N(100, 40)$, and not significantly 
% and not much 
higher % compared with that
than using the 
two exponential 
models % even
when $\xi_S \sim Exp(100)$. 
% This demonstrates a relatively robust
% performance of ETO based on the normal distribution % here
% in the face of misspecification error % to
% % downstream decisions in the
% for the considered
% newsvendor problem.
%\hl{I don't understand this discussions -- needs to be clarified.}

We also assess the empirical benefits of OIC in decision selection. % Specifically, we use
We use EM, 10-CV, AIC, and OIC to select the
best decisions among SAA, Normal and Exponential decision rules (ETO) over
500 problem instances where the true distribution is generated from a
misspecified normal distribution (i.e., $\xi_S \sim N(100, 40)$ above). In
\Cref{fig:main_nv}(d), we plot \tyw{the optimality gap % across
  of the decision rules selected by four evaluation procedures (EM, 10-CV, AIC, OIC) as a
  function of the sample size~$n$. Here, the optimality gap is
  % calculated
  defined
  as the difference between the average cost of the selected
  decisions and the oracle cost, % where the oracle always selects the
  i.e., the cost of the
  optimal decision 
  for
  the true stochastic model. {We find that the optimality gap of OIC
  % achieves strong performance: it's optimality gap 
  is the smallest among
  all evaluation procedures across all sample sizes~$n$, i.e., the % ts selected 
  decision rule selected by OIC consistently
  outperforms the rules selected by EM, AIC, and 10-CV approaches.} } 

% \wty{TODO: model selection in ETO vs IEO vs SAA?}

%Surprisingly, when sample size is large, under normal DGP, the two exponential models would lead to comparable accurate decision makings, i.e. we estimate the misspecification error to be -0.057 (0.006) and -0.058 (0.007) for exp without and with OS when $n = 100$ while normal models would be 0.398 (0.004). It can be explained by the fact similar optimal decisions under such misspecified model. However, in the exponential DGP, neither Normal nor Exponential models would attain good decisions, where the model misspecification error would be 0.905 (0.014) for exponential, 0.896 (0.014) for Exp-OS models. And error of normal models would be 1.074 (0.014) [CHECK AGAIN FOR POSITIVE / NEGATIVE SIGN].

\subsection{Real-World Regression}\label{subsec:regression}
In this section, we report the results of our numerical experiments with the UCI wine dataset consisting of $N = 6,498$ red and white wines~\citep{cortez2009modeling,duchi2023distributionally}, 
%% with 6498 examples
% andwith $12$ features, % in total,
where each wine is associated with a feature $z \in \R^{12}$ that
corresponds to chemical properties, and a label $\xi \in \R$ that denotes a subjective quality assessment. \wty{Note that we are in the contextual optimization setting. We consider $4$
different decision rules $x^*(\theta;z)$, where the $\hat\theta$ of each
decision rule is estimated via the squared error of the average
performance: $\hat\theta \in \argmin_{\theta} \sum_{i \in [n]}\paran{(\xi_i -
x^*(\theta;z_i))^2 + \alpha \|\theta\|_2^2}$ and the ridge 
term $\alpha \|\theta\|_2^2$ regularizes the prediction function. We
specify each $x^*(\theta;z)$ as follows: 
\begin{enumerate}[(i)]
\item  Linear regression, i.e., $x^\ast(\theta,z) = \theta^{\top}z$. We consider two cases: $\alpha = 0$ (Linear) and $\alpha = 1$ (Linear-R1).
\item Linear regression with a second-degree polynomial of the features,
  $x^{*}(\theta,z) = \theta^{\top}p_2(z)$, where $p_2(z)$ denotes a vector that includes all polynomial combinations of the features up to
  degree $\leq 2$. We consider three cases: $\alpha = 0$~(Quad), $\alpha =
  1$~(Quad-R1) and $\alpha = 5$~(Quad-R5).
\item Neural networks with
2 hidden layers~(NN), i.e. $x(\theta;z) =
  \theta^{(2)}\sigma(\theta^{(1)}\sigma((\theta^{(0)})^{\top} z))$,
  where 
  $\sigma(\cdot)$ is the \texttt{softplus} activation function, and
  $\theta =(\theta^{(0)}, \theta^{(1)}, \theta^{(2)})$ and $\alpha = 0$.
\end{enumerate}
The details of model fitting and influence functions for each decision
rule are discussed in \Cref{app:regression}. The goal is to evaluate each decision rule $x^\ast(\theta; z)$ in terms of the expected value of the loss
function 
\[
  h\big(x^*(\theta,z);\xi\big) = \big((\xi - x^*(\theta,z))^2 -
  \beta\big)^+.
\]
The loss function $h$ only penalizes the squared error larger than
$\beta$, i.e., the tail risk, and is different from the function used to calibrate the decision. In this regard, the objective $h$ is
suitable for risk management in avoiding a large prediction error (e.g., in the electricity market, small errors in demand are tolerable while large prediction errors may lead to costly emergency purchase) and
similar in spirit to conditional value-at-risk~\citep{rockafellar2002conditional}.}
  % i.e. 
%   (Quad), and  corresponding ridge
% regression with the ridge parameter % ridge
% % parameter 
% % Quad-R: 2-Poly Feat. + Ridge Regression
% $\alpha = 1, 5$ (Quad-R1 and Quad-R5), \wty{where $s_{\theta}(\xi^u) = \theta^{\top}p_2(\xi^u)$ with $p_2(\xi^u)$ representing a feature vector including all polynomial combinations of the features with degree $\leq 2$}; 
% For each % following model
% decision
% rule,  % over
% % the standard MSE criterion, 
% we are interested in evaluating % over
% % the tail risks specifically, 
% the squared prediction error above some threshold
% $\beta$, \wty{which is suitable for risk management in avoiding a large prediction error and shares similar formulations as the conditional value-at-risk objective \citep{rockafellar2002conditional}.} Therefore,  
% %we may prefer the desired quality prediction system with robustness to bad predictions instead of average predictions. Then
% % our target
% the
% cost function % would be:
% $h\big(x^*(\theta);(\xi^u, \xi^v)\big) = \big((\xi^v - s_{\theta}(\xi^u))^2 -
% \beta\big)^+$, where $s_{\theta}(\cdot)$ denotes a mapping from the feature to the predicted label. We consider the following decision rules:(ii)
% (iii)

\begin{table}[htbp]
\centering
\small
\caption{Estimated objective values and computation times for the real-world regression experiment, averaged over 100 random seeds. The Oracle is approximated by evaluating each fitted decision rule on the held-out \(N-n\) observations. In Panel A, boldface indicates the estimate closest to the Oracle within each column. Panel B reports the average evaluation time in seconds.}
\label{tab:reg}
\resizebox{\textwidth}{!}{
\begin{tabular}{l|cccc|cccccc}
\toprule
& \multicolumn{4}{c|}{Training fraction \(0.1\) (\(n=650\))}
& \multicolumn{6}{c}{Training fraction \(0.3\) (\(n=1950\))} \\
\midrule
Decision Rule
& Linear & Linear-R1 & Quad & Quad-R1
& Linear & Linear-R1 & Quad & Quad-R1 & Quad-R5 & NN \\
Parameter Num.
& 13 & 13 & 92 & 92
& 13 & 13 & 92 & 92 & 92 & 185 \\
\midrule
\multicolumn{11}{c}{{A: Estimated objective value (unit: \%)}} \\
\midrule
EM
& \(34.5 \pm 0.2\) & \(34.2 \pm 0.2\) & \(24.2 \pm 0.2\) & \(24.4 \pm 0.2\)
& \(35.3 \pm 0.1\) & \(35.1 \pm 0.1\) & \(28.2 \pm 0.1\) & \(28.2 \pm 0.1\) & \(28.4 \pm 0.1\) & \(27.2 \pm 0.4\) \\

5-CV
& \(37.3 \pm 0.3\) & \(37.2 \pm 0.2\) & \(83.8 \pm 13.7\) & \({72.9 \pm 7.6}\)
& \(36.2 \pm 0.1\) & \(\mathbf{36.0 \pm 0.1}\) & \(40.7 \pm 0.9\) & \(41.2 \pm 1.4\) & \(41.6 \pm 1.2\) & \(39.7 \pm 2.1\) \\

OIC
& \(36.6 \pm 0.2\) & \(36.2 \pm 0.2\) & \(41.0 \pm 0.3\) & \(40.4 \pm 0.3\)
& \(\mathbf{36.1 \pm 0.1}\) & \(35.9 \pm 0.1\) & \(35.5 \pm 0.1\) & \(35.4 \pm 0.1\) & \(35.4 \pm 0.1\) & \(34.7 \pm 1.2\) \\

IFC-OIC
& \(\mathbf{36.7 \pm 0.3}\) & \(\mathbf{37.0 \pm 0.3}\) & \(\mathbf{50.7 \pm 0.3}\)
& \({48.6 \pm 0.3}\) & \(35.7 \pm 0.2\) & \(35.8 \pm 0.2\) 
& \(\mathbf{37.5 \pm 0.2}\) & \(37.7 \pm 0.2\) & \(36.9 \pm 0.2\) & \(\mathbf{34.9 \pm 0.6}\) \\

ALO-IJ
& \(\mathbf{36.7 \pm 0.3}\) & \(37.2 \pm 0.4\) & \(75.7 \pm 3.6\) & \(\mathbf{61.5 \pm 0.5}\)
& \(35.7 \pm 0.2\) & \(36.0 \pm 0.2\) & \(36.1 \pm 0.5\) 
& \(\mathbf{39.6 \pm 1.3}\) & \(\mathbf{40.2 \pm 1.0}\) & \(34.6 \pm 1.8\) \\

\midrule
Oracle
& \(36.8 \pm 0.0\) & \(36.7 \pm 0.0\) & \(62.0 \pm 1.8\) & \(56.8 \pm 1.5\)
& \(35.9 \pm 0.1\) & \(36.0 \pm 0.1\) & \(38.1 \pm 0.6\) & \(39.2 \pm 0.6\) & \(39.1 \pm 0.5\) & \(37.1 \pm 0.5\) \\

\midrule
\multicolumn{11}{c}{{B: Average evaluation time (unit: seconds)}} \\
\midrule
5-CV
& 0.006 & 0.002 & 0.084 & 0.087
& 0.015 & 0.011 & 0.015 & 0.019 & 0.018 & \(\approx 830\) \\

OIC
& 0.004 & 0.004 & 0.011 & 0.007
& 0.011 & 0.012 & 0.019 & 0.016 & 0.016 & \(\approx 180\) \\

IFC-OIC
& 0.008 & 0.008 & 0.019 & 0.015
& 0.012 & 0.011 & 0.034 & 0.041 & 0.052 & \(\approx 4\)\\

ALO-IJ
& 0.025 & 0.026 & 0.034 & 0.031
& 0.078 & 0.077 & 0.088 & 0.086 & 0.084 & \(\approx 190\) \\

\bottomrule
\end{tabular}}
\end{table}

We consider two settings: small training set $n = 0.1N =  650$ and large training
set $n = 0.3N = 1950$. \tyw{Since the number of parameters is large
  compared with the sample size, we introduce  IFC-OIC, an
  influence-function-corrected variant of OIC. This variant introduces an additional correction factor to the empirical influence function that improves finite-sample performance~\citep{wilson2020approximate}~(see Appendix~\ref{app:ifc-oic} for details). Specifically, for the neural network model, it replaces the potentially unstable
  Hessian estimate with a gradient outer-product
  approximation~\citep{martens2010deep}. } 

\tyw{We report the estimated out-of-sample performance for each of these decision rules
using EM,  5-CV, OIC, IFC-OIC, and ALO-IJ in \Cref{tab:reg}.} The table
reports the number of parameters for each decision rule (first row), the
objective value computed by each bias correction method for each decision
rule, and the running time for 5-CV, OIC, IFC-OIC, and ALO-IJ. For the
linear models, the performance of OIC is similar to or slightly better
than 5-CV, and the Oracle value lies within the confidence
region. \tyw{For the quadratic and neural network models, OIC
  underestimates the bias because the empirical
  influence-function estimate is unstable in high-dimensional regimes. In contrast,
  IFC-OIC provides more accurate estimates than the original OIC for
  quadratic and neural network models, indicating that improving the
  influence function approximation is an effective way to extend OIC. }On
the other hand, 5-CV overestimates the bias (also 
observed by~\cite{beirami2017optimal}).  % This is 
\Cref{tab:reg} also shows that the biases of the
quadratic and neural network decision rules are very large % comparing
compared to the difference between the oracle and EM.  
% In general, we may prefer a simple linear model in this case and OIC can
% identify correct the large bias term from 2-poly features and neural net
% regressions.  
% Furthermore, the % bias correction term in
% OIC estimates are more stable, i.e., they have a
% smaller standard error when compared with 5-CV.
% \hl{Isn't this bad for the quadratic and NN case? The confidence interval
%   is falsely too tight.}
%The left and right part of the table shows the evaluated decision
%performance under training samples to be 10\% and 30\% of total samples
%respectively. 
% We leave the investigation as our future work.
%Due to fewer numbers of calls of $\paranx$ in the evaluation, 
Also, the computational time of OIC is lower than that of ALO and 5-CV for most problem configurations. Specifically, for the neural network model in the large training set setting, i.e., $n = 0.3N$, the average computation time for OIC is approximately $3$ minutes,  and  5-CV is $14$ minutes. \tyw{IFC-OIC further reduces the computational cost substantially because it avoids the explicit Hessian computation.} % samples are used in training.
% and 
% % under 30\% samples and 
% 1.2 minutes and 9 minutes, respectively, when  % under
% 10\% samples % , respectively.
% are used for training. And further computational efficiency
% %We can even adopt better tools to accelerate the Hessian evaluations as provided in \cite{koh2017understanding}. 
% % In all,
To summarize, OIC is closer to the true decision performance compared to ALO and 5-CV in settings where the number of parameters is not too large. \tyw{IFC-OIC further mitigates the limitations of OIC in high-dimensional settings, leading to slightly improved statistical performance and substantially better computational efficiency for neural network models}. Moreover, we expect that OIC will gain further advantages as more efficient methods for approximating the influence function become available. 

\bibliographystyle{informs2014}
\bibliography{references}

\ECSwitch
%\ECTheoremsNumberedThrough
%\ECDisclaimer
%%%%%%%%%%%%%%%%%%%%%%%%%%%%%%%%%%%%%%%%%%%%%%%%%%%%%%%%%%

%%% Main head for the e-companion
\ECRUNTITLE{OIC: Dissecting and Correcting Bias in Data-Driven Optimization}
% \ECRUNAUTHOR{Iyengar, Lam and Wang}
\ECHead{Appendices}
%\section{Discussion of }

We provide further discussions on our assumptions and alternative evaluation methods in Appendix \ref{sec:further discussions}, details of assumptions and proofs of statements in Appendices \ref{sec:proofs main},~\ref{sec:proofs extension} and~\ref{app:challenge}, and additional experimental results in Appendix \ref{app:numeric}.

\section{Further Discussions in \Cref{sec:main}}\label{sec:further discussions}
Across this section, we denote one function is $C^1$ (or $C^2$) when it is continuously differentiable (or twice continuously differentiable) with respect to all variables unless mentioned. 

\subsection{Verification of Assumption~\ref{asp:nonsmooth}}\label{app:discuss}
\wty{
%\subsection{Implicit Function Theorem}
To verify Assumption~\ref{asp:nonsmooth}, we first state the general implicit function theorem as follows:
\begin{lemma}[Implicit Function Theorem, from Theorem 3.9 in \cite{folland2023calculus}]\label{lemma:implicitfunction}
    Consider $f(\theta;x)$ as an $\R^{D_x}$-value function of class $C^1$ on some neighborhood of a point $(\theta_0, x_0) \in \R^{D_{\theta} + D_x}$ and let $B_{i,j} = \frac{\partial f_i}{\partial x_j}(\theta_0, x_0)$. Suppose that $f(\theta_0, x_0) = 0$ and $\det B \neq 0$. Then there exists positive numbers $r_0, r_1$ such that the following conclusions are valid:
    \begin{itemize}
        \item For each $\theta$ in the ball $\|\theta - \theta_0\|_2 < r_0$, there exists a unique $x$ such that $\|x - x_0\|_2 \leq r_1$ and $f(\theta;x) = 0$. Denote this $x$ by $x^*(\theta)$. In particular $x^*(\theta_0) = x_0$;
        \item The function thus defined is of class $C^1$; and its partial derivative $\partial_{\theta_j} x^*$ can be computed by differentiating the equation $f(\theta, x^*(\theta)) = 0$ with respect to $\theta_j$ and solving the resulting linear system of equation of $\partial_{\theta_j} x_1^*, \ldots, \partial_{\theta_j} x_{D_x}^*$. 
    \end{itemize}
    Furthermore, if $f(\theta;x)$ is $C^2$, i.e., twice continuously differentiable, then $\nabla_{\theta\theta}^2 x^*(\theta)$ exists.
\end{lemma}}

\wty{Since Assumption~\ref{asp:nonsmooth} builds upon Assumption~\ref{asp:nonsmooth}(i) and (ii), we focus our discussion on how to verify Assumption~\ref{asp:nonsmooth}(i) since \Cref{asp:nonsmooth}(ii) is with respect to the structure for $f(z)$ and relatively easy to verify. }

% For $\paranx$ in Definition~\ref{ex:ddo}$(b)$, i.e., when $\paranx$ is expressed as some explicit closed-form function of $\theta$, we verify \wty{the correctness of it} directly from the closed-form of $\paranx$. The verification of \Cref{asp:nonsmooth}(i) can be carried out directly based on definitions of $g(\paranx;\xi)$.

\subsubsection{Standard Conditions that Allow \Cref{asp:nonsmooth}(i) to Hold}
Before we provide standard conditions that allow \Cref{asp:nonsmooth}(i) to hold, we first mention that for $\paranx$ in Definition~\ref{ex:ddo}$(b)$, i.e., when $\paranx$ is expressed as some explicit closed-form function of $\theta$, we verify \wty{the correctness of it} directly from the closed-form of $\paranx$. The verification of \Cref{asp:nonsmooth}(i) can be carried out directly based on definitions of $g(\paranx;\xi)$. 

\wty{For general decision rules, we provide a set of sufficient conditions for \Cref{asp:nonsmooth}(i). This set of sufficient conditions consists of Assumptions~\ref{asp:smooth-cost-objective} and~\ref{asp:decision}:}
% i.e, $g(x^*(\theta);\xi)$ is $C^2$ with respect to $\theta$. This one set of the sufficient condition consists of Assumptions~\ref{asp:decision} and~\ref{asp:smooth-cost-objective}}:
% that $h(x;\xi)$ is twice continuously differentiable with respect to $x$ \wty{and $\paranx$ is twice differentiable}. 

\begin{assumption}[Smoothness of Cost Objective]\label{asp:smooth-cost-objective}
$g(x;\xi)$ is $C^2$ with respect to $x$ for any $x \in \Xscr$.
\end{assumption}

\begin{assumption}[Smoothness of Decision Rule]\label{asp:decision}
    $x^*(\theta)$ is $C^2$ with respect to $\theta$ for any $\theta$
\end{assumption}

% In this part, we provide some sufficient conditions by different optimization problem instances satisfying Assumptions \ref{asp:func} and \ref{asp:represent}.

In the above, \Cref{asp:smooth-cost-objective} can be explicitly verified. On the other hand, for each optimization procedure in Definition~\ref{ex:ddo}, we need to verify \Cref{asp:decision} holds for the corresponding $\paranx$. 
%Therefore, we mainly discuss when the gradient $\nabla_{\theta} x^*(\theta)$ and hessian $\nabla_{\theta\theta}^2 x^*(\theta)$ in Example~\ref{ex:ddo} $(a)$ exist and how we compute them.
%In fact, throughout the whole analysis, due to Taylor expansion, we only need $h(\paranx;\xi)$ is twice continuous differentiable around $\theta^*$.
%ii) \underline{When $\paranx$ is not an explicitly analytic function of $\theta$, such as an implicit function (ETO)}, 
Therefore, for $\paranx$ in Definition~\ref{ex:ddo}$(a)$, i.e., when $\paranx$ is not a closed-form function in general, we apply the following \Cref{asp:smooth-decision-rule}, which is a sufficient condition that ensures Assumption~\ref{asp:decision} holds:
\begin{assumption}[Smoothness of Parametrized Decision Rule: Extended]\label{asp:smooth-decision-rule}
    Consider $\alpha(\theta)$ and $B({\theta})$ are the Lagrangian multipliers and indices of binding constraints under $\P_{\theta}$ for the solution $\paranx$ in Definition~\ref{ex:ddo}$(a)$. We assume:
    \begin{enumerate}
        \item  Each distribution in $\P_{\theta}$ has a smooth density denoted by $p_{\theta}(\xi)$. $p_{\theta}(\xi) \in C^3$ and its up to third-order derivative with respect to $\theta$ is $L_1$-integrable, i.e., $\nabla_{\theta}p_{\theta}(\xi), \nabla_{\theta\theta}^2 p_{\theta}(\xi), \nabla_{\theta\theta\theta}^3 p_{\theta}(\xi) \in L^1(\xi)$;
        \item $h(x;\xi)$ satisfies Assumption~\ref{asp:nonsmooth};
        \item In $\Xscr$, $F_j(x) \in C^3$ for each $j \in J$;
        \item \tyw{The KKT matrix $\begin{pmatrix}
\nabla_{xx}^2 \mathcal L(x^*(\theta),\alpha(\theta);\theta)
& \nabla F_B(x^*(\theta))\\
\nabla F_B(x^*(\theta))^{\top} & 0
\end{pmatrix}$ is nonsingular, where $\Lscr(x,\alpha;\theta):= \E_{\P_\theta}[h(x;\xi)] +\sum_{j\in B(\theta)}\alpha_j F_j(x)$ and $F_B(x) = (F_{j_1}(x), \ldots, F_{j_m}(x))^{\top}$ if $B({\theta}) = \{j_1,\ldots, j_m\}$;
\item Strict Complementarity: Every active inequality constraint has a strictly positive Lagrangian multiplier, i.e., $\alpha_j(\theta) > 0$, $\forall j \in \{j \in J_1: F_j(x^*(\theta)) = 0\}$.}
    \end{enumerate}
\end{assumption}
In \Cref{asp:smooth-decision-rule}, the first condition ensures that the vector function $\nabla_x \E_{\P_{\theta}}[h(x;\xi)]$ is $C^2$ with respect to $\theta$ and $x$, which is commonly satisfied by exponential distribution families as we will discuss in \Cref{app:smooth-example}. \tyw{The last two conditions are regularity conditions for constrained problems and hold for many objectives~\citep{bonnans2000perturbation}.  For example, the nonsingularity of the KKT matrix is typically satisfied when the objective of the optimization problem defining $x^*(\theta)$ is strictly convex. We will provide further explanations and examples in \Cref{app:smooth-example}.} Then combining them with other conditions, we have $\nabla_x \E_{\P_{\theta}}[h(x;\xi) + \sum_{j \in B_{\theta}}\alpha_j(\theta)F_j(x)]$ is $C^2$ with respect to $\theta$ and $x$ in constrained problems.

Finally, we comment that Assumptions~\ref{asp:decision} (or~\ref{asp:smooth-decision-rule}) and~\ref{asp:smooth-cost-objective} only provide one set of sufficient conditions for \Cref{asp:nonsmooth}. When $x^*(\theta)$ has a closed-form solution, one can directly verify the correctness of \Cref{asp:nonsmooth} via checking the form of $h(x^*(\theta);\xi)$ such that Assumptions~\ref{asp:decision} and~\ref{asp:smooth-cost-objective} may not necessarily hold. For example, if $x^*(\theta)$ is a piecewise linear function and $h(x;\xi)$ is $C^2$ with respect to $x$ for any $x \in \Xscr$, one can check that the composite function still satisfies Assumption~\ref{asp:nonsmooth} while $x^*(\theta)$ may not satisfy Assumption~\ref{asp:decision}.

\subsubsection{Computation of Gradients and Hessians in \Cref{asp:nonsmooth}}
Denote $x = (x_1, \ldots, x_{D_x})^{\top}$. To compute $\nabla_{\theta} h(\paranx;\xi)$ and $\nabla_{\theta\theta}^2 h(\paranx;\xi)$, we consider different decision rules calibrated in \Cref{ex:ddo}.  Applying the standard chain rule to $h(\paranx;\xi)$, we have:
% when $\paranx$ is a vector, we can apply the first-order condition $f(\hat{\theta};\datax) = 0$ with all $D_x$-th components of $\bm x = (x_1, \ldots, x_{D_x})^{\top}$ such that:
\begin{equation}\label{eq:datax}
\begin{aligned}
\frac{\partial h(\paranx;\xi)}{\partial \theta} &=    \sum_{i = 1}^{D_x}\frac{\partial h(x;\xi)}{\partial x_i} \frac{\partial x_i}{\partial \theta},\\
\frac{\partial^2 h(\paranx;\xi)}{\partial \theta^2} &= \sum_{i = 1}^{D_x}\Para{\frac{\partial^2 h(x;\xi)}{\partial x_i^2}\frac{\partial x_i}{\partial \theta}\cdot \Para{\frac{\partial x_i}{\partial \theta}}^{\top} + \frac{\partial h(x;\xi)}{\partial x_i}\frac{\partial^2 x_i}{\partial \theta^2}} + 2\sum_{i = 1}^{D_x}\sum_{j \neq i}\frac{\partial^2 h(\paranx;\xi)}{\partial x_i \partial x_j} \frac{\partial x_i}{\partial \theta}\cdot\Para{\frac{\partial x_j}{\partial \theta}}^{\top}.
\end{aligned}
\end{equation}

Below, we show how the gradient $\nabla_{\theta} x^*(\theta)$ and Hessian $\nabla_{\theta\theta}^2 x^*(\theta)$ in \Cref{ex:ddo}$(b)$. First, we consider $D_x = 1$, and then the vector equation becomes a scalar one, i.e., 
% We focus on the computation of the $i$-th  component of $\nabla_{\theta} x^*(\theta)$ and $\nabla_{\theta\theta}^2 x^*(\theta)$, i.e., $\frac{\partial x_i}{\partial \theta}$ and $\frac{\partial^2 x_i}{\partial \theta^2}$. 
$f(\theta;x):=\nabla_x \E_{\P_{\theta}}[h(x;\xi) + \sum_{j \in B_{\theta}}\alpha_j(\theta)F_j(x)] = 0$. We also denote $f_{x\theta} = \nabla_x \nabla_{\theta}f(\theta;x), f_{\theta} = \nabla_{\theta}f(\theta;x)$ and $f_{x} = \nabla_x f(\theta;x), f_{\theta\theta} = \nabla_{\theta\theta}^2 f(\theta;x)$. Applying the standard implicit function theorem to the $i$-th component of $x$, we have:
\begin{equation}\label{eq:inverse}
    \begin{aligned}
    \frac{\partial x}{\partial \theta} & = - \frac{f_{\theta}}{f_x} \in \R^{D_{\theta}},\\
   \frac{\partial^2 x}{\partial \theta^2} &= \frac{-f_x^2 f_{\theta \theta} + f_x (f_{\theta}f_{x\theta} + f_{\theta x}f_{\theta}^{\top}) - f_{xx} f_{\theta}f_{\theta}^{\top}}{f_{x}^3} \in \R^{D_{\theta}\times D_{\theta}},
\end{aligned}
\end{equation}
where $f_{x\theta}\in \R^{1\times D_{\theta}}, f_{\theta}\in \R^{D_{\theta}}, f_{\theta x} \in \R^{D_{\theta}\times 1}$. 

For general $D_x$ and $D_{\theta}$, we can compute:
\begin{equation}\label{eq:inverse-general}
    \begin{aligned}
            \nabla_{\theta}x^*(\theta)& = -(\nabla_x f(\theta, x))^{-1} \nabla_{\theta} f(\theta,x);\\
    \nabla_{\theta\theta}^2x^*(\theta)[u, v]& = A(\theta)(\nabla_{\theta x}^2 f(\theta, x)[v] + \nabla_{xx}^2 f(\theta, x)[\nabla_{\theta}x^*(\theta) v] A(\theta) B(\theta) u\\
    &~- A(\theta)(\nabla_{\theta\theta}^2 f(\theta, x)[v] + \nabla_{x\theta}^2 f(\theta, x)[\nabla_{\theta}x^*(\theta) v])u,
    \end{aligned}
\end{equation}
where $u, v \in \R^{D_{\theta}}$ and $\nabla_{\theta\theta}^2 x^*(\theta)[u, v] \in \R^{d_x}$, $A(\theta) = \nabla_x f(\theta,x) \in \R^{D_x\times D_x},  B(\theta) = \nabla_{\theta}f(\theta, x) \in \R^{D_{\theta}\times D_x}$. That is, when $u = e_i, v = e_j$, we obtain the $(i,j)$-entry of Hessian, which is a vector of $\R^{D_xx}$. Here we apply the direction gradient definition such that:
\[\nabla_{\theta x}^2 f(\theta, x)[v] = \frac{d}{dt}|_{t = 0}\nabla_x f(\theta + tv, x), ~\nabla_{xx}^2 f(\theta, x)[\tilde v] = \frac{d}{dt}|_{t = 0}\nabla_x f(\theta, x + t\tilde v),\]
Computing~\eqref{eq:inverse-general} only involves $\nabla_{\theta x}^2, \nabla_{xx}^2, \nabla_{x\theta}^2$, which can be obtained as long as $f(\theta;x)$ is twice continuous differentiable.

% \subsubsection{Relaxation of \Cref{asp:nonsmooth}}
% In fact, to show \Cref{thm:main}, we can relax the conditions of Assumptions~\ref{asp:nonsmooth} from every $\theta \in \Theta$ to a neighborhood around $\theta^*$. We require Assumption~\ref{asp:nonsmooth} holds for every $\theta \in \Theta$ since we need to estimate $\gradp h(\datax;\xi)$ (and $\hessianp h(\datax;\xi)$) such that OIC is well-defined everywhere.

\wty{\subsubsection{Examples satisfying Assumption~\ref{asp:nonsmooth}}\label{app:smooth-example}
We list some examples satisfying Assumption~\ref{asp:nonsmooth}, where each example consists of a concrete cost function $h(x;\xi)$ and decision rule $x^*(\theta)$. } 
%\tyw{We end the section with a discussion of how assumptions are satisfied in constrained optimization problems.}

\noindent\underline{For the optimization procedures in \Cref{ex:ddo}$(b)$}, when $x^*(\theta) = \theta$ or some decision rules that are twice continuously differentiable with respect to $\theta$, 
%\Cref{asp:func} holds.}
the following cost functions satisfy Assumption~\ref{asp:nonsmooth} (with $f(x) = x$) besides \Cref{ex:portfolio} in the main body:
\begin{example}[Mean-Variance Portfolio Allocation]\label{ex:mean-variance-portfolio}
    Consider $h(x;\xi) = x^{\top} (\xi - \E\xi)(\xi - \E\xi)^{\top} x - \xi^{\top} x + 2 x^{\top}x$, where $\xi$ denotes the assets' return; and $x$ denotes the portfolio weight.
\end{example}
%For \Cref{asp:nonsmooth}, t
\begin{example}[Multi-product Newsvendor Loss]\label{ex:mtl-newsvendor}
    Consider $h(x;\xi) = \sum_{i = 1}^{D_{\xi}}\Para{c_i x_i  - p_i \min\{\xi^{(i)}, x_i\}}$, where $c_i, p_i$ are the (known) cost and price per unit for the $i$-th product. Here $g(x;\xi) = (x_1 - \xi^{(1)}, \ldots, x_{D_{\xi}} - \xi^{(D_{\xi})})^{\top}$.
\end{example}
\begin{example}[Conditional-Value-at-Risk (CVaR) Portfolio Allocation]\label{ex:cvar-portfolio}
    Consider $h((x, v);\xi) = v + \frac{1}{\epsilon}(-\xi^{\top}x - v)^+$, where $\epsilon$ denotes the quantile value and $v$ denotes the additional parameter to optimize. Here $g((x,v);\xi) = (v, -\xi^{\top}x - v)^{\top}$. 
\end{example}
\begin{example}[Quantile Regression]\label{ex:quantileregression}
    Consider $h(x;\xi) = \tau(x^{\top}\xi^u - \xi^v)_+ + (1-\tau)(\xi^v - x^{\top}\xi^u)_+$, where $\tau \in (0, 1)$ denotes the quantile percentage of the loss, $\xi^u$ and $\xi^v$ denote the feature and outcome respectively. Here $g(x;\xi) = x^{\top}\xi^u - \xi^v$.
\end{example}
\tyw{% Here under
  For
  unconstrained problems, $x^*(\theta)$ can be specified as any twice-differentiable function of $\theta$. For constrained problems, the feasibility and differentiability of $x^*(\theta)$ are enforced through the decision rule designed by the user. Specifically, the decision-maker can choose a differentiable parametrization whose output lies in $\Xscr$ for every $\theta$. For simple constraints, one may use a smooth parametrization, such as softmax mappings for simplex constraints or positive transformations for nonnegativity constraints. For more general constraints, one may use a smooth feasible parametrization or a twice-differentiable optimization layer that maps parameters into feasible decisions~\citep{amos2017optnet,donti2021dc3}. These constructions generate feasible and differentiable decision rules so that our results can apply.}

\noindent\underline{For the optimization procedures in \Cref{ex:ddo}$(a)$}, \tyw{in unconstrained problems, \Cref{asp:nonsmooth} is satisfied for all the cost functions mentioned in Examples~\ref{ex:mean-variance-portfolio} to~\ref{ex:quantileregression} with the following distributions parametrized by $\theta$:} 
\begin{example}[Exponential Family Distributions]\label{ex:exponential}
    $p(\xi;\theta) = h(\xi)\exp(\eta(\theta)^{\top} T(\xi) - A(\theta))$, where $T(\xi), h(\xi), \eta(\theta)$ and $A(\theta)$ are some known functions. $\|T(\xi)\|_2^k h(\xi)\exp(\eta(\theta)^{\top} T(\xi))\in L^1(\xi)$ for $k = 1,2,3$; and $\eta(\theta), A(\theta) \in C^3$.
\end{example}
\Cref{ex:exponential} ensures the first condition of \Cref{asp:smooth-decision-rule} holds with respect to $\P_{\theta}$. The exponential family distributions cover many of the most common distributions modeled in operations management areas, such as Gaussian, beta, and exponential distributions. \tyw{The second and third condition of \Cref{asp:smooth-decision-rule} can be verified directly.}

\tyw{In constrained problems, \Cref{asp:smooth-decision-rule} remains valid under some regularity conditions, e.g., a convex loss with a binding constraint set that does not change at $x^*(\theta)$. Below, we verify that \Cref{asp:smooth-decision-rule} holds using the newsvendor objective in \Cref{ex:mtl-newsvendor} and the CVaR portfolio objective in \Cref{ex:cvar-portfolio} with constraints}. \tyw{Compared with the unconstrained case, we only need to verify that the fourth and fifth conditions in \Cref{asp:smooth-decision-rule} hold.}
%with $\Xscr = \R^{D_x}$ and exponential distribution class with independent margins $\P_{\theta} = \{\theta \in \R^{D_{\theta}}|\prod_{i \in [D_{\theta}]} \text{exp}(\theta_i)\}$. In this case, $p_{\theta}(\xi) = \sum_{i \in [D_{\theta}]}\theta_i \exp(-\sum_{i \in [D_{\theta}]} \xi_i \theta_i)$ and its 1,2,3-th derivative is in $L_1(\xi)$. And $\text{det}(\nabla_{xx}\E_{\P_{\theta}}[h(x;\xi)]) = \prod_{i \in [D_{\theta}]}\Para{p_i \theta_i \exp(-\theta_i x_i)} \neq 0$ for every $\theta$ and $x^*(\theta)$. Therefore, \Cref{asp:smooth-decision-rule} holds. 

{\begin{proposition}[Multi-product Newsvendor with a Capacity Constraint]\label{ex:smooth-newsvendor-capacity}
Consider the multi-product newsvendor objective in Example~\ref{ex:mtl-newsvendor} with $\Xscr=\left\{x\in\mathbb R_+^{D_x}:\sum_{i=1}^{D_x}x_i\leq C
\right\}$. Assume the demand variables $\xi$ are independent exponential random variables with parameters $\theta_i>0$, i.e., $p_\theta(\xi)
=
\prod_{i=1}^{D_x}(\theta_i e^{-\theta_i \xi_i})$, for $\xi\geq \bm{0}$. Then \Cref{asp:smooth-decision-rule} is satisfied if $x_i^*(\theta) > 0$ for all $i$ and $\sum_{i = 1}^{D_x} \log(p_i/c_i)/\theta_i > C$.
\end{proposition}
\textit{Proof of \Cref{ex:smooth-newsvendor-capacity}.} We verify the fourth and fifth conditions in \Cref{asp:smooth-decision-rule}.
% For the first condition of \Cref{asp:smooth-decision-rule}, we can directly check that the distribution family satisfies this condition.

For the fourth condition of \Cref{asp:smooth-decision-rule}, define $D(\theta, x) := \nabla_{xx}^2\E_{\P_\theta}[h(x;\xi)]$. Since $x^*(\theta)>0$, we have $\frac{\partial}{\partial x_i}
\E_{\P_\theta}[h(x;\xi)]
=
c_i-p_i\P_{\theta}(\xi_i\geq x_i)
=
c_i-p_i e^{-\theta_i x_i}$, and $\frac{\partial^2}{\partial x_i^2} \E_{\P_\theta}[h(x;\xi)] =
p_i\theta_i e^{-\theta_i x_i}>0$, $D(\theta, x) = \text{diag}
\left(
p_i\theta_i e^{-\theta_i x_i}
\right)_{i=1}^{D_x}$, which is positive definite for every interior $x$. Consider the following two cases:

(i) The capacity constraint is inactive, then the unconstrained
first-order condition
\begin{equation*}
c_i-p_i e^{-\theta_i x_i}=0,\qquad i=1,\ldots,D_x,
\end{equation*}
defines an optimizer map $x^*(\theta)$ in $C^2$, because $D(\theta,x^*(\theta))$ is nonsingular.

(ii) The capacity constraint is active, then the KKT system is
\begin{equation*}
c_i-p_i e^{-\theta_i x_i}+\lambda=0,
\qquad i=1,\ldots,D_x,
\end{equation*}
together with
\begin{equation*}
\sum_{i=1}^{D_x}x_i=C,
\end{equation*}
where $\lambda$ is the Lagrange multiplier for the capacity constraint. The Jacobian of this KKT system with respect to $(x,\lambda)$ is the KKT matrix
\begin{equation*}
K(\theta,x)
=
\begin{pmatrix}
D(\theta,x) & \mathbf 1\\
\mathbf 1^\top & 0
\end{pmatrix}.
\end{equation*}
We show that $K(\theta,x)$ is nonsingular. If $K(\theta, x)(v, a)^{\top} = 0$, then
\begin{equation*}
D(\theta,x)v+a\mathbf 1=0, \mathbf 1^{\top} v = 0.
\end{equation*}
Multiplying the first equation by $v^\top$ gives $v^\top D(\theta,x)v+a\mathbf 1^\top v = v^\top D(\theta,x)v = 0$.
Because $D(\theta,x)$ is positive definite, this implies $v=0$, and hence $a=0$. Therefore, $K(\theta,x)$ is nonsingular.

For the fifth condition of \Cref{asp:smooth-decision-rule}, from the KKT system, we can compute $x_i^* = \frac{1}{\theta_i}
\log\left(\frac{p_i}{c_i+\lambda}\right)$, where $\lambda$ solves $\sum_{i=1}^{D_x}
\frac{1}{\theta_i}
\log\left(\frac{p_i}{c_i+\lambda}\right)
= C$. The solution satisfies $\lambda > 0$ since the map $\lambda\mapsto \sum_{i=1}^{D_x}
\frac{1}{\theta_i}\log\left(\frac{p_i}{c_i+\lambda}\right)$ is monotonically decreasing and $\sum_{i =1}^{D_x}\log(p_i/c_i)/\theta_i > C$. Therefore, this condition holds. $\hfill \square$

We comment that if some products are stocked at zero, the fifth condition of \Cref{asp:smooth-decision-rule} further requires the multiplier of the nonnegativity constraint $\beta_i = c_i + \lambda - p_i > 0$ whenever $x_i^*(\theta) = 0$, which is less straightforward to satisfy.

\begin{proposition}[CVaR Portfolio Optimization with No-Short-Sale Constraints]\label{ex:smooth-cvar-simplex}
Consider the CVaR portfolio optimization objective in \Cref{ex:cvar-portfolio} with $\Xscr = \left\{
x\in\mathbb R_+^{D_x}:
\mathbf 1^\top x=1
\right\}$ and $\P_{\theta} = N(\theta, \Sigma)$ for some fixed $\Sigma \succ 0$. Let $H_{\theta}(x, v) = \E_{\P_{\theta}}[h((x, v);\xi)]$ and let $\lambda$ be the multiplier corresponding to the constraint $\mathbf 1^{\top} x = 1$. Suppose that $\left[\nabla_x H_{\theta}(x^*(\theta),v^*(\theta))\right]_i+\lambda>0$ whenever $x_i^*(\theta) = 0$. Then \Cref{asp:smooth-decision-rule} is satisfied.
\end{proposition}

\textit{Proof of \Cref{ex:smooth-cvar-simplex}.} We verify the fourth and fifth conditions of \Cref{asp:smooth-decision-rule}. In this problem, the decision variable is $(x,v)$ and the constraints are
\[
\mathbf 1^\top x-1=0,
\qquad
-x_i\leq 0,\quad i=1,\ldots,D_x.
\]

The population CVaR objective $H_\theta(x,v)$ is twice continuously differentiable in $(\theta,x,v)$. Let $(x^*(\theta),v^*(\theta))$ denote a locally unique optimizer. Let $B(\theta) := \{i\in\{1,\ldots,D_x\}:x_i^*(\theta)=0\}$ denote the active no-short-sale constraints. The Lagrangian is
\[
\mathcal L(x,v,\lambda,\nu_B;\theta)
=
H_\theta(x,v)
+
\lambda(\mathbf 1^\top x-1)
-
\sum_{i\in B(\theta)}\nu_i x_i,
\]
where $\lambda$ is the multiplier for the simplex constraint and $\nu_B=(\nu_i)_{i\in B(\theta)}$ are the multipliers for the no-short-sale constraints. The active-set KKT system is
\[
\nabla_x H_\theta(x,v)
+
\lambda \mathbf 1
-
\sum_{i\in B(\theta)}\nu_i e_i
=
0,
\]
\[
\nabla_v H_\theta(x,v)=0,
\]
\[
\mathbf 1^\top x=1,
\qquad
x_i=0,\quad i\in B(\theta).
\]
Equivalently, define the active-set KKT mapping
\[
G(\theta,x,v,\lambda,\nu_B)
=
\begin{pmatrix}
\nabla_x H_\theta(x,v)+\lambda\mathbf 1-\sum_{i\in B(\theta)}\nu_i e_i\\
\nabla_v H_\theta(x,v)\\
\mathbf 1^\top x-1\\
(x_i)_{i\in B(\theta)}
\end{pmatrix}.
\]
For the fourth condition of \Cref{asp:smooth-decision-rule}, we take the derivative of $G$ with respect to $(x, v, \lambda, \nu_B)$ and obtain the KKT matrix:
\[K(\theta)=\begin{pmatrix} Q & A^{\top}\\ A & 0\end{pmatrix},\]
where $Q=\nabla^2_{(x,v)}H_{\theta}(x^*,v^*)$, $A=\begin{pmatrix}\mathbf 1^{\top} & 0\\ E_B^{\top} & 0\end{pmatrix}$, $E_B$ is the matrix whose columns are $\{e_i:i\in B(\theta)\}$, and $(x^*,v^*)=(x^*(\theta),v^*(\theta))$. We now show that $K(\theta)$ is nonsingular. 

To see this, it suffices to show that (i) $A$ has full row rank and (ii) $w^{\top}Qw>0$ for every nonzero $w\in\ker(A)$. When these two conditions hold, if $K(\theta)(w,a)^{\top}=0$, then $Qw+A^{\top}a=0$ and $Aw=0$; multiplying the first equation by $w^{\top}$ gives $w^{\top}Qw=0$. Therefore, $w = 0$ and $a = 0$. 

For (i), since $\mathbf 1^{\top}x^*=1$ and $x_i^*=0$ for $i\in B(\theta)$,
there exists $j\notin B(\theta)$. Here, for any linear combination $c\mathbf{1} + \sum_{i \in B(\theta)}\beta_i e_i = 0$, we have $c = 0$ since $c$ is the $j$-th coordinate of the left-hand side. Then $\beta_i = 0$, implying that $A$ has full row rank.

For (ii), let $Z=-\xi^{\top}x^*\sim N(-\theta^{\top}x^*, x^{*\top}\Sigma x^*)$. Since the mapping $(x,v)\mapsto-\xi^{\top}x-v$ is linear
in $(x,v)$, its Hessian takes the following form:
\[\nabla_{(x,v)}^2 \E_{\P_{\theta}}[(-\xi^{\top}x - v)_+] = p_Z(v^*)\E_{\P_{\theta}}[(\xi;1)^{\top}(\xi;1)| Z = v^*].\]
Therefore, recall $Q = \nabla_{(x,v)}^2 H_{\theta}(x^*, v^*)$ and $h((x,v);\xi) = v + \frac{1}{\epsilon}(-\xi^{\top}x - v)^+$, for any $(d_x,d_v) \in \R^{D_x}\times \R$,
\[
(d_x,d_v)^{\top}Q\,(d_x,d_v)
=\frac{p_Z(v^*)}{\epsilon}\,
 \E\!\left[(\xi^{\top}d_x+d_v)^2 \,\middle|\, Z=v^*\right]
=\frac{p_Z(v^*)}{\epsilon}\!\left[
 \big(\E[\xi\mid Z=v^*]^{\top}d_x+d_v\big)^2+d_x^{\top}\Sigma_c d_x\right],
\]
where $\Sigma_c=\Sigma-\Sigma x^*x^{*\top}\Sigma/(x^{*\top}\Sigma x^*)$ is
the conditional covariance of $\xi$ given $Z=v^*$. Since $\Sigma\succ0$, we have: $\Sigma_c \succ 0$, then $d_x^{\top}\Sigma_c d_x=0$ if and only if $d_x$ is proportional to $x^*$.
However, for $(d_x,d_v)\in\ker(A)$ we have $\mathbf 1^{\top}d_x=0$, whereas
$\mathbf 1^{\top}x^*=1$; hence $d_x=\mathbf{0}$, and the quadratic form reduces to
$p_Z(v^*)d_v^2/\epsilon$, which is positive unless $d_v = 0$. That is, $w^{\top}Qw > 0$ for every $w \in \ker(A)$.

For the fifth condition of \Cref{asp:smooth-decision-rule}, since $\left[\nabla_x H_{\theta}(x^*(\theta),v^*(\theta))\right]_i+\lambda>0$ whenever $x_i^*(\theta) = 0$, it holds as well.$\hfill \square$

In the above derivation, when the strict complementarity condition is satisfied, the active no-short-sale set $B(\theta)$ remains unchanged around the point $\theta$.
%the simplex constraint and active no-shortsale constraints can be handled through the KKT system above. 
Otherwise, if an asset weight switches between zero and positive, $(x^*(\theta),v^*(\theta))$ may fail to be twice differentiable. Such transitions are commonly associated with a failure of strict complementarity or other strong-regularity conditions. These transition cases are not directly covered by our results without a nonsmooth or directionally differentiable extension. 
}

\wty{\subsection{Verification of \Cref{asp:theta-hat}}
\label{subsec:comp}
% Going further, the condition that the first-order error being
% $o_p(n^{-1})$ in \Cref{asp:theta-hat} can be satisfied when the
% iteration number in an iterative optimization algorithm is polynomial
% with respect to $n$.  
%WHICH IS THE EMPHASIS OF ALL THE STOCHASTIC OPTIMIZATION LITERATURE FOCUS ON. 
% Across different methods in
% When the procedures listed in \Cref{ex:ddo} % under enough
% are sufficiently 
% smooth, % conditions,
% if we run
For each procedure listed in \Cref{ex:ddo}, the number of iterations of 
appropriately defined (variants of) first-order stochastic approximation algorithms
% with $K$ iterations, where $K$ is polynomial in $n$,
% the condition of the
% optimality error being
required to achieve
$o_p(n^{-1})$ error (i.e., \Cref{asp:theta-hat}) is polynomial in $n$ as long as the following assumption is satisfied:
%(AGAIN, DO WE HAVE THEOREMS TO JUSTIFY THESE CLAIMS?)
\begin{assumption}[% First-order Error in Computation Procedure
  Computational Complexity for Performance Gap]\label{asp:first-order-comp}
  % For each procedure, consider the solution $x_K^*(\hat\theta)$ as the
  % output of some approximation algorithm with $K$ iterations, then the
  % first-order error is: $\E_{\hat\P_n}[G(x_K^*(\hat\theta);\xi)] =
  % o_p(K^{-\beta})$ for each corresponding $G$ in \Cref{asp:theta-hat}.
  Let $R(\datax)$ denote the appropriate optimality gap for each of the cases in  \Cref{asp:theta-hat} (e.g., $\E_{\hat\P_n}[h(\datax;\xi)] - \min_{x \in \Xscr}\E_{\hat\P_n}[h(\paranx;\xi)]$ in \Cref{asp:theta-hat}$(ii)$). Then there
  exists an iterative algorithm such that the output $x^*_K(\hat\theta)$ after $K$ iterations satisfies $R(x_K^*(\hat\theta)) = o_p(K^{-\beta_1})$ for some $\beta_1 > 0$.
\end{assumption}
% \begin{proposition}[Performance Guarantees with a large $K$]
%     In each procedure, consider the solution $x_K^*(\hat\theta)$ as the
%     output of an approximation algorithm with $K$ iterations,
%     corresponding to the first-order stochastic (sub) gradient descent
%     methods (with $K = \omega(n^4)$) for ETO or E2E or R-E2E; the
%     stochastic bilevel first-order method (with $K = \omega(n^5)$) for
%     IEO, For DR-E2E, applying stochastic gradient descent (with $K =
%     \omega(n^{1/2})$) iterations, then \wty{under some additional smooth
%     conditions (each procedure listed in \Cref{app:apply})},
%     Assumption~\ref{asp:theta-hat} is satisfied when replacing
%     $x^*(\hat\theta)$ with $x_K^*(\hat\theta)$. 
%     % Therefore, all the bias formulas of $A$ and $\hat A$ holds with
%     their properties under other same conditions. \wty{I THINK THIS
%     DETAILS SHOULD BE MOVED TO APPENDIX. DO WE THINK WE NEED TO TO WRITE
%     EXPLICITLY THE DETAILED ALGORITHM.} 
% \end{proposition}
% Each procedure of the detailed stochastic approximation algorithms and
% analytical results can be found in stochastic optimization literature. 
\wty{\Cref{asp:first-order-comp} % , setting
implies that 
$K = \omega(n^{\frac{1}{\beta_1}})$ ensures \Cref{asp:theta-hat} holds for each procedure.}  For
ETO or E2E or R-E2E, standard first-order (sub)-gradient descent methods
satisfy \Cref{asp:first-order-comp}  with $\beta_1 = \frac{1}{2}$ (see
Corollary 2.2 in \citet{ghadimi2013stochastic} for smooth objectives) for IEO, bilevel approximation methods satisfy \Cref{asp:first-order-comp} with $\beta = 1$ (see \cite{ghadimi2018approximation} for convex objectives); for DR-E2E, mirror descent methods satisfy
\Cref{asp:first-order-comp} with $\beta_1 = \frac{1}{2}$
(see~\citet{namkoong2016stochastic})}
%and stochastic gradient descent methods satisfy~\Cref{asp:first-order-comp} with $\beta = \frac{1}{4}$ even for some non-convex objectives (see~\citet{jin2021non}).}
%Despite the requires some convexity.
% to verify Assumption 4, while the validity of M-estimator is tolerable
% such that it allows some approximation error in the computation of
% $o_p(n^{-1})$, e.g., when the first-order optimality condition
% $\nabla_{\theta}\E_{\hat\P_n}[h(x^*(\hat\theta);\xi)] = o_p(n^{-1})$ for
% quantile-based problem (Theorem 5.23 in \cite{van2000asymptotic}) or
% generally $o_p(n^{-1/2})$ (Theorem 5.31 in \cite{van2000asymptotic}).  
%Furthermore, even it is not i.e., condition is satisfied with some tolerant computational error; 
%our bias can be further augmented with some computational error. 
%When we have an optimization problem, our analyses can also be extended to the following cases, 
% Based on 

\wty{Besides, for the unconstrained optimization, one can verify the first-order optimality condition alternatively via \Cref{asp:theta-hat2} as follows. \Cref{asp:theta-hat2} serves as a replacement for Assumption~\ref{asp:theta-hat} and ensures that the conclusions of Theorems~\ref{thm:main} and~\ref{coro:main-moment} still hold. 
\begin{assumption}[Computing $x^*(\hat\theta)$ via First-order Optimality Error]\label{asp:theta-hat2}
    When $\Xscr = \R^{D_x}$, each $\hat\theta$ and $x^*(\hat\theta)$ satisfies the following condition:
    \begin{enumerate}[(i),leftmargin=*]
        \item ETO: $\E_{\hat{\P}_n}[\nabla_{\theta}\phi(\hat\theta;\xi)] =
          o_p(n^{-1/2})$ and $\left.\nabla_{x}\big(\E_{\P_{\hat\theta}}[h(x;\xi)]\big)\right|_{{x = x^*(\hat\theta)}} = o_p(n^{-1/2})$;
        \item IEO / E2E: $\wty{\nabla_{\theta} \E_{\hat\P_n}[h(\datax;\xi)]} = o_p(n^{-1/2})$; 
        \item R-E2E: $\gradp \E_{\hat\P_n}[h(\datax;\xi)] + \lambda \gradp R(x^*(\hat\theta)) = o_p(n^{-1/2})$;
    %     $\gradp \E_{\hat\P_n}[h(\datax;\xi)] + \lambda \gradp R(x^*(\hat\theta)) + \sum_{j \in \hat{B}}
    % \hat{\alpha}_j \nabla_{\theta} F_j(\datax)= o_p(n^{-1})$;
        \item DR-E2E: 
        %$\sup_{d(\P, \hat\P_n)\leq \epsilon}\E_{\P}[h(\datax;\xi)] = \min_{\theta \in \Theta}\sup_{d(\P, \hat\P_n)\leq \epsilon} \E_{\hat\P}[h(\paranx;\xi)] + o_p(n^{-1})$.
        $\gradp \sup_{d(\P, \hat\P_n)\leq \epsilon} \E_{\hat\P_n}[h(\datax;\xi)]  = o_p(n^{-1/2})$.
    \end{enumerate}
\end{assumption}
\Cref{asp:theta-hat2} is easier to satisfy for the unconstrained nonconvex optimization problem in the literature:
\begin{assumption}[% First-order Error in Computation Procedure
  Computational Complexity for First-order Optimality Error]\label{asp:first-order-comp2}
  % For each procedure, consider the solution $x_K^*(\hat\theta)$ as the
  % output of some approximation algorithm with $K$ iterations, then the
  % first-order error is: $\E_{\hat\P_n}[G(x_K^*(\hat\theta);\xi)] =
  % o_p(K^{-\beta})$ for each corresponding $G$ in \Cref{asp:theta-hat}.
  Let $G(x^*(\hat\theta))$ denote the appropriate first-order error for each of the cases in  \Cref{asp:theta-hat2} (e.g., $\nabla_{\theta}\E_{\hat\P_n}[h(\datax;\xi)]$ in \Cref{asp:theta-hat2}$(ii)$). Then there exists an iterative algorithm such that the output $x^*_K(\hat\theta)$ after $K$ iterations satisfies $G(x_K^*(\hat\theta))= o_p(K^{-\beta_2})$ for some $\beta_2 > 0$.
\end{assumption}
\wty{\Cref{asp:first-order-comp2} % , setting
implies that 
$K = \omega(n^{\frac{1}{2\beta_2}})$ ensures \Cref{asp:theta-hat2} is
satisfied for each procedure.} Besides the references mentioned above, for
ETO or E2E or R-E2E, standard first-order (sub)-gradient descent methods
satisfy \Cref{asp:first-order-comp2}  with $\beta_2 = \frac{1}{4}$ (see Theorem 2.5 in \citet{davis2018stochastic} for nonsmooth weakly-convex
functions), for IEO, stochastic bi-level first-order methods satisfy
\Cref{asp:first-order-comp2} with $\beta_2 = \frac{1}{5}$ (see Corollary 4.2
(b) in \cite{kwon2023fully}); for DR-E2E, stochastic gradient descent
methods 
satisfy~\Cref{asp:first-order-comp2} with $\beta_2 = \frac{1}{4}$ even for
some non-convex objectives (see~\citet{jin2021non}).}

\wty{Certainly, Assumption~\ref{asp:theta-hat} or~\ref{asp:theta-hat2}  is one
sufficient condition that our theoretical results are valid. Even if the
iteration number $K$ is too small such that the $o_p(n^{-1})$ error
condition holds, our bias % formulation
expressions may still remain valid for
the % approximate solution
iterate
$x_K^*(\hat\theta)$ when the optimization procedure satisfies the following assumption.}  
% Below, 
%In many scenarios, if we solve the optimization problem based on SGD, then 
% In this case, we can incorporate such variability of computational error due to the stochastic gradient descent into the bias framework. 
% %This helps obtain a per-step evaluation to add 
% %For those computational issues, 
% %if we have $T$ rounds to compute, we consider SGD bias (Liwei and JohnDuchi-Ruanfengwhich is the computational bias into the whole evaluation framework. 
% However, the obtained bias is different, depending on where the optimization computational challenge occurs:
\wty{\begin{assumption}[Central Limit Theorem in Computation
  Procedure]\label{asp:clt-computation}
  % Let $\E_{\hat\P_n}[G(x^*(\hat\theta);\xi)]$ denote the appropriate first-order error
  % condition for each case in  \Cref{asp:theta-hat}, and 
  Let $x_E^*(\hat\theta)$ denote the
  exact solution of $T(\hat\P_n)$. There exists an
  iterative algorithm such that the $K$-th iterate $x_K^*(\hat\theta)$ satisfies:
  % Conditioned on the realization of $\Dscr_n$ (where $\datax$ denotes the exact solution such that the right-hand side becomes zero in \Cref{asp:theta-hat}), for some $H(\hat\theta)$ and approximated solution $x_K^*(\hat\theta)$, we have:
  \begin{equation}\label{eq:sgd-asymptotics}  
    \E[x_K^*(\hat\theta) - x_E^*(\hat\theta)|x_E^*(\hat\theta)] = o(1/K),~\sqrt{K}(x_K^*(\hat\theta)- x_E^*(\hat\theta)) \overset{d}{\to} N(0, H(\hat\theta)),
  \end{equation}
  where the expectation and probability is taken with respect to the randomization of the iterative algorithm and $H(\hat{\theta})$ is some positive semidefinite covariance matrix.
\end{assumption}
% We describe how
The standard stochastic gradient descent
procedure~\citep{robbins1951stochastic} % its within this framework.
appropriately applied to the procedures in \Cref{ex:ddo} satisfies
\Cref{asp:clt-computation}. 
Specifically, at the $k$-th iteration, we randomly sample one point from
$\hat\P_n$ (in \Cref{ex:ddo}$(b)$) or $\hat\P_{\hat\theta}$ (in
\Cref{ex:ddo}$(a)(1)$) and perform a gradient descent step with
step length of the order $\frac{1}{k}$.
%in solving the optimization problem. 
{For ETO,~\eqref{eq:sgd-asymptotics} holds with $H(\hat\theta) = \nabla_{xx} \E_{\P_{\theta}}[h(x_E^*(\hat\theta);\xi)] \text{Cov}_{\xi \sim \P_{\hat\theta}}(\nabla_x h(x_E^*(\hat\theta);\xi)) \nabla_{xx} \E_{\P_{\theta}}[h(x_E^*(\hat\theta);\xi)]$ when
% suppose the computational challenge arises in solving $x^*(\hat\theta) \in \argmin_{x \in \Xscr}\E_{\P_{\hat\theta}}[h(x;\xi)]$ and we only perform $K$ iterations of gradient descent. 
% % Then the computational challenge arises in solving: $x^*(\hat\theta) \in \argmin_{x \in \Xscr}\E_{\P_{\hat\theta}}[h(x;\xi)]$.
% More specifically, we 
$x_K^*(\hat\theta)$ % s the average of the solutions amon
is defined as the running average 
$K$ iterates of the stochastic gradient descent % when computing
% $x^*(\hat\theta)$.
applied to an appropriate cost function. 
The results for the standard strongly convex and smooth functions were
established in \cite{polyak1992acceleration} and generalized to nonsmooth
functions in \cite{davis2024asymptotic}. For E2E,~\eqref{eq:sgd-asymptotics}
also holds for  $x_K^* = x^*(\hat\theta_K)$ where 
$\hat\theta_K$ % being the average of the solutions among
is the average of the first $K$ iterates when a stochastic gradient
algorithm is used to compute $\hat\theta$. 
%and~\eqref{eq:sgd-asymptotics} still holds.
%Regardless of each scenario when $K = \omega(n)$, 
% When $K = \omega(n)$, we can show that
The following result shows that 
the expected bias and variability
term of OIC remains unchanged when replacing $x^*(\hat\theta)$ with
$x_K^*(\hat\theta)$ for $K = \omega(n)$ when \ref{asp:clt-computation} holds.
% However, the asymptotic normality in~\eqref{eq:sgd-asymptotics} introduces additional variance when evaluating the performance of $x_K^*(\hat\theta)$, specifically, $\hat A_c(x^*(\hat\theta)) - \hat A_c(x_K^*(\hat\theta)) = O_p(\frac{1}{\sqrt{K}})$. 
\begin{proposition}[Performance Guarantee with a small $K$]\label{prop:compare-smallK}
Suppose Assumptions~\ref{asp:nonsmooth},~\ref{asp:moment},~\ref{asp:clt-computation}, Equations~\eqref{eq:theta-if} and~\eqref{eq:empirical-if} hold. Then the conclusion of
Theorem~\ref{thm:main} hold when $x^*(\hat\theta)$
in~\eqref{eq:oic} is replaced by $x_K^*(\hat\theta)$ provided $K = \omega(n)$.
\end{proposition}
% \hl{Is this correct? Did you want $\omega(n^2)$?} \wty{The bias is small order compared with the expected performance?}
Intuitively, the result holds since conditioned on $\Dscr_n$, the stochastic term $x_K^*(\hat\theta) - x^*(\hat\theta) = O_p(1/\sqrt{K})$ under \Cref{asp:clt-computation}. The stochastic term $x_K^*(\hat\theta) - x^*(\hat\theta)$ is independent with the realization of $\Dscr_n$ and becomes negligible compared to the order of $A_c$ when $K  = \omega(n)$.}}

\wty{\textit{Proof of \Cref{prop:compare-smallK}.} In the proof of \Cref{thm:main}, we take the second-order Taylor expansion and expand $x_K^*(\hat\theta)$ with respect to $x^*(\theta^*)$. In this case, 
\[x_K^*(\hat\theta) - x^*(\theta^*) = \Para{x_K^*(\hat\theta) - x^*(\hat\theta)} + \Para{x^*(\hat\theta) - x^*(\theta^*},\]
where the first term in RHS above is due to the sampling uncertainty in the iterate algorithm conditioned on $\Dscr_n$ and the second term in RHS is because each realization of $\Dscr_n$ is random. Conditional on $\Dscr_n$, these two terms in RHS are independent. For the additional terms in~\eqref{eq:t1-expand} and~\eqref{eq:t2-expand} due to the iterate algorithm (in the proof of \Cref{thm:main}), conditioned on $\Dscr_n$, from \Cref{asp:clt-computation}, we have: 
\begin{align*}
    & \E[\|x_K^*(\hat\theta) - x^*(\hat\theta)\|_2^2] = \Theta(1/K^2) = o(1/n),\\
    & \E[x_K^*(\hat\theta) - x^*(\hat\theta)] = o(1/K) = o(1/n),\\
    & \E[(x_K^*(\hat\theta) - x^*(\hat\theta))^{\top}(x^*(\hat\theta) - x^*(\theta))] = o(1/n),
\end{align*}
Therefore, the conclusions of \Cref{thm:main} hold since these additional terms are $o(1/n)$ if we focus on the bias of the order $\Theta(1/n)$. $\hfill \square$}

\subsection{Discussion on Alternative Estimators}\label{app:btsp-jcnf}
% We first list the missing details in \Cref{sec:discuss}.
% \textbf{Cross Validation.} The bias of K-Fold CV estimator $(R_{KCV})$ would be:
% \[\E_{\Dscr_n}\E_{\P^*}[h(\datax;\xi)] = \E[R_{KCV}] - \frac{B}{(K - 1)n},\]
% that is why $K$-fold CV generally overestimates the bias. This is obviously seen from $\E[R_{KCV}] = \E_{\Dscr_{(K - 1)n/K}}\E_{\P^*}[h(x^*(\hat{\theta}(\Dscr_{(K - 1)n/K}));\xi)] = \E_{\Dscr_n}\E_{\P^*}[h(\bestx;\xi)] + \frac{B(K-1)}{Kn} + o\Para{\frac{1}{n}}$.
Besides cross-validation, we discuss the computational procedure of the bootstrap and jackknife methods:
%given in \Cref{thm:main} and existing literature naturally. 
%No matter which method is used, recall $\hat{\theta}$ as the parametric estimator using all the samples $\Dscr_n = \{\xi_i\}_{i = 1}^n$. 

%And we give the following debiasing result of the jackknife approach under such case. 
\textbf{Bootstrap.} Denote the number of bootstrap replications as $B_T$. For $b \in [B_T]$, we resample $n$ samples with replacement independently from $\Dscr_n$ as $\{\xi_{b,i}\}_{i \in [n]}$, and compute the corresponding estimator $\hat{\theta}_b$. Then we obtain:
\[\hat{A}_B = 2\hat{A}_o - \frac{1}{B_T}\sum_{b = 1}^{B_T}\hat{A}_{o,b},\]
where $\hat{A}_{o, b} = \frac{1}{n}\sum_{i = 1}^n h(x^*(\hat{\theta}_b);\xi_{b,i}), \forall b \in [B_T]$.

\textbf{Jackknife.} Recall $\hat{\theta}_{-i}$ as the parametric estimator using samples $\Dscr_n \backslash \{\xi_i\}$. Then we obtain:
\[\hat{A}_J = n \hat{A}_o - \frac{n - 1}{n}\sum_{k = 1}^n A_{o,k},\]
where $\hat{A}_{o,k} = \frac{1}{n - 1}\sum_{i \in [n], i \neq k} h(x^*(\hat{\theta}_{-k});\xi_i), \forall k \in [n]$.

Since $\E[\hat{A}_o] = A + \frac{\E_{\P^*}[\nabla_{\theta}h(\bestx;\xi)^{\top}\IFx(\xi)]}{n}  + o\Para{\frac{1}{n}}$ from \Cref{thm:main}, we obtain the following result, which follows directly from \cite{quenouille1956notes,hall1994methodology}:
\begin{lemma}[Debiasing effects of Bootstrap and Jackknife Approaches]\label{lemma:debias}
The bootstrap estimator $\hat{A}_B$ and jackknife estimator $\hat{A}_J$ satisfy: $\E[\hat{A}_{B}] = A + o\Para{\frac{1}{n}}, \E[\hat{A}_{J}] = A + o\Para{\frac{1}{n}}$.
\end{lemma}

\subsection{Comparison with Debiasing in Linear Optimization}\label{subsec:compare-linear}

%When the objective is linear in the decision variable,
%e.g., $h(x;\xi) = \xi^{\top} u(x)$,  remove the evaluation bias through Stein's lemma or sensitivity analysis \wty{from Danskin's Theorem}. 

% To explain the above,
\wty{In order to carefully discuss the issue between linear and nonlinear optimization, 
we first review the existing methods and results for debiasing linear
optimization. For simplicity, we consider the standard linear optimization
problem and describe one representative debiasing procedure synthesized
from the results in \cite{ito2018unbiased} and \cite{gupta2022debiasing}. In particular,
we restate the non-asymptotic bounds % derived
in \cite{gupta2022debiasing}
and the asymptotic bias bounds in \cite{ito2018unbiased} to our setup in
% this paper
order 
to facilitate comparison with our results.} 
\wty{\begin{lemma}[Bias and Variance for Debiasing Data-Driven Linear Optimization]\label{lemma:debias-linear}
  % Consider
  Suppose
  $h(x;\xi) = \xi^{\top}x$ and $\Xscr$ is a polyhedron. Then % across
  for 
  optimization procedures in \Cref{ex:ddo}$(a)(1)$ with $T(\P) = \E_{\P}[\xi]$, consider $S$ perturbed current solutions $\{x^*(\hat\theta_{\lambda}^{(j)})\}_{j \in [S]}$,  with $\hat\theta_{\lambda}^{(j)} \sim N (\hat\theta, (\lambda^2 + 2\lambda)\Sigma/n)$ with $\lambda$ being the perturbation step size and $\Sigma$ being the
asymptotic variance of $\hat\theta$ such that $\sqrt{n}(\hat\theta -
\theta^*) \overset{d}{\to} N(0, \Sigma)$.
  % \hl{What is $s$? Where is it defined?}
  We compute the bias term as:  
  \begin{equation}\label{eq:linear-debias}
    \hat A_{c,\lambda} = \frac{1}{\lambda S}\sum_{j \in [S]}\Para{(\hat\theta_{\lambda}^{(j)})^{\top} x^*(\hat\theta_{\lambda}^{(j)}) - \hat\theta^{\top}\datax}, 
\end{equation}
Then for $\hat A_{\lambda} =
\hat\theta^{\top}x^*(\hat\theta)+ \hat A_{c,\lambda}$, we have: $\E[\hat
A_{\lambda}] = A + O(\lambda)$ and $\text{Var}[\hat A_{c,\lambda}] =
O\Para{\frac{1}{n^2 \lambda^2}}$.
% \hl{The paragraph above is not clear and needs to be rephrased.} \wty{TW: done.}
\end{lemma}
%One can show that 
For $\lambda$ of order $n^{-2/3}$, 
%we can ensure 
the bias of the resulting $\hat A$, i.e., $\E[\hat A_{\lambda}] - A$, is $o(n^{-1/2})$ while the variability term satisfies $\text{Var}[\hat A_{c,\lambda}] = \text{Var}[\hat A_o](1 + o(1/n))$. This means that the dominating term of the bias can be removed while ensuring a similar variability to the empirical estimate.}
%\[O(h) + O\Para{\frac{1}{n^2 h}} = o(n^{-1/2}),\]
%This perturbation removes the dominating term of the bias and achieves the goal of exceeding the variability compared with the empirical estimate.  

% Moreover, these proposed methods
% require solving additional optimization problems,  %\wty{TODO: add clarification} 

\wty{Comparing \Cref{lemma:debias-linear} with our results, we note that
  there are three significant differences between linear and (smooth)
  nonlinear optimizations problems: 
\begin{enumerate}[(i)]
    \item The expressions for the % statistical
      bias and variance
      in~\eqref{eq:linear-debias} % (relying on the
      are functions of the 
      step size $\lambda$, which are
      different from those for  nonlinear
      optimization problems (see \Cref{thm:main}). This is because in the linear optimization
      setting, the second-order derivative of the cost function is zero and, therefore, the
      solution $x^*(\hat\theta)$ % breaks the
      % asymptotic normality for
      is no longer asymptotically normal. A non-zero second-order derivative is also required to remove the
      bias of cross-validation~\citep{fushiki2011estimation}. 
      % to remove their bias
      % ~\citep{fushiki2011estimation}. 
      As noted in
      Theorem 5.1 in~\cite{gupta2021small}, the standard properties of
      cross-validation do not necessarily hold for linear optimization. 
    \item The bias removal procedures require different
      techniques. % depending on the properties of the objective
      % function. 
      The debias % properties in
      result~\eqref{eq:linear-debias} leverages
      Stein's lemma~\citep{gupta2021small} or sensitivity
      analysis~\citep{ito2018unbiased,gupta2022debiasing} from Danskin's
      Theorem. % , utilizing decoupled structures.
      These methods are
      well-suited for decision rules for  linear cost objectives
      or weakly coupled objectives in general, where the estimated uncertainty
      % from data $\Dscr_n$
      and decisions are decoupled or the problem can be reformulated into a linear objective with one auxiliary variable and additional constraint, i.e., Section 4.1 in \cite{gupta2022debiasing}. However, these techniques do not apply to % our settings of 
      general nonlinear
      objectives, e.g., the newsvendor problem and the portfolio
      optimization problem with exponential utility in
      \Cref{ex:portfolio}. % For general nonlinear objectives, we apply
      Second-order Taylor series expansion and influence functions used to debias nonlinear objectives do not work in the linear setting. % to
      % remove the bias. 
      % Notably
      Thus, neither approach is effective for the
      other class of objectives. 
    \item % Thirdly,
      Computing
      $\hat A_{c,\lambda}$ in~\eqref{eq:linear-debias}
      requires solving additional optimization problems even when $S = 1$
      \citep{gupta2024decision}, i.e., solving two optimization problems to evaluate the in-sample bias for a fixed policy. % as opposed to our explicit
      % characterization of the bias based on direct estimation from
      % $\hat\theta$.
      The nonsmoothness of linear optimization makes
      evaluating the true performance of the linear
      optimization solution % is challenging 
      without solving additional
      optimization problems % due to.
      challenging.
      More specifically, a
      single sample can % lead to significant changes in
      significantly impact
      $x^*(\hat\theta)$, and therefore, 
      % making the perturbation 
      difference $x^*(\hat\theta) - x^*(\theta^*)$
      cannot be reliably 
      % unreliable when approximating it with the
      approximated by an
      influence
      function. Consequently, the current solution $x^*(\hat\theta)$ is
      ineffective as an approximation of the true optimal solution. 
\end{enumerate}}

\section{Proofs in Section~\ref{sec:main}}\label{sec:proofs main}

Before going to the main proofs, we present the following technical lemma:
\begin{lemma}\label{lemma:matrix-expectation}
If $x \in \R^d$ is a random vector, $B \in \R^{d\times d}$ is a deterministic matrix, then $\E[x^{\top}B x] = \text{Tr}[BC]$, where $C = \E[x x^{\top}]$. 
\end{lemma}
We introduce several notations throughout this section:
\begin{equation}\label{eq:bias-notation}
    \begin{aligned}
    \hat A_o &= \frac{1}{n}\sum_{i = 1}^n h(x^*(\hat{\theta});\xi_i),\\
    A_c & = - \frac{\E_{\P^*}[\gradp h(\bestx;\xi)^{\top} \IFx(\xi)]}{n},\\
    \hat A_c^* &= -\frac{1}{n^2}\sum_{i = 1}^n \gradp h(\bestx;\xi_i)^{\top} \IFx(\xi_i),\\
    \hat A_c & =-\frac{1}{n^2}\sum_{i = 1}^n \gradp h(\datax;\xi_i)^{\top} IF_{\hat\theta}(\xi_i).
\end{aligned}
\end{equation}

In the following, to show Theorems~\ref{thm:main} and~\ref{coro:main-moment} hold for each procedure in \Cref{ex:ddo}, we first show the results on the basis of Lemmas~\ref{asp:theta-asymptotics} and~\ref{asp:empirical-if} in Appendix~\ref{app:main-1} and~\ref{app:main2} (In fact, we show a stronger version, i.e., \Cref{prop:oic-general} holds). Then in \Cref{app:apply}, we show Lemmas~\ref{asp:theta-asymptotics} and~\ref{asp:empirical-if}.

\subsection{Proof of Theorem~\ref{thm:main}}\label{app:main-1}
We divide the proof into two cases depending on the smoothness of the composite function $h(x;\xi)$:
\subsubsection{Cases when $f(z) = z$ in \Cref{asp:nonsmooth}}
Recall the notations in \eqref{eq:bias-notation}. 
%\Cref{thm:main} is equivalent to showing $\E[\hat{A}_o + \hat A_c^*] = A + o\Para{\frac{1}{n}}$. 
Notice that $\hat A_c^*$ is an unbiased estimator of $A_c$ ($\E[\hat{A}_c^*] = A_c$) since each $\xi_i$ is i.i.d. sampled from $\P^*$. Therefore, the conclusion of \Cref{thm:main} is equivalent to the following equation:
\begin{equation}\label{eq:general-eq}
    \E[\hat A_o + \hat A_c^*] = A+ o\Para{\frac{1}{n}}.
\end{equation}
We show this equation holds through the following two steps.

\textbf{Step 1: Translating bias to optimality gaps.}~For any solution $\datax$, we can decompose its true performance $A(:=\E_{\Dscr_n}\E_{\P^*}[h(\datax;\xi)])$ to:
\begin{equation*}
    \begin{aligned}
        A &= A' + \E_{\P^*}[h(\bestx;\xi_i)] - \frac{1}{n}\sum_{i = 1}^n h(\bestx;\xi_i),\\
    \end{aligned}
\end{equation*}
where $A' = \E_{\Dscr_n}\E_{\P^*}[h(x^*(\hat{\theta});\xi) - h(x^*(\theta^*);\xi)] + \frac{1}{n}\sum_{i = 1}^n h(x^*(\theta^*);\xi_i)$. We have $\E[A] = \E[A']$ by:
\begin{equation}\label{eq:unbias-best}
    \E\Paran{\frac{1}{n}\sum_{i = 1}^n h(\bestx;\xi_i)} - \E_{\P^*}[h(\bestx;\xi)]  = \E_{\P^*}[h(\bestx;\xi)]  - \E_{\P^*}[h(\bestx;\xi)]  = 0.
\end{equation}
Therefore, to show~\eqref{eq:general-eq}, it is equivalent to showing that $\E[\hat A_o + \hat A_c^* - A'] = o\Para{\frac{1}{n}}$. First, we further decompose $A'$:
\begin{equation}
    \begin{aligned}
        A' & = \frac{1}{n}\sum_{i = 1}^n h(x^*(\hat{\theta});\xi_i) + \underbrace{\frac{1}{n}\sum_{i = 1}^n h(x^*(\theta^*);\xi_i)  - \frac{1}{n}\sum_{i = 1}^n h(x^*(\hat{\theta});\xi_i)}_{T_1}\\
&+\underbrace{\E_{\Dscr_n}\E_{\P^*}[h(x^*(\hat{\theta});\xi) - h(x^*(\theta^*);\xi)]}_{T_2}.\\
    \end{aligned}
\end{equation}
Denote the two terms $T_1$ and $T_2$ as the sample ``optimistic gap'' and true ``optimistic gap'' respectively. Here $T_1$ is random and $T_2$ is deterministic. Then showing \eqref{eq:general-eq} is equivalent to showing that: 
\[\E[T_1] + T_2 = \E[\hat{A}_c^*] + o\Para{\frac{1}{n}}.\]
 
\textbf{Step 2: Analyses on optimality gaps.}~We analyze the two terms $T_1$ and $T_2$ respectively.

For the term $T_1$, following a second-order Taylor expansion with the Peano's remainder at the center $\theta^*$, we have:
\begin{equation}\label{eq:t1-expand}
    \begin{aligned}
        T_1 & = \frac{1}{n}\sum_{i = 1}^n \gradp h(\bestx;\xi_i)^{\top}(\theta^* - \hat{\theta}) - \frac{1}{2}(\theta^* - \hat{\theta})^{\top}\Para{\frac{1}{n}\sum_{i = 1}^n \hessianp h(\bestx;\xi_i)} (\theta^* - \hat{\theta}) + o\Para{\|\theta^* - \hat\theta\|^2}.
    \end{aligned}
\end{equation}
Then we plug the asymptotic expression of $\hat{\theta} - \theta^*$ in~\Cref{asp:theta-asymptotics} and take expectation over $T_1$. Recall that $\E[\|\hat{\theta} - \theta^*\|_2^2] = O\Para{\frac{1}{n}}$. Then we have:
\begin{equation}\label{eq:oic-t1}
    \E[T_1] = \E\Paran{-\frac{1}{n}\sum_{i = 1}^n \nabla_{\theta} h(\bestx;\xi_i)^{\top}\Para{\frac{1}{n}\sum_{i = 1}^n \IFx(\xi_i) + \Delta}} - \frac{1}{2n}\text{Tr}[I_h(\theta^*)\Psi(\theta^*)] + o\Para{\frac{1}{n}},\\
\end{equation}
where $\Delta = \hat{\theta} - \theta^* - \frac{1}{n}\sum_{i = 1}^n \IFx(\xi_i)$ denotes the higher order term of the estimation error with $\E[\Delta \Delta^{\top}] = o\Para{\frac{1}{n}}$ from \Cref{asp:theta-asymptotics}. 
The second term of the right-hand side in \eqref{eq:oic-t1} follows by:
\begin{equation}\label{eq:second-order-asymp-theta}
\begin{aligned}
    &\quad \E\Paran{(\theta^* - \hat{\theta})^{\top}\E_{\hat{\P}_n}[\hessianp h(\bestx;\xi)](\theta^* - \hat{\theta})}\\
    & = \E\Paran{(\theta^* - \hat{\theta})^{\top}\E_{\P^*}[\hessianp h(\bestx;\xi)](\theta^* - \hat{\theta})} + o\Para{\frac{1}{n}} \\
    & = \frac{1}{n}\text{Tr}[I_h(\theta^*) \Psi(\theta^*)]+ o\Para{\frac{1}{n}},
\end{aligned}
\end{equation}
where the first equality above in \eqref{eq:second-order-asymp-theta} follows the Holder inequality that:
\begin{align*}
    & \quad \E_{\Dscr_n}\Paran{(\theta^* - \hat{\theta})^{\top}|(\E_{\hat{\P}_n} - \E_{\P^*})[\hessianp h(\bestx;\xi)]|(\theta^* - \hat{\theta})} \\
    & \leq \|\|\hat\theta - \theta^*\|^2\|_{L^1, \Dscr_n} \|(\E_{\hat{\P}_n} - \E_{\P^*})[\hessianp h(\bestx;\xi)]\|_{L^{\infty}, \Dscr_n} \leq o(1) \E_{\Dscr^n}[\|\hat\theta  - \theta^*\|_2^2] = o\Para{\frac{1}{n}},
\end{align*}
where the second inequality follows from: $\E_{\hat{\P}_n}[\hessianp h(\bestx;\xi)] \overset{a.s.}{\to} \E_{\P^*}[\hessianp h(\bestx;\xi)]$ such that $\|(\E_{\hat{\P}_n} - \E_{\P^*})[\hessianp h(\bestx;\xi)]\|_{L^{\infty}} = o(1)$ and the last equality follows from $\E[\|\hat{\theta} - \theta^*\|_2^2] = O\Para{\frac{1}{n}}$. 

The second equality above in \eqref{eq:second-order-asymp-theta} follows by observing $\|\E_{\Dscr_n}\Paran{(\hat{\theta} - \theta^*) (\hat{\theta} - \theta^*)^{\top}} - \Para{\frac{\Psi(\theta^*)}{n}}\|_1 = o\Para{\frac{1}{n}}$ in \Cref{asp:theta-asymptotics}, where $\|\cdot\|_1$ is the 1-norm of matrix induced by the vector norm such that $\E[(\hat{\theta} - \theta^*)(\hat{\theta} - \theta^*)^{\top}] = \frac{\Psi(\theta^*)}{n} + o\Para{\frac{1}{n}}$ and~\Cref{lemma:matrix-expectation}. 
% Then we compare the difference between the left-hand side in~\eqref{eq:second-order-asymp-theta} and the second term in the right-hand side in~\eqref{eq:t1-expand} under expectation, i.e.:
% \begin{equation}
%     \begin{aligned}
%         & \E\Paran{(\theta^* - \hat{\theta})^{\top}\Para{\frac{1}{n}\sum_{i = 1}^n \hessianp h(\bestx;\xi_i) - \E_{\P^*}[\hessianp h(\bestx;\xi)]} (\theta^* - \hat{\theta})}\\
%         & = \E\Paran{(\theta^* - \hat{\theta})^{\top}\Para{\E_{\P_n}[\hessianp h(x^*(\tilde{\theta});\xi_i)] - \E_{\P_n}[\hessianp h(\bestx;\xi)]} (\theta^* - \hat{\theta})}  + o\Para{\frac{1}{n}}\\
%         & = o\Para{\frac{1}{n}} + o\Para{\frac{1}{n}} = o\Para{\frac{1}{n}},
%     \end{aligned}
% \end{equation}
% where the first equality follows by noticing $\E_{\P_n}[\hessianp h(\bestx;\xi)] = \E_{\P^*}[\hessianp h(\bestx;\xi)] + O_p\Para{\frac{1}{\sqrt{n}}}$ by central limit theorem. And the second equality follows by the definition of $\tilde{\theta} \convp \theta^*$ as $n \to \infty$, which implies~\eqref{eq:second-order-asymp-theta}.

For the second term in $T_2$, following a second-order Taylor expansion at the center $\theta^*$, we have:
\begin{equation}\label{eq:t2-expand}
\begin{aligned}
    T_2 &= \E_{\P^*}[\nabla_{\theta}h(x^*(\theta^*);\xi)]^{\top}\E\Paran{\frac{1}{n}\sum_{i = 1}^n \IFx(\xi_i) + \Delta}+ \frac{1}{2n}\text{Tr}[I_h(\theta^*)\Psi(\theta^*)] + o\Para{\frac{1}{n}}\\
    & = \E_{\P^*}[\nabla_{\theta}h(x^*(\theta^*);\xi)]^{\top} \E_{\Dscr_n}[\Delta]+ \frac{1}{2n}\text{Tr}[I_h(\theta^*)\Psi(\theta^*)] + o\Para{\frac{1}{n}},
\end{aligned}
\end{equation}
where the second equality in \eqref{eq:t2-expand} is due to $\E[\IFx(\xi)] = 0$. Combining the results of~\eqref{eq:oic-t1} and~\eqref{eq:t2-expand}, we have:
\begin{equation}\label{eq:t1-t2}
    \begin{aligned}
        \E[T_1] + T_2  & = \E\Paran{-\frac{1}{n}\sum_{i = 1}^n \nabla_{\theta} h(\bestx;\xi_i)^{\top}\frac{1}{n}\sum_{i = 1}^n \IFx(\xi_i)} -\E\Paran{\frac{1}{n}\sum_{i = 1}^n \nabla_{\theta} h(\bestx;\xi_i)^{\top} \Delta}\\ &\quad + \E_{\P^*}[\nabla_{\theta}h(x^*(\theta^*);\xi)]^{\top} \E_{\Dscr_n}[\Delta]\\
        & = -\E\Paran{\frac{1}{n}\sum_{i = 1}^n \nabla_{\theta} h(\bestx;\xi_i)^{\top}\frac{1}{n}\sum_{i = 1}^n \IFx(\xi_i)} + o\Para{\frac{1}{n}}.
    \end{aligned}
\end{equation}
Denote $g(\xi_i) := \nabla_{\theta}h(x^*(\theta^*);\xi_i)$. Then the second equality in \eqref{eq:t1-t2} is proved in the following:
%from the following result (here 
\begin{equation}\label{eq:if12}
\begin{aligned}
    &\E_{\P^*}[\nabla_{\theta}h(x^*(\theta^*);\xi)]^{\top} \E_{\Dscr_n}[\Delta] - \E_{\Dscr_n}\Paran{\frac{1}{n}\sum_{i = 1}^n \nabla_{\theta} h(\bestx;\xi_i)^{\top} \Delta}\\
    =& \E_{\Dscr_n}\Para{\Paran{\E_{\P^*}[\nabla_{\theta}h(x^*(\theta^*);\xi)] - \E_{\hat{\P}_n}[\nabla_{\theta}h(x^*(\theta^*);\xi)]}^{\top}\Delta}\\
    \leq & \sqrt{\E_{\Dscr_n}[\|\E_{\P^*}[g(\xi)] - \E_{\hat{\P}_n}[g(\xi)]\|_2^2]}\sqrt{\E_{\Dscr_n}[\|\Delta\|_2^2]}\\
    = & O\Para{\frac{1}{\sqrt{n}}}\sqrt{ \E_{\Dscr_n}\Para{\text{Tr}[\Delta\Delta^{\top}]}} =o\Para{\frac{1}{n}}.
\end{aligned}
\end{equation}
where the first inequality in \eqref{eq:if12} follows by $\E[x^{\top}y] \leq \E[\|x\|_2\|y\|_2] \leq \sqrt{\E[\|x\|_2^2]}\sqrt{\E[\|y\|_2^2]}$. Recall the mean-squared error of the sample mean is $O\Para{\frac{1}{n}}$. Then the second inequality in \eqref{eq:if12} follows by:
\[\E_{\Dscr_n}\Paran{\|\E_{\P^*}[g(\xi)] - \E_{\hat{\P}_n}[g(\xi)]\|_2^2} = O\Para{\frac{1}{n}}.\]
The last equality in \eqref{eq:if12} that bounds $\text{Tr}[\Delta \Delta^{\top}]$ follows by $\E_{\Dscr_n}(\text{Tr}[\Delta \Delta^{\top}]) = \text{Tr}\Paran{\E_{\Dscr_n}[\Delta \Delta^{\top}]} = o\Para{\frac{1}{n}}$ from \Cref{asp:theta-asymptotics}.
%And the second inequality follows by:
%the first term $\|\E_{\P^*}[\nabla_{\theta}h(x^*(\theta^*);\xi)] - \E_{\hat{\P}_n}[\nabla_{\theta}h(x^*(\theta^*);\xi)]\|_{\infty}$ is bounded by Chebyshev inequality such that $\sqrt{\E_{\Dscr_n}[\|\E_{\P^*}[\nabla_{\theta}h(x^*(\theta^*);\xi)] - \E_{\hat{\P}_n}[\nabla_{\theta}h(x^*(\theta^*);\xi)]\|_{\infty}^2]} = O\Para{\frac{1}{\sqrt{n}}}$ while the second term is of order $O\Para{\frac{1}{n}}$ given by \Cref{asp:theta-asymptotics}. 
Therefore, the left-hand side of~\eqref{eq:if12} is bounded by $o\Para{\frac{1}{n}}$ and the second equality in \eqref{eq:t1-t2} holds.

Going back to the first term in the right-hand side of~\eqref{eq:t1-t2}, i.e.,
\begin{equation}\label{eq:debias-term}
\begin{aligned}
    \E_{\Dscr_n}\Paran{\Para{\frac{1}{n}\sum_{i = 1}^n g(\xi_i)}^{\top}\frac{1}{n}\sum_{i = 1}^n \IFx(\xi_i)}& = \E\Paran{\frac{1}{n^2}\sum_{i = 1}^n g(\xi_i)^{\top} \IFx(\xi_i)} + \E\Paran{\frac{1}{n^2}\sum_{i = 1}^n \sum_{j \neq i} g(\xi_i)^{\top}\IFx(\xi_j)}\\
    & = \E\Paran{\frac{1}{n^2}\sum_{i = 1}^n g(\xi_i)^{\top} \IFx(\xi_i)} + \frac{1}{n^2}\sum_{i = 1}^n \sum_{j \neq i} \E[g(\xi_i)]^{\top}\E[\IFx(\xi_j)]\\
    & = \E\Paran{\frac{1}{n^2}\sum_{i = 1}^n g(\xi_i)^{\top} \IFx(\xi_i)} + 0,\\
\end{aligned}
\end{equation}
where the second equality in~\eqref{eq:debias-term} follows by the independence of $g(\xi_i)$ and $IF[\xi_j]$ for $i \neq j$, and the third equality follows by $\E[\IFx(\xi)] = 0$ such that $\E\Paran{\frac{1}{n^2}\sum_{i = 1}^n \sum_{j \neq i} g(\xi_i)^{\top}\IFx(\xi_j)} = 0$. 

 % above in~\eqref{eq:debias-term} above 
 
Recall the definitions of $\hat{A}_c^*$ and $\hat A_o$ in~\eqref{eq:bias-notation}. Comparing these terms in~\eqref{eq:t1-t2} and~\eqref{eq:debias-term}, we have:
\[\E[\hat{A}_c^* + \hat{A}_o] = \E[T_1] + T_2 + \E[\hat{A}_o] = A + o\Para{\frac{1}{n}}.\quad \hfill \square \]
\subsubsection{Cases for general functions in \Cref{asp:nonsmooth}}
We borrow the approximation lemma with a sequence of functions $\{f_m(z)\}_{m \geq 1}$ to approximate $f(z)$, which is extracted from Lemma C.1 in \cite{wang2018approximate}:
\begin{lemma}\label{lemma:nonsmooth}
Suppose \Cref{asp:nonsmooth} holds for $f(\cdot)$. Given $\phi(\cdot)$ being a kernel with compact support, and is smooth and symmetric around 0 on $\R$, the smoothed objective function $f_m(z): = m \int f(u) \phi(m (z - u)) du, \forall m \geq 1$ satisfies the following properties:
\begin{enumerate}[(1),leftmargin=*]
    \item $f_m(z) \geq f(z), \forall z \in \R$, and $\lim_{m \to \infty}f_m(z) = f(z), \forall z \in \R$.
    \item $\forall z \in \R \backslash K$, and $\forall m$ large enough, 
    \[\nabla_z f_m(z) = m \int \nabla_u f(u) \phi(m(z-u)) du, \nabla_{zz} f_m(z) = m \int \nabla_{uu}^2 f(u) \phi(m (z-u)) du.\]
    \item $\forall z \in K$, we have:
    \[\lim_{m \to \infty} \nabla_z f_m(z) = \frac{\nabla_z f^-(z) + \nabla_z f^+(z)}{2}, \lim_{m \to \infty} \nabla_{zz}^2 f_m(z) = + \infty.\]
    \item For any compact set $\Zscr$, we have: $f_m(z)$ converges to $f(z)$ uniformly.
    %$\lim_{m \to \infty}\E_{z\sim\P}[f_m(z)] = \E_{z\sim \P}[f(z)]$ for any probability distribution $\P$ such that $\E_{z\sim \P}[f(z)] < \infty$. \wty{Use the argument that uniform convergence implies L1 convergence and then converge in expectation. }
    %($f_m(z)$ uniformly converges to $f(z)$ under any compact set).
\end{enumerate}
\end{lemma}

Examples of the proper kernel include any box kernel such as $\phi(x) = \frac{1}{2}\mathbf{1}_{\{\|x\|\leq 1\}}$ and the Epanechnikov kernel $\phi(x) = \frac{3}{4}(1-\|x\|^2)\mathbf{1}_{\{\|x\|\leq 1\}}$. 

%\textit{Proof of \Cref{thm:nonsmooth}.}~
Under such construction, it is easy to see that $h_m(\paranx;\xi) = f_m(g(\paranx;\xi))$ satisfies $\lim_{m \to \infty}h_m(\paranx;\xi) = h(\paranx;\xi), \forall \xi, \theta$ if we set $z = g(\paranx;\xi)$ in \Cref{lemma:nonsmooth}(1). Besides, since the second-order derivative of $Z(\theta)$ exists for $\theta$ near $\theta^*$, following the definition of derivative as well as \Cref{lemma:nonsmooth}(4), since $f_m(z)$ converges to $f(z)$ uniformly for all $z \in \Zscr$, $\lim_{m \to \infty}\E_{P}[|f_m(z) - f(z)|] \to 0$ since $\|f_m(z) - f(z)\|_{L^1} \to 0$ for any distribution $P$.

Consider the distribution of the random variable $g(\paranx;\xi)$ and set $z = g(\paranx;\xi)$. Then we immediately have $ \lim_{m \to \infty}\E_{\P^*}[h_m(\paranx;\xi)] = \E_{\P^*}[h(\paranx;\xi)]$. And this argument above works for any $\theta \in \Theta$. Furthermore, through the definition of the gradient, we have $\lim_{m \to \infty}\nabla_{\theta}\E_{\P^*}[h_m(\paranx;\xi)] = \lim_{m \to \infty}\lim_{k \to \infty}\frac{\E_{\P^*}[h_m(x^*(\theta + 1/k);\xi)] - \E_{\P^*}[h_m(\paranx;\xi)]}{1/k} = \lim_{k \to \infty} \frac{\E_{\P^*}[h(x^*(\theta + 1/k);\xi)] - \E_{\P^*}[h(\paranx;\xi)]}{1/k} = \nabla_{\theta} \E_{\P^*}[h(\paranx;\xi)]$, which leads to the following:
\begin{equation}\label{eq:expect}
    \begin{aligned}
        %\lim_{m \to \infty}\E_{\P^*}[h_m(\paranx;\xi)] &= \E_{\P^*}[h(\paranx;\xi)].\\
        \lim_{m \to \infty}\nabla_{\theta}\E_{\P^*}[h_m(\paranx;\xi)] &= \nabla_{\theta}\E_{\P^*}[h(\paranx;\xi)].\\
        \lim_{m \to \infty}\hessianp\E_{\P^*}[h_m(\paranx;\xi)] &= \hessianp\E_{\P^*}[h(\paranx;\xi)],\\
    \end{aligned}
\end{equation}
Denote $A_m = \E_{\Dscr_n}\E_{\P^*}[h_m(\datax;\xi)]$. Then for any fixed $m$, we have that OIC for the objective $h_m$, i.e., $\hat{A}_m = \frac{1}{n}\sum_{i = 1}^n h_m (\datax;\xi_i) - \frac{1}{n^2}\sum_{i = 1}^n\gradp h_m(\datax;\xi_i)^{\top}\widehat{\IFx}(\xi_i)$ satisfies $\E[\hat{A}_m] = A_m + o\Para{\frac{1}{n}}$. This follows the same proof structure as in \Cref{thm:main}, i.e., applying the Taylor expansion to $\frac{1}{n}\sum_{i = 1}^n h(\paranx;\xi_i)$ and $\E_{\P^*}[h(\paranx;\xi)]$ at the center $\theta^*$.

Then we observe $\lim_{m \to \infty}A_m = A + o\Para{\frac{1}{n}}$. We take a second-order Taylor expansion with respect to the function $\E_{\P^*}[h_m(\datax;\xi) - h(\datax;\xi)]$ at the center $\theta^*$ with Peano's remainder such that:
\begin{equation}
    \begin{aligned}
       A_m - A &= \nabla_{\theta}\E_{\P^*}[h_m(\bestx;\xi) - h(\bestx;\xi)]^{\top}\E_{\Dscr_n}[\hat{\theta} - \theta^*]\\
       &+ \frac{1}{2}\E_{\Dscr_n}[(\hat{\theta} - \theta^*)^{\top}\hessianp\E_{\P^*}[h_m(\bestx;\xi) - h(\bestx;\xi)](\hat{\theta} - \theta^*)] + o\Para{\|\hat{\theta} - \theta^*\|^2}. 
    \end{aligned}
\end{equation}
Then we apply \eqref{eq:expect} when $m \to \infty$ and obtain $\lim_{m \to \infty}A_m  = A + o\Para{\frac{1}{n}}$.

On the other hand, from \eqref{eq:expect}, as $m \to \infty$, we can similarly show that:
\[\E\Paran{\frac{1}{n^2}\sum_{i = 1}^n\gradp h_m(\datax;\xi_i)^{\top}\widehat{\IFx}(\xi_i) - \frac{1}{n^2}\sum_{i = 1}^n\gradp h(\datax;\xi_i)^{\top}\widehat{\IFx}(\xi_i)} = o\Para{\frac{1}{n}},\]
where $\gradp h$ represents the subgradient of $h$ if the function is not differentiable at that point.

Therefore, combining these previous arguments above, 
for $\hat{A}$ in \eqref{eq:oic}, we have:
\[\E[\hat{A}] = \E[\hat{A}_m] + o\Para{\frac{1}{n}} = A_m + o\Para{\frac{1}{n}} = A + o\Para{\frac{1}{n}},\]
which finishes the proof. $\hfill \square$
\subsection{Proof of Theorem~\ref{coro:main-moment}}\label{app:main2}
\quad \textbf{Part (1)}. Recall the result in \Cref{thm:main} and notations in~\eqref{eq:bias-notation}. We only need to show:
% Here we want to control the estimation error between $\hat{A}_c$ and $\hat{A}_c^*$. And we only need to show $\E[T_1] + T_2 = \E[\hat{A}_c] + o\Para{\frac{1}{n}}$, which reduces to the verification of the following equation:
\begin{equation}\label{eq:oic-unbiased0}
    (\E[\hat{A}_c^* - \hat{A}_c] = )\E\Paran{\frac{1}{n^2}\sum_{i = 1}^n \Para{\nabla_{\theta} h(\bestx;\xi_i)^{\top} \IFx(\xi_i) - \nabla_{\theta} h(\datax;\xi_i)^{\top} \widehat{\IFx}(\xi_i)}} = o\Para{\frac{1}{n}}.
\end{equation}
For now, we only assume the two terms $\gradp h(\bestx;\xi), \IFx(\xi)$ are scales instead of vectors since both vectors are of finite dimensions and adding each component does not affect the result when terms are of the order $o\Para{\frac{1}{n}}$. Since we are in a fixed dimensional problem setup,  %\footnote{This works in our setups, i.e. we are in a fixed dimensional problem setup, i.e. $D_{\xi}/n = o(1)$.}.
if \eqref{eq:oic-unbiased0} holds for the scales, then left-hand side of \eqref{eq:oic-unbiased0} in the vector version is of the order $D_{\xi}\times o\Para{\frac{1}{n}} = o\Para{\frac{1}{n}}$ too. Therefore due to the additivity of the expectation, to show \eqref{eq:oic-unbiased0}, it is sufficient to show:
\begin{equation}\label{eq:oic-unbiased1}
    \E_{\Dscr_n}\Paran{\gradp h(\bestx;\xi) \IFx(\xi) - \gradp h(\datax;\xi) \widehat{\IFx}(\xi)} = o(1), \forall \xi \in \Xi.
\end{equation}
We decompose the left-hand side of \eqref{eq:oic-unbiased1} as:
\begin{equation}\label{eq:oic-unbiased2}
    \begin{aligned}
        &\quad\E_{\Dscr_n}\Paran{\gradp h(\datax;\xi) \widehat{\IFx}(\xi)} - \gradp h(\bestx;\xi) \IFx(\xi) \\
        & = \E_{\Dscr_n}\Paran{\gradp h(\datax;\xi) \Para{\widehat{\IFx}(\xi) - \IFx(\xi)}} + \E_{\Dscr_n}\Paran{\Para{\gradp h(\datax;\xi) - \gradp h(\bestx;\xi)}\IFx(\xi)} \\
        & + \E_{\Dscr_n}\Paran{\Para{\widehat{\IFx}(\xi) - \IFx(\xi)}\Para{\gradp h(\datax;\xi) - \gradp h(\bestx;\xi)}}.\\
    \end{aligned}
\end{equation}
From \Cref{asp:empirical-if}, we have:
\begin{equation}\label{eq:if1}
    \E_{\Dscr_n}[\|\widehat{\IFx}(\xi) - \IFx(\xi)\|^2] = o(1).
\end{equation}
Furthermore, since $h(\paranx;\xi)$ is twice continuous differentiable, we have:
\begin{equation}\label{eq:if2}
    \E_{\Dscr_n}[(\gradp h(\datax;\xi) - \gradp h(\bestx;\xi))^2] = \E_{\Dscr_n}[O(\|\hat{\theta} - \theta^*\|_2^2)]  = o(1).
\end{equation}
Applying the Cauchy-Schwartz inequality to each of the three terms on the right-hand side in \eqref{eq:oic-unbiased2} with the form $\E[xy]\leq \sqrt{\E[x^2]}\sqrt{\E[y^2]} = o(1)$, and plugging either~\eqref{eq:if1} or~\eqref{eq:if2} to each of three terms there, \eqref{eq:oic-unbiased1} holds and we finish the proof of part $(1)$ of \Cref{thm:main}.
%Then plugging either~\eqref{eq:if1} or~\eqref{eq:if2} to each of three terms, the  through the Cauchy-Schwartz inequality, 
% it is easy to see that each of these three terms is of the order $o(1)$. This replies that 

%Next, we show the moment and asymptotic property of $\hat{A}_c$ appeared in \Cref{thm:main}.
%\subsection{Proof of~\Cref{prop:moment}}
\textbf{Part (2).}
Through~\eqref{eq:debias-term} in the proof of \Cref{thm:main}, we have $\E[\hat{A}_c] = \E[\hat{A}_c^*]+ o\Para{\frac{1}{n}}$. Furthermore: 
\begin{align*}
   \E[\hat{A}_c^*] = -\frac{1}{n^2}\sum_{i = 1}^n \E[\nabla_{\theta} h(\bestx;\xi_i)^{\top}\IFx(\xi_i)] = -\frac{1}{n}\E[\nabla_{\theta}h(\bestx;\xi)^{\top}\IFx(\xi)] = A_c. 
\end{align*}
% Then we see that:
% \begin{align*}
%     n \hat{A}_c & = -\frac{1}{n}\sum_{i =1}^n \nabla_{\theta} h(\datax;\xi_i)^{\top} IF_{\hat{\theta}}(\xi_i)\\
%     & = - \frac{1}{n}\sum_{i =1}^n \nabla_{\theta} h(\bestx;\xi_i)^{\top} \IFx(\xi_i) + o_p(1)\\
%     & \convp -\E[\nabla_{\theta} h(\bestx;\xi)^{\top} \IFx(\xi)],
% \end{align*}
% where the second equality is trivially from the expectation result in \eqref{eq:oic-unbiased0} and the third line follows by the WLLN.

\textbf{Part (3).} Consider the second moment term.
% , if we assume $\E_{\P^*}[\Para{\gradp h(\paranx;\xi)^{\top} \IFx(\xi)}^2] < \infty$,
% %, denote $f(\hat{\theta};\xi) = \nabla_{\theta} h(\datax;\xi_i)^{\top} IF_{\hat{\theta}}(\xi_i)$, 
Then we have:
\begin{align*}
    \E[\hat{A}_{c}^2] &= \frac{1}{n^4}\Para{\sum_{i = 1}^n \E\Paran{\nabla_{\theta} h(\datax;\xi_i)^{\top} \widehat{\IFx}(\xi_i)}}^2\\
    &\leq \frac{1}{n^2}\E\Paran{(\nabla_{\theta} h(\datax;\xi_i)^{\top} \widehat{\IFx}(\xi_i))^2} = O\Para{\frac{1}{n^2}},
\end{align*}
where the inequality follows by $(\sum_{i = 1}^n x_i)^2 \leq n \sum_{i = 1}^n x_i^2$ and the second equality follows by the assumption that $\E_{\P^*}  \Paran{\Para{\nabla_{\theta}h(\paranx;\xi)^{\top} \IFx(\xi)}^2} < \infty, \forall \theta \in \Theta$. Combining the argument of $(\E[\hat{A}_c])^2 = O\Para{\frac{1}{n^2}}$, we obtain $\var[\hat{A}_c] =\E[\hat{A}_{c}^2] - \E^2[\hat{A}_c]  = O\Para{\frac{1}{n^2}}$. $\hfill \square$

\subsection{Proofs of Equations~\eqref{eq:theta-if} and~\eqref{eq:empirical-if} hold across procedures in \Cref{ex:ddo}}\label{app:apply}
We prove equations~\eqref{eq:theta-if} and~\eqref{eq:empirical-if} hold 
%Lemmas~\ref{asp:theta-asymptotics} and~\ref{asp:empirical-if} hold 
for each procedure. Before that, we first summarize additional regularity conditions across each procedure:
\begin{assumption}[Other Regularity Assumptions]\label{asp:additional-regular}
~~~~~~
\begin{enumerate}[(i),leftmargin=*]
    \item ETO: \Cref{ex:theta-asymptotics} in
     Appendix~\ref{app:eto} holds;
     \item IEO / E2E: When $\Xscr = \R^{D_x}$, \Cref {asp:ierm-additional}in Appendix~\ref{app:ieo} holds; for general $\Xscr$, \Cref{asp:const} in Appendix~\ref{app:constraint} hold; To compute $\widehat{\IFx}(\xi)$ using the empirical Hessian, $f(z) = z$ in Assumption~\ref{asp:nonsmooth}; and Assumption~\ref{asp:ierm-additional-eif} in Appendix~\ref{app:ieo} holds; 
    \item R-E2E: $\lambda \in \R$ and $R(\cdot)$ is twice differentiable. Other assumptions of $(ii)$ hold when replacing $h(x^*(\cdot);\xi) $ with $h(x^*(\cdot);\xi) + \lambda R(x^*(\cdot))$;
      \item DR-E2E: Assumption~\ref{asp:dro} in Appendix~\ref{app:chi2dro} holds. To compute $\widehat{\IFx}(\xi)$ using the empirical Hessian, $f(z) = z$ in Assumption~\ref{asp:nonsmooth}; Assumption~\ref{asp:ierm-additional-eif} in Appendix~\ref{app:ieo} holds. For general $\Xscr$, Assumptions~\ref{asp:const} in Appendix~\ref{app:constraint} and~\ref{asp:dro} in Appendix~\ref{app:chi2dro} hold.
\end{enumerate}
\end{assumption}

\subsubsection{ETO.}\label{app:eto}
% We list additional assumptions such that, which is classical in the asymptotic statistics.
Denote $\psi(\theta;\xi) = \nabla_{\theta}\phi(\theta;\xi)$.
\begin{assumption}[Additional Conditions of ETO]\label{ex:theta-asymptotics}
For any $\xi \in \Xi$ almost surely, the vector-valued function $\psi(\theta;\xi)$ is differentiable at $\theta$ with a nonsingular derivative matrix $\nabla_{\theta}\psi(\theta;\xi)$ for any $\theta \in \Theta$ almost surely. $\sup_{\theta \in \Theta}|\E_{\hat{\P}_n}[\psi(\theta;\xi)] - \E_{\P^*}[\psi(\theta;\xi)] | \convp 0$; $\forall \epsilon > 0$, $\inf_{\theta \in \Theta, \|\theta - \theta^*\|\geq \epsilon}\E_{\P^*}[\psi(\theta;\xi)] > 0$. For $\theta$ around $\theta^*$, $\nabla_{\theta}\psi(\theta;\xi)$ is continuously differentiable with respect to $\theta$. Besides, for every $\theta_1$ and $\theta_2$ in a neighborhood of $\theta^*$, we have a measurable function $K(\xi)$ with $\E_{\P^*}[K^2(\xi)] < \infty$ such that:
\[|\E_{\P^*}[\psi(\theta_1;\xi)] - \E_{\P^*}[\psi(\theta_2;\xi)]|\leq K(\xi) \|\theta_1 - \theta_2\|_2.\]
\end{assumption}

\textit{Proof.} 
Following Assumption~\ref{ex:theta-asymptotics}, we have $\hat\theta \overset{p}{\to}\theta^*$ along with Assumption~\ref{asp:nonsmooth}. Furthermore, all the conditions in Theorem 5.21 (or Theorem 5.23) in \cite{van2000asymptotic} can be verified through Assumptions~\ref{asp:nonsmooth},~\ref{asp:theta-star}$(1)$,~\ref{asp:moment}, which implies that:
\begin{equation}\label{eq:eto-asymptotics}
    \sqrt{n} (\hat\theta - \theta^*) = -(\nabla_{\theta}\E_{\P^*}[\psi(\theta;\xi)])^{-1} \frac{1}{\sqrt{n}}\sum_{i = 1}^n \psi(\theta;\xi_i) + o_p(1).
\end{equation}
Therefore, we have $\IFx(\xi) = -(\nabla_{\theta}\E_{\P^*}[\psi(\theta;\xi)])^{-1} \psi(\theta;\xi)$, $\widehat{\IFx}(\xi) = -((\E_{\hat{\P}_n}[\nabla_{\theta}\psi(\hat{\theta};\xi)])^{-1} \psi(\hat{\theta};\xi)$. Then we show $\IFx(\cdot)$ and $\widehat{\IFx}(\cdot)$ satisfy~\eqref{eq:theta-if} and~\eqref{eq:empirical-if}. This is because:
\begin{align*}
    \E_{\Dscr_n}[\|\widehat{\IFx}(\xi)\|_2^2] &= \E_{\Dscr_n}[(\E_{\hat{\P}_n}[\nabla_{\theta}\psi(\hat{\theta};\xi)])^{-2}\psi(\hat\theta;\xi) \psi(\hat\theta;\xi)^{\top}]\\
    & = \E_{\Dscr_n}[(\E_{\P^*}[\nabla_{\theta}\psi(\hat\theta;\xi)])^{-2} \psi(\hat\theta;\xi)\psi(\hat\theta;\xi)^{\top}] + o(1)\\
    & = \E_{\Dscr_n}[(\E_{\P^*}[\nabla_{\theta}\psi(\theta^*;\xi)])^{-2} \psi(\theta^*;\xi)\psi(\theta^*;\xi)^{\top}] + o(1)
\end{align*}
where the second equality follows by $\E_{\Dscr_n}[|\E_{\hat \P_n}[\nabla_{\theta}\psi(\hat\theta;\xi)] - \E_{\P^*}[\nabla_{\theta}\psi(\hat\theta;\xi)]|]  = o(1)$ by the law of large number. The third equality follows by the fact that $\psi(\theta;\xi)$ and $\nabla_{\theta}\psi(\theta;\xi)$ are both continuously differentiable with respect to $\theta$ around $\theta^*$ and $\E_{\Dscr_n}[\|\hat\theta - \theta\|_2^2] \to 0$. 
%Then the expression of $\hat{A}_c$ in \Cref{coro:oic-procedure}$(1)$ is obtained through a direct plug-in of the empirical influence function $\widehat IF_{\hat\theta}(\cdot)$ here. 
$\hfill \square$

%We can incorporate the regularized version in \Cref{subsec:reg} here with the same estimation structure.
\subsubsection{Unconstrained IEO / E2E.}\label{app:ieo}
% We list additional assumptions such that \Cref{asp:theta-asymptotics} holds for $\hat{\theta}$ obtained from \Cref{ex:ddo}$(b_1)$, which is classical in the asymptotic analysis.

\begin{assumption}[Additional Conditions of Unconstrained IEO / E2E]\label{asp:ierm-additional}
For an open set $\Theta$, we have:
\begin{enumerate}
    \item Consistency condition: $\sup_{\theta}|\E_{\hat{\P}_n}[h(\paranx;\xi)] - \E_{\P^*}[h(\paranx;\xi)]| \convp 0$; $\forall \epsilon > 0$, $\inf_{\theta \in \Theta, \|\theta - \theta^*\|\geq \epsilon}\E_{\P^*}[h(\paranx;\xi)] > \E_{\P^*}[h(\bestx;\xi)]$.
    \item Regularity condition: For any $\theta_1, \theta_2$ in a neighborhood of $\theta^*$, there exists a measurable function $K$ with $\E_{\P^*}[K^2(\xi)] < \infty$ such that $|h(x^*(\theta_1);\xi) - h(x^*(\theta_2);\xi)| \leq K(\xi)\|\theta_1 - \theta_2\|$. 
\end{enumerate}

\end{assumption}
\begin{assumption}[Estimated Influence Function Condition of IEO]\label{asp:ierm-additional-eif}
    $\hessianp h(\paranx;\xi)$ exists for any $\theta \in \Theta$.
\end{assumption}

\textit{Proof.}~First, following Assumptions~\ref{asp:nonsmooth} and \ref{asp:theta-star}$(2)$, the consistency condition holds $\hat{\theta} \convp \theta^*$. Then from \Cref{asp:ierm-additional}, applying Theorem 5.23 in \cite{van2000asymptotic} for $M$-estimators and comparing with the form of influence function in \Cref{asp:theta-asymptotics}, we obtain:
\begin{equation}\label{eq:if-ierm}
    \begin{aligned}
        \IFx(\xi) &= - \Paran{\hessianp\E_{\P^*}[h(\bestx;\xi)]}^{-1}\gradp h(\bestx;\xi).\\
    \end{aligned}
\end{equation}
And the sequence $\sqrt{n}(\hat\theta - \theta^*)$ is asymptotically normal, which verifies that the influence function satisfies~\eqref{eq:theta-if}.
%\Cref{asp:theta-asymptotics} holds. 
%Plugging this back into the result of \Cref{thm:main}, we obtain $A_c = \frac{1}{n}\text{Tr}[I_h(\theta^*)^{-1} J_h(\theta^*)]$ for $I_h(\theta) = \hessianp\E_{\P^*}[ h(\paranx;\xi)]$ and $J_h(\theta) = \E_{\P^*}[\gradp h(\paranx;\xi)\gradp h(\paranx;\xi)^{\top}]$, which shows that the first part of \Cref{coro:oic-procedure}$(2)$ holds.

Second, when $f(z) = z$ in Assumption~\ref{asp:nonsmooth} and Assumption~\ref{asp:ierm-additional-eif} holds, we consider the following empirical influence function:
\begin{equation}\label{eq:eif-ierm}
    \begin{aligned}
         \widehat{\IFx}(\xi_i) = -\Paran{\frac{1}{n}\sum_{i = 1}^n\nabla_{\theta\theta}^2 h(\datax;\xi_i)}^{-1} \nabla_{\theta} h(\datax;\xi_i),
    \end{aligned}
\end{equation}    
and then verify that this empirical estimator satisfies the consistency assumption in \Cref{asp:empirical-if}. To see this, denoting $B(\theta) = \nabla_{\theta} h(\paranx;\xi),  A(\theta) = \frac{1}{n}\sum_{i = 1}^n \nabla_{\theta\theta}^2 h(\paranx;\xi_i)$, we have:
$B(\hat\theta) = B(\theta^*) + O(\|\hat\theta - \theta^*\|_2), A(\hat\theta) = A(\theta^*) + O(|\hat\theta - \theta^*|)$. From \Cref{asp:ierm-additional}, we have: $(A(\hat\theta))^{-2} = (A(\theta^*))^{-2} + O(\|\hat\theta - \theta^*\|_2)$. Therefore, we obtain: $\E_{\Dscr_n}[B(\hat\theta)^{\top} A(\hat\theta)^{-2} B(\hat\theta)] = \E_{\Dscr_n}[B(\theta^*)^{\top} A(\theta^*)^{-2} B(\theta^*)] + o(1) = B(\theta^*)^{\top} \E_{\Dscr_n}[A(\theta^*)^{-2}] B(\theta^*)$. Furthermore, notice that: $\|\widehat{\IFx}(\xi)\|_2^2 = B(\hat\theta)^{\top} A(\hat\theta)^{-2} B(\hat\theta)$ and $\|\IFx(\xi)\|_2^2 = B(\theta^*)^{\top} (\E_{\P^*}[\nabla_{\theta\theta}^2 h(\bestx;\xi)])^{-2} B(\theta^*) = B(\theta^*)^{\top} \E_{\Dscr_n}[A(\theta^*)^{-2}] B(\theta^*) + o(1)$. Therefore, we have: $\E_{\Dscr_n}[\|\widehat{\IFx}(\xi)\|_2^2] = \|\IFx(\xi)\|_2^2 + o(1)$ such that the empirical influence function estimator satisfies equation~\eqref{asp:empirical-if}.  $\hfill \square$

\subsubsection{Constrained IEO / E2E.}\label{app:constraint}
% \paragraph{Constraints.} We also investigate the following problem with some inequality constraints. Specifically, we consider the following problem:
% \begin{equation}\label{eq:constrain-prob}
%     \min_{x} \E_{\P^*}[h(x;\xi)]~\text{s.t.}~F_j(x) \leq 0, \forall j \in J.
% \end{equation}

% We would like the following sets of assumptions to be held for Problem~\eqref{eq:constrain-prob}, which is also prevalent in previous stochastic optimization literature \citep{lam2021impossibility,duchi2021asymptotic}.

We denote $Z(\theta) = \E_{\P^*}[h(\paranx;\xi)]$ in this part:
\begin{assumption}[Conditions of Constrained Problem]\label{asp:const}
We assume the following under the optimization problem~\eqref{eq:constrain-prob}:
\begin{enumerate}
    \item Lagrangian Optimality Condition: For the optimal $\theta^*$, the solution $\bestx$ to~\eqref{eq:constrain-prob} is a unique solution to solving the problem:
    \[\nabla_{\theta} Z(\theta^*) + \sum_{j \in B_{\theta^*}} \alpha_j(\theta^*) \nabla_{\theta} F_j(x^*(\theta^*)) = 0,\]
    where $\alpha_j(\theta) > 0$ are the Lagrange multipliers and $B$ denotes the index of binding active set of constraints. $\{\nabla_{\theta} Z(\theta)\} \cup \{\nabla_{\theta} F_j(x^*(\theta))\}_{j \in B}$ are linearly independent $\forall \theta \in \Theta$. $\alpha_j(\theta), F_j(\paranx)$ is twice differentiable with respect to $\theta$ with $\theta^*$. Besides, for the optimal solution $\bestx$, we denote $B$ without subscript to be the corresponding active constraints and $\alpha_j^* = \alpha_j(\theta^*)$.
    Furthermore, $w^{\top}\Para{\hessianp Z(\theta^*) + \sum_{j \in B}\alpha_j^* \hessianp F_j(\bestx)}w \geq \mu\|w\|_2^2$ for any $w \in \{w \in \R^{D_{\theta}}: \nabla F_i(x)^{\top}w = 0, \forall i \in J, \text{s.t.} F_i(x) = 0\}$ and $\alpha(\theta) F_j(\paranx) = 0, \forall j \in B$.
    \item Active Binding Constraints: The data-driven solution $\datax$ satisfies $F_j(\datax) = 0, \forall j \in B$ almost surely.
    \item Linear Independence Constraint Qualification (LICQ): $-Z(\theta^*)$ is a relative interior point of $\{v: v^{\top}(\theta - \theta^*) \leq 0, \forall \theta \in \Theta\}$. 
    \item Regularity Condition: There exists $C_1 > 0$ such that:
    \[\|\gradp Z(\theta) - \gradp Z(\theta^*)\|_2 \leq C_1 \|\theta - \theta^*\|.\]
    And there exists $C_2, \varepsilon > 0$ such that $\forall \|\theta - \theta^*\| \leq \varepsilon$:
    \[\|\gradp Z(\theta) - \gradp Z(\theta^*) - \hessianp Z(\theta^*)(\theta - \theta^*)\| \leq C_2 \|\theta - \theta^*\|^2.\]
    And there exists $C_3 > 0$ such that $\forall \theta \in \Theta$:
    \[\E_{\P^*}[\|\gradp h(\paranx;\xi) - \gradp h(\bestx;\xi)\|^2] \leq C_3 \|\paranx - \bestx\|_2^2.\]
    \item Recovery Condition: Denote the constrained set matrix $P = I - C^{\top}(C C^{\top})^{\dagger} C$ and $C \in \R^{|B| \times D_{\theta}}$ denotes the matrix with rows $\{\nabla F_j(\bestx)^{\top}: F_j(\bestx) = 0 \}$. For the empirical multiplier $\alpha_j(\hat{\theta})$ and the empirical matrix $\hat P$ defined in \eqref{defn:estimate-if}, we have: 
    \begin{align*}
        & \alpha_j(\hat{\theta}) \overset{a.s.}{\to} \alpha_j^*, \quad \hat P \overset{a.s.}{\to} P, \\
        &  \hat{P} (\hat{I}_{h,\alpha}(\hat{\theta}))^{\dagger} \hat{P} = P (I_{h,\alpha}(\theta^*))^{\dagger} P+ O(\|\hat\theta - \theta\|_2),
    \end{align*}
    where $I_{h,\alpha}(\theta^*)= \hessianp Z(\bestx) + \sum_{j \in B}\alpha_j^* \hessianp F_j(\bestx)$.
\end{enumerate}
\end{assumption}

\textit{Proof.}~Since~\Cref{asp:const} holds, we can apply Proposition 1 and Corollary 1 in \cite{duchi2021asymptotic} and obtain:
\[\hat{\theta} - \theta^* = -P (I_{h,\alpha}(\theta^*))^{\dagger} P\Para{\frac{1}{n}\sum_{i = 1}^n \nabla_{\theta} h(\bestx;\xi_i)} + o_p\Para{\frac{1}{\sqrt{n}}}.\]
Comparing it with the condition in \Cref{asp:theta-asymptotics}, the influence function for the constrained optimization under this problem is:
\[\IFx(\xi_i) = -P (I_{h,\alpha}(\theta^*))^{\dagger} P \nabla_{\theta} h(\bestx;\xi_i).\]
This influence function form satisfies~\eqref{eq:theta-if}.

Besides, the estimated influence function is:
\[\widehat{\IFx}(\xi) = -\hat{P} (\hat{I}_{h,\hat{\alpha}}(\hat\theta))^{\dagger} \hat{P} \nabla_{\theta} h(\datax;\xi).\]
Then we verify $\widehat{\IFx}(\cdot)$ satisfies the equation~\eqref{eq:empirical-if}. Denote $B(\theta;\alpha, P) := P (\hat{I}_{h,\alpha}(\theta))^{\dagger} P$. Then we have:
\begin{align*}
    \E[\|\widehat{\IFx}(\xi)\|_2^2] & = \E[\|B(\hat\theta;\alpha^*,\hat P)\nabla_{\theta} h(\datax;\xi)\|_2^2] + o(1)\\
    & = \E[\|B(\theta^*;\alpha^*,P)\nabla_{\theta} h(\bestx;\xi)\|_2^2] + o(1)\\
    & = \E[\|P (I_{h,\alpha}(\theta^*))^{\dagger} P\nabla_{\theta} h(\bestx;\xi)\|_2^2] + o(1) = \IFx(\xi) + o(1),
\end{align*}
where the second equality follows by $B(\hat\theta;\alpha, \hat P) = B(\theta^*;\alpha^*,P) + O(\|\hat\theta - \theta^*\|)$ from \Cref{asp:const} and the third equality follows by $\E_{\hat\P_n}[\nabla_{\theta\theta}^2 h(\bestx;\xi)] \overset{a.s.}{\to} \E_{\P^*}[\nabla_{\theta\theta}^2 h(\bestx;\xi)]$. 
$\hfill \square$

\subsubsection{R-E2E.}
For R-E2E, the condition and proof are the same as the standard E2E in \Cref{coro:oic-procedure}$(2)$ as long as we replace $h(\paranx;\xi)$ there by $h(\paranx;\xi) + \lambda R(\paranx)$, so we do not mention here. $\hfill \square$

\subsubsection{(Unconstrained) DR-E2E.}\label{app:chi2dro}
% We list additional results of the $f$-divergence DRO here since the result of R-E2E is straightforward.
% \textit{Proof of \Cref{coro:oic-procedure}$(3)$.} 
For the DR-E2E case with $d$ set as an $f$-divergence, we suppose the following assumption holds:
%$f$-divergence DRO problem under the following assumptions.
\begin{assumption}[Conditions of $f$-divergence DRO, extracted from \cite{lam2021impossibility}]\label{asp:dro}
We assume the following set of conditions holds under the DRO optimization problem:
\begin{enumerate}
    \item Lagrangian Multiplier Condition: $\inf_{\theta \in \Theta} \var_{\P^*}[h(\paranx;\xi)] > 0$; $\forall \theta \in \Theta, \max_{d(\Q, \hat{\P}_n)\leq \epsilon} \E_{\Q}[h(\paranx;\xi)] \neq ess \sup_{\theta}\E_{\hat{\P}_n} h(\paranx;\xi)$. Besides, the following optimization problem has a unique solution $(\hat{\theta}, \hat{\alpha}, \hat{\beta})$ satisfying KKT condition with $\hat{\theta}$ in the interior of $\Theta$:
    \[\min_{\theta \in \Theta, \alpha \geq 0, \beta \in R}\Para{\alpha \E_{\hat{\P}_n}\Paran{f^*\Para{\frac{h(\paranx;\xi) - \beta}{\alpha}}}} + \alpha \epsilon + \beta.\]
    \item First-order Optimality Condition: $\theta^*$ is the solution to $\nabla_{\theta}\E_{\P^*}[h(\paranx;\xi)] = 0$ and $\inf_{\theta \in \Theta: \|\theta - \theta^*\|\geq \varepsilon}\|\nabla_{\theta}\E_{\P^*}[h(\paranx;\xi)]\| > 0, \forall \varepsilon > 0$.
    \item Regularity Condition: $h(\paranx;\xi)$ and $\gradp h(\paranx;\xi)$ are uniformly bounded over $\theta \in \Theta$ and $\xi \in \Xi$. And $\|\gradp h(x^*(\theta_1);\xi) - \gradp h(x^*(\theta_2);\xi)\| \leq K(\xi)\|\theta_1 - \theta_2\|$ with $\E_{\P^*}[K^2(\xi)] < \infty$. $\hessianp h(\paranx;\xi)$ is \wty{invertible} for any $\theta \in \Theta$.
    \item Function Complexity: $\{h(\paranx;\cdot), \theta \in \Theta\}, \{h^2(\paranx;\cdot), \theta \in \Theta\}, \{\gradp h(\paranx;\cdot), \theta \in \Theta\}$ are Glivenko-Cantelli. And Donsker property holds for $\{(f^*)'(\alpha h(\paranx;\cdot) - \beta)\gradp h(\paranx;\cdot): 0 \leq \|\theta - \theta^*\|, \alpha,\beta\leq \delta\}$.
\end{enumerate}
\end{assumption}
The following lemma simply generalizes the result from $h(x^*;\cdot)$ in \cite{lam2021impossibility} to a model-based version $h(x^*(\theta);\cdot)$.
\begin{lemma}[Extracted from Theorem 4 in \cite{lam2021impossibility}]\label{lemma:dro-if}
Recall $\hat{\theta}$ from the $f$-divergence-based DRO model in~\Cref{ex:ddo}$(b)(3)$ and $\theta^* \in \argmin_{\theta}\E_{\P^*}[h(x^*(\theta);\xi)]$. Suppose Assumptions~\ref{asp:nonsmooth},~\ref{asp:moment} and~\ref{asp:dro} hold. Then we have:
\begin{equation}\label{eq:dro-theta-asymptotics}
\begin{aligned}
    \hat{\theta} - \theta^* & = \frac{1}{n}\sum_{i = 1}^n \IFx(\xi_i) - \sqrt{\epsilon (f^{*})''(0)}(\nabla_{\theta\theta}^2 \E_{\P^*}[ h(\bestx;\xi)])^{-1}\frac{\cov_p(h(x^*(\theta^*),\xi), \nabla_{\theta}h(x^*(\theta^*),\xi))}{\sqrt{\var_{\P^*}[h(x^*(\theta^*);\xi)]}}\\
    & + o_p\Para{\frac{1}{\sqrt{n}} + \sqrt{\epsilon}},
\end{aligned}
\end{equation}
where the $\IFx(\xi)$ is the same as that in the empirical optimization, i.e.~\eqref{eq:if-ierm}.
\end{lemma}
\textit{Proof.}~ From \Cref{lemma:dro-if}, denoting another new limiting point of $\hat{\theta}$ as:
\begin{equation}\label{eq:dro-new-limit}
    \theta_{\epsilon}^*:= \theta^* - \sqrt{\epsilon (f^{*})''(0)}(\E_{\P^*}[\nabla_{\theta\theta}^2 h(x^*(\theta);\xi)])^{-1}\frac{\cov_p(h(x^*(\theta^*),\xi), \nabla_{\theta}h(x^*(\theta^*),\xi))}{\sqrt{\var_{\P^*}[h(x^*(\theta^*);\xi)]}}.
\end{equation}
%This $\theta_{\epsilon}^*$ is set since we want to follow the form of $\hat\theta - \theta^*$ satisfies \Cref{asp:theta-asymptotics}. 
Consider the asymptotic limit between $\hat{\theta}$ and $\theta_{\epsilon}^*$ and set $\epsilon = O\Para{\frac{1}{n}}$ in~\eqref{eq:dro-theta-asymptotics} and~\eqref{eq:dro-new-limit}, then:
\[\hat{\theta} - \theta_{\epsilon}^* = \frac{1}{n}\sum_{i = 1}^n \IFx(\xi_i) + o_p\Para{\frac{1}{\sqrt{n}}}.\]
This influence function form satisfies equation~\eqref{eq:theta-if}. Since the influence function form is the same as that in the empirical optimization, and the difference between these two limit points is small $\E[\|\theta_{\epsilon}^* - \theta^*\|_2] = O(1/n)$. Then we can use the same estimated influence function $\widehat{\IFx}(\xi)$ in equation~\eqref{eq:empirical-if} and have the same bias term as that in~\Cref{coro:oic-procedure}$(2)$. $\hfill \square$

We only discuss the unconstrained DR-E2E here since the constrained DR-E2E results follow the same structure after imposing \Cref{asp:const}. 

%\section{Missing Proofs in~\Cref{sec:extension}}

% \subsection{Proofs in \Cref{subsec:cv}}

%\subsection{Computing Hessian Inverse}\label{subsec:compute-if}
\section{Further Assumptions and Proofs in Section~\ref{sec:extension}}\label{sec:proofs extension}
\subsection{Proofs in Section~\ref{subsec:application-if}}\label{app:application-if}

\textit{Proof of \Cref{prop:oic-entropic-lp}.}
%\label{app:entropic-lp-proof}
For simplicity, we denote $\hat\mu = \E_{\hat\P_n}[\xi]$ and $\mu = \E_{\P^*}[\xi]$. We also denote the limiting parameter of the entropic-regularized problem as:
\[\theta^*(\eta_n) \in \argmin_{\theta}\paran{\mu^{\top}\theta + \eta_n^{-1}\paran{\sum_{i}\theta_i \log\theta_i}, \text{s.t.} F\theta = g}.\]

We assume the following assumptions hold, which is natural for most linear optimization problems:
\begin{assumption}\label{asp:additional-entropic-lp}
    The linear optimization satisfies the following conditions:
    \begin{enumerate}
        \item The solution $\theta^* \in \argmin_{\theta}\{\mu^{\top}\theta, \text{s.t.}~F \theta = g\}$ exists and is unique;
        \item $\Xscr = \{x: Fx = g, x \geq 0\}$ has non-empty relative interior and there exists $x \in \text{relint}~\Xscr$ such that $x > 0$;
        \item The constraint matrix $F$ has full rank.
    \end{enumerate}
\end{assumption}

For this, we apply the results of \Cref{prop:oic-general}. To apply \Cref{prop:oic-general}, we only need to prove equations~\eqref{eq:theta-if} and~\eqref{eq:empirical-if} hold for some $\IFx(\xi)$ and $\widehat{\IFx}(\xi)$.

First, we verify \eqref{eq:theta-if} and calculate $\IFx(\xi)$. To see this, we need to compute $\frac{\partial \theta^*(\eta_n)}{\partial \mu}$ since $\IFx(\xi) =\frac{\partial \theta^*(\eta_n)}{\partial \mu} (\xi - \mu)$. We compute the KKT condition of the entropic-regularized linear optimization. Consider the Lagrangian multiplier $\alpha$ corresponding to the constraint $F\theta = g$, we have:
\[\nabla_{\theta}\Para{\mu^{\top}\theta + \eta_n^{-1} \sum_i \theta_i^* \log\theta_i^*} - F^{\top} \alpha = 0,\]
which implies that:
\[\mu + \eta_n^{-1}(\sum_i \log \theta_i^* + 1) = F^{\top}\alpha.\]
Taking derivative with respect to $\mu$ for this, we have:
\[I + \text{diag}(\theta^*(\eta_n))^{-1}\frac{\partial \theta^*(\eta_n)}{\partial \mu} = F^{\top} \frac{\partial \alpha}{\partial \mu},\]
which gives $\frac{\partial \theta^*(\eta_n)}{\partial \mu} = \text{diag}(\theta^*(\eta_n))(F^{\top}\frac{\partial \alpha}{\partial \mu} - I).$
Then differentiating the constraint of $F\theta = g$,  we have: $F \frac{\partial \theta^*(\eta_n)}{\partial \mu} = 0$. Plugging the expression of $\frac{\partial \theta^*(\eta_n)}{\partial \mu}$ in, we obtain $\frac{\partial \alpha}{\partial \mu} = (F\text{diag}(\theta^*(\eta_n)F^{\top})^{-1} F \text{diag}(\theta^*(\eta_n))$. And we can take back to obtain: $\frac{\partial \theta^*(\eta_n)}{\partial \mu} = -\text{diag}(\theta^*(\eta_n))(I - F^{\top}(F\text{diag}(\theta^*(\eta_n)F^{\top})^{-1} F\text{diag}(\theta^*(\eta_n))$. Since $\tilde h(x;\xi) = \xi^{\top}x+ \eta_n^{-1}\Para{\sum_i x_i \log x_i}$, \Cref{asp:const} directly follows. Then we apply Corollary 1 in \cite{duchi2021asymptotic} and obtain $\sqrt{n}(\hat\theta(\eta_n) - \theta^*(\eta_n))$ is asymptotically normal and:
\[\hat\theta(\eta_n) - \theta^*(\eta_n) = \frac{1}{n}\sum_{i \in [n]}\IFx(\xi_i) + o_p\Para{\frac{\log n}{\sqrt{n}}}.\]

Then, we verify \eqref{eq:empirical-if} for $\widehat{\IFx}(\xi)$ in \Cref{prop:oic-entropic-lp}. Note that the difference between $\widehat{\IFx}(\xi)$ in \Cref{prop:oic-entropic-lp} and $\IFx(\xi)$ here is only with respect to (i) $\text{diag}(\hat\theta(\eta_n))$ and $\text{diag}(\theta^*(\eta_n))$, which follows from the convergence of $\hat\theta(\eta_n)$ to $\theta^*(\eta_n)$; (ii) $\E_{\hat\P_n}[\xi]$ and $\E_{\P^*}[\xi]$, which follows from the standard convergence result. 

Given the forms of $\IFx(\xi)$ and $\widehat{\IFx}(\xi)$, we apply \Cref{prop:oic-general} to the objective $\tilde h(x;\xi) = \xi^{\top}x + \eta_n^{-1}\sum_j x_j \log x_j$, we have:
\[\E_{\hat\P_n}[\xi]^{\top}\hat\theta(\eta_n)  - \frac{1}{n^2}\sum_{i = 1}^n \xi_i^{\top}\widehat{\IFx}(\xi_i) = \E_{\Dscr_n}\E_{\P^*}[\tilde h(x^*(\hat\theta(\eta_n));\xi)] + o\Para{\frac{\log^2 n}{n}},\]
where the remainder term is of order $o(\log^2 n/n)$ since $\hat\theta(\eta_n) - \theta^*(\eta_n) \sim O_p\Para{\frac{\log n}{\sqrt{n}}}$ instead of $O_p\Para{\frac{1}{\sqrt{n}}}$ due to the order $\eta_n$.

Then subtracting the regularized term $\eta_n^{-1}\sum_j \hat\theta_j(\eta_n)\log \hat\theta_j(\eta_n)$ from both sides, we have:
\begin{equation}\label{eq:entropic-add}
\begin{aligned}
    \hat A(\eta_n) & = \E_{\Dscr_n}\E_{\P^*}[h(x^*(\hat\theta(\eta_n));\xi)] + o(\log^2 n / n)\\
    & = \E_{\Dscr_n}\E_{\P^*}[h(x^*(\hat\theta);\xi)] +  \Para{\E_{\Dscr_n}\E_{\P^*}[h(x^*(\hat\theta(\eta_n));\xi)] - \E_{\Dscr_n}\E_{\P^*}[h(x^*(\hat\theta);\xi)]} + o(\log^2 n/n)\\
    & = \E_{\Dscr_n}\E_{\P^*}[h(x^*(\hat\theta);\xi)] + o(\log^2 n / n).
    \end{aligned}
\end{equation}
where the third equality follows from the fact:
\[|\E_{\P^*}[h(x^*(\hat\theta(\eta_n);\xi)] - \E_{\P^*}[h(x^*(\hat\theta);\xi)]| \leq \|\E_{\P^*}[\xi]\|_{\infty}\|\hat\theta(\eta_n) - \hat\theta\|_1 = o(1/n),\]
with the last inequality given by $\|\hat\theta(\eta_n) - \hat\theta\|_1 \leq C_1\exp(-C_2\eta_n)$ in Corollary 9 of \cite{weed2018explicit}. Therefore, the conclusion holds.

$\hfill \square$

\wty{\textit{Computing Hessian inverse.} In a high-dimensional setting, 
computing $\hat
I_h(\hat\theta)^{-1}$ is computationally expensive, and therefore, % Instead of
% explicitly inverting the matrix, we can approximate the influence
% function by solving the linear system
the estimated influence function should be computed by approximately solving linear
systems 
$\hat I_h(\hat\theta)
\cdot\widehat{\IFx}(\xi) = -\gradp h(\datax;\xi)$ using the conjugate gradient
  method.% which are particularly efficient when $\hat I_h(\hat\theta)$ is
  % large. 
%or  approximation~\citep{asmussen2007stochastic}. (ARE THESE CLAIMS JUSTIFIED?)
\begin{proposition}[Chapter 6 in \cite{saad2003iterative}]
    For each $\xi \in \Xi$, denote $\widehat{\IFx}^{(K)}(\xi)$ as the output
    of Algorithm~\ref{alg:cg-solve} when inputs $F = \hat I_h(\hat\theta)$
    and $g = -\gradp h(\datax;\xi)$. Then \Cref{asp:empirical-if} holds if
    $K = \omega(\log n)$. 
\end{proposition}
\begin{algorithm}
\caption{Conjugate Gradient Method to Solve $Fu = g$}
\label{alg:cg-solve}
\begin{algorithmic}[1]
\REQUIRE Symmetric positive-definite matrix $F \in \mathbb{R}^{d \times d}$, vector $g \in \mathbb{R}^d$, tolerance $\varepsilon > 0$, maximum iterations $K$
\ENSURE Approximate solution $u$ to $F u = g$
\STATE Initialize $u^{(0)} \leftarrow 0$, $r^{(0)} \leftarrow g - F u^{(0)}$, $p^{(0)} \leftarrow r^{(0)}$
\FOR{$k = 0$ to $K-1$}
    \STATE $\alpha_k \leftarrow \frac{\langle r^{(k)}, r^{(k)} \rangle}{\langle p^{(k)}, F p^{(k)} \rangle}$
    \STATE $u^{(k+1)} \leftarrow u^{(k)} + \alpha_k p^{(k)}$
    \STATE $r^{(k+1)} \leftarrow r^{(k)} - \alpha_k F p^{(k)}$
    \IF{$\|r^{(k+1)}\|_2 < \varepsilon$}
        \STATE \textbf{break}
    \ENDIF
    \STATE $\beta_k \leftarrow \frac{\langle r^{(k+1)}, r^{(k+1)} \rangle}{\langle r^{(k)}, r^{(k)} \rangle}$
    \STATE $p^{(k+1)} \leftarrow r^{(k+1)} + \beta_k p^{(k)}$
\ENDFOR
\STATE \textbf{return} $u^{(K)}$
\end{algorithmic}
\end{algorithm}}

\subsection{Further Assumptions and Proofs in Section~\ref{sec:general-risk}}\label{app:general-risk}
% We list some technical conditions such that. Note that these conditions are not necessary to ensure that \Cref{thm:risk-oic} holds.
\begin{assumption}\label{asp:additional-risk}
    The domain $\Theta$ is bounded. And $\forall \theta \in \Theta$, we have: $\E[(\frac{1}{n}\sum_{i = 1}^n h(\paranx;\xi_i)- Z(\paranx))^{\gamma}] = O(n^{-\gamma/2})$ for $\gamma = 2, 3$.
\end{assumption}
The second condition in~\Cref{asp:additional-risk} is the moment convergence version of the central limit theorem of $\frac{1}{n}\sum_{i = 1}^n h(\paranx;\xi_i)- Z(\paranx)$.

Before the main proof of \Cref{thm:risk-oic}, we first present the following lemmas:
\begin{lemma}\label{lemma:gap}
    Suppose the same conditions in \Cref{coro:main-moment} and \Cref{asp:additional-risk} hold, we have: $\E[(Z(\datax) - Z(\bestx))^{\gamma}] = O(n^{-\gamma}), \forall \gamma$.
\end{lemma}
\textit{Proof of \Cref{lemma:gap}.} This is a result following the key proofs in \cite{lam2021impossibility} and \cite{elmachtoub2023estimatethenoptimize} in deriving the performance gap such that: $Z(\datax) - Z(\bestx) = C \|\hat\theta - \theta^*\|^2 + o_p(\|\hat\theta- \theta^*\|^2)$. Then the result follows by \Cref{asp:theta-asymptotics} such that: $\E[\|\hat\theta - \theta\|_2^{2\gamma}] = O(n^{-\gamma})$ by asymptotic normality. $\hfill \square$
\begin{lemma}\label{lemma:joint-gap}
    Suppose the same conditions in \Cref{coro:main-moment} and \Cref{asp:additional-risk} hold. We have: $\E[(Z(\datax - Z(\bestx)))(\hat A_o - Z(\datax))^{\gamma}] = o(1/n)$ for $\gamma = 1, 2$.
\end{lemma}
\textit{Proof of \Cref{lemma:joint-gap}.} This result follows by the Holder inequality such that for $\gamma  = 1, 2$:
\[\E[Z(\datax - Z(\bestx)))(\hat A_o - Z(\datax))^{\gamma}] \leq (\E[(Z(\datax) - Z(\bestx))^{1 + \frac{1}{\gamma}}])^{\frac{\gamma}{\gamma + 1}} (\E[(\hat A_o - Z(\datax))^{\gamma + 1}])^{\frac{1}{\gamma + 1}}.\]
Then we apply \Cref{lemma:gap} and \Cref{asp:additional-risk} to bound each term on the right-hand side above respectively.
$\hfill \square$
\begin{lemma}\label{lemma:stoc-donsker}
    For the variance estimator $\hat \sigma^2 = \frac{1}{n}\sum_{i = 1}^n (h(\datax;\xi_i) - \hat A_o)^2$, we have: $\E[\hat\sigma^2] = \E[(\hat A_o - Z(\datax))^2] + o(1) = \E[(h(\bestx;\xi) - Z(\bestx))^2] +o(1)$.
\end{lemma}
% This is implied by when $x^*(\theta)$ is of some parametric function, $\{h(\paranx;\xi), \theta \in \Theta\}$ is Donsker and asymptotic equicontinuity satisfies such that this lemma holds \citep{vaart2023empirical}. 
This result of \Cref{lemma:stoc-donsker} follows the same idea as in Proposition 1 (iii) from \cite{bayraksan2006assessing} since $\hat\theta \overset{p}{\to}\theta^*$ with probability 1 and the uniform convergence result there still applies.

\textit{Proof of \Cref{thm:risk-oic}.} In the following, we assume the equalities are equivalent ignoring terms of order $o\Para{1/n}$. We take a second-order Taylor expansion with the Peano remainder for the difference $u(Z(\datax)) - u(\hat A_o)$:
\begin{equation}\label{eq:risk-taylor}
\begin{aligned}
    &\quad \E[u(Z(\datax)) - u(\hat A_o)] \\
    & = \E[u'(\hat A_o)(Z(\datax) - \hat A_o)] + \frac{1}{2}\E[u''(\hat A_o)(Z(\datax) - \hat A_o)^2] + o(|\hat A_o - Z(\datax)|^2).\\
    & \overset{(a)}{=} \E\Paran{u'(Z(\bestx))(Z(\datax) - \hat A_o)} - \frac{1}{2}\E\Paran{u''(\hat A_o)(\hat A_o - Z(\datax))^2} \\
    & \overset{(b)}{=} \E\Paran{u'(Z(\bestx))(Z(\datax) - \hat A_o)} - \frac{1}{2}\E\Paran{u''(Z(\bestx))(\hat A_o - Z(\datax))^2} \\
    & = u'(Z(\bestx))\E[Z(\datax) - \hat A_o] - \frac{1}{2}u''(Z(\bestx))\E[(\hat A_o - Z(\datax))^2]\\
    & \overset{(c)}{=} u'(Z(\bestx)) A_c - \frac{1}{2}u''(Z(\bestx)) \E[(h(\bestx;\xi) - Z(\bestx))^2]\\
    & \overset{(d)}{=} u'(Z(\bestx))\E[\hat A_c] - \frac{1}{2n}u''(Z(\bestx)) \E[\hat\sigma^2]\\
    & \overset{(e)}{=} \E[u'(\hat A_o)\hat A_c] - \frac{1}{2n}\E[u''(\hat A_o)\hat\sigma^2].
\end{aligned}
\end{equation}
Notice that $(a)$ above follows from: 
\begin{align*}
    u'(\hat A_o) &= u'(Z(\bestx)) - u''(\hat A_o)(Z(\bestx) - \hat A_o) + o(|Z(\bestx) - \hat A_o|).
\end{align*}
And then $\E[(Z(\bestx) - \hat A_o)(Z(\datax) - \hat A_o)] = \E[(Z(\datax) - \hat A_o)^2]+ o(1/n)$ from \Cref{lemma:joint-gap}. And the remaining terms can be ignored since $\E[(\hat A_o - Z(\datax))^2] = O(1/n)$ from \Cref{lemma:gap}. Then, $(b)$ above in~\eqref{eq:risk-taylor} follows from the fact that $g''$ is continuous such that $|u''(\hat A_o) - u''(Z(\bestx))| =  O(|\hat A_o - Z(\bestx)|)$. Then $\E[(\hat A_o - Z(\bestx))(Z(\datax) - \hat A_o)^2] = o(1/n)$ following \Cref{lemma:gap} and \Cref{lemma:joint-gap}. $(c)$ above in~\eqref{eq:risk-taylor} follows from \Cref{thm:main} and \Cref{lemma:stoc-donsker} respectively. And $(d)$ follows by $\E[\hat A_c] = A_c + o\Para{\frac{1}{n}}$ from \Cref{coro:main-moment}.
%$\hat A_o - Z(\datax) = O_p\Para{\frac{1}{\sqrt{n}}}$.
The final step $(e)$ in~\eqref{eq:risk-taylor} follows by noticing that $\E[(\hat A_o - Z(\bestx))^2] = O(1/n)$ such that we can replace $u'(Z(\bestx))$ and $u''(Z(\bestx))$ with $u'(\hat A_o)$ and $u''(\hat A_o)$ without changing the bias order $o\Para{\frac{1}{n}}$ since $\E[\hat \sigma^2] = n\E[(\hat A_o - Z(\datax))^2] + o\Para{1}$. Therefore, $\hat A_u := u(\hat A_o) + u'(\hat A_o)\hat A_c - \frac{1}{2n}u''(\hat A_o)\hat\sigma^2$ debiases the decision performance up to $o(1/n)$.  $\hfill \square$

% \textit{Proof of \Cref{coro:data}.} When the given condition holds, following the right-hand side of step $(c)$ in~\eqref{eq:risk-taylor}, we have: $\E[u(Z(\datax)) - u(\hat A_o)] = o(1/n)$ and vice versa. $\hfill \square$

\textit{Proof of \Cref{coro:risk-loocv}.} Recall \Cref{prop:loocv}, we have: $\E[\hat A_{loocv}] = A + o\Para{\frac{1}{n}}$. Similarly, we take the second-order Taylor expansion while ignoring the term of order $o\Para{\frac{1}{n}}$:
\begin{equation}
    \begin{aligned}
        &\quad \E[u(Z(\datax)) - u(\hat A_{loocv})]\\
        & = \E[u'(\hat A_{loocv})(Z(\datax) - \hat A_{loocv})] + \frac{1}{2}\E[u''(\hat A_{loocv})(Z(\datax) - \hat A_{loocv})^2] + o(|\hat A_{loocv} - Z(\datax)|^2)\\
        & \overset{(a)}{=} \E[u'(Z(\bestx))(Z(\datax) - \hat A_{loocv})] - \frac{1}{2}\E[u''(\hat A_{loocv})(Z(\datax) - \hat A_{loocv})^2]\\
        & \overset{(b)}{=}\E[u'(Z(\bestx))(Z(\datax) - \hat A_{loocv})] - \frac{1}{2}\E[u''(Z(\bestx))(Z(\datax) - \hat A_{loocv})^2]\\
        & \overset{(c)}{=} u'(Z(\bestx))(A - \E[\hat A_{loocv}]) - \frac{1}{2n}u''(Z(\bestx)\E[\hat\sigma_{loocv}^2],
    \end{aligned}
\end{equation}
where the first two equalities $(a)(b)$ follow the same arguments as in~\eqref{eq:risk-taylor}. The last equality $(c)$ is due to the following chain equality:
\begin{align*}
    \E[(Z(\datax) - \hat A_{loocv})^2] &= \E[(Z(\bestx) - A)^2] + o(1) = (1/n)\text{Var}[h(\bestx;\xi)] + o(1) \\
    & = (1/n) \E[\hat \sigma_{loocv}^2] + o(1). \quad \hfill \square
\end{align*}

\subsection{Proofs in Section~\ref{sec:POIC}}
To show~\Cref{prop:pf-fit}, we consider a more general case than \eqref{eq:theta-if}, where we express a second-order bias term of order $O\Para{\frac{1}{n}}$ for the estimation of $\hat{\theta}$. This gives a more general bias result:
\begin{assumption}[Augmented Statistical Properties of Parameter]\label{asp:theta-asymptotics-strong}
\Cref{eq:theta-if} holds and $\E[\hat \theta - \theta^*] = C(\P^*, \theta^*)/n + o(1/n)$ for some $C(\P^*, \theta^*) \in (0, \infty)$.
%Denote $\hat{\theta}$ to be the parameter estimated under $\Dscr_n$, 
% We have $\hat{\theta}\convp \theta^*$ and:
% \[\hat{\theta} - \theta^* = \frac{1}{n}\sum_{i = 1}^n \IFx(\xi_i)  + \frac{C(\P^*, \theta^*)}{n} +  o_p\Para{\frac{1}{n}}.\]
% Additionally, we have: $\sqrt{n}(\hat{\theta} - \theta^*) \convd N(0, \Psi(\theta^*))$.
%$\E_{\Dscr_n}\Paran{(\hat{\theta} - \theta^*) (\hat{\theta} - \theta^*)^{\top}} = O\Para{\frac{\Psi(\theta^*)}{n}} + o\Para{\frac{1}{n}}$. 
\end{assumption}
This \Cref{asp:theta-asymptotics-strong} holds for general moment estimation problem as follows:
\begin{example}[Moment methods extracted from \cite{rilstone1996second}]\label{ex:theta-asymptotics-strong}
Suppose the parameters $\hat{\theta}$ and $\theta^*$ are attained by $\E_{\hat{\P}_n}[\psi(\theta;\xi)] = 0$ and $\E_{\P^*}[\psi(\theta;\xi)] = 0$ respectively. If $\psi$ is third continuously differentiable with respect to $\theta$, then:
\begin{equation}\label{eq:asymptotics-theta}
    \hat{\theta} - \theta^* = a_{-1/2} + a_{-1} +o_p\Para{\frac{1}{n}},
\end{equation}
where $a_{-1/2} = -Q \E_{\hat{\P}_n}[\psi(\theta^*;\xi)] =  O_p(n^{-\frac{1}{2}}),a_{-1} = -Q V a_{-1/2} - \frac{1}{2}Q H [a_{-1/2}\otimes a_{-1/2}] = O_p(n^{-1})$,
and $Q = (\E_{\P^*}[\nabla_{\theta}\psi(\theta^*;\xi))])^{-1}, V = \E_{\P_n}[\nabla_{\theta}\psi(\theta^*;\xi)] - \E_{\P^*}[\nabla_{\theta}\psi(\theta^*;\xi)]$, and $H = \E_{\P^*}[\nabla_{\theta\theta}^2 \psi(\theta^*;\xi))]$, $\otimes$ represents the Kronecker product.
\end{example}

% We prove a stronger result indicating the higher-order bias $C(\P^*, \theta^*)$. And it
\begin{proposition}[Augmented Parametric-OIC]\label{prop:pf-fit-stronger}
If $\P^* = \P_{\theta^*}\in \Pscr_{\Theta}$, and we obtain $\datax$ 
from~\Cref{ex:ddo}(a)(1) and $\nabla_{\theta} \E_{\P_{\hat{\theta}}}[h(\datax;\xi)] \neq 0$. Suppose the same conditions in \Cref{thm:main} and the optimality condition (\Cref{asp:pf-fit-optimal}) holds. Then:
\begin{equation}\label{eq:pf-fit-strong}
    \hat{A}_p = \E_{\P_{\hat{\theta}}}\Paran{h(x^*(\hat{\theta});\xi)} + \frac{1}{2n}\text{Tr}\Paran{I_h(\hat{\theta}) \Psi(\hat{\theta})} - \frac{\nabla_{\theta}\Para{\E_{\P_{\hat{\theta}}}[h(\datax;\xi)]}^{\top} C(\P^*, \theta^*)}{n}
\end{equation}
satisfies $\E[\hat{A}_p] = A + o\Para{\frac{1}{n}}$, where the definition of $I_h(\theta)$ is the same as in~\Cref{coro:oic-procedure}$(2)$.
\end{proposition}
If $\hat{\theta}$ is an unbiased estimator of $\theta^*$, i.e., $\E[\hat\theta] = \theta^*$, then $C(\P^*, \theta^*)$ = 0 in \Cref{asp:theta-asymptotics-strong} and \Cref{prop:pf-fit-stronger} reduces to \Cref{prop:pf-fit}. That is, $\hat{A}_p = \E_{\P_{\hat{\theta}}}[h(\datax;\xi)] + \frac{1}{2n}\text{Tr}[I_h(\hat{\theta})\Psi(\hat{\theta})]$.

% \begin{assumption}[Optimality Condition]\label{asp:pf-fit-optimal}
% The first-order optimality condition holds for the Problem~\eqref{eq:pf-fit}. Namely, $\bestx$ uniquely solves the equation $\nabla_x \E_{\P_{\theta^*}}[h(x;\xi)] = 0$ and $\datax$ uniquely solves the equation $\nabla_x \E_{\P_{\hat{\theta}}}[h(x;\xi)] = 0$ respectively. Besides, the second order optimality condition holds with $\nabla_{xx}^2 \E_{\P_{\theta^*}}[h(x;\xi)]$ positive definite.
% \end{assumption}

\textit{Proof of \Cref{prop:pf-fit-stronger}.} Throughout this proof, we abbreviate $x^*(\hat{\theta})$ (or $x^*(\theta^*))$ as $\hat{x}$ (or $x^*)$ respectively since we only consider the decision under the parametric distribution $\P_{\hat{\theta}}$ (or $\P_{\theta^*})$. We have $\E[\|\hat{x} - x^*\|_2^2] = O(\E[\|\hat\theta - \theta^*\|_2^2]) =  O\Para{\frac{1}{n}}$ from Assumption~\ref{asp:theta-asymptotics}.

%We have the following strategies to construct model selection through nearly unbiased decision quality estimator.

%First of all, since the true distribution $\P^*$ is the same as $\P_{\theta^*}$,
We first decompose the target quantity of evaluation as:
\begin{equation}\label{eq:param-decomp}
\begin{aligned}
    \E_{\Dscr_n}\E_{\P^*}[h(\hat{x};\xi)]=&\E_{\Dscr^n} \E_{\P_{\hat{\theta}}}[h(\hat{x};\xi)] + \E_{\Dscr_n}\E_{\P_{\hat{\theta}}}\Paran{h(x^*;\xi) - h(\hat{x};\xi)}  \\
    +&\E_{\Dscr_n}\E_{\P^*}\Paran{h(\hat{x};\xi) - h(x^*;\xi)} + \Para{\E_{\P^*}[h(x^*;\xi)] - \E_{\Dscr_n}\E_{\P_{\hat{\theta}}}[h(x^*;\xi)]}
\end{aligned}
\end{equation}

%\underline{(1) Idea 1: [Follow the steps in deriving AIC from KL divergence].} 
In the decomposition above, the second and third terms on the right-hand side of~\eqref{eq:param-decomp} involve the difference between $\hat{x} - x^*$, where the fourth term on the right-hand side of~\eqref{eq:param-decomp} captures the effects of the downstream objective due to the distribution misspecification error. 
% Denote the first term in the right-hand of~\eqref{eq:param-decomp} to be the base estimator, 
To calculate the bias, we analyze the other terms besides the first term on the right-hand side of~\eqref{eq:param-decomp}.

For the second term of the right-hand side in~\eqref{eq:param-decomp}, following a second-order Taylor expansion with Peano's remainder at the center $\hat x$ for the function $\E_{\P_{\theta}}[h(\cdot;\xi)]$, we expand $\E_{\P_{\hat{\theta}}}[h(\hat{x};\xi)]$ as follows:
\begin{equation}
    \begin{aligned}
    \E_{\P_{\hat{\theta}}}[h(x^*;\xi)]& = \E_{\P_{\hat{\theta}}}[h(\hat{x};\xi)] +  \Para{\nabla_{x}\E_{\P_{\hat{\theta}}}[h(\hat{x};\xi)]}^{\top}(x^* - \hat{x}) + \frac{1}{2}(x^* - \hat{x})^{\top}\Paran{\nabla_{xx}^2\Para{\E_{\P_{\hat{\theta}}}[h(\hat{x};\xi)]}}(x^* - \hat{x}) \\
    & + o(\|x^* - \hat{x}\|^2)\\
    & = \E_{\P_{\hat{\theta}}}[h(\hat{x};\xi)] + \frac{1}{2}(x^* - \hat{x})^{\top}\Paran{\nabla_{xx}^2\Para{\E_{\P_{\hat{\theta}}}[h(x^*;\xi)]}}(x^* - \hat{x}) + o(\|x^* - \hat{x}\|^2),
    \end{aligned}
\end{equation}
%where $\tilde{x} = \hat{x} + a(x^* -\hat{x})$ with $|a|\leq 1$. 
And the second equality follows by applying the first-order optimality condition from \Cref{asp:pf-fit-optimal} such that $\nabla_{x}\E_{\P_{\hat{\theta}}}[h(\hat{x};\xi)] = 0$. Then taking expectations, we have:
\begin{equation}\label{eq:param-fit-2}
    \E_{\Dscr_n}\E_{\P_{\hat{\theta}}}[h(\hat{x};\xi)] = \E_{\Dscr_n}\E_{\P_{\hat{\theta}}}[h(x^*;\xi)] - \frac{1}{2}\E_{\Dscr_n}\Paran{(\hat{x} - x^*)^{\top} \Para{\nabla_{xx}^2\E_{\P_{\hat{\theta}}}[h(x^*;\xi)]} (\hat{x} - x^*)} + o\Para{\frac{1}{n}}.
\end{equation}

For the third term of the right-hand side in~\eqref{eq:param-decomp}, following the same step above and noticing that $\nabla_x \E_{\P_{\theta^*}}[h(x^*;\xi)] = 0$, we obtain:
\begin{equation}\label{eq:param-fit-3}
    \E_{\Dscr_n}\E_{\P^*}[h(\hat{x};\xi))] = \E_{\P^*}[h(x^*;\xi)] + \frac{1}{2}\E_{\Dscr_n}\Paran{(\hat{x} - x^*)^{\top} \Para{\nabla_{xx}^2 \E_{\P^*}[h(x^*;\xi)]} (\hat{x} - x^*)} + o\Para{\frac{1}{n}}.
\end{equation}
%where $\tilde{x}' = x^* + b(\hat{x} - x^*)$ with $|b|\leq 1$.
We can further compute $\E\Paran{(\hat{x} - x^*)^{\top} \Para{\nabla_{xx}^2 \E_{\P^*}[h(x^*;\xi)]} (\hat{x} - x^*)} = \frac{1}{n}\text{Tr}[I_h(\theta^*)\Psi(\theta^*)] + o\Para{\frac{1}{n}}$ by algebraic manipulation (i.e. $\gradp^{\top}\bestx \nabla_{xx}^2 \E_{\P^*}[h(x^*;\xi)]\gradp \bestx = \hessianp \E_{\P^*}[h(\bestx;\xi)]$).

For the fourth term of the right-hand side in~\eqref{eq:param-decomp}, denoting $u(\theta;x) = \E_{\P_{\theta}}[h(x;\xi)]$, then taking a second-order Taylor expansion over $u(\theta;x)$ at the center of $\theta^*$, we have: 
\begin{align*}
   &\quad \E_{\P^*}[h(x^*;\xi)] - \E_{\Dscr_n}\E_{\P_{\hat{\theta}}}[h(x^*;\xi)] \\
   & = u(\theta^*;x^*) - \E_{\Dscr_n}[u(\hat{\theta};x^*)]\\
   & = -\E_{\Dscr_n}[\hat{\theta} - \theta^*]^{\top} \nabla_{\theta} u(\theta^*;x^*) - \frac{1}{2}\E_{\Dscr_n}[(\hat{\theta} - \theta^*)^{\top} \nabla_{\theta\theta}^2 u(\theta^*;x^*) (\hat{\theta} - \theta^*)] + o\Para{\frac{1}{n}},
\end{align*}
where $\E_{\Dscr_n}[\hat{\theta} - \theta^*] = \frac{C(\P^*, \theta^*)}{n}+ o\Para{\frac{1}{n}}$ from~\Cref{asp:theta-asymptotics-strong}. 
%And $\tilde{\theta} = \theta^* + c(\hat{\theta} - \theta^*)$ for $|c|\leq 1$.

%$|\E_{\P^*}[h(x^*;\xi)] - \E_{\Dscr_n}\E_{\P_{\hat{\theta}}}[h(x^*;\xi)]| \to 0$, 

%TODO: equivalence of xx and \theta\theta
Furthermore from~\Cref{lemma:matrix-expectation}, one can see that:
\begin{align*}
 \E_{\Dscr_n}[(\hat{\theta} - \theta^*)^{\top} \nabla_{\theta\theta}^2 u(\theta^*;x^*) (\hat{\theta} - \theta^*)] &= \E_{\Dscr_n}\Paran{(\hat{x} - x^*)^{\top}\E_{\P_{\theta^*}}\Paran{\nabla_{xx}^2 h(x^*;\xi)}(\hat{x} - x^*)}\\
 & = \frac{1}{n}\text{Tr}\Paran{I_h(\theta^*)\Psi(\theta^*)} + o\Para{\frac{1}{n}}.
\end{align*}

Combining all the arguments above into~\eqref{eq:param-decomp}, we have: 
\begin{align*}
    \E_{\Dscr_n}\E_{\P^*}[h(\hat{x};\xi)] &= \E_{\Dscr_n}\E_{\P_{\hat{\theta}}}[h(\hat{x};\xi)] + \E_{\Dscr_n}[(\hat{x} - x^*)^{\top}\nabla_{xx}^2 \E_{\P_{\theta^*}}[h(x^*;\xi)](\hat{x} - x^*)] \\
    & -\frac{1}{2}\E_{\hat{\theta}}[(\hat{\theta} - \theta^*)^{\top} \nabla_{\theta\theta}^2 u(\theta^*;x^*) (\hat{\theta} - \theta^*)]- \nabla_{\theta}\Para{\E_{\P_{\theta^*}}[h(x^*;\xi)]} \frac{C(\P^*, \theta^*)}{n} + o\Para{\frac{1}{n}}\\
    &= \E_{\Dscr_n}\E_{\P_{\hat{\theta}}}[h(\hat{x};\xi)] \\
    & +\frac{1}{2n}\text{Tr}\Paran{I_h(\theta^*)\Psi(\theta^*)}- \nabla_{\theta}\Para{\E_{\P_{\theta^*}}[h(\bestx;\xi)]}^{\top}\frac{C(\P^*, \theta^*)}{n} + o\Para{\frac{1}{n}}.
\end{align*}
Furthermore, if we replace $\theta^*$ with $\hat{\theta}$ above, this does not affect the bias order of $\hat A_p$ since $\E[\text{Tr}[I_h(\theta^*) \Psi(\theta^*)]] = \text{Tr}[I_h(\hat\theta) \Psi(\hat\theta)] + o\Para{\frac{1}{n}}$. Then we obtain the result of \Cref{prop:pf-fit-stronger}.$\hfill \square$

\textit{Proof of \Cref{coro:eto-misspecification}.}  From \Cref{prop:pf-fit-stronger} and the discussion in the main body, we have: 
\[\E[\hat A_p] = \E_{\Dscr_n}\E_{\P_{\theta^*}}[h(\datax;\xi)] + o(1/n).\]
Besides, from \Cref{coro:main-moment}, we have:
\[\E[\hat A] = \E_{\Dscr_n}\E_{\P^*}[h(\datax;\xi)] + o(1/n).\]
Putting these two together, we obtain $\E[\hat A - \hat A_p] = \Escr(\paranx;\Pscr_{\Theta}) + o(1/n)$.
$\hfill \square$

%Below, we first illustrate the case where the problem is with constraints.

\subsection{Proofs in Section~\ref{sec:context}}
\textit{Proof of \Cref{coro:context}.}~For any data-driven solution, we decompose the true performance $A_{con} = \E_{\Dscr_n}\E_z \E_{\P_{\xi|z}^*}[h(x^*(\hat{\theta},z);\xi)]$ as:
\begin{equation*}
    \begin{aligned}
        A_{con} &= A_{con}' + \E_{z}\E_{\P_{\xi|z}^*}[h(\bestx;\xi_i)] - \frac{1}{n}\sum_{i = 1}^n h(x^*(\theta^*,z_i);\xi_i)\\
    \end{aligned}
\end{equation*}
where $A_{con}' = \E_{\Dscr_n}\E_z \E_{\P_{\xi|z}^*}[h(x^*(\hat{\theta},z);\xi) - h(x^*(\theta^*,z);\xi)] + \frac{1}{n}\sum_{i = 1}^n h(x^*(\theta^*,z_i);\xi_i)$. It is easy to see that:
\begin{equation}\label{eq:unbias-best2}
\begin{aligned}
    &\quad \E\Paran{\frac{1}{n}\sum_{i = 1}^n h(x^*(\theta^*,z_i);\xi_i)} - \E_{z}\E_{\P_{\xi|z}^*}[h(x^*(\theta^*,z);\xi)] \\
    &= \E_{z}\E_{\P_{\xi|z}^*}[h(x^*(\theta^*,z);\xi)] - \E_{z}\E_{\P_{\xi|z}^*}[h(x^*(\theta^*,z);\xi)]  = 0. 
\end{aligned}
\end{equation}
Therefore, in order to show $\E[\hat{A}_{con}] = A + o\Para{\frac{1}{n}}$, we only need to show $\E[\hat{A}_{con} - A_{con}'] = o\Para{\frac{1}{n}}$. We further decompose the difference by:
\begin{equation}
    \begin{aligned}
        A' & = \frac{1}{n}\sum_{i = 1}^n h(x^*(\hat{\theta},z_i);\xi_i) + \underbrace{\frac{1}{n}\sum_{i = 1}^n h(x^*(\theta^*,z_i);\xi_i)  - \frac{1}{n}\sum_{i = 1}^n h(x^*(\hat{\theta},z_i);\xi_i)}_{T_1}\\
&+\underbrace{\E_{\Dscr_n}\E_{\P^*}[h(x^*(\hat{\theta},z);\xi) - h(x^*(\theta^*,z);\xi)]}_{T_2}\\
    \end{aligned}
\end{equation}

The following steps maintain the same structure as the analysis in the proof of \Cref{thm:main}.
$\hfill \square$
% Assuming that we can get access to the solution $x^*(\hat{\theta},z_i)$ for each $z_i$ quickly under the given parametric model. 

% Meanwhile, we can evaluate the effects of the ``best" expected model misspecification error to the downstream optimization problem for the data-driven solution $\hat{x}$ (abbr. of $x^*(\hat{\theta};z)$), i.e.  with:
% \[\hat{B} = \hat{A} - \frac{1}{n}\sum_{i = 1}^n A_{z_i},\]
% such that $\E[\hat{B}] = B + o\Para{\frac{1}{n}}$.
%\subsection{Model-Agnostic Approach}

\textit{Proof of \Cref{prop:contextual-misspecified}.} Based on the same arguments as \Cref{prop:pf-fit},  we have:
\[\E[\hat A_z] = \E_{\Dscr_n}\E_{\xi \sim \P_{\theta^*|z}} [h(x^*(\hat\theta, z);\xi)] + o\Para{\frac{1}{n}}, \forall z \in \Zscr.\]
This is because if we fix the conditional covariate $z$, then the problem reduces to the standard stochastic optimization problem. Then we have:
\begin{align*}
    \E\Paran{\frac{1}{n}\sum_{i = 1}^n \hat A_{z_i}}& = \E_{\{z_i\}_{i \in [n]}}\Paran{\frac{1}{n}\sum_{i = 1}^n \E_{\Dscr_n}\E_{\xi \sim \P_{\theta^*|z_i}}[h(x^*(\theta, z);\xi)]} + o\Para{\frac{1}{n}}\\
    & = \E_{\Dscr_n}\E_{z\sim \P_z^*}\E_{\xi \sim \P_{\theta^*|z}}[h(x^*(\theta,z);\xi)] + o\Para{\frac{1}{n}},
\end{align*}
where the second equality follows from the fact that each $z_i$ is drawn i.i.d. from $\P_z^*$. Recall that from \Cref{coro:context}, we have:
\[\E[\hat A_{con}] = \E_{\Dscr_n}\E_{{z}} \E_{\P_{\xi|{z}}^*}[h(x^*(\hat\theta,z);\xi)] + o\Para{\frac{1}{n}}.\]
Combining the differences between the two equations above, we have: 
\[\E\Paran{\hat A_{con} - \frac{1}{n}\sum_{i = 1}^n A_{z_i}} = \Escr(x^*(\theta,z);\Pscr_{\Theta|z}) + o\Para{\frac{1}{n}}.~\hfill \square\]

\section{Proofs in \Cref{sec:challenge}}\label{app:challenge}
\textit{Proof of~\Cref{coro:K-Fold}.}
The performance of $K$-fold CV is evaluated as $\hat A_{kcv} = \frac{1}{K}\sum_{i=1}^K \sum_{i \in [\Dscr_{(k)}]}h(x^*(\hat\theta_{-k});\xi_i)$, where $\Dscr_n$ is partitioned into $K$ folders $\Dscr_{(k)}$ with $|\Dscr_{(k)}| = n / K,\forall k \in [K]$. Denote $\hat\theta_{-k} = T(\Dscr_n \backslash \Dscr_{(k)})$. Then we have: $\E[\hat A_{kcv}] = \E_{\Dscr_n \backslash \Dscr_{(k)}}\E_{\P^*}[h(x^*(\hat\theta_{-k});\xi)]$ and the performance difference between $\E[\hat A_{kcv}]$ and $A$ is given by:
\begin{align*}
    \E[\hat A_{kcv}] - A & = \E_{\Dscr_n \backslash \Dscr_{(k)}}\E_{\P^*}[h(x^*(\hat\theta_{-k});\xi)] - \E_{\Dscr_n}\E_{\P^*}[h(x^*(\hat\theta);\xi)]\\
    & = (\E_{\Dscr_n \backslash \Dscr_{(k)}}\E_{\P^*}[h(x^*(\hat\theta_{-k});\xi)] - \E_{\Dscr_n}\E_{\P^*}[h(x^*(\theta^*);\xi)]) \\
    & - (\E_{\Dscr_n}\E_{\P^*}[h(x^*(\hat\theta);\xi)] - \E_{\Dscr_n}\E_{\P^*}[h(x^*(\theta^*);\xi)]).
\end{align*}
Each term above uses the same argument as analyzing $T_2$ from \eqref{eq:t2-expand} in the proof of Theorem~\ref{thm:main} with the corresponding sample sizes $(K - 1)n/K$ and $n$. Specifically, for the true expected performance gap $T_2$ when the sample size is $n$ (denoted as $PG(n)$), we have:
\begin{align*}
    PG(n) & = \frac{\text{Tr}[I_h(\theta^*)\Psi(\theta^*)]}{2n} + (\gradp \E_{\P^*}[h(\bestx;\xi)])^{\top}\E_{\Dscr_n}[\Delta] + o\Para{\frac{1}{n}} \\
    & = \frac{\text{Tr}[I_h(\theta^*)\Psi(\theta^*)]}{2n} + (\gradp \E_{\P^*}[h(\bestx;\xi)])^{\top}(\E[\hat\theta] - \theta^*) + o\Para{\frac{1}{n}}.
\end{align*}
That is, $PG(n) = \frac{B}{n} + o\Para{\frac{1}{n}}$ with $B = \frac{\text{Tr}[I_h(\theta^*)\Psi(\theta^*)]}{2} + (\gradp \E_{\P^*}[h(\bestx;\xi)])^{\top}n(\E[\hat\theta] - \theta^*)$. Then we can obtain the result with:
\[\E[\hat A_{kcv}] - A = PG\Para{\frac{(K-1)n}{K}} - PG(n)  = \frac{B}{(K-1)n} + o\Para{\frac{1}{n}}. \quad \hfill \square\]

\textit{Proof of Corollary~\ref{prop:loocv}.} 
% The moment equivalence argument immediately holds since 
% %, $\E[\hat A - \hat A_{loocv}] = o\Para{\frac{1}{n}}$ holds by noticing 
% $\E[\hat A_{loocv}] = A + o\Para{\frac{1}{n}}$ and $\E[\hat A] = A + o\Para{\frac{1}{n}}$. 
First, we show the probabilistic equivalence argument between OIC and LOOCV, i.e., $n(\hat A - \hat A_{loocv}) \overset{p}{\to}0$. Following the first-order Taylor expansion  for each $h(x^*(\hat\theta_{-i});\xi_i)$ at the center of $\hat\theta$, we have:
\begin{equation}\label{eq:taylor-loocv}
\sum_{i = 1}^n h(x^*(\hat{\theta}_{-i});\xi_i) = \sum_{i = 1}^n h(\datax;\xi_i) + \sum_{i = 1}^n(\hat{\theta}_{-i} - \hat{\theta})^{\top} \nabla_{\theta} h(x^*(\tilde{\theta}_{-i});\xi_i),
\end{equation}
where $\tilde{\theta}_{-i}: = \hat{\theta} + a_i (\hat{\theta}_{-i} - \hat{\theta})$ with $|a_i|\leq 1,\forall i \in [n]$. The first term on the right-hand side of~\eqref{eq:taylor-loocv} is $n \hat A_o$. Therefore, we only need to analyze the second term on the right-hand side of~\eqref{eq:taylor-loocv} and compare it with the asymptotics of $n \hat A_c$.

%Then following~\Cref{asp:theta-asymptotics} and specifc condition, we have:
Based on the expansion from \Cref{asp:theta-asymptotics}, we have:
\begin{equation}\label{eq:loocv-theta}
    \hat{\theta}_{-i} - \hat\theta = - \frac{\IFx(\xi_i)}{n} + \frac{1}{n(n-1)}\sum_{j = 1, j \neq i}^n \IFx(\xi_j) + o_p\Para{\frac{1}{\sqrt n}}.
\end{equation}
Plugging~\eqref{eq:loocv-theta} back into the second part of the right-hand side in~\eqref{eq:taylor-loocv}, we then obtain:
\begin{equation}\label{eq:loocv-final}
    \begin{aligned}
        \sum_{i = 1}^n(\hat{\theta}_{-i} - \hat{\theta})^{\top} \nabla_{\theta}h(x^*(\tilde{\theta}_{-i});\xi_i) & = \sum_{i = 1}^n \frac{1}{n (n - 1)}\sum_{j =1, j\neq i}^n \IFx(\xi_j)^{\top} \nabla_{\theta}h(x^*(\tilde{\theta}_{-i});\xi_i)\\
        &- \frac{1}{n}\sum_{i = 1}^n \IFx(\xi_i)^{\top}\nabla_{\theta} h(x^*(\tilde{\theta}_{-i});\xi_i) + o_p(1),
    \end{aligned}
\end{equation}
where the last term converges in probability to zero because $\sum_{i = 1}^n o_p(n^{-1/2}) \nabla_{\theta} h(x^*(\tilde \theta_{(-i)});\xi_i) \leq o_p(n^{-1/2})\sum_{i = 1}^n \nabla_{\theta} h(x^*(\tilde \theta_{(-i)});\xi_i) = o_p(n^{-1/2}) \times O_p(\sqrt n) = o_p(1)$. Then starting from~\eqref{eq:loocv-final}, we immediately obtain the following arguments from~\Cref{asp:theta-asymptotics}:
\begin{itemize}
    \item $\hat{\theta}_{-i} \convp \theta^*$ as $n \to \infty$ for each $i \in [n]$; It follows simply by observing $\hat{\theta} \convp \theta^*$ as $n \to \infty$.
    \item $\tilde{\theta}_{-i} \convp \hat{\theta}$ since $|a_i|\leq 1$.
\end{itemize}
Given the properties above, we analyze the two terms of the right-hand side of~\eqref{eq:loocv-final} respectively.

For the first term of the right-hand side in~\eqref{eq:loocv-final}, we have:
\begin{equation*}
\begin{aligned}
    \sum_{i = 1}^n \frac{1}{n (n - 1)}\sum_{j =1, j\neq i}^n \IFx(\xi_j)^{\top}h(x^*(\tilde{\theta}_{-i});\xi_i) &=\frac{1}{n(n- 1)}\sum_{i = 1}^n\sum_{j = 1}^n \IFx(\xi_j)^{\top}h(x^*(\tilde{\theta}_{-i});\xi_i)\\
    &- \sum_{i = 1}^n \frac{1}{n(n - 1)}\IFx(\xi_i)^{\top}h(x^*(\tilde{\theta}_{-i});\xi_i) \\
    &\overset{p}{\to} 0 - 0 = 0,
\end{aligned}
\end{equation*}
where the first term of the third line is based on the fact that $\E_{\P^*}[\IFx(\xi)] = 0$ and therefore, $\frac{1}{n}\sum_{i = 1}^n \IFx(\xi_i) = o_p(1)$. And the second term $ \sum_{i = 1}^n \frac{1}{n(n - 1)}\IFx(\xi_i)^{\top}h(x^*(\tilde{\theta}_{-i});\xi_i) \overset{p}{\to} 0$ above is due to:
\[\frac{1}{n}\sum_{i = 1}^n \IFx(\xi_i)^{\top} \nabla_{\theta} h(x^*(\tilde{\theta}_{-i});\xi_i) \convp \frac{1}{n}\sum_{i = 1}^n \IFx(\xi_i)^{\top}\nabla_{\theta} h(x^*(\theta^*);\xi_i) \convp \E_{\P^*}[\nabla_{\theta} h(x^*(\theta);\xi) ^{\top}\IFx(\xi)] < \infty.\]

For the second term of the right-hand side in~\eqref{eq:loocv-final}, recalling the definition of $\hat A_c^*$ in~\eqref{eq:bias-notation}, we have $\frac{1}{n}\sum_{i = 1}^n \IFx(\xi_i) h(x^*(\tilde{\theta}_{-i});\xi_i) - \hat A_c^* \overset{p}{\to} 0$. Therefore, 
\begin{equation*}
    \frac{1}{n}\sum_{i = 1}^n h(x^*(\tilde{\theta}_{-i});\xi_i)^{\top}\IFx(\xi_i) \convp \E_{\P^*}[\nabla_{\theta} h(x^*(\theta^*);\xi) ^{\top}\IFx(\xi)].
\end{equation*}
Combining the previous two result and plugging them into \eqref{eq:loocv-final}, we then obtain:
\[\sum_{i = 1}^n(\hat{\theta}_{-i} - \hat{\theta})^{\top}h(x^*(\tilde{\theta}_{-i});\xi_i) \convp -\E_{\P^*}[\nabla_{\theta} h(x^*(\theta^*);\xi) ^{\top}\IFx(\xi)].\] 
which is the same as the asymptotics of $n\hat{A}_c$. Therefore, we have: $n(\hat{A} - \hat{A}_{loocv}) \convp 0$. 

For the probabilistic difference argument between the empirical objective and LOOCV, it is easy to see:
\[n(\hat A_o - \hat A_{loocv}) = \E_{\P^*}[\nabla_{\theta} h(\bestx;\xi)^{\top}\IFx(\xi)] + o_p(1).\]
Therefore, $\hat A_o - \hat A_{loocv} = O_p(1/n)$. $\hfill \square$

\textit{Proof of~\Cref{coro:alo-bias}.}~The expected performance of the ALO estimator is given as:
\begin{align*}
    \E[\hat A_{alo}] & = \E\Paran{\frac{1}{n}\sum_{i = 1}^n h(x^*(\tilde\theta_{-i});\xi_i)} \\
    & = \E\Paran{\frac{1}{n}\sum_{i = 1}^n h(\datax;\xi_i) + \gradp h(\datax;\xi)^{\top}(\tilde\theta_{-i} - \hat\theta) + o_p(|\tilde\theta_{-i} - \hat\theta|)}\\
    & = \E\Paran{\frac{1}{n}\sum_{i = 1}^n h(\datax;\xi_i) + \frac{1}{n^2}\sum_{i = 1}^n\gradp h(\datax;\xi)^{\top}\widehat{\IFx}(\xi_i)} + o\Para{\frac{1}{n}} = \E[\hat A]+ o\Para{\frac{1}{n}},
\end{align*}
where we take a second-order Taylor expansion at the center of $\hat\theta$ for each $i \in [n]$ and plug in the corresponding one-step approximation $\tilde\theta_{-i} - \hat\theta = \frac{1}{n}\widehat{\IFx}(\xi_i)$ to obtain the result.
$\hfill \square$

\section{Further Numerical Details in \Cref{sec:numerical}}\label{app:numeric}
Each experiment was run on a cluster using 24 cores from an Intel Xeon Gold 6126 Processor and 16 GB of memory. We use \texttt{cvxpy} with the \texttt{MOSEK} solver to solve the optimization problems in Sections~\ref{subsec:portfolio} and \ref{subsec:numerical-newsvendor}.

%All the codes presented in this paper are available at: \url{https://github.com/wangtianyu61/oic_codes}.

% \subsection{Discussion on Influence Functions}
% Throughout the two case studies, we consider the moment estimators $\hat{\mu} = \frac{1}{n}\sum_{i = 1}^n \xi_i$ and $\hat{\sigma}^2 =\frac{1}{n}\sum_{i = 1}^n (\xi_i - \hat{\mu})^2$. Then it is easy to see the influence functions of mean, variance estimators from i.i.d. samples, e.g.:
% \begin{equation}\label{eq:if-mean-var}
% \begin{aligned}
%     IF_{\hat{\mu}}(\xi_i) &= \xi_i - \mu \\
%     IF_{\hat{\sigma}^2}(\xi_i) & = (\xi_i - \mu)^2 - \sigma^2,
% \end{aligned}
% \end{equation}
% where $\mu = \E[\xi]$ and $\sigma^2 = \var[\xi]$. We can approximate them by the empirical influence functions:
% \begin{equation}\label{eq:eif-mean-var}
% \begin{aligned}
%     \widehat{IF}_{\hat{\mu}}(\xi_i) &= \xi_i - \hat{\mu}\\
%     \widehat{IF}_{\hat{\sigma}^2}(\xi_i) & = (\xi_i - \hat{\mu})^2 - \hat{\sigma}^2.
% \end{aligned}
% \end{equation}
% We can check $\E[\|\widehat{IF}_{\hat{\mu}}(\xi)\|^2] = \|IF_{\mu}(\xi)\|^2 + o(1), \E[\|\widehat{IF}_{\hat{\sigma}^2}(\xi)\|^2] = \|IF_{\sigma^2}(\xi)\|^2 + o(1)$ as long as $\xi$ has fourth moments. 
%For simplicity, we remove constraints for now and investigate the pure objective performance.

% \subsection{Gradient Evaluation in General Models}

% We use \texttt{torch} to compute the gradients and hessian when it is difficult to manipulate by hand.

\subsection{Portfolio Allocation}\label{app:portfolio}
\subsubsection{Detailed Setups.}\label{app:setup}
%\paragraph{DGP.}~
%The asset returns $\xi = (\xi_A, \xi_B)^{\top}$ with $\xi_A \sim N(\mu_A, \Sigma_A)$, $\xi_B \sim N(\mu_B, \Sigma_B)$, $\xi_A \perp \xi_B$, $D_{\xi_A} = D_{\xi_B} = \frac{1}{2}D_{\xi}$ denoting two classes of assets. 
Recall the notations $\mu_A, \mu_B, \Sigma_A, \Sigma_B$ in \Cref{subsec:portfolio}, for each instance, we generate each entry of $\mu_A, \mu_B$ by $U(0, 4)$ and $\Sigma_A, \Sigma_B = CC^{\top}$ with each entry $c_{ij} \sim U(0, \frac{1}{2})$ from $C = \{c_{ij}\}_{i,j \in [D_{\xi_A}]}$. We further denote $A(\xi) = (\xi - \E\xi)(\xi - \E\xi)^{\top}\in \R^{D_{\xi_A}\times D_{\xi_A}}$. 

% \paragraph{Evaluation.} For the model bias evaluated in the main body, we compute the true one with $A - \hat{A}_o$, where $A$ is the true cost of the decision and $\hat{A}_o$ is the empirical cost of the decision,  and bias of other evaluation methods with $A - \hat{A}$ where $\hat{A}$ is the cost of the considered decision.

\subsubsection{Decision Rules.} We detail the following decision rules and their bias computations:

\textbf{(i) E2E under different mappings.} More specifically, we consider the following three mappings:

For SAA-U, we restrict the space of $\Xscr$ such that the allocation amount of each asset is the same, i.e., $x^*(\theta) = \theta {1}_{D_{\xi}}$ with $\theta \in \R$. Then the space $\Theta = \R$. Plugging in the expression of the decision rule, we have:
$$ h(x^*(\theta);\xi) = (\sum_{i,j}^d A(\xi)_{i,j} + \lambda_2 d)\theta^2 - (\lambda_1 \xi^{\top} \mathbf{1}_d)\theta.$$ 

In this case, $\text{Tr}[I_h(\theta^*)^{-1}J_h(\theta^*)]$ can be expressed (and approximated) by:
\begin{align*}
    \text{Tr}[I_h(\theta^*)^{-1}J_h(\theta^*)] &= \frac{\E_{\P^*}[((\sum_{i,j}^d A(\xi)_{i,j} + \lambda_2 d) 2\theta^* - \lambda_1 \xi^{\top}\mathbf{1}_d)^2]}{\E_{\P^*}[{2(\sum_{i,j}^d A(\xi)_{i,j} + \lambda_2 d)}]},\\
    &\approx \frac{\E_{\hat{\P}_n}[((\sum_{i,j}^d A(\xi)_{i,j} + \lambda_2 d) 2\hat{\theta} - \lambda_1 \xi^{\top}\mathbf{1}_d)^2]}{\E_{\hat{\P}_n}\left[2(\sum_{i,j}^d A(\xi)_{i,j} + \lambda_2 d)\right]}.
\end{align*}

For SAA-B, we have: $\paranx = (\theta_1 {1}_{D_{\xi_A}}^{\top}, \theta_2 {1}_{D_{\xi_B}}^{\top})^{\top}$ with $\theta_1, \theta_2 \in \R$. We can plug in the corresponding expression of $\paranx$ and obtain the bias formula by approximating $\text{Tr}[I_h(\theta^*)^{-1} J_h(\theta^*)]$ too.

For SAA, we have: $x^*(\theta) = \theta \in \Theta$, i.e., the space $\Theta = \R^{D_{\xi}}$. And the bias can be approximated by:
\begin{align*}
    \text{Tr}[I_h(\theta^*)^{-1}J_h(\theta^*)] &= \text{Tr}\left(\frac{\E_{\P^*}[(A(\xi) + \lambda_2 I)^{-1}]}{2} \E_{\P^*}[(2(A(\xi) + \lambda_2 I)\theta^* - \lambda_1 \xi)(2(A(\xi) + \lambda_2 I)\theta^* - \lambda_1 \xi)^{\top}]\right)\\
    &\approx \text{Tr}\left(\frac{\E_{\hat{\P}_n}[(A(\xi) + \lambda_2 I)^{-1}]}{2} \E_{\hat{\P}_n}[(2(A(\xi) + \lambda_2 I)\hat{\theta} - \lambda_1 \xi)(2(A(\xi) + \lambda_2 I)\hat{\theta} - \lambda_1 \xi)^{\top}]\right).
\end{align*}

%[TDOO: HOW SAA INVOLVES WITH COMPLEXITY, SHOW IT INCREASES WITH $D_{\xi}^2$]

\textbf{(ii) ETO set to be Gaussian models with independent margins.} Empirically, to obtain $\paranx$ in~\Cref{ex:portfolio}$(a)$, we sample 100$\times \texttt{sample size}$ points from $\P_{\theta}$ and call the standard optimization solver. To calculate the bias term $\hat{A}_c$, we obtain the formula $\nabla_{\theta} \paranx$ through~\eqref{eq:inverse} and $\widehat{\IFx}(\xi) = (\xi - \hat{\mu}_1,\ldots,\xi - \hat{\mu}_{D_{\xi}}, (\xi - \hat{\mu}_1)^2 - \hat{\sigma}_1,\ldots, (\xi - \hat{\mu}_{D_{\xi}})^2 - \hat{\sigma}_{D_{\xi}}^2)^{\top}$. Then we plug them into~\eqref{eq:oic}. 

\textbf{(iii) $\chi^2$-divergence-based DR-E2E.} For $\chi^2$-divergence-based DRO, we set $f(t) = t^2 - 1$ in $d_f(\P, \Q) = \int f\Para{\frac{d\P}{d\Q}} d\Q$ in \Cref{ex:ddo}$(b)(3)$. We have: $x^*(\theta) = \theta$. Since $h(x;\xi)$ is convex, we can directly apply results in \cite{ben2013robust,duchi2019variance} to obtain convex reformulations for this type of problem and call the standard convex solvers. And the bias for this DR-E2E is directly obtained from \Cref{coro:oic-procedure}$(2)$. 

\subsubsection{Additional Results.}\label{app:experiment-risk}
First, we evaluate the performances with problem instances varying sample size $n$ and $D_{\xi}$ to understand the variability of the bias across different decision rules in Tables~\ref{tab:add-bias1},~\ref{tab:add-bias2} and~\ref{tab:add-bias3}. Due to computational considerations, we only report results of OIC against 2, 5, 10-fold cross-validation approaches. In these three tables, the observation is consistent with Table~\ref{tab:portfolio-complexity} in the main body. OIC can estimate the optimistic bias quite closely to the true bias across all different methods compared with those cross-validation methods. Besides, $\chi^2$-DRO incurs less bias than the empirical counterpart across different setups. And the intrinsic optimistic bias of ETO under Gaussian models with independent margins is quite small.
 
%from the two Figures that no method can outperform OIC.

\begin{table}[htbp]
\centering
\caption{Estimated bias of each method with $n = 100, D_{\xi} = 10, \rho = 3$}
\label{tab:add-bias1}
\begin{tabular}{l|ccccc}
\toprule
 & Oracle       & OIC & 2-CV & 5-CV & 10-CV \\
 \midrule
SAA&0.045\scriptsize $\pm 0.030$&0.047\scriptsize $\pm 0.013$&0.078\scriptsize $\pm 0.028$&0.055\scriptsize $\pm 0.017$&0.051\scriptsize $\pm 0.015$\\
SAA-U&0.027\scriptsize $\pm 0.027$&0.023\scriptsize $\pm 0.007$&0.040\scriptsize $\pm 0.018$&0.026\scriptsize $\pm 0.009$&0.025\scriptsize $\pm 0.008$\\
SAA-B&0.039\scriptsize $\pm 0.029$&0.037\scriptsize $\pm 0.011$&0.062\scriptsize $\pm 0.024$&0.043\scriptsize $\pm 0.014$&0.040\scriptsize $\pm 0.012$\\
ETO&0.026\scriptsize $\pm 0.056$&0.030\scriptsize $\pm 0.010$&0.018\scriptsize $\pm 0.009$&0.026\scriptsize $\pm 0.007$&0.028\scriptsize $\pm 0.008$\\
DRO&0.044\scriptsize $\pm 0.028$&0.042\scriptsize $\pm 0.011$&0.083\scriptsize $\pm 0.027$&0.055\scriptsize $\pm 0.016$&0.051\scriptsize $\pm 0.014$\\
 \bottomrule
\end{tabular}
\end{table}
 
\begin{table}[htbp]
\centering
\caption{Estimated bias of each method with $n = 100, D_{\xi} = 20, \rho = 3$}
\label{tab:add-bias2}
\begin{tabular}{l|ccccc}
\toprule
 & Oracle      & OIC & 2-CV & 5-CV & 10-CV \\
 \midrule
SAA&0.326\scriptsize $\pm 0.108$&0.296\scriptsize $\pm 0.061$&0.561\scriptsize $\pm 0.130$&0.396\scriptsize $\pm 0.091$&0.370\scriptsize $\pm 0.081$\\
SAA-U&0.106\scriptsize $\pm 0.081$&0.113\scriptsize $\pm 0.024$&0.204\scriptsize $\pm 0.058$&0.147\scriptsize $\pm 0.037$&0.139\scriptsize $\pm 0.032$\\
SAA-B&0.178\scriptsize $\pm 0.090$&0.177\scriptsize $\pm 0.038$&0.322\scriptsize $\pm 0.081$&0.231\scriptsize $\pm 0.056$&0.217\scriptsize $\pm 0.051$\\
ETO&0.294\scriptsize $\pm 0.562$&0.498\scriptsize $\pm 0.114$&-0.018\scriptsize $\pm 0.196$&0.318\scriptsize $\pm 0.081$&0.402\scriptsize $\pm 0.107$\\
DRO&0.288\scriptsize $\pm 0.090$&0.236\scriptsize $\pm 0.047$&0.483\scriptsize $\pm 0.109$&0.345\scriptsize $\pm 0.077$&0.323\scriptsize $\pm 0.068$\\
 \bottomrule
\end{tabular}
\end{table}

\begin{table}[htbp]
\centering
\caption{Estimated bias of each method with $n = 50, D_{\xi} = 20, \rho = 3$}
\label{tab:add-bias3}
\begin{tabular}{l|ccccc}
\toprule
 & Oracle & OIC & 2-CV & 5-CV & 10-CV \\
 \midrule
SAA&0.168\scriptsize $\pm 0.055$&0.154\scriptsize $\pm 0.027$&0.256\scriptsize $\pm 0.053$&0.187\scriptsize $\pm 0.036$&0.175\scriptsize $\pm 0.034$\\
SAA-U&0.062\scriptsize $\pm 0.048$&0.059\scriptsize $\pm 0.012$&0.090\scriptsize $\pm 0.028$&0.069\scriptsize $\pm 0.015$&0.064\scriptsize $\pm 0.014$\\
SAA-B&0.097\scriptsize $\pm 0.049$&0.092\scriptsize $\pm 0.017$&0.147\scriptsize $\pm 0.037$&0.109\scriptsize $\pm 0.023$&0.102\scriptsize $\pm 0.022$\\
ETO&0.211\scriptsize $\pm 0.289$&0.247\scriptsize $\pm 0.055$&-0.081\scriptsize $\pm 0.073$&0.148\scriptsize $\pm 0.038$&0.195\scriptsize $\pm 0.045$\\
DRO&0.151\scriptsize $\pm 0.048$&0.130\scriptsize $\pm 0.022$&0.232\scriptsize $\pm 0.044$&0.170\scriptsize $\pm 0.031$&0.158\scriptsize $\pm 0.029$\\
 \bottomrule
\end{tabular}
\end{table}

\begin{figure}
    \centering
    \subfloat[$(n, D_{\xi}) = (50, 30)$]
    {
        \begin{minipage}[t]{0.33\textwidth}
            \centering
            \includegraphics[width = 0.9\textwidth]{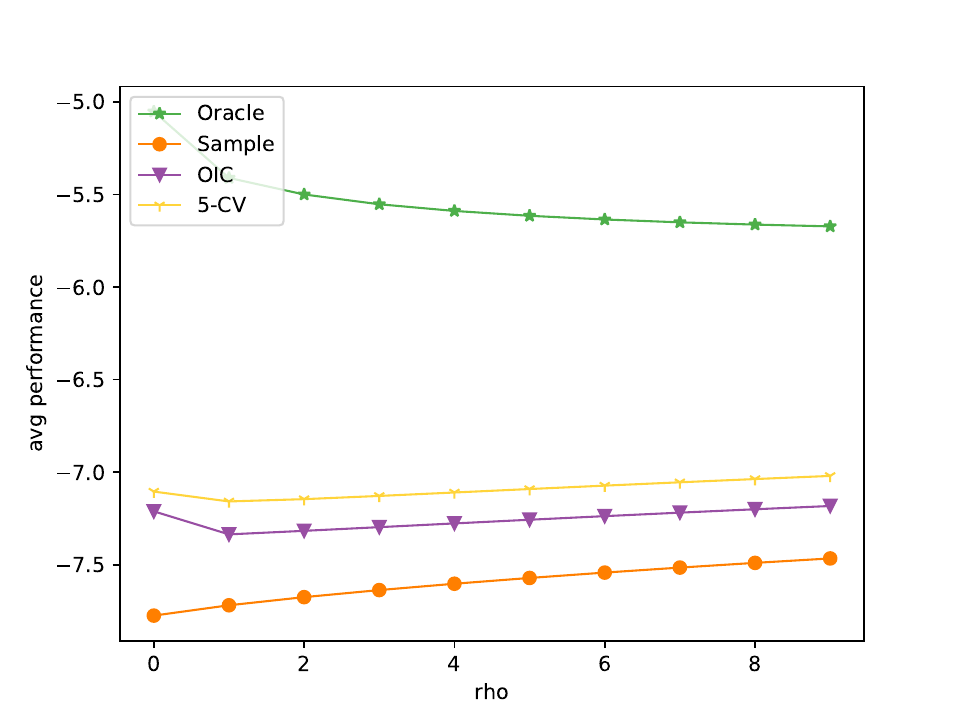}
        \end{minipage}
    }
    \subfloat[$(n, D_{\xi}) = (100, 20)$]
    {
        \begin{minipage}[t]{0.33\textwidth}
            \centering
            \includegraphics[width = 0.9\textwidth]{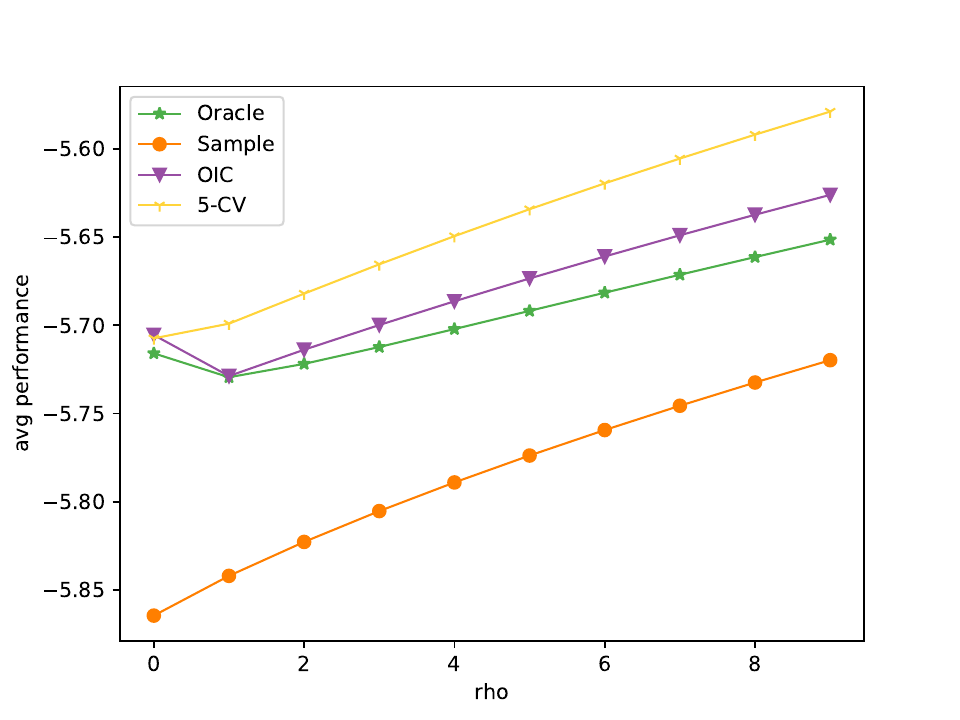}
        \end{minipage}
    }
    \subfloat[$(n, D_{\xi}) = (100, 40)$]
    {
        \begin{minipage}[t]{0.33\textwidth}
            \centering
            \includegraphics[width = 0.9\textwidth]{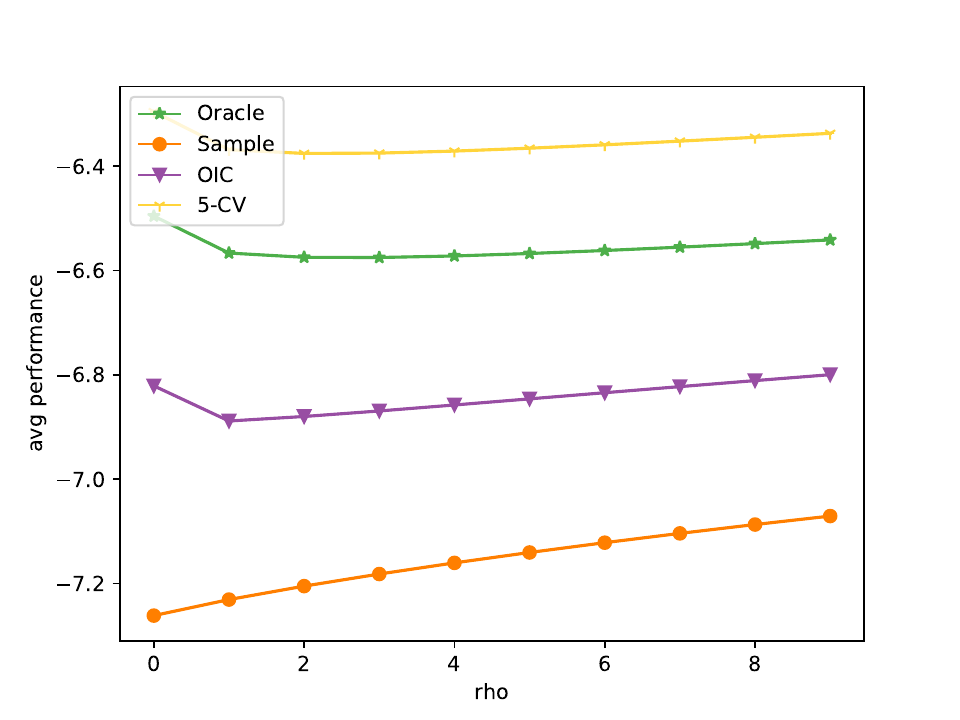}
        \end{minipage}
    }
    \caption{Estimated performances of $\chi^2$-DRO varying ambiguity levels $\rho$}
    \label{fig:app-dro}
\end{figure}

Then we report the performances of OIC in evaluating $\chi^2$-DRO varying sample size $n$ and feature dimension $D_{\xi}$ with respect to the ambiguity level $\epsilon = \frac{\rho}{n}$ in \Cref{fig:app-dro}. In all these scenarios, OIC can approximately identify ambiguity levels $\rho$ that DRO methods outperform the empirical counterpart (the method with $\rho = 0$). OIC helps in making better decisions and achieving performance that is comparable to or better than other evaluation criteria. 

\begin{figure}
    \centering
    \includegraphics[width=0.5\linewidth]{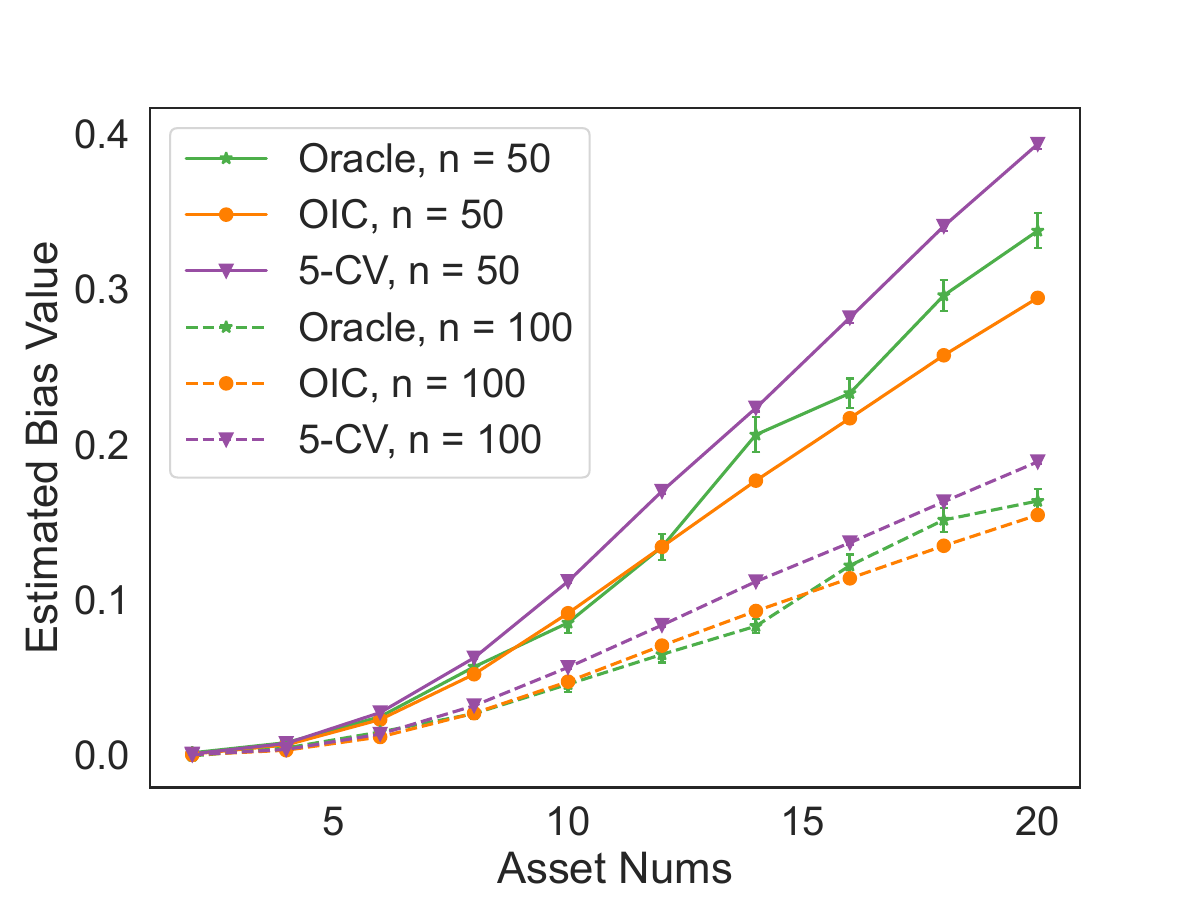}
    \caption{Estimated Bias of SAA varying $D_{\xi}$ and $n$}
    \label{fig:estimate-bias-saa}
\end{figure}

% \subsubsection{Other models}
% For $\gamma_0, \gamma_1, \gamma_2 > 0$, we consider the exponential utility function for the portfolio case with
% \[h(x;\xi) =\gamma_0 \exp(-\gamma_1 \xi^{\top}x) + \gamma_2 x^{\top}x.\]
% Under Gaussian models with independent margins, we have:
% \[\E_{\P_{\theta}}[h(x;\xi)] = \exp\Paran{-\gamma_1 \mu^{\top}x + \frac{1}{2}\gamma_1^2\Para{\sum_{i = 1}^d \sigma_i^2 x_i^2}}  + \gamma_2 x^{\top}x.\]

\wty{We also show the bias performances of OIC in evaluating SAA. From \Cref{fig:estimate-bias-saa}, it is clear that the bias of SAA grows with $D_{\xi}$ nonlinearly. Our OIC can approximate the true bias well across each problem specification.}

% \subsubsection{Results for General Risk Problem}
Finally, we calculate the estimated biases of the decision rules for general risk functions $g$ in \Cref{sec:general-risk} in Table~\ref{tab:portfolio-risk-complexity} as follows.

\begin{table}[htb]
    \centering
    \caption{Estimated Risk Bias in $A_u - \E_{\Dscr_n}[u(\hat A_o)]$ with $n = 50, D_{\xi} = 10, \rho = 3$ for Risk-OIC against the true risk bias.}
    \label{tab:portfolio-risk-complexity}
    \resizebox{\textwidth}{!}{
    \begin{tabular}{c|c|ccccc}
    \toprule
        Method & $u(x)$ & SAA & SAA-U & SAA-B & ETO & $\chi^2$-DRO\\
        \midrule
       Oracle Risk Bias  & \multirow{2}{*}{$(50 - (1 - x)^2)^+$} & 0.917&0.464&0.726&0.246&0.851\\
       Risk-Incorporated OIC & & 0.901\scriptsize $\pm 0.100$&0.288\scriptsize $\pm 0.041$&0.682\scriptsize $\pm 0.050$&0.180\scriptsize $\pm 0.023$&0.761\scriptsize $\pm 0.080$\\
       \bottomrule
    \end{tabular}}
\end{table}

\subsection{Newsvendor Problem}\label{app:newsvendor}
\subsubsection{Detailed Setups.}\label{app:newsvendor-setup}
% For the single-item newsvendor objective, the standard problem is given by:
% \[h(x;\xi) = c x - p \min\{\xi,x\},\]
% where $x$ is the order quantity, $\xi$ is the random demand. 
% % 
% We consider $p = 5, c = 2$ here.
In the evaluation, we consider cross-validation with the fold number being $K = 2, 3, 4, 5, 10$ and $K = n$ (LOOCV). We consider other model-free approaches including the bootstrap with $B = 10, 50$ and the jackknife, which are discussed in Appendix~\ref{app:btsp-jcnf}.

\subsubsection{Additional Results.}\label{app:newsvendor-additional}
We conduct the same evaluation setups as \cite{siegel2021profit}, i.e., we conduct 100 independent runs to report the decision performance as one batch and average over 100 independent batches to calculate the associated error compared with the true performance. 

\begin{figure}
    \centering
    \subfloat[SAA, $n = 25$]
    {
        \begin{minipage}[t]{0.24\textwidth}
            \centering
            \includegraphics[width = 0.98\textwidth]{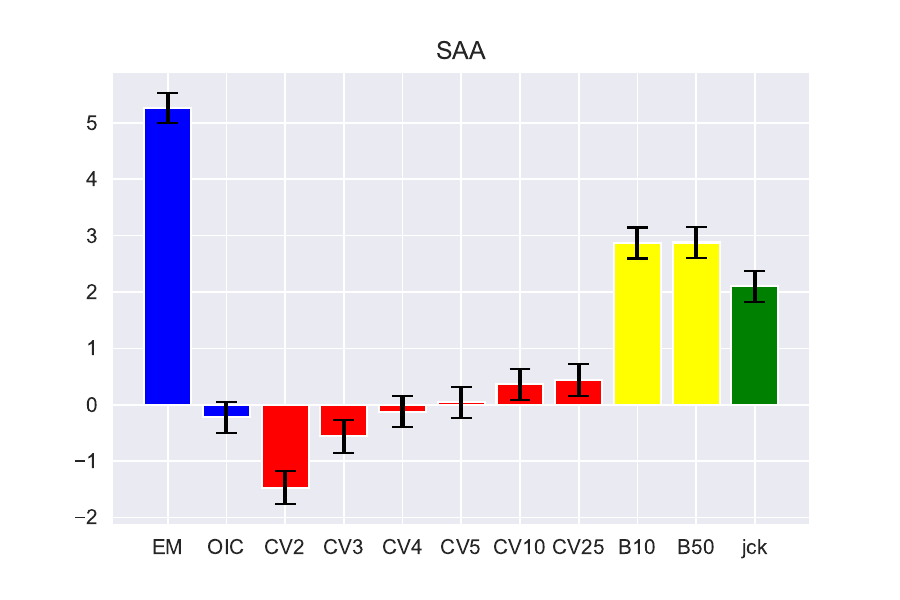}
        \end{minipage}
    }
    \subfloat[Normal, $n = 25$]
    {
        \begin{minipage}[t]{0.24\textwidth}
            \centering
            \includegraphics[width = 0.98\textwidth]{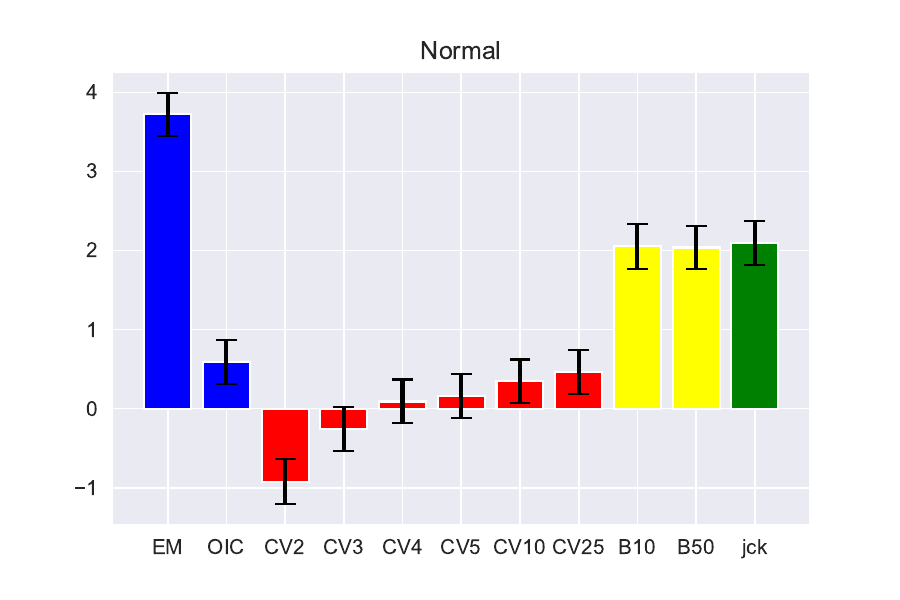}
        \end{minipage}
    }
    \subfloat[Exponential, $n = 25$]
    {
        \begin{minipage}[t]{0.24\textwidth}
            \centering
            \includegraphics[width = 0.98\textwidth]{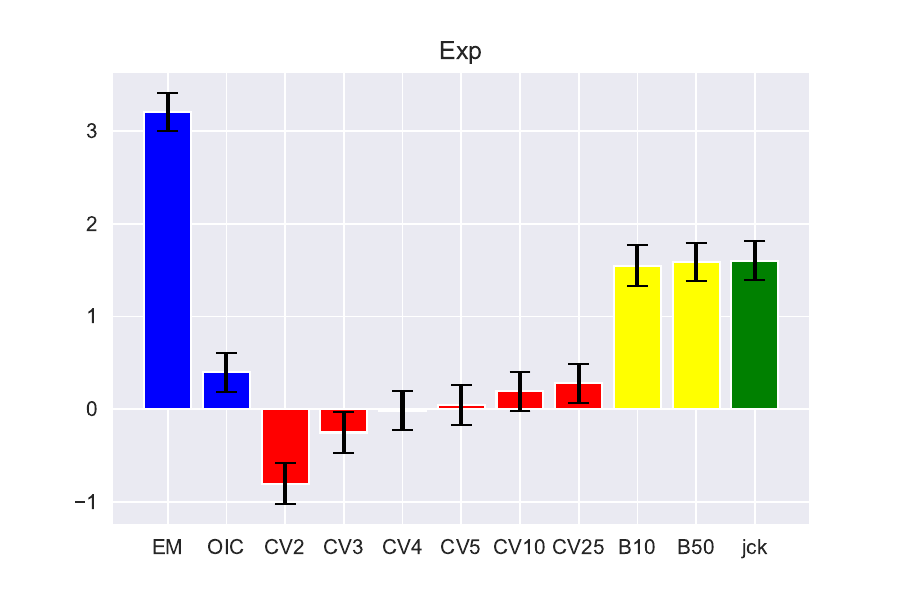}
        \end{minipage}
    }
    \subfloat[Exp-OS, $n = 25$]
    {
        \begin{minipage}[t]{0.24\textwidth}
            \centering
            \includegraphics[width = 0.98\textwidth]{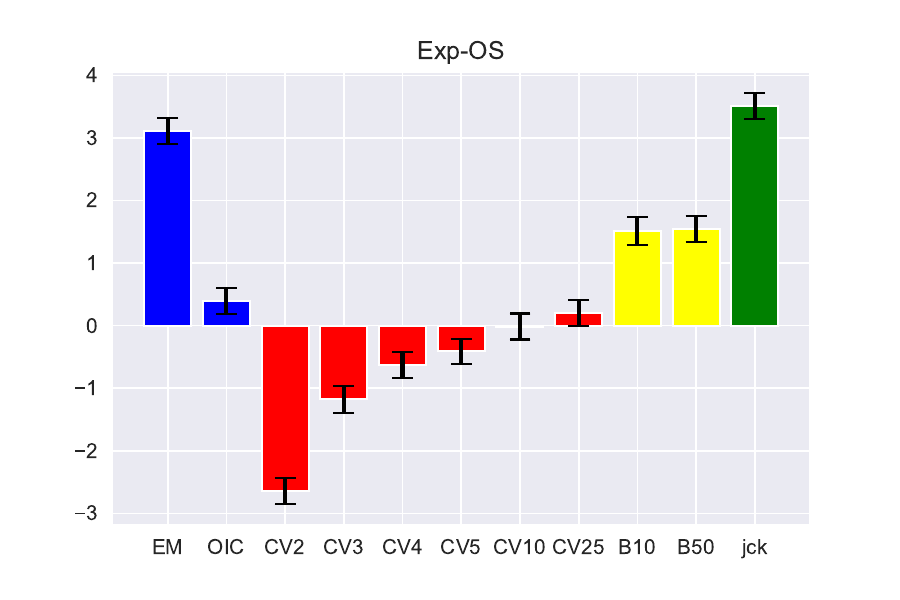}
        \end{minipage}
    }
    
    \subfloat[SAA, $n = 50$]
    {
        \begin{minipage}[t]{0.24\textwidth}
            \centering
            \includegraphics[width = 0.98\textwidth]{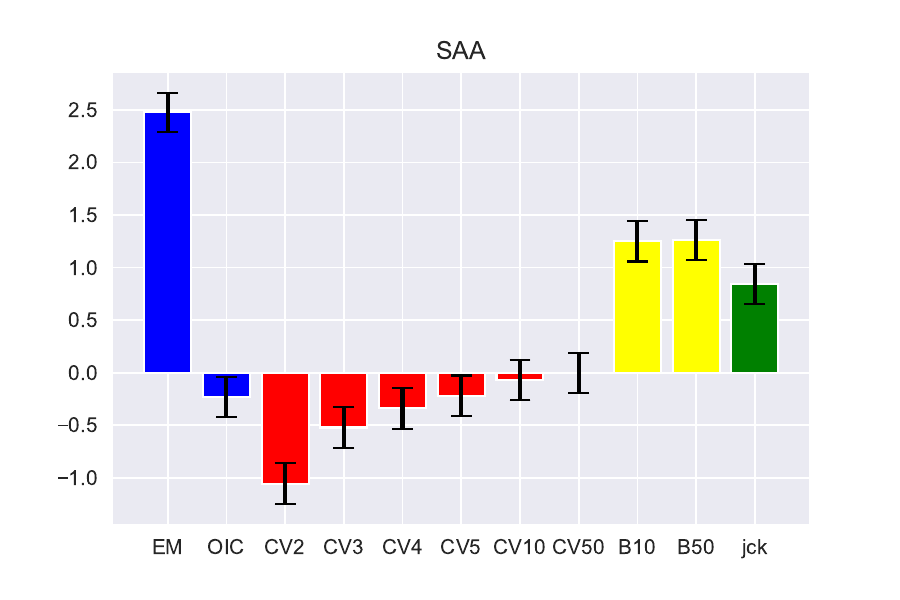}
        \end{minipage}
    }
    \subfloat[Normal, $n = 50$]
    {
        \begin{minipage}[t]{0.24\textwidth}
            \centering
            \includegraphics[width = 0.98\textwidth]{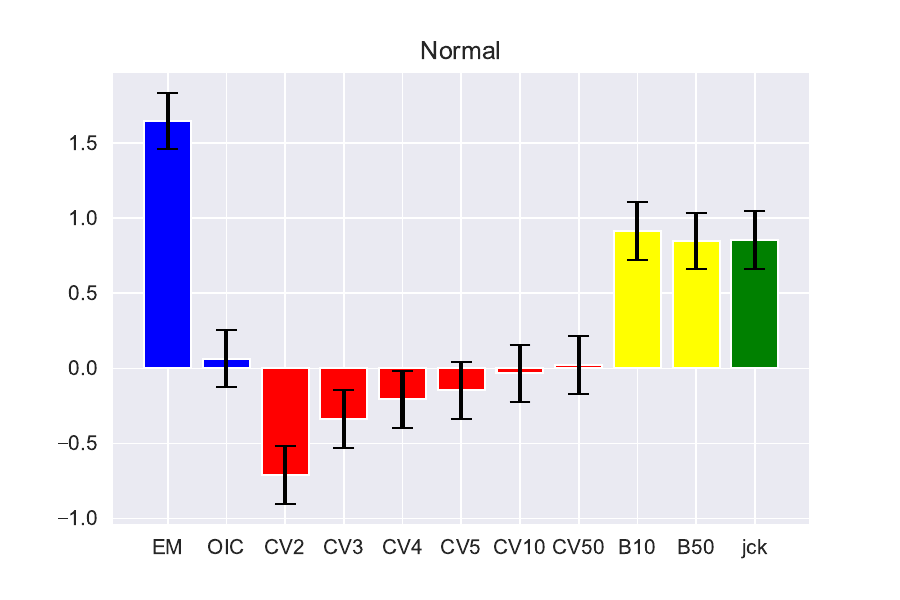}
        \end{minipage}
    }
    \subfloat[Exponential, $n = 50$]
    {
        \begin{minipage}[t]{0.24\textwidth}
            \centering
            \includegraphics[width = 0.98\textwidth]{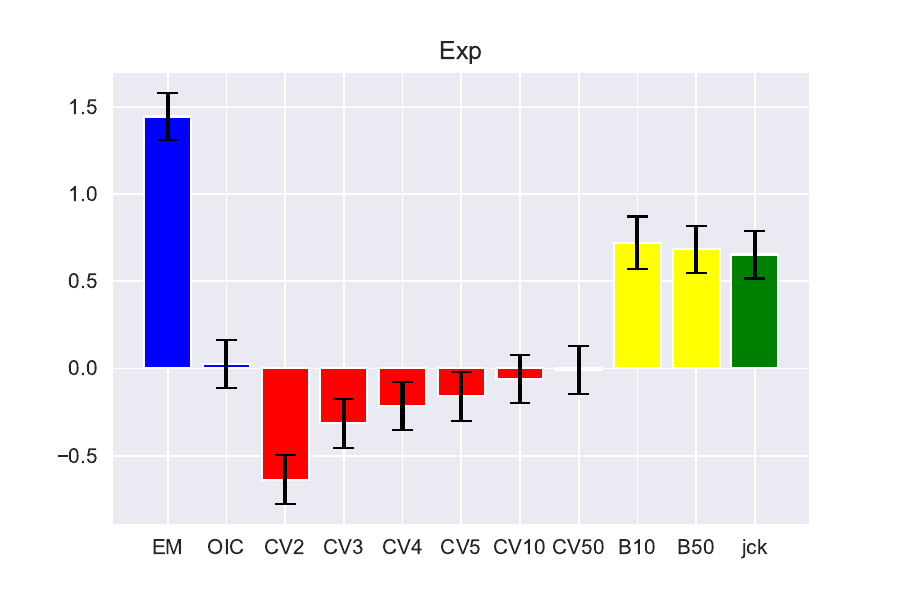}
        \end{minipage}
    }
    \subfloat[Exp-OS, $n = 50$]
    {
        \begin{minipage}[t]{0.24\textwidth}
            \centering
            \includegraphics[width = 0.98\textwidth]{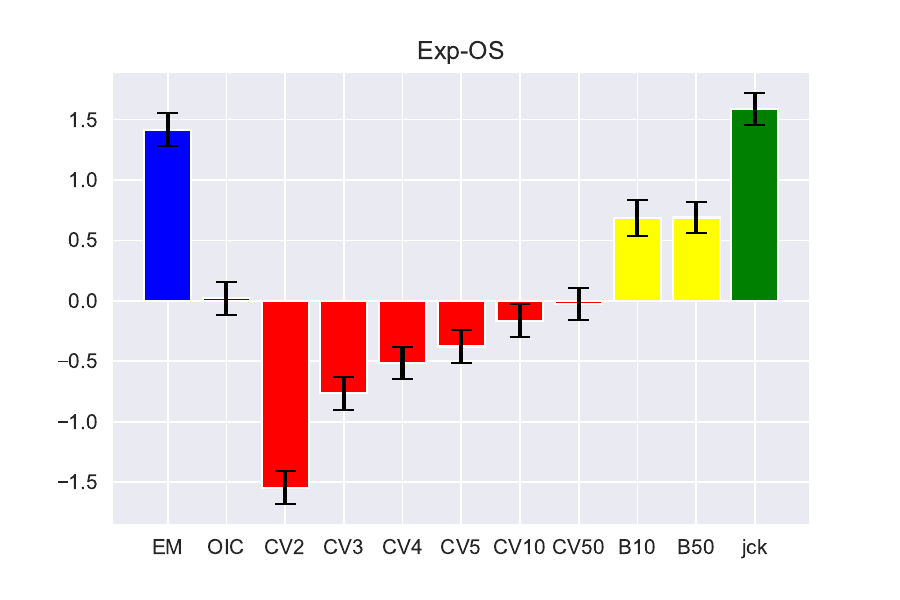}
        \end{minipage}
    }
    
    \subfloat[SAA, $n = 75$]
    {
        \begin{minipage}[t]{0.24\textwidth}
            \centering
            \includegraphics[width = 0.98\textwidth]{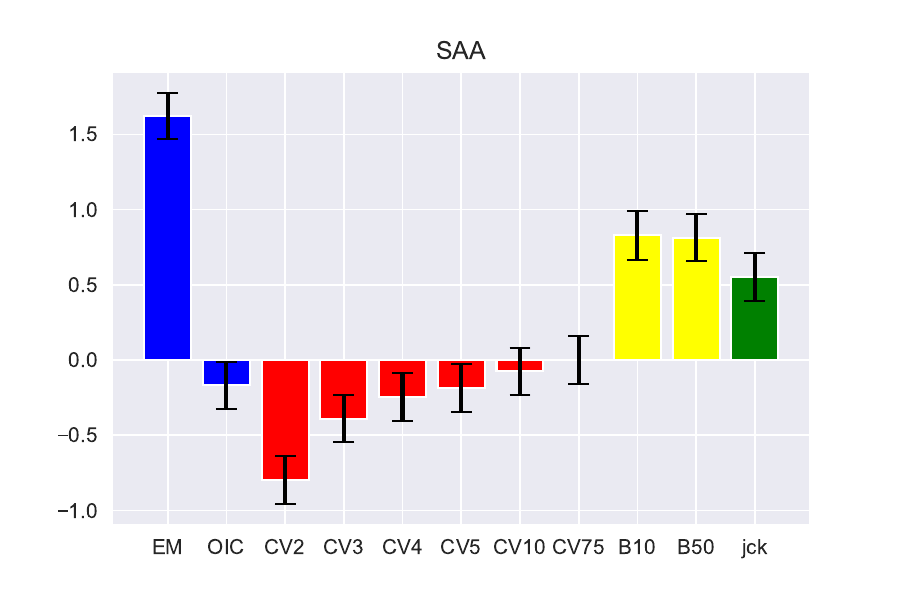}
        \end{minipage}
    }
    \subfloat[Normal, $n = 75$]
    {
        \begin{minipage}[t]{0.24\textwidth}
            \centering
            \includegraphics[width = 0.98\textwidth]{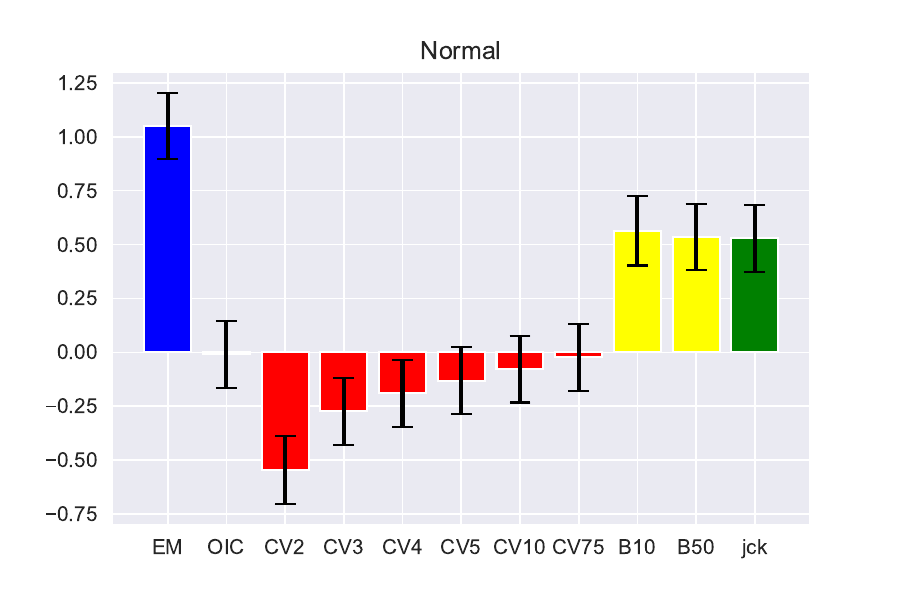}
        \end{minipage}
    }
    \subfloat[Exponential, $n = 75$]
    {
        \begin{minipage}[t]{0.24\textwidth}
            \centering
            \includegraphics[width = 0.98\textwidth]{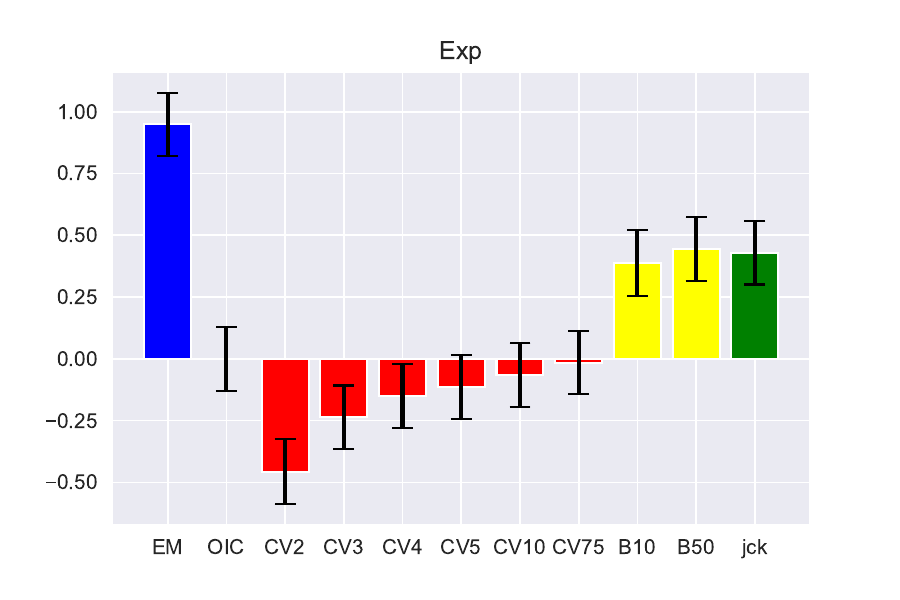}
        \end{minipage}
    }
    \subfloat[Exp-OS, $n = 75$]
    {
        \begin{minipage}[t]{0.24\textwidth}
            \centering
            \includegraphics[width = 0.98\textwidth]{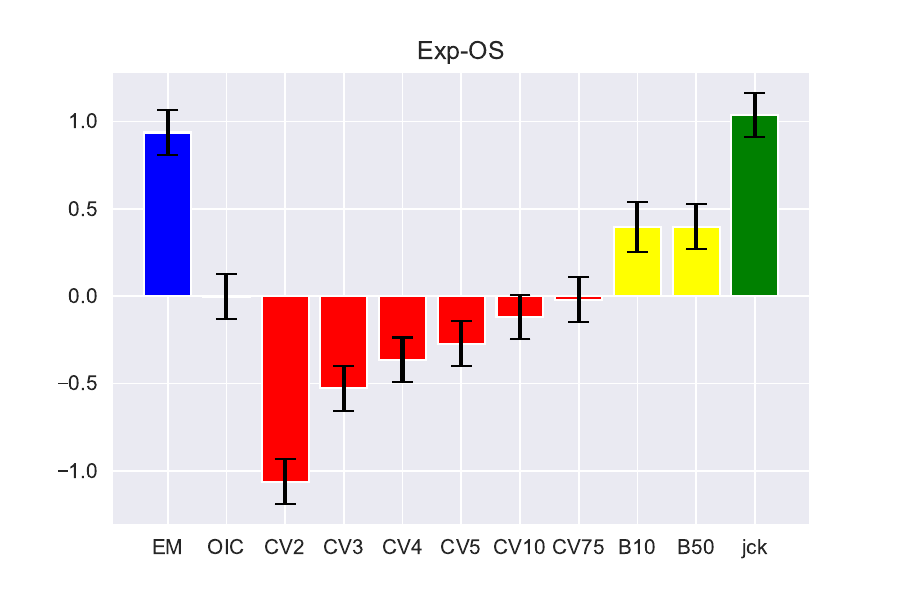}
        \end{minipage}
    }
    
    \subfloat[SAA, $n = 100$]
    {
        \begin{minipage}[t]{0.24\textwidth}
            \centering
            \includegraphics[width = 0.98\textwidth]{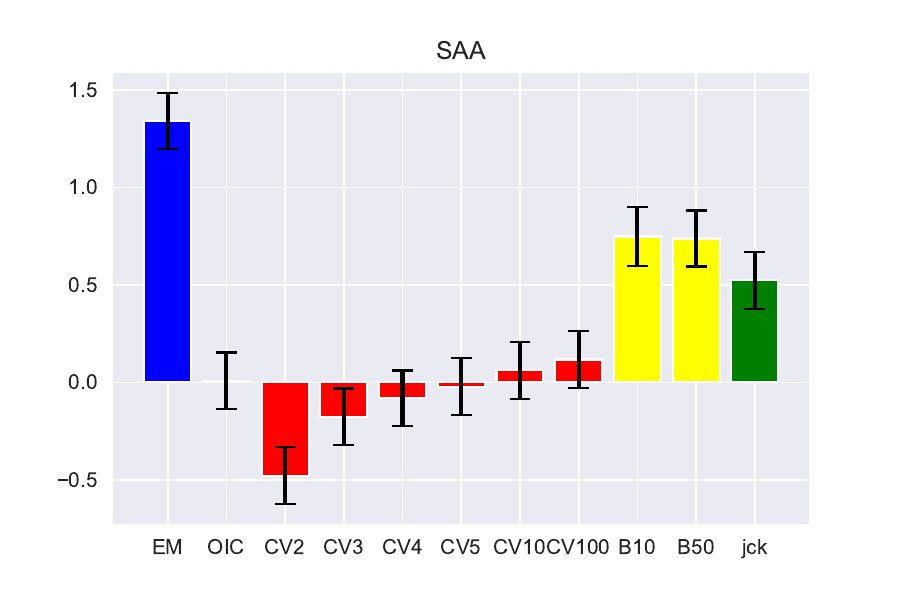}
        \end{minipage}
    }
    \subfloat[Normal, $n = 100$]
    {
        \begin{minipage}[t]{0.24\textwidth}
            \centering
            \includegraphics[width = 0.98\textwidth]{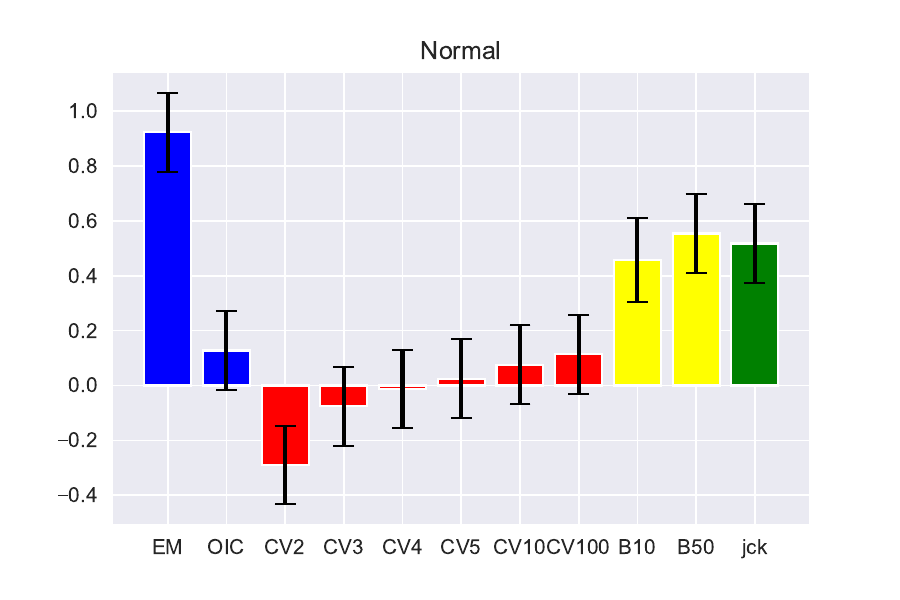}
        \end{minipage}
    }
    \subfloat[Exponential, $n = 100$]
    {
        \begin{minipage}[t]{0.24\textwidth}
            \centering
            \includegraphics[width = 0.98\textwidth]{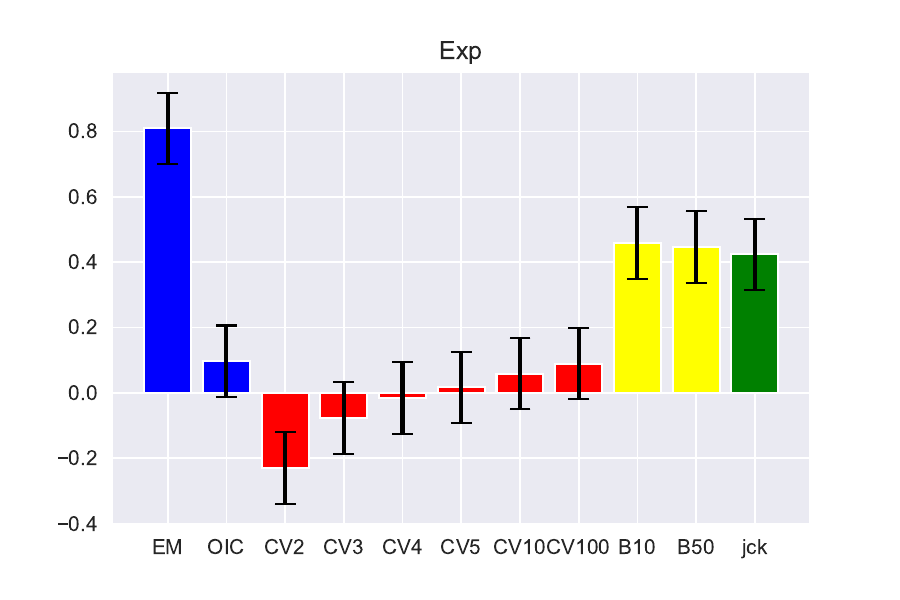}
        \end{minipage}
    }
    \subfloat[Exp-OS, $n = 100$]
    {
        \begin{minipage}[t]{0.24\textwidth}
            \centering
            \includegraphics[width = 0.98\textwidth]{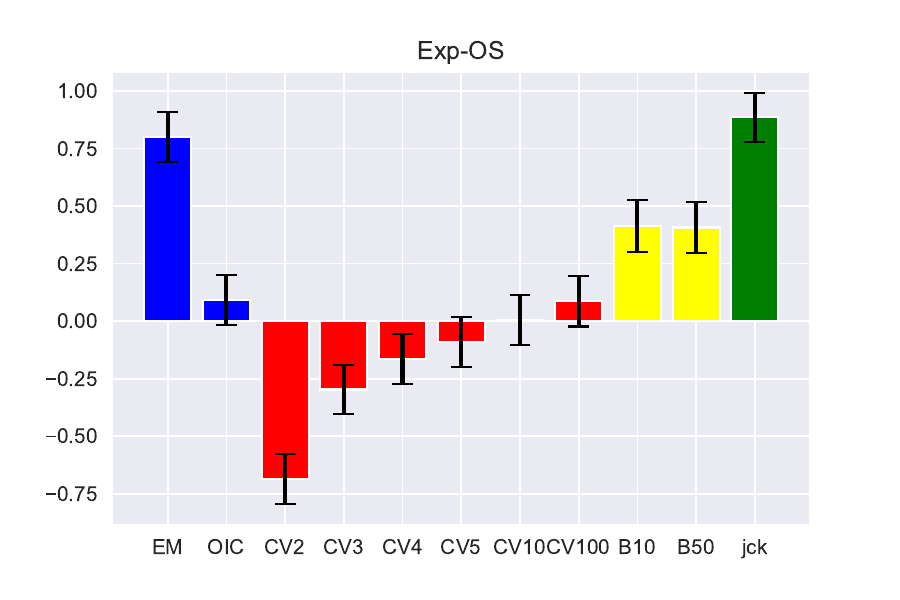}
        \end{minipage}
    }
    
    \subfloat[SAA, $n = 125$]
    {
        \begin{minipage}[t]{0.24\textwidth}
            \centering
            \includegraphics[width = 0.98\textwidth]{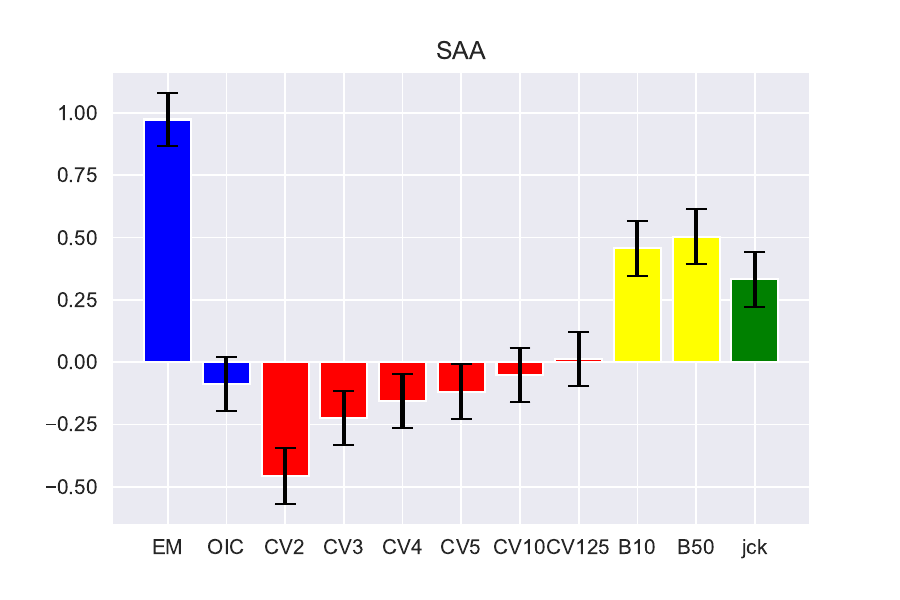}
        \end{minipage}
    }
    \subfloat[Normal, $n = 125$]
    {
        \begin{minipage}[t]{0.24\textwidth}
            \centering
            \includegraphics[width = 0.98\textwidth]{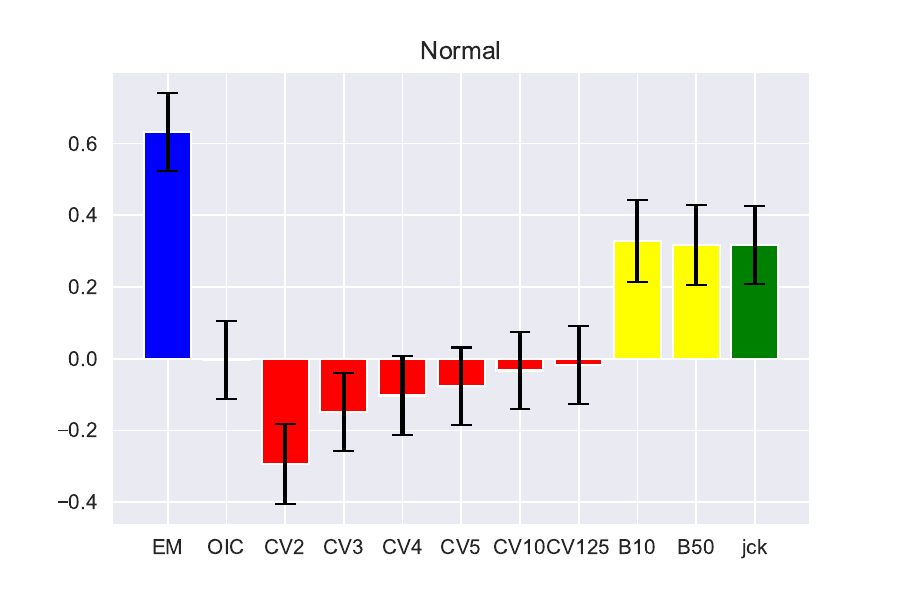}
        \end{minipage}
    }
    \subfloat[Exponential, $n = 125$]
    {
        \begin{minipage}[t]{0.24\textwidth}
            \centering
            \includegraphics[width = 0.98\textwidth]{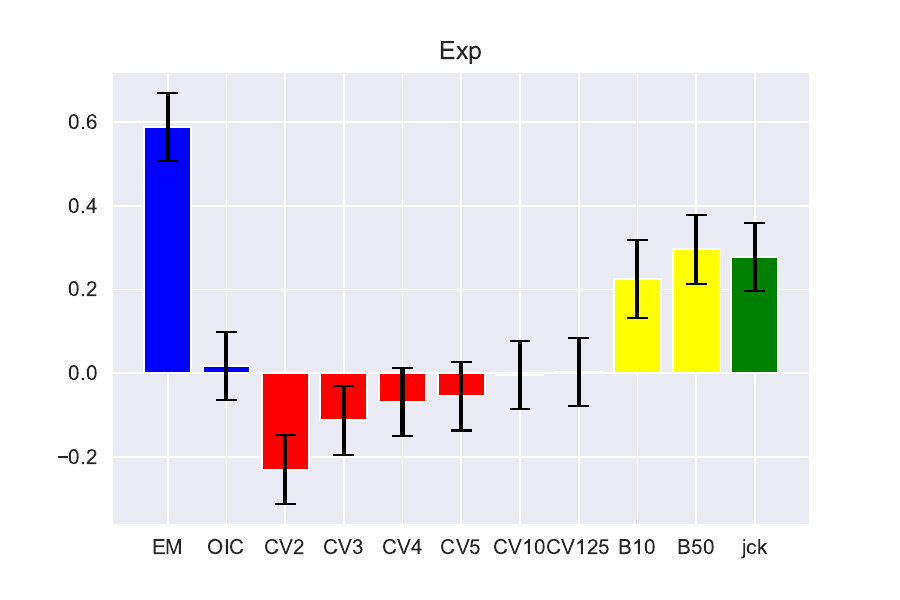}
        \end{minipage}
    }
    \subfloat[Exp-OS, $n = 125$]
    {
        \begin{minipage}[t]{0.24\textwidth}
            \centering
            \includegraphics[width = 0.98\textwidth]{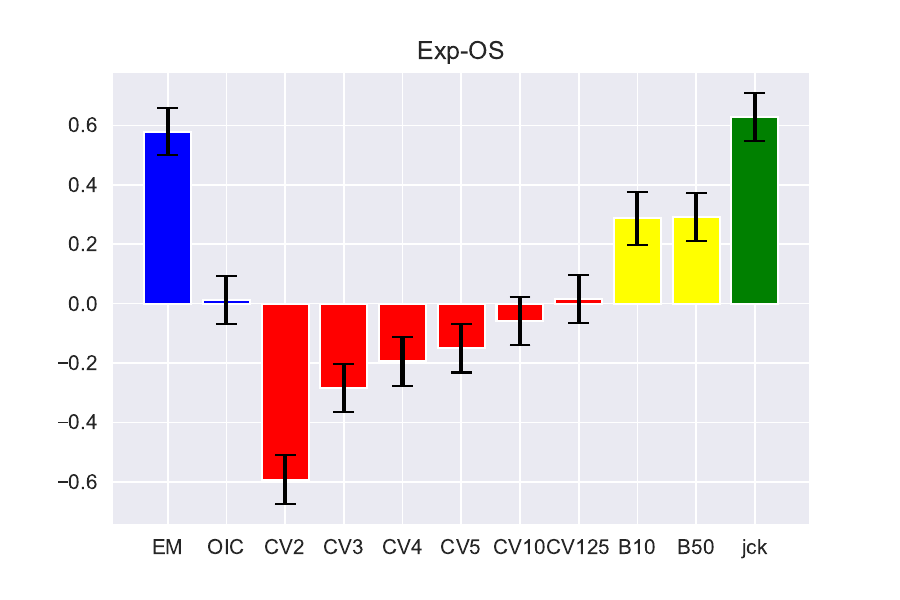}
        \end{minipage}
    }
    \caption{Evaluation results of different data-driven solutions under normal-based distributions}
    \label{fig:newsvendor_normal}
\end{figure}

\begin{figure}
    \centering
    \subfloat[SAA, $n = 25$]
    {
        \begin{minipage}[t]{0.24\textwidth}
            \centering
            \includegraphics[width = 0.98\textwidth]{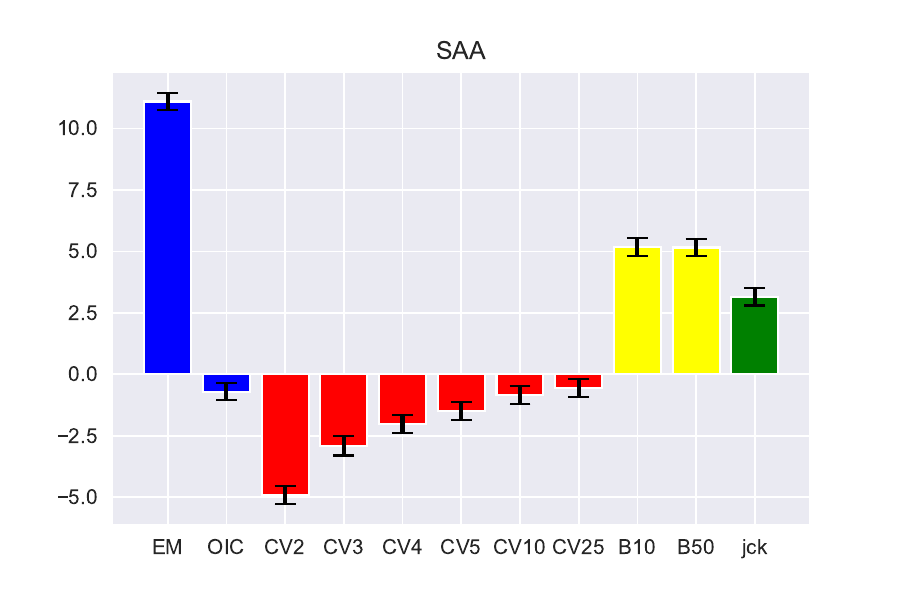}
        \end{minipage}
    }
    \subfloat[Normal, $n = 25$]
    {
        \begin{minipage}[t]{0.24\textwidth}
            \centering
            \includegraphics[width = 0.98\textwidth]{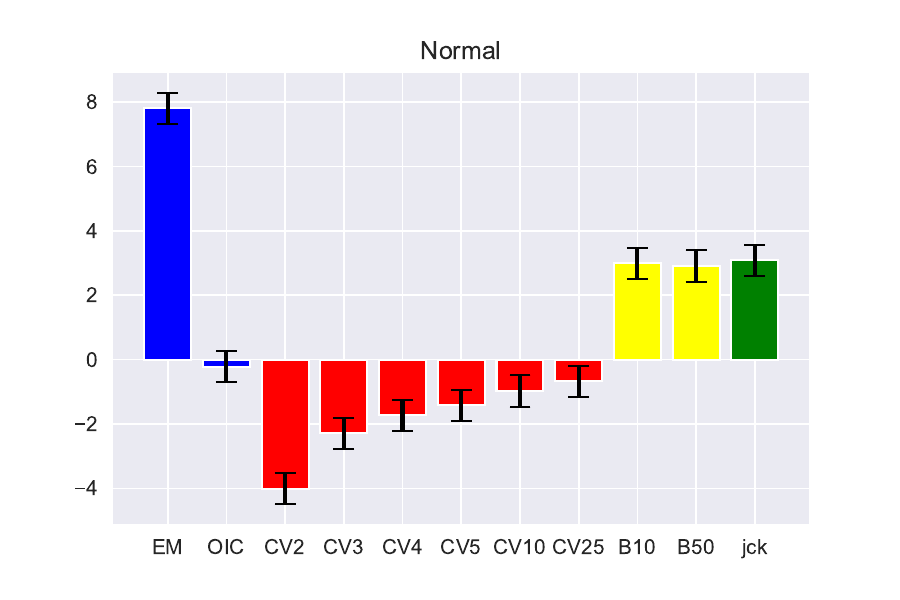}
        \end{minipage}
    }
    \subfloat[Exponential, $n = 25$]
    {
        \begin{minipage}[t]{0.24\textwidth}
            \centering
            \includegraphics[width = 0.98\textwidth]{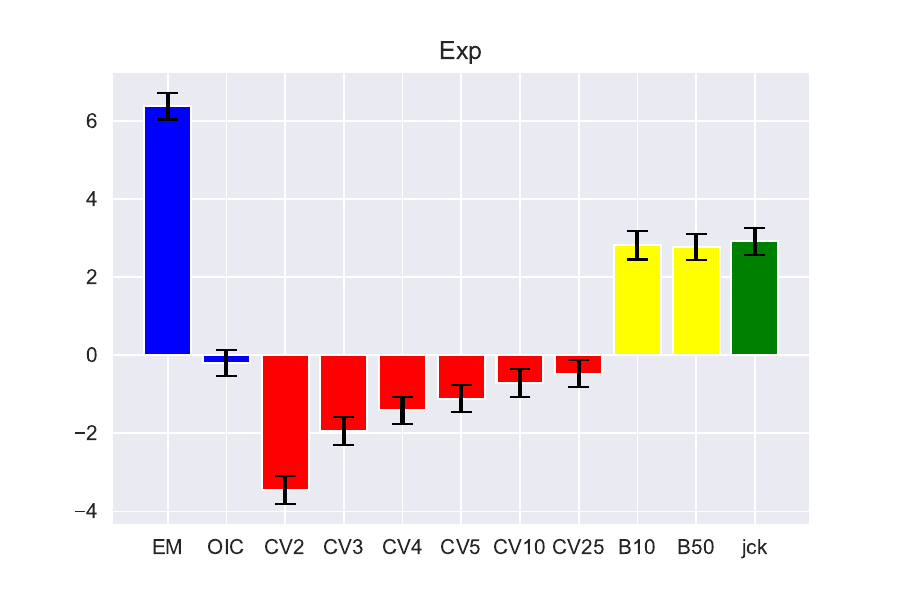}
        \end{minipage}
    }
    \subfloat[Exp-OS, $n = 25$]
    {
        \begin{minipage}[t]{0.24\textwidth}
            \centering
            \includegraphics[width = 0.98\textwidth]{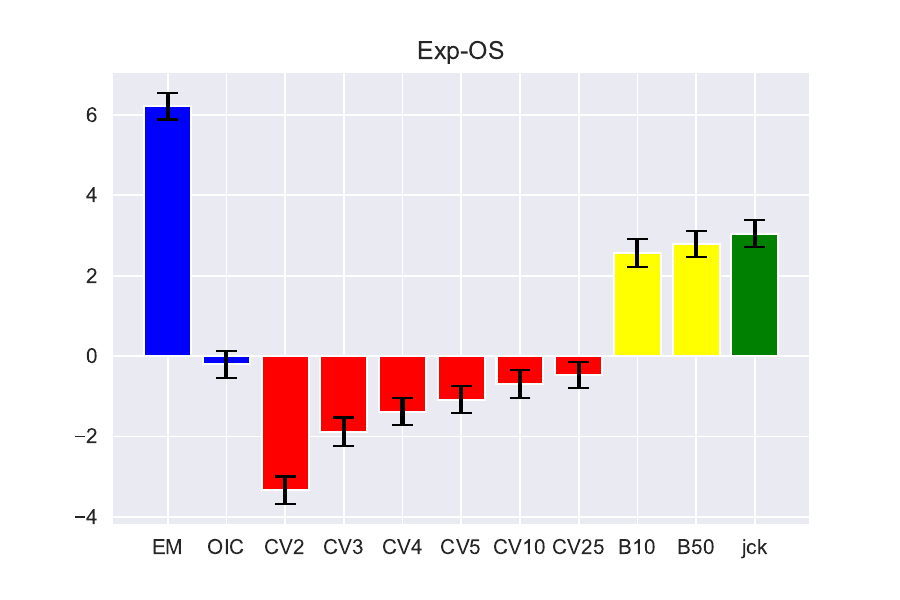}
        \end{minipage}
    }
    
    \subfloat[SAA, $n = 50$]
    {
        \begin{minipage}[t]{0.24\textwidth}
            \centering
            \includegraphics[width = 0.98\textwidth]{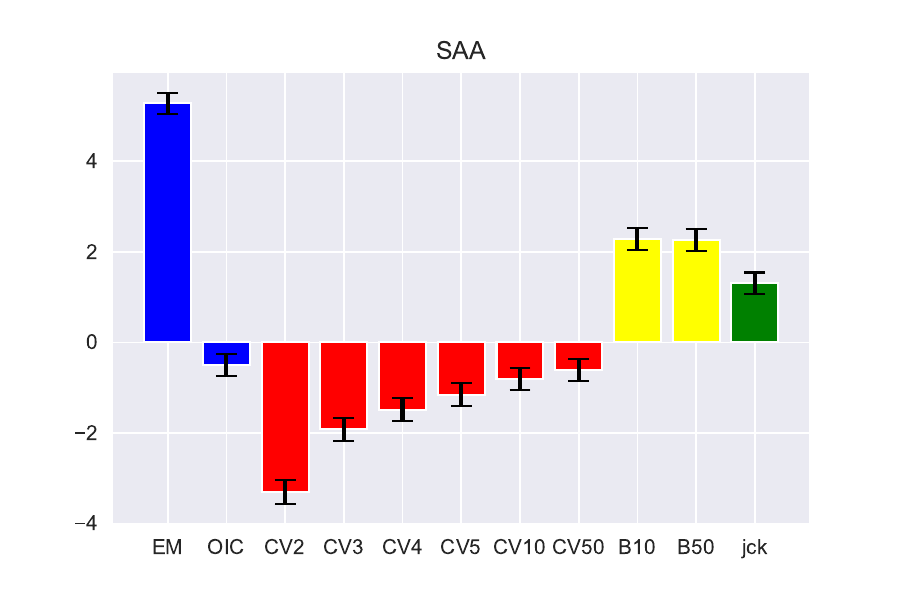}
        \end{minipage}
    }
    \subfloat[Normal, $n = 50$]
    {
        \begin{minipage}[t]{0.24\textwidth}
            \centering
            \includegraphics[width = 0.98\textwidth]{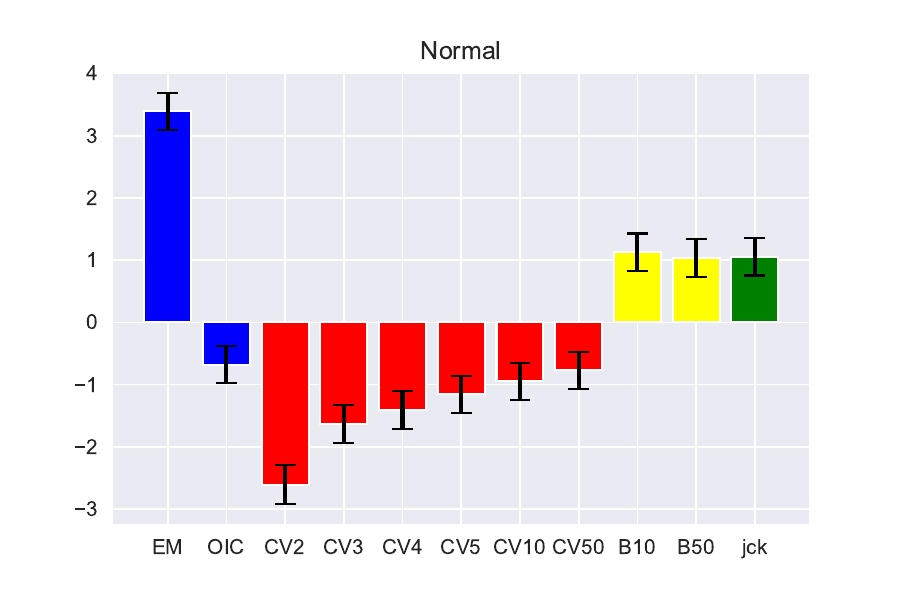}
        \end{minipage}
    }
    \subfloat[Exponential, $n = 50$]
    {
        \begin{minipage}[t]{0.24\textwidth}
            \centering
            \includegraphics[width = 0.98\textwidth]{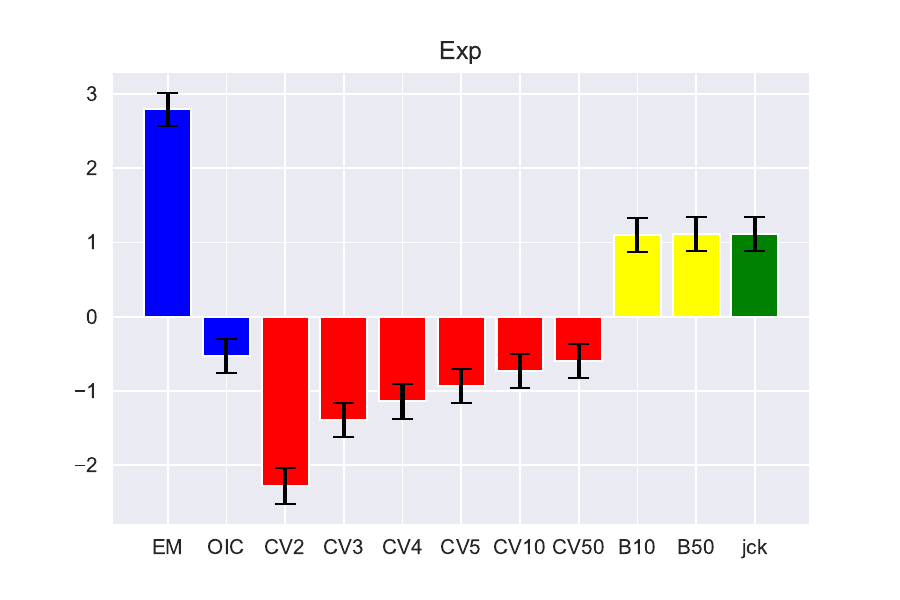}
        \end{minipage}
    }
    \subfloat[Exp-OS, $n = 50$]
    {
        \begin{minipage}[t]{0.24\textwidth}
            \centering
            \includegraphics[width = 0.98\textwidth]{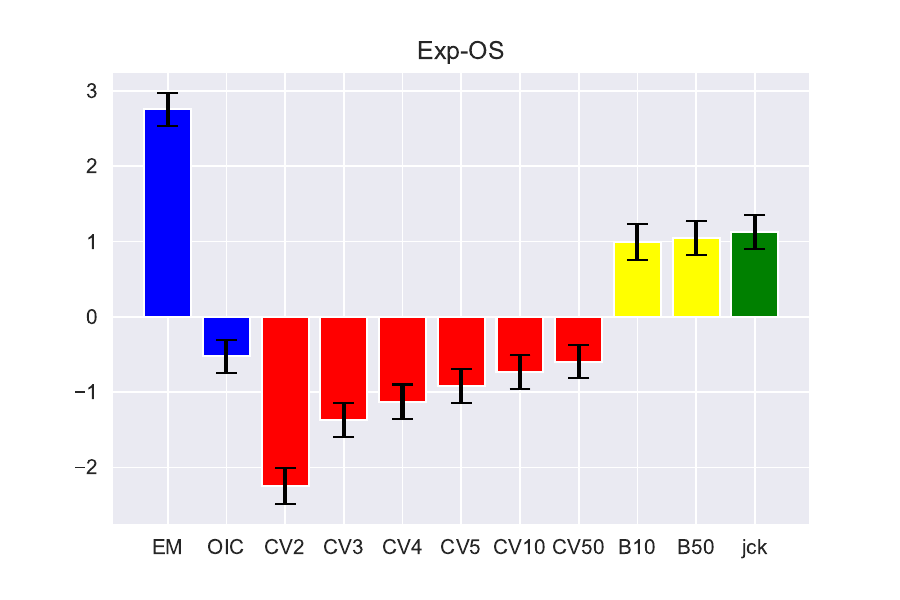}
        \end{minipage}
    }
    
    \subfloat[SAA, $n = 75$]
    {
        \begin{minipage}[t]{0.24\textwidth}
            \centering
            \includegraphics[width = 0.98\textwidth]{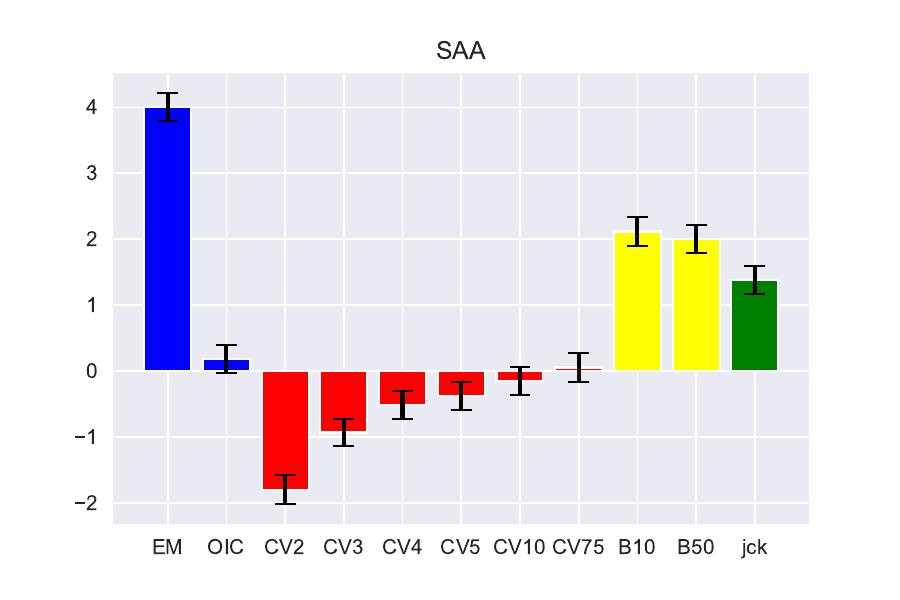}
        \end{minipage}
    }
    \subfloat[Normal, $n = 75$]
    {
        \begin{minipage}[t]{0.24\textwidth}
            \centering
            \includegraphics[width = 0.98\textwidth]{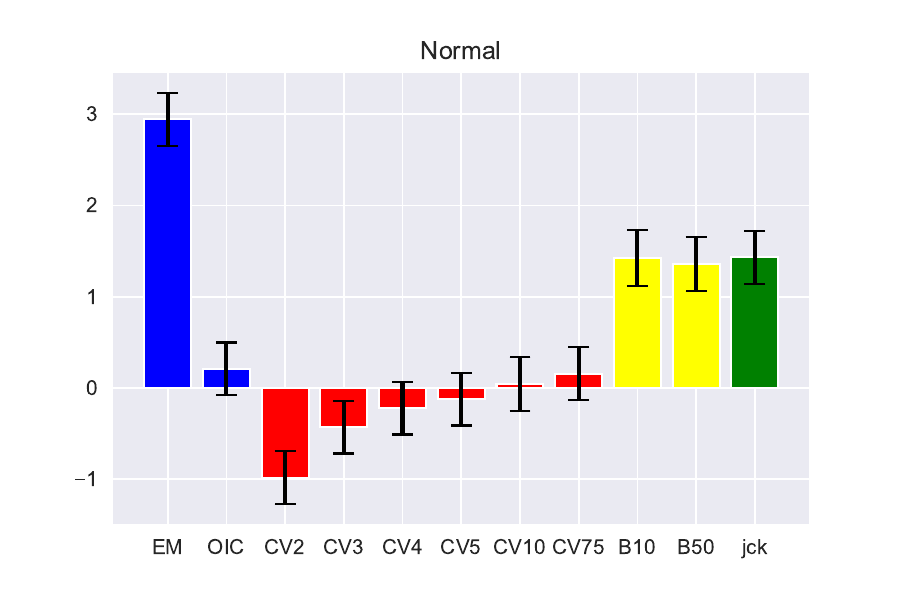}
        \end{minipage}
    }
    \subfloat[Exponential, $n = 75$]
    {
        \begin{minipage}[t]{0.24\textwidth}
            \centering
            \includegraphics[width = 0.98\textwidth]{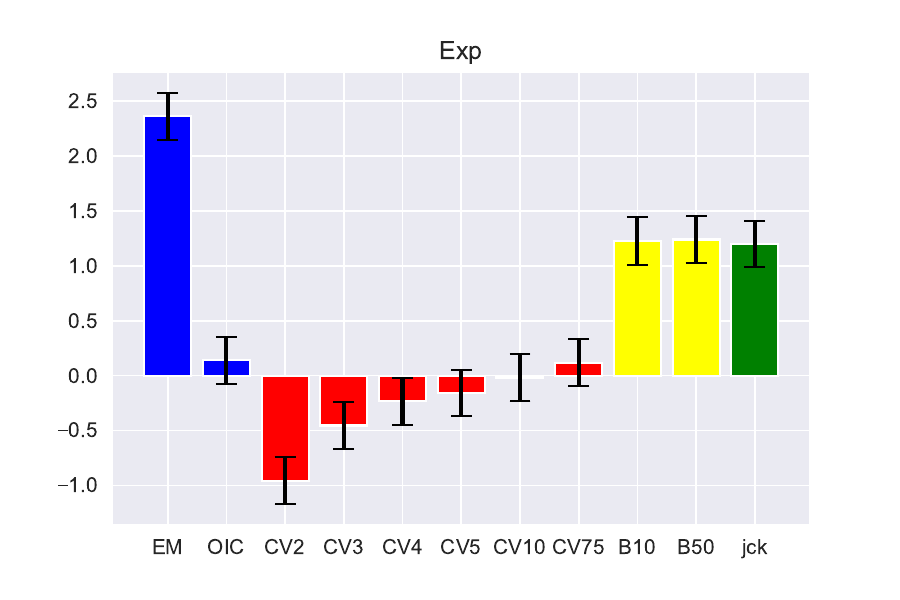}
        \end{minipage}
    }
    \subfloat[Exp-OS, $n = 75$]
    {
        \begin{minipage}[t]{0.24\textwidth}
            \centering
            \includegraphics[width = 0.98\textwidth]{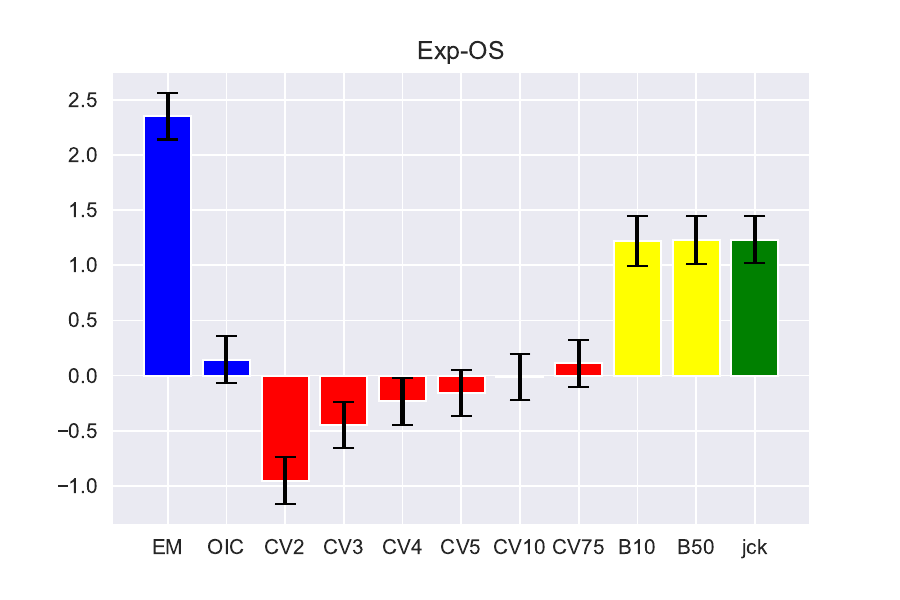}
        \end{minipage}
    }
    
    \subfloat[SAA, $n = 100$]
    {
        \begin{minipage}[t]{0.24\textwidth}
            \centering
            \includegraphics[width = 0.98\textwidth]{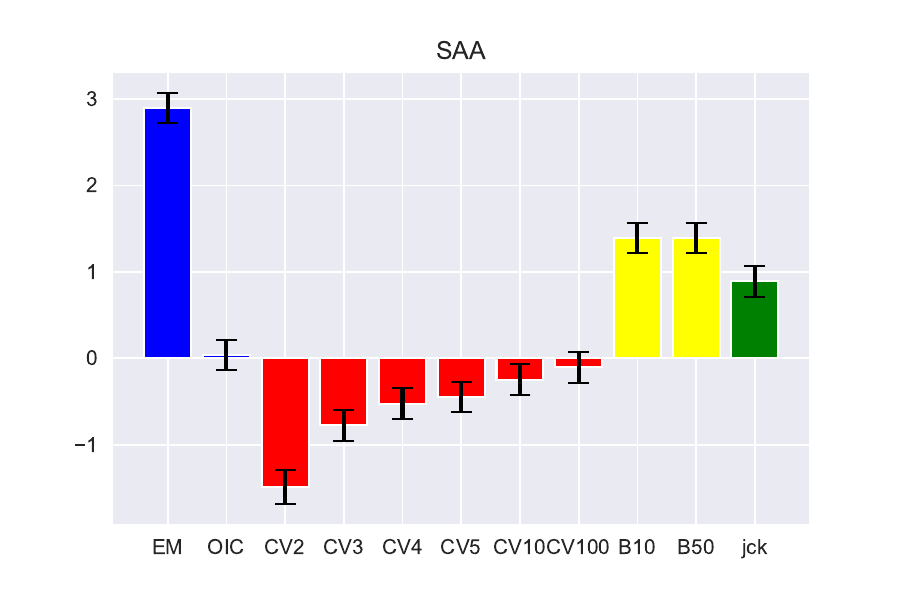}
        \end{minipage}
    }
    \subfloat[Normal, $n = 100$]
    {
        \begin{minipage}[t]{0.24\textwidth}
            \centering
            \includegraphics[width = 0.98\textwidth]{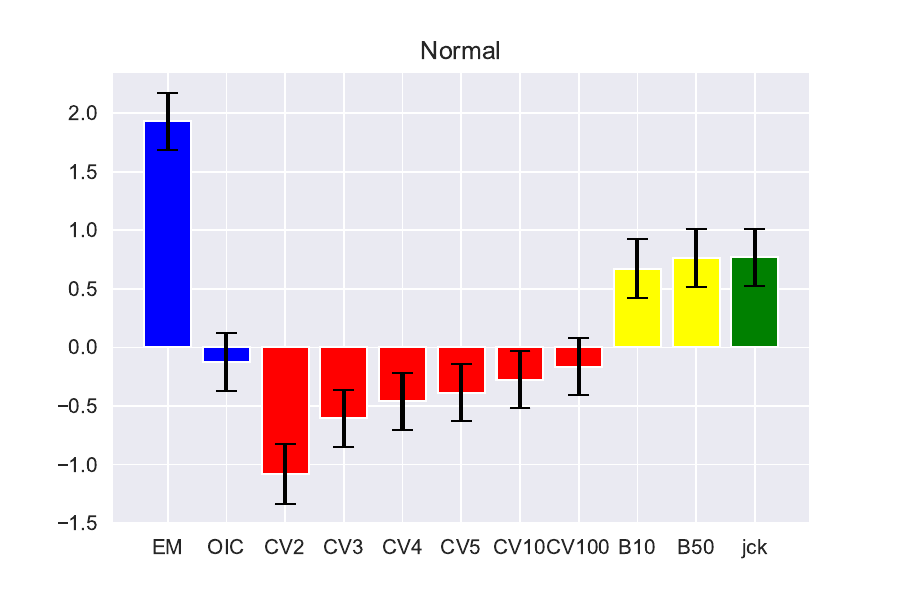}
        \end{minipage}
    }
    \subfloat[Exponential, $n = 100$]
    {
        \begin{minipage}[t]{0.24\textwidth}
            \centering
            \includegraphics[width = 0.98\textwidth]{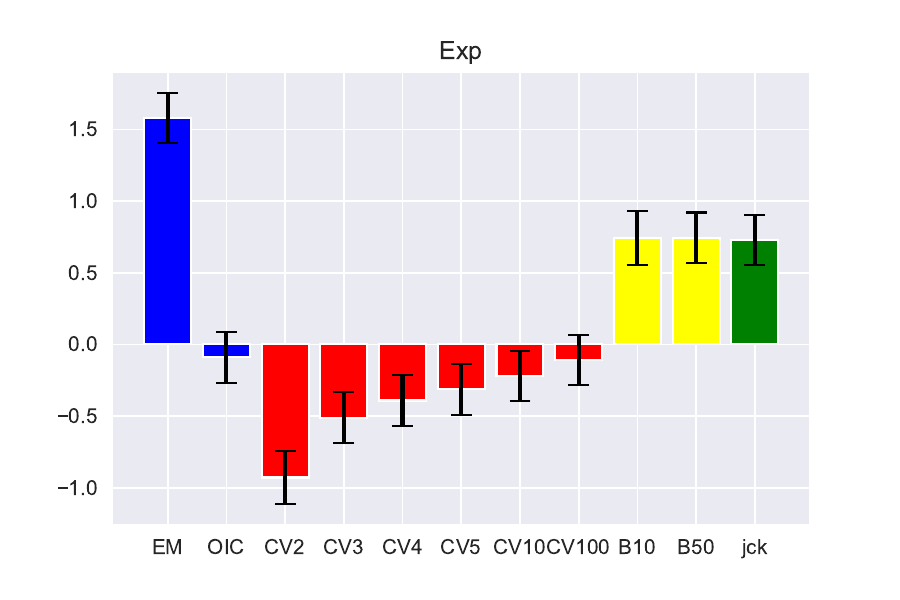}
        \end{minipage}
    }
    \subfloat[Exp-OS, $n = 100$]
    {
        \begin{minipage}[t]{0.24\textwidth}
            \centering
            \includegraphics[width = 0.98\textwidth]{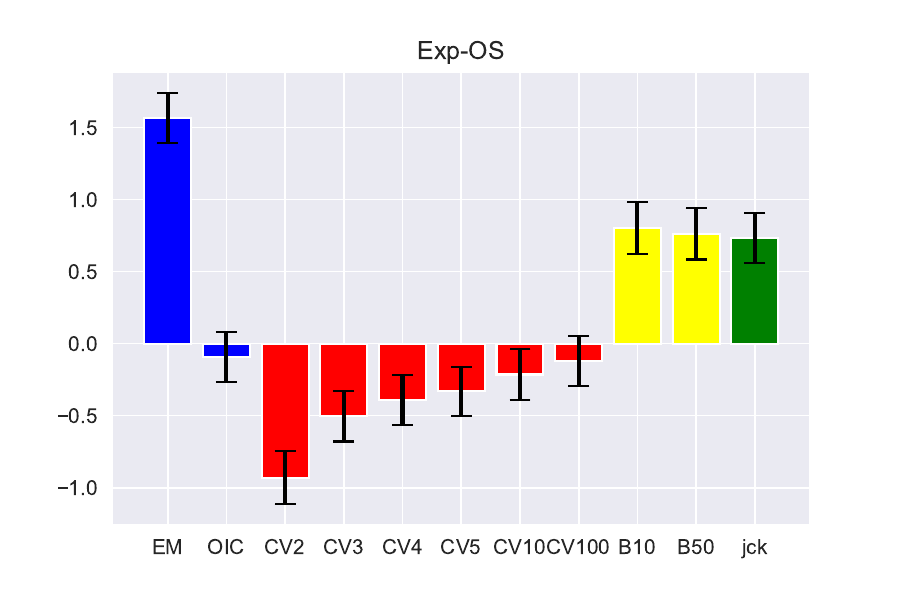}
        \end{minipage}
    }

    \subfloat[SAA, $n = 125$]
    {
        \begin{minipage}[t]{0.24\textwidth}
            \centering
            \includegraphics[width = 0.98\textwidth]{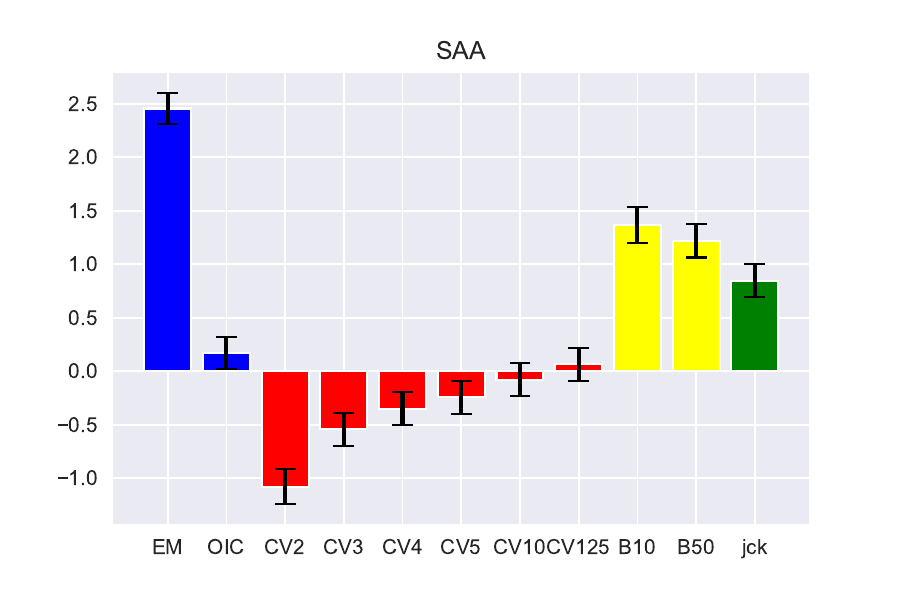}
        \end{minipage}
    }
    \subfloat[Normal, $n = 125$]
    {
        \begin{minipage}[t]{0.24\textwidth}
            \centering
            \includegraphics[width = 0.98\textwidth]{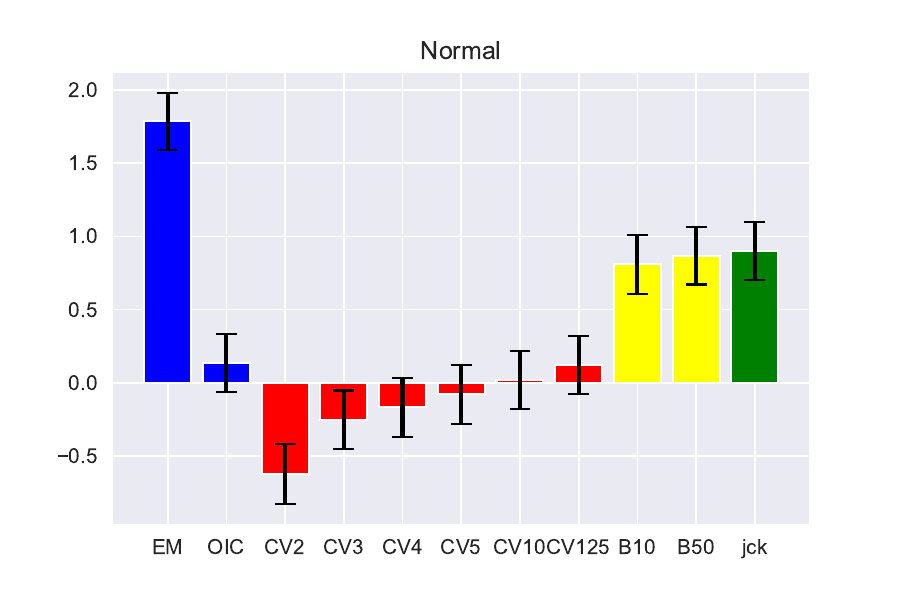}
        \end{minipage}
    }
    \subfloat[Exponential, $n = 125$]
    {
        \begin{minipage}[t]{0.24\textwidth}
            \centering
            \includegraphics[width = 0.98\textwidth]{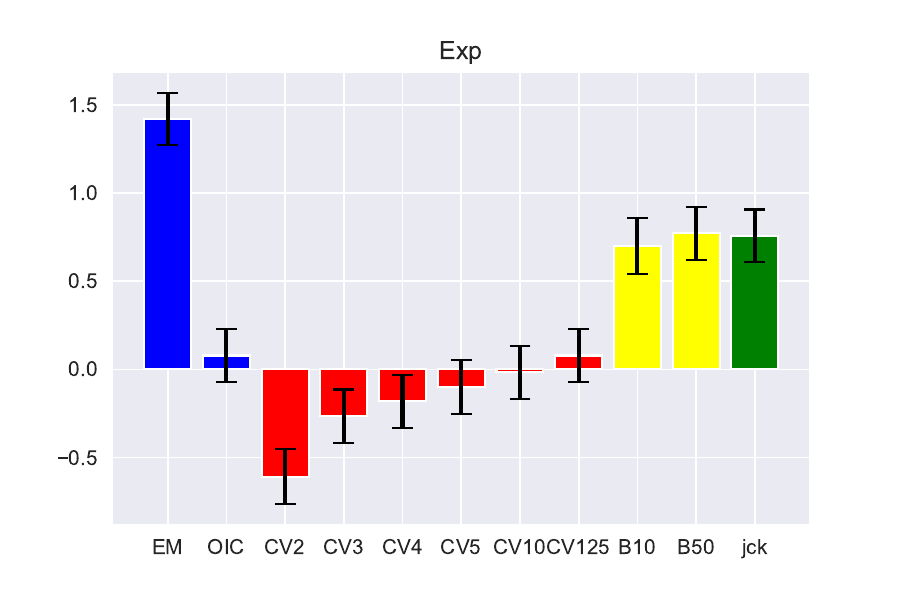}
        \end{minipage}
    }
    \subfloat[Exp-OS, $n = 125$]
    {
        \begin{minipage}[t]{0.24\textwidth}
            \centering
            \includegraphics[width = 0.98\textwidth]{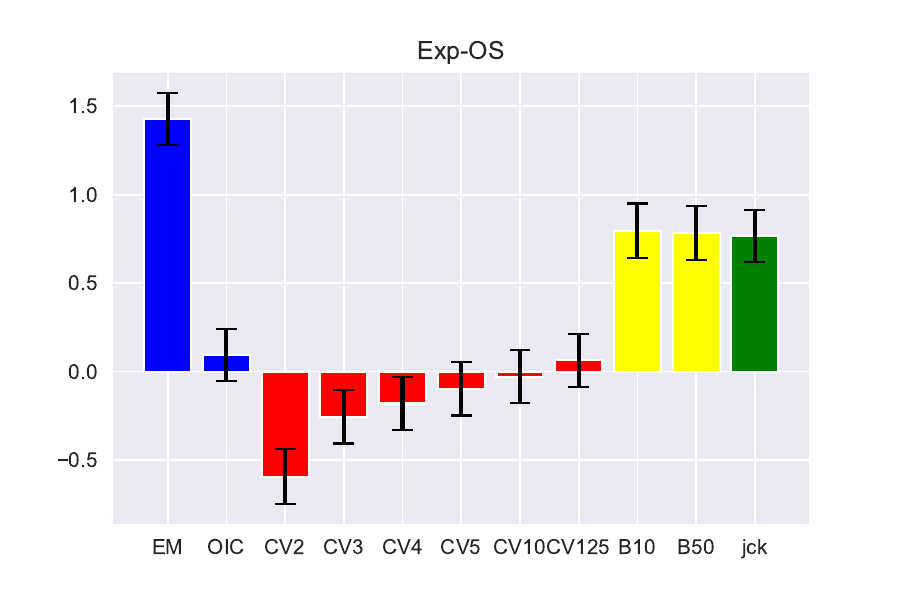}
        \end{minipage}
    }
    \caption{Evaluation results of different data-driven solutions under exponential-based distributions}
    \label{fig:newsvendor_exponential}
\end{figure}

We evaluate different decision rules based on evaluation procedures varying the sample size $n$ in Figures \ref{fig:newsvendor_normal} and \ref{fig:newsvendor_exponential}, each with a different DGP. In general, the closer an evaluation procedure is to zero, the better that evaluation criterion is considered to be.

Across different instances, it is evident that EM consistently leads to an optimistic bias, while the bootstrap and jackknife approaches can eliminate this bias. Additionally, most $K$-fold cross-validation methods tend to incur a pessimistic bias, which aligns with the empirical observation reported in \cite{fushiki2011estimation}. OIC usually performs well when compared with other evaluation approaches. Although some cross-validation approaches may outperform OIC in some instances (e.g. 5-CV in \Cref{fig:newsvendor_normal} $(o)$ and \Cref{fig:newsvendor_exponential} $(t)$), based on results in all instances from the two Figures, no method can uniformly outperform OIC.

\subsection{Real-World Regression}\label{app:regression}
%\paragraph{Models.}

%When the neural network is deep with large model parameters, the OIC is computed in the degenerate case by only taking into account the eigenvalues of the Hessian of significant magnitude while cutting all eigenvalues smaller than one threshold.

\subsubsection{Detailed Setups.} We randomly select 10\% and 30\% of all samples but stratify over red and white wine across different random seeds since the two kinds of wine appear to be quite different, which is reported in \cite{duchi2023distributionally}.

The evaluation performance of EM, OIC, and 5-CV is calculated the same as before in Appendices~\ref{app:setup} and~\ref{app:newsvendor-setup}. However, in this real-world dataset, the true (oracle) decision performance $A$ is unknown and therefore we approximate it directly from the test set. Given the total sample size $N (=6,498)= n + n_{te}$ and the whole dataset $\{\xi_i\}_{i \in [n_{all}]}$, in each problem instance (where $n = 0.1 N$ or $n = 0.3 N$), we index $\{1,\ldots, n\}$ as the training set and $\{n + 1, \ldots, n_{all}\}$ as the test set. That is, we fit each decision rule with the data $\{\xi_j\}_{j \in [n]}$ and evaluate the decision performance via $\{\xi_{j + n}\}_{j \in [n_{te}]}$ approximately. That is to say, for each model $\paranx$, we approximate the true decision performance $A \approx \frac{1}{n_{te}}\sum_{j \in [n_{te}]}h(\paranx;\xi_{j + n})$.

\subsubsection{Decision Rules.}\label{app:regression-decision-rule}
The decision rules of linear and Ridge regression with 2-polynomial features with $\alpha = 0, 1, 5$ are implemented in the \texttt{sklearn.linear\_model.LinearRegression} and  \texttt{sklearn.linear\_model.ridge} module (with polynomial features generated from \texttt{sklearn.preprocessing.PolynomialFeature}). 
The decision rule of the neural network is implemented through a two-hidden-layer architecture in \texttt{torch}, one with 8 hidden neurons, and use \texttt{nn.Softplus()} as the activation unit of each layer, total of 185 parameters. We set the learning rate to be 0.1, and batch size to be 128 during the training procedure of the neural network. 

We estimate the bias of each decision rule according to \Cref{def:oic}. More specifically, we want to compute $\nabla_{\theta} h(x^*(\hat\theta, z);\xi)$ and $\widehat{\IFx}(\xi_i)$ for each decision rule. Recall:
\[h(x^*(\theta, z);\xi) = ((\xi - x^*(\theta, z))^2 - \beta)^+.\]
Then the (sub)gradient $\gradp h(x^*(\theta, z);\xi)$ above can be directly computed since $x^*(\theta, z)$ is a closed-form function in the linear, quadratic or the neural network with $\theta$ being the parameter in the decision rule as we defined in \Cref{subsec:regression}.

In terms of $\widehat{\IFx}(\xi)$, no matter in each decision rule, $\hat\theta \in \argmin_{\theta}\sum_{i \in [n]}\tilde{h}(x^*(\theta,z_i);\xi_i)$ for some losses written as $\tilde h(x^*(\theta,z_i),\xi_i) = (\xi_i - x^*(\theta, z_i))^2+ \alpha/n \|\theta\|_2^2$ , then the estimated influence function can be computed as: $\widehat{\IFx}(\xi_i) = -H_{\hat\theta}^{-1}\nabla_{\theta\theta}^2\tilde h(x^*(\hat\theta,z_i);\xi_i))$, where $H_{\hat\theta} = \frac{1}{n}\sum_{i \in [n]}\nabla_{\theta\theta}^2 \tilde h(x^*(\hat\theta, z_i);\xi_i)$. For linear and quadratic regressions, we can compute the exact formulas.

In the neural network, we directly use \texttt{torch.gradient} and \texttt{torch.hessian} to compute gradients and Hessians in the neural networks and compute the empirical influence function with the same formulation as in \cite{koh2017understanding}. However, the estimated influence function in \cite{koh2017understanding} is not accurate because modern neural networks rarely satisfy the first and second-order optimality conditions at the end of the training period. This issue has motivated refinements in neural networks, as discussed in \cite{yeh2018representer,pruthi2020estimating}. We leave developing better estimators in these deep learning models as future work. 

% In terms of the estimated influence function $\widehat{\IFx}(\cdot)$, since the linear regression as $\theta(\xi^u) = \theta^{\top}\xi^u$, then the empirical influence function of $\hat{\theta}$ is:
% \begin{equation}\label{eq:if-lr}
% \begin{aligned}
%     \widehat{\IFx}(\xi_i) & = (\xi_i^v - \hat{\theta}^{\top} \xi_i^u) \hat{\Sigma}^{-1}\xi_i^u,
% \end{aligned}
% \end{equation}
% where $\hat{\Sigma} = \frac{1}{n}\sum_{i = 1}^n \xi_i^u (\xi_i^u) ^{\top}$. For ridge regression, we can compute the empirical influence function accordingly as:
% \begin{equation}\label{eq:if-lr2}
% \begin{aligned}
%     \widehat{\IFx}(\xi_i) & =  (\hat{\Sigma} + \frac{\alpha}{n} I)^{-1}\Para{ (\xi_i^v - \hat{\theta}^{\top} \xi_i^u) \xi_i^u  + 2\alpha \hat\theta},
% \end{aligned}
% \end{equation}

% When applying to quadratic regression, we can replace $\xi_i^u$ in Equations~\eqref{eq:if-lr} and~\eqref{eq:if-lr2} to $p_2(\xi_i^u)$ for each case. 

{\subsubsection{IFC-OIC.}\label{app:ifc-oic}
  This section provides the details of constructing the influence-function-corrected OIC (IFC-OIC) used in the real-world regression experiment. Let $\tilde h_i(\theta)= (\xi_i - x^*(\theta, z_i))^2 + \alpha/n \|\theta\|_2^2$ and $g_i = \nabla_\theta \tilde h_i(\theta)$ denote the squared-error fitting loss associated with the $i$-th sample and its gradient. From Section~\ref{app:regression-decision-rule}, the estimated influence function % can be written in
  has 
the generic form:
\begin{equation}\label{eq:eif}
    \widehat{\IFx}(\xi_i) = -\widehat H_i^{-1} g_i,
\end{equation}
where $\widehat H_i$ is an estimated Hessian matrix. In the standard OIC implementation from \Cref{coro:oic-procedure-2}, % $\widehat H_i$ is taken to be the empirical Hessian:
\[\widehat H_i = \frac{1}{n}\sum_{j = 1}^n \nabla_{\theta\theta}^2\tilde h_j(\hat\theta), \forall i \in [n].\] 
However, in high-dimensional finite-sample regimes, this raw empirical Hessian can be ill-conditioned and can lead to unstable estimates of $\widehat{\IFx}(\xi_i)$. % Our
IFC-OIC keeps the same bias-correction formula as OIC, but replaces the raw empirical Hessian matrix in the empirical influence function with a more stable Hessian estimate empirically. 
We describe the specific choices of $\widehat H_i$ used in IFC-OIC below.
%formulation as~\eqref{eq:eif} but use a different estimated hessian matrix, which achieves a better accuracy-efficiency frontier. The IFC-OIC is then plugged in with that corrected-influence function. We describe the setup of $H_i$ in IFC-OIC across different decision rules in the following:
\paragraph{Linear and quadratic decision rules.} % In the l
For linear or quadratic regression, we set:
\[\widehat H_i = \Para{1 - \phi_i^\top
  \left(\Phi^\top \Phi+\alpha I\right)^\dagger \phi_i} \frac{1}{n}\sum_{j = 1}^n \nabla_{\theta\theta}^2\tilde h_j(\hat\theta),\forall i \in [n] \]
where the factor $1-\phi_i^\top
  \left(\Phi^\top \Phi+\alpha I\right)^\dagger \phi_i \in (0, 1)$ accounts for the fact that the $i$-th sample affects both the fitted parameter and its own fitted values,
  %before the empirical Hessian matrix corrects for the fact that the $i$-th sample affects both the fitted parameter and its own fitted value, 
  with $\Phi$ being the design matrix and $\phi$ denoting the $i$-th row of $\Phi$. More specifically, $\phi_i = z_i$ % and $\phi_i=z_i$ 
  for the linear model and $\phi_i = p_2(z_i)$ for the
quadratic model. In simple linear models, this form of correction is the standard exact leave-one-out adjustment~\citep{giordano2019swiss,wilson2020approximate}, i.e., $\widehat{\IFx}(\xi_i) = n(\hat\theta - \theta_{-i})$, and differs from the influence function estimate with the raw empirical Hessian by a term of order $\Theta(D_{\theta}/n)$, which vanishes as $n \to \infty$.

\paragraph{Neural-network decision rule.}
For the neural-network model, we use the following approximation
\[
  \widehat H_i
  =
  \frac{1}{n}\sum_{j=1}^n g_j g_j^\top, \forall i \in [n].
\]
This
avoids explicitly forming second derivatives of the neural network and gives a positive semidefinite curvature approximation, and is closely related to the Gauss-Newton and empirical Fisher approximations used in second-order neural-network optimization~\citep{martens2010deep,pascanu2014revisiting}.

Note that we do not claim that IFC-OIC entirely addresses the underestimation issue in high-dimensional problems, as it is not intended as a unified high-dimensional correction for arbitrary cost functions. The above corrections exploit the specific regression structure of the decision rules considered in \Cref{subsec:regression}. Developing a unified theory for such influence-function corrections is an interesting direction for future work. 

}

\end{document}